%% file: main.tex
\documentclass[a4paper, 12pt, oneside, openright]{memoir}
\usepackage[T1]{fontenc}
\usepackage[english]{babel}
\usepackage[pdftex]{graphicx}   
\usepackage{epsfig}             
\usepackage{amsmath}            
\usepackage{amssymb}            
\usepackage{amsthm}
\usepackage{mathtools}
\usepackage{natbib}
\usepackage{multicol}
\usepackage[bookmarks=true, breaklinks=true]{hyperref}
\usepackage[x11names]{xcolor}
\usepackage{empheq}
\usepackage{bm}
\usepackage{enumitem}
\usepackage{nicematrix}
\usepackage{nicefrac, xfrac}
\usepackage[many]{tcolorbox}
\usepackage[noabbrev]{cleveref}
\usepackage{cancel}
\usepackage[normalem]{ulem}
\usepackage{minted}
\usepackage{units}
\usepackage{gensymb}
\usepackage{multirow} 
\usepackage{float}

\usepackage{wrapfig}
\usepackage{lscape}
\usepackage{rotating}
\usepackage{epstopdf}

\usepackage{algorithm2e}
\NoCaptionOfAlgo

\def\author{Jan Michalczyk}
\def\place{Klagenfurt}
\def\submitdate{September 2025}

\usepackage{tabularray}
\UseTblrLibrary{booktabs}
\usepackage{array}

\usepackage{acro}
\input{acro/acro.sty}
\acsetup{make-links}

\input{preamble/preamble}

\input{preamble/notation_defs}
\input{styles/helpers}
\input{styles/math}
\input{styles/robotics}

\SetAlFnt{\small}

\input{styles/titlepage}

\input{styles/informal_titlepage}
\begin{document}

\setsecnumdepth{subsection}
\maxsecnumdepth{subsection}

\maxtocdepth{subsection}


\chapterstyle{madsen}
\renewcommand*{\chaptitlefont}{\normalfont\Huge\bfseries\raggedleft}
\renewcommand*{\chapnamefont}{\normalfont\LARGE\scshape\raggedleft}
\renewcommand*{\printchapternum}{%
    \makebox[0pt][l]{\hspace{0.4em}%
      \resizebox{!}{4ex}{%
        \chapnamefont\bfseries\thechapter}%
    }%
  }%

\makeatletter
\newcommand{\figrefcaption}[1]{\renewcommand\fnum@figure{\figurename~\thefigure~(from~#1)}}
\newcommand{\tabrefcaption}[1]{\renewcommand\fnum@table{\tablename~\thetable~(from~#1)}}
\makeatother


\frontmatter
\titlepage
\clearpage
\input{chapters/affidavit}
\clearpage
\input{chapters/disclosure}
\clearpage
\input{chapters/acknowledgments}
\clearpage
\input{chapters/abstract}
\clearpage
\input{chapters/publications}
\clearpage
\printacronyms
\clearpage
\tableofcontents*
\clearpage
\listoffigures*
\clearpage
\listoftables*

\mainmatter
\input{chapters/introduction}
\input{chapters/related_work}
\input{chapters/radar_fundamentals}
\input{chapters/ekf_rio}
\input{chapters/fg_rio}
\input{chapters/dl_matching}
\input{chapters/conclusion}

\appendix
\input{chapters/appendix}

\backmatter
\bibliographystyle{plain}
\bibliography{bib/rio}

\end{document}

%% file: preamble/preamble.tex
\usepackage[T1]{fontenc}

\usepackage{MnSymbol}
\usepackage{mathdots}

%% file: preamble/notation_defs.tex
\providecommand{\R}{\mathbb{R}}

\usepackage{accents}
\makeatletter
\providecommand{\scirc}{%
    \hbox{\fontfamily{\rmdefault}\fontsize{0.4\dimexpr(\f@size pt)}{0}\selectfont{\raisebox{-0.52ex}[0ex][-0.52ex]{$\circ$}}}}

\makeatother

\mathchardef\mhyphen="2D

%% file: styles/math.tex
\theoremstyle{plain}

\theoremstyle{definition}

\theoremstyle{remark}

\tcbset{
    box/.style={
        enhanced,
        breakable=true,
        colback=black!2.5,
        colframe=black!65,
        fonttitle=\bfseries,
    }
}

\newtcbtheorem[number within=chapter, number freestyle={(E\noexpand\thechapter.\noexpand\arabic{\tcbcounter})}, crefname={example}{examples}]{boxexample}{Example}{box}{th}

\newtcbtheorem[number within=chapter, number freestyle={(I\noexpand\thechapter.\noexpand\arabic{\tcbcounter})}, crefname={intuition}{intuitions}]{boxintuition}{Intuition}{box}{th}

\newtcbtheorem[number within=chapter, number freestyle={(Q\noexpand\thechapter.\noexpand\arabic{\tcbcounter})}, crefname={question}{quetions}]{boxquestion}{Question}{box}{th}

\newtcbtheorem[number within=chapter, number freestyle={(A\noexpand\thechapter.\noexpand\arabic{\tcbcounter})}, crefname={algorithm}{algorithms}]{boxalgorithm}{Algorithm}{box}{th}

\newtcbtheorem[number within=chapter, number freestyle={(S\noexpand\thechapter.\noexpand\arabic{\tcbcounter})}, crefname={snippet}{snippets}]{boxsnippet}{Snippet}{box}{th}

\newcommand{\norm}[1]{\lVert #1 \rVert}

\newcommand{\chrulefill}{%
  \leaders\hrule height \dimexpr\fontdimen22\scriptscriptfont2+0.2pt\relax
                 depth -\dimexpr\fontdimen22\scriptscriptfont2-0.2pt\relax
          \hfill}
\newcommand{\xoverline}[2]{%
  \mathord{
    \vbox{\offinterlineskip
      \halign{##\cr
        \chrulefill$\,\scriptscriptstyle#1\,$\chrulefill\cr
        \noalign{\kern.2ex}
        $#2$\cr
      }%
    }%
  }%
}

\newcommand\inverse{{\mathsf{-\!1}}}
\newcommand\transpose{{\mathsf{T}}}

\newcommand\bomega{{\boldsymbol\omega}}

\newcommand\bsigma{{\boldsymbol\sigma}}
\newcommand\bSigma{{\mathbf{\Sigma}}}

\newcommand\btheta{\boldsymbol{\theta}}

\newcommand\bPhi{\boldsymbol{\Phi}}

\newcommand\bZero{\boldsymbol{0}}

\makeatletter
\newcommand{\generateAlphabet}[4]{%
  \def\@tempa{#1} 
  \count@=`#3
  \loop
  \begingroup\lccode`?=\count@
  \lowercase{\endgroup\@namedef{\@tempa ?}{#2{?}}}%
  \ifnum\count@<`#4
  \advance\count@\@ne
  \repeat
}

\generateAlphabet{c}{\mathcal}{A}{Z}
\generateAlphabet{cc}{\mathcal}{a}{z}
\generateAlphabet{cs}{\scriptscriptstyle\mathcal}{A}{Z}

\generateAlphabet{v}{\mathbf}{A}{Z}
\generateAlphabet{v}{\mathbf}{a}{z}

\newcommand\referencediag[4]{{}^{{#2}}_{{#4}}{{#1}}^{}_{{#3}}}   
\newcommand\referencediagt[5]{{}^{{#2}}_{{#4}}{{#1}}^{#5}_{{#3}}}   
\newcommand\reference[4]{{\referencediag{#1}{#2}{#3}{#4}}}  
\newcommand\referencet[5]{{\referencediagt{#1}{#2}{#3}{#4}{#5}}}  

\newcommand{\skewmat}[1]{\left[#1\right]_{{\scriptscriptstyle \!\times}}}

%% file: styles/robotics.tex
\newcommand{\Vector}[3]{\prescript{#1}{}{\bm{#2}}_{#3}}

\newcommand{\Rot}[2]{\prescript{#1}{}{\mathbf{R}}_{#2}}

\newcommand{\frameofref}[1]{\{$#1$\}}



%% file: styles/titlepage.tex
\newcommand*{\titlepage}{%
\thispagestyle{empty}
\begingroup%
\centering
{\huge\bfseries\scshape Radar-Inertial Odometry For Computationally Constrained Aerial Navigation}\par
\vspace{1.0\baselineskip}
{\large\mdseries \author, M.Sc.}\par
\vspace{1.0\baselineskip}
{\large\mdseries\scshape DISSERTATION}\par
\vspace{0.5\baselineskip}
{\small submitted in fulfillment of the requirements for the degree of\\[0.25\baselineskip]
{\itshape Doctor of Technical Sciences}}\par
\vspace{1.0\baselineskip}
\includegraphics[width=0.33\linewidth]{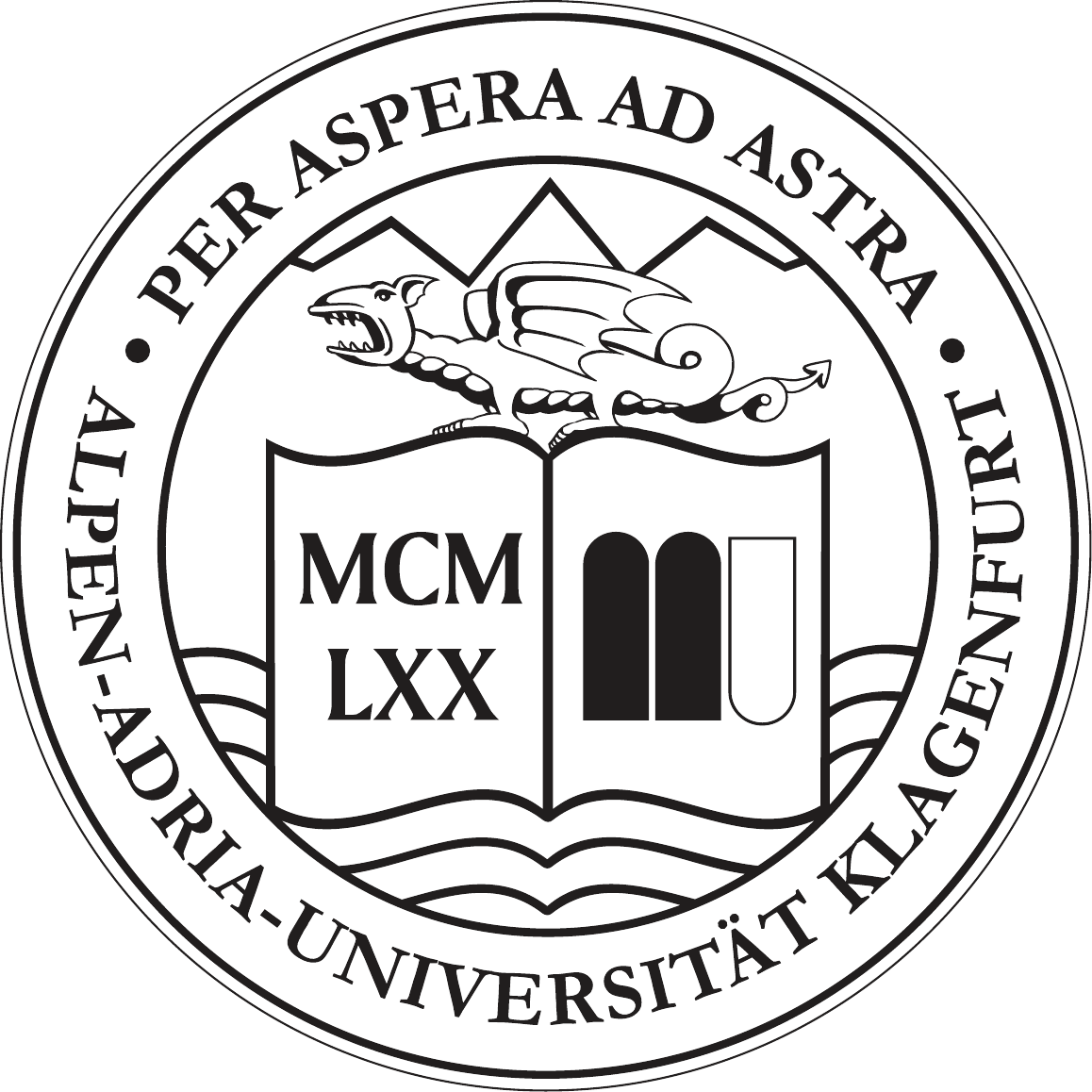}\\[1.0\baselineskip]
{University of Klagenfurt\\[0.25\baselineskip]
Faculty of Technical Sciences}\par
\vspace{1.0\baselineskip}
\begin{tblr}
{
    rows = {m},
    column{1} = {valign=t, halign=c, wd=0.45\textwidth, font=\small},
    column{2} = {valign=t, halign=c, wd=0.45\textwidth, font=\small},
}
\SetCell[c=2]{c}{Supervisors}\\[0.25\baselineskip]
{Univ.-Prof. Dr. Stephan Weiss\\University of Klagenfurt\\Institute of Smart Systems Technologies} & 
{Assoc. Prof. Dr. Jan Steinbrener\\University of Klagenfurt\\Institute of Smart Systems Technologies}\\[1.0\baselineskip]
\SetCell[c=2]{c}{Expert Reviewers}\\[0.25\baselineskip]
{Prof. Konstantinos Alexis, Ph.D.\\Norwegian University of Science and Technology\\Department of Engineering Cybernetics} & 
{Prof. Martin Saska, Ph.D.\\Czech Technical University in Prague\\Department of Cybernetics}\\
\end{tblr}\\[1.5\baselineskip]
{\place, \submitdate}\par
\vfill
\endgroup}

%% file: chapters/affidavit.tex
\chapter*{Affidavit}

I hereby declare in lieu of an oath that
\begin{itemize}
    \item the submitted academic paper is entirely my own work and that no auxiliary materials have been used other than those indicated,
    \item I have fully disclosed all assistance received from third parties during the process of writing the thesis, including any significant advice from supervisors,
    \item any contents taken from the works of third parties or my own works that have been included either literally or in spirit have been appropriately marked and the respective source of the information has been clearly identified with precise bibliographical references (e.g. in footnotes),
    \item I have fully and truthfully declared the use of generative models (Artificial Intelligence, e.g. ChatGPT, Grammarly Go, Midjourney) including the product version,
    \item to date, I have not submitted this paper to an examining authority either in Austria or abroad and that
    \item when passing on copies of the academic thesis (e.g. in printed or digital form), I will ensure that each copy is fully consistent with the submitted digital version.
\end{itemize}
I am aware that a declaration contrary to the facts will have legal consequences.\\[0.25\baselineskip]
\begin{tblr}{
    rows = {m},
    width = \linewidth,
    colspec = {X[1,l] X[1,r]}
}
\author~m.p. & \place, \submitdate
\end{tblr}

%% file: chapters/disclosure.tex
\chapter*{Disclosure}

I hereby declare the use of AI-assisted technologies, including Grammarly and GPT 3.5, as online spell-checking tools during the preparation of this manuscript. Note that none of the aforementioned technologies were directly used to generate complete paragraphs within this thesis.

%% file: chapters/acknowledgments.tex
\chapter*{Acknowledgments}

\noindent I would like to thank my supervisors, Professor Stephan Weiss and Professor Jan Steinbrener for their support and guidance. My gratitude to you both is beyond words.\\

\noindent I want to thank all my colleagues from the Control of Networked Systems (CNS) group for interesting research discussions as well as great time and fun together.\\

\noindent I wish to thank my parents for sparking my curiosity of the world which eventually led me to the path of a researcher.\\

\noindent Last but not least, I would like to express my deepest appreciation to Professor Martin Saska and Professor Konstantinos Alexis for reviewing this dissertation.\\

\noindent I wish to dedicate this work to my wife Julieta and our daughter Gali\ensuremath\heartsuit\\

%% file: chapters/abstract.tex
\chapter*{Abstract}
Radar or RAdio Detection And Ranging, which is a process consisting in transmission and reception of electromagnetic waves enabling the measurement of distance, bearing angle and radial velocity of reflecting objects, has a longstanding history across various domains of engineering. Recently, the progress in the radar sensing technology consisting in the miniaturization of the packages and increase in their measuring precision has drawn the interest of the robotics research community. Indeed, a crucial task enabling autonomy in robotics is to precisely determine the pose of the robot in space. To fulfill this task sensor fusion algorithms are often used, in which data from one or several exteroceptive sensors like, for example, LiDAR, camera, laser ranging sensor or GNSS are fused together with the Inertial Measurement Unit (IMU) measurements to obtain an estimate of the navigation states of the robot. Nonetheless, owing to their particular sensing principles, some exteroceptive sensors are often incapacitated in extreme environmental conditions, like extreme illumination or presence of fine particles in the environment like smoke or fog. Radars are largely immune to aforementioned factors thanks to the characteristics of electromagnetic waves they use. In this thesis, we present Radar-Inertial Odometry (RIO) algorithms to fuse the information from IMU and radar in order to estimate the navigation states of a (Uncrewed Aerial Vehicle) UAV capable of running on a portable resource-constrained embedded computer in real-time and making use of inexpensive, consumer-grade sensors. We present novel RIO approaches relying on the multi-state tightly-coupled Extended Kalman Filter (EKF) and Factor Graphs (FG) fusing instantaneous velocities of and distances to 3D points delivered by a lightweight, low-cost, off-the-shelf Frequency Modulated Continuous Wave (FMCW) radar with IMU readings. We also show a novel way to exploit advances in deep learning to retrieve 3D point correspondences in sparse and noisy radar point clouds. This thesis advances the knowledge of how lightweight and consumer-grade FMCW millimeter-wave radar sensors can be used together with IMU sensors to build localization systems for UAVs.

%% file: chapters/publications.tex
\chapter*{Publications}
\addcontentsline{toc}{chapter}{Publications}

The results presented in this dissertation are included in multiple papers in internationally recognized peer-reviewed conference proceedings.

\section*{Conference papers}
\begin{itemize}[label=$\blacktriangle$, leftmargin=*]
\item J. Michalczyk, C. Sch\"offmann, A. Fornasier, J. Steinbrener, and S. Weiss, “Radar-Inertial State-Estimation for UAV Motion in Highly Agile Manoeuvres”, in 2022 International Conference on Unmanned Aircraft Systems (ICUAS), 2022.  \href{https://ieeexplore.ieee.org/abstract/document/9836130}{[IEEExplore]}

\item J. Michalczyk, R. Jung, and S. Weiss, “Tightly-Coupled EKF-Based Radar-Inertial Odometry”, 2022 IEEE/RSJ International Conference on Intelligent Robots and Systems (IROS), 2022.  \href{https://ieeexplore.ieee.org/document/9981396}{[IEEExplore]}

\item J. Michalczyk, R. Jung, C. Brommer, and S. Weiss, "Multi-State Tightly-Coupled EKF-Based Radar-Inertial Odometry With Persistent Landmarks" 2023 IEEE International Conference on Robotics and Automation (ICRA), 2023. \href{https://ieeexplore.ieee.org/document/10160482}{[IEEExplore]}

\item J. Michalczyk, M. Scheiber, R. Jung, and S. Weiss, "Radar-Inertial Odometry for Closed-Loop Control of Resource-Constrained Aerial Platforms," 2023 IEEE International Symposium on Safety, Security, and Rescue Robotics (SSRR), 2023.  \href{https://ieeexplore.ieee.org/document/10499937}{[IEEExplore]}

\item J. Michalczyk, J. Quell, F. Steidle, M. G. M\"uller, and S. Weiss, "Tightly-Coupled Factor Graph Formulation For Radar-Inertial Odometry," 2024 IEEE/RSJ International Conference on Intelligent Robots and Systems (IROS), 2024.  \href{https://ieeexplore.ieee.org/document/10801945}{[IEEExplore]}

\item J. Michalczyk, S. Weiss, and J. Steinbrener, "Learning Point Correspondences In Radar 3D Point Clouds For Radar-Inertial Odometry," 2025 IEEE/RSJ International Conference on Intelligent Robots and Systems (IROS), 2025.  Accepted in June 2025 (not yet published). Preprint: \href{https://arxiv.org/abs/2506.18580}{[ArXiv]}

\end{itemize}

%% file: chapters/introduction.tex
\chapter{Introduction}\label{chap:intro}

\bigskip

\noindent Achieving accurate spatial awareness in \ac{gnss}-denied environments is a key task for a reliably operating autonomous \ac{uav} in many scenarios. Most notably, it is a prerequisite for controlling the motion of a \ac{uav} in space. This task is commonly approached by the fusion of measurements from a combination of sensors within a state estimation framework.

In the area of \ac{uav}s localization without\ac{gnss} support, a prominent place is held by \ac{vio} approaches employing a camera and an \ac{imu} \cite{geneva2020openvins, Leutenegger, MSCKF-Paper}. This prominence is partly due to the rich information provided by camera sensors which paved the way to the significant body of research devoted to using computer vision and image processing techniques in robot navigation in recent years. Nonetheless, camera sensors exhibit drawbacks rendering them vulnerable in certain environmental conditions such as extreme illumination (both very low and high), lack of features, or presence of air obscurants like fog or smoke. Other limiting factors for camera-based perception systems include sharp motions causing image blur, lack of metric measurements, and in some contexts, privacy issues related to grabbed images.   

LiDAR-based odometry systems for small-sized robots are mentioned in \cite{fast_lio2} and \cite{fastlio}. LiDAR uses time-of-flight of laser pulses from the visible spectrum to generate fine-grained 3D point clouds. Given the spectrum of used pulses, LiDAR is susceptible to same environmental factors as cameras. Using shorter wavelengths also reduces the range of measurements.  

Recently, fusing millimeter-wave \ac{fmcw} radar and \ac{imu} measurements has gained popularity in the \ac{uav}s autonomy research. A setup composed of these two sensors offers not only robustness against low-visibility environments, but also long-range perception thanks to the properties of electromagnetic waves used in radars \cite{kramer2020radar, kim_2_fog, morten}, and allows for \ac{imu} drift reduction, enabling accurate ego-motion estimation even in conditions challenging to camera or LiDAR-based setups.

Several kinds of \ac{fmcw} radars have been used in the context of autonomous navigation. The most common ones are scanning radars \cite{petillot, Burnett2021RadarOC, barnes2020under, park2020pharao} and \ac{soc} radars \cite{jan_icra, previous_iros, DCLoc, ContinuousRIO, kramer2020radar, 4Dradarslam, kim_2_fog, doer2020ekf, almalioglu2020milli}. Scanning radars are bulky and expensive mechanically rotating sensors. After performing a \unit[360]{\degree} scan they provide polar images of the environment with a high angular resolution. They typically do not provide the relative velocity (Doppler) information. \ac{soc} radars are usually much smaller in size and require less power. \ac{soc} radars output distance, relative radial velocity, azimuth (and sometimes elevation) angles of reflecting points in the environment in the form of a 4D pointcloud (3D position and Doppler velocity). Their accuracy and resolution vary broadly depending on the antenna array characteristics and on-chip processing algorithms. Using millimeter-wave technology in automotive industry \cite{hasch2012millimeter, rohling2001waveform, schneider2005automotive} brought about the miniaturization of the radar sensors and boosted their accuracy. This in turn, paved the way for using them onboard small \ac{uav}s and \ac{ugv}s for fusion with the \ac{imu} sensors. Interestingly, current generation \ac{soc} radars price tags vary significantly as a function of the sensor characteristics and range between several tens (consumer-grade) to several thousands (industry-grade) of euros.

It seems both challenging and appealing to try to build an accurate localization system for a \ac{uav} exploiting solely consumer-grade sensors. Hence,
in this dissertation we explore methods to leverage measurements from consumer-grade millimeter-wave \ac{fmcw} \ac{soc} radar and fuse them together with \ac{imu} sensor measurements in \ac{rio} state estimation frameworks with the goal of endowing small \ac{uav}s with accurate localization capabilities in unknown and potentially visually degraded environments.

\section{Motivation}\label{sec:motiv}
Since several years, a standard approach to estimating the odometry of a small \ac{uav} or \ac{ugv}, a setup consisting of a camera and an \ac{imu} has been used with impressive results. Nonetheless, there exist potential deployment conditions which incapacitate camera sensors rendering the state estimates unusable. These conditions are, however, not as challenging to radar as they are to a camera. When aiming at the environmental resilience, it is therefore purposeful to build robot localization systems based on radar. Given the research community interest, and the increasing prevalence of millimeter-wave \ac{fmcw} radars lending themselves to usage in robotics due to their cost, size and power consumption, there is also a need to research effective methods to use these sensors in robotic tasks. Recently, works which aim at closing this gap started to appear such as \cite{doer2020ekf, doer2020radar, 9843326, doer_manh}. Nonetheless, given the rich information about the environment which radar sensors perceive and the large knowledge in state estimation theory applied to other sensor setups such as \ac{vio} and \ac{lio}, there exists large room for improvement and still many promising research avenues to explore in the area of \ac{rio} research. In this dissertation we aim at expanding the knowledge on fusing together consumer-grade \ac{fmcw} radar and \ac{imu} data in order to estimate the navigation state of small-sized \ac{uav}s as well as pushing the boundaries of the state-of-the-art in terms of accuracy obtained by designed state estimators. 

\section{Contributions}\label{contrib}
The core contributions of this thesis are:

\begin{sloppypar}
\begin{itemize}
\item \emph{Tightly-coupled formulation to include both radar distance and velocity measurements for \ac{imu} integration correction in an \ac{ekf} framework, allowing accurate 3D velocity and 6DoF pose estimation.}
\item \emph{Application of stochastic cloning for inclusion of single and multiple past robot poses used for formulating an update equation on the accurately measured distances (rather than on the full 3D point positions polluted by the highly imprecise azimuth and elevation angular measurements) to multiple points in radar measurement trails allowing the extension of the initial single-state \ac{ekf} approach into a multi-state \ac{ekf} \ac{rio}.}
\item \emph{Improved 3D point matching across sparse, noisy radar scans in full 6DoF motion allowing ad-hoc point correspondence generation for point-distance based observation inclusion with only past radar poses (or a single radar pose) in the state in contrast to maintaining many 3D point vectors in the state.}
\item \emph{Efficient trail generation/handling for using past measurement-trails and extension of the radar feature matching to work with the trails of features as well as extension of the distance measurement update to work with the measurement trails in the \ac{ekf}.}
\item \emph{Introduction of the online estimation of the extrinsic calibration parameters and thus simplifying the use of the \ac{ekf} framework as well as improving its accuracy and consistency.}
\item \emph{Tightly-coupled formulation of a sliding window factor graph-based \ac{rio} making maximal use of all the noisy measurements from a light-weight, inexpensive \ac{soc} \ac{fmcw} radar sensor to correct the \ac{imu} drift.}
\item \emph{Implementation of persistent radar landmarks for increased estimation accuracy in both \ac{ekf} and factor graph method as well as their inclusion in the update equation of the \ac{ekf} and as a factor in the factor graph.}
\item \emph{\textit{One-to-one} comparison of the factor graph \ac{rio} with the multi-state \ac{ekf}-based \ac{rio}.}
\item \emph{Implementation of both \ac{rio} frameworks capable of executing in real-time on a portable resource-constrained credit-card sized onboard computer allowing the evaluation of the implemented framework in closed-loop control flights.}
\item \emph{Deep learning framework for predicting robust correspondences in sparse and noisy \ac{fmcw} \ac{soc} radar 3D point clouds, in which the formulation of the learning problem is posed as multi-label classification which allows training (using self-supervised method not requiring hand annotated ground-truth data) on the sparse and noisy 3D point clouds yet results in unambiguous matches.}
\end{itemize}
\end{sloppypar}

The source code of the implementation of both the \ac{ekf} and factor graph methods is made open-source and available here: \texttt{\url{https://github.com/aau-cns/aaucns_rio}}. 

\section{Outline}\label{outline}
The remaining chapters of this thesis are organized as follows. In \textbf{\cref{chap:related_work} and~\cref{chap:fund_radar}} we introduce the related work and the fundamentals of radar sensing. In particular,\textbf{~\cref{chap:related_work}} introduces the most recent advances in the area of radar-based state estimation discussing methods based on the \ac{soc} as well as spinning \ac{fmcw} radars, methods making use of the Doppler velocity only as well as methods utilizing also 3D point matches. We mention both classical and learning-based \ac{rio} methods. In some cases, mostly the case of imaging (mechanically spinning) radars, we introduce methods which rely solely on radar and are thus \ac{ro} methods. In the \textbf{\cref{chap:fund_radar}} we outline the necessary principles related to the radar sensing. We also show the signal processing pipeline in a \ac{fmcw} \ac{soc} radar leading to the measurement of 3D points and Doppler velocities. In \textbf{~\cref{chap:ekf_rio}} we present our progressive research within the \ac{ekf}-based \ac{rio} which led to the final formulation of the multi-state \ac{rio} with self-calibration capabilities. We show how this final implementation is used in closed-loop flights and how it successfully deals with deployment in the dense artificial fog where a state-of-the-art \ac{vio} fails. In\textbf{~\cref{chap:fg_rio}} we present an implementation of \ac{rio} based on the optimization of factor graphs. We compare and contrast this state estimation formalism with the presented earlier \ac{ekf}-based one. In~\textbf{\cref{chap:dl_matching}} we propose a learning framework based on the transformer architecture to learn point correspondences between pairs of sparse and noisy 3D point clouds measured by \ac{soc} \ac{fmcw} radar sensor. At the end, in the \textbf{\cref{chap:concl_chp}}, we offer closing discussion and conclusions.

\section{Notation}\label{notation}
In this section we introduce the notation used throughout this dissertation.

Vector spaces are denoted with blackboard letters such as ${\mathbb{X}}$. Vectors expressing elements of some $n$-dimensional vector space $\mathbb{X} \subset \R^n$ are denoted by lowercase bold letters such as $\vx$. Matrices representing linear operators on vector spaces are denoted by capital bold letters such as $\vX$.

Vectors describing quantities $\vx$ of a target object \frameofref{B}, observed from a \emph{frame of reference} \frameofref{O} and expressed in a \emph{frame of reference} \frameofref{A}, are denoted by $\reference{\vx}{A}{B}{O}$. Note that when the observer frame and the coordinate frame coincide, the left subscript is dropped ${\reference{\vx}{A}{B}{A} = \reference{\vx}{A}{B}{}}$.

Thus, to denote a the translation between a frame of reference \frameofref{A} and a frame of reference \frameofref{B} expressed in the frame of reference \frameofref{A} we write $\reference{\vp}{A}{B}{}$. To denote the relative velocity of a moving frame of reference \frameofref{B} expressed in a frame of reference \frameofref{A} we write a vector $\reference{\vv}{A}{B}{}$. 

Rotation matrices encoding the orientation of a frame of reference \frameofref{B} with respect to a reference \frameofref{A} are denoted by $\Rot{A}{B}$. In particular, $\Rot{A}{B}$ is the matrix whose columns are the orthonormal vectors defining the axes of \frameofref{B} expressed in \frameofref{A}:
\begin{equation*}
    \Rot{A}{B} = 
    \begin{bmatrix}
        \reference{\ve}{A}{B_x}{B} & \reference{\ve}{A}{B_y}{B} & \reference{\ve}{A}{B_z}{B}
    \end{bmatrix} \in \R^{3 \times 3} .
\end{equation*}

Pose matrices encode simultaneously the orientation and the translation of a frame of reference \frameofref{B} with respect to a reference \frameofref{A}, and are denoted by $\reference{\vT}{A}{B}{}$, as in
\begin{equation*}
    \reference{\vT}{A}{B}{} = 
    \begin{bmatrix}
        \Rot{A}{B} & \reference{\vp}{A}{B}{}\\
        \Vector{}{0}{1 \times 3} & 1
    \end{bmatrix} \in \R^{4 \times 4} .
\end{equation*}

The transformation of a vector $\reference{\vp}{C}{P_1}{C}$ pointing from the origin of the reference frame \frameofref{C} to a point $P_1$, expressed in \frameofref{C}, can be transformed into the frame \frameofref{A} by $\begin{bmatrix} \reference{\vp}{A}{P_1}{A} \\ 1 \end{bmatrix} = \reference{\vT}{A}{C}{} \begin{bmatrix} \reference{\vp}{C}{P_1}{C} \\ 1 \end{bmatrix}$ (read as $\reference{\vx}{from}{to}{~~~in}$). A normally distributed multivariate variable is defined as $\vX_{i} \sim \cN(\vx_{i}, \bSigma_{ii})$, with a mean $\vx_i$ and covariance (uncertainty) $\bSigma_{ii}$, which is called the belief of $i$. We often express rotations as unit quaternion $\bar{\vq} \in \mathrm{SO}(3)$ with $ \norm{\bar{\vq}} = 1$ allowing a direct mapping between rotation matrices and unit Hamiltonian quaternions by $\reference{\vR}{A}{B}{} = \vR \left\{ \reference{\bar{\vq}}{A}{B}{} \right\} \in \mathrm{SO}^{3}$ and $\reference{\bar{\vq}}{A}{B}{} = \bar{\vq} \left\{ \reference{\vR}{A}{B}{}  \right\}$~\cite{sommer2018and}. $\vI$ is the identity matrix.
The \textit{a priori} and \textit{a posteriori} of a belief are indicated by a $\{\bullet\}^{(-)}$ and $\{\bullet\}^{(+)}$, respectively. $\{\bullet\}^{\#}$ specifies measured (perturbed) quantities. For vectors and block matrices, semicolons and colons improve the readability such that $[\vA; \vB] \equiv \begin{bmatrix} \vA \\ \vB \end{bmatrix}$ and $[\vA, \vB] \equiv \begin{bmatrix} \vA & \vB \end{bmatrix}$. We use capital calligraphic letters to denote sets, for example, $\cL$ is a set of landmarks, while $\cR$ is a set of extrinsic calibration state variables corresponding to the position and orientation of the radar sensor with respect to the IMU. Sometimes we use small calligraphic letters to denote elements of sets, for example, $\ell$ is a single landmark.

%% file: chapters/related_work.tex
\chapter[Related Work][Related Work]{Related Work}\label{chap:related_work}

\bigskip

\noindent This chapter presents the relevant recent scientific work in the area of radar-based state estimation. Given the diversity in measurements quality and density among radar sensors, some approaches presented in this chapter will rely only on radar for the task of state estimation. This is mostly the case with mechanically spinning radars, but also with some higher-end \ac{soc} radars. In these approaches, the \ac{imu} sensor is omitted rendering the respective methods radar-only, thus \ac{ro}. Though mechanically spinning radars have the same underlying principle of sensing as their \ac{soc} counterparts, given their highly superior perception quality, they can be considered a different class of sensors. Spinning radars are sometimes called \textit{imaging} radars, due to their finer spatial granularity. Although in this thesis the focus is set on using consumer-grade, inexpensive and lightweight \ac{soc} radar variants, it is still meaningful to consider relevant works with spinning radars, as some concepts from the state estimation using spinning radars can and are still be exploited with \ac{soc} radar sensors.

\begin{figure}[h!tbp]
  \centering
  \includegraphics[width=1.\columnwidth]{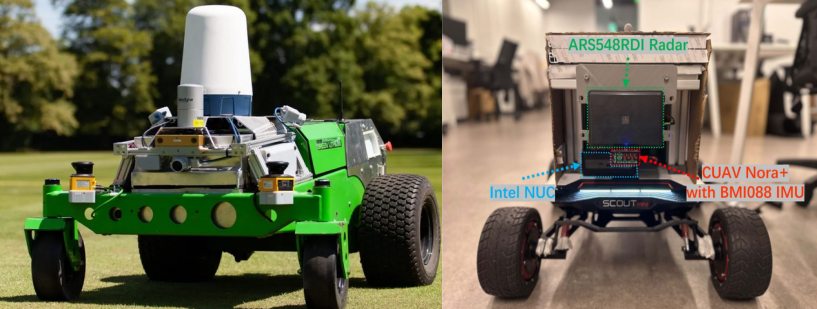}
  \caption[Spinning and SOC radar sensor examples]{Examples of a spinning (left) and \ac{soc} (right) radar sensors mentioned in the following sections. The depicted \ac{soc} radar is a high-end automotive radar (Continental ARS548) used in \cite{poin_uncert, lessismore}. For an example of consumer-grade radar sensors (used in this work) see~\cref{fig:platform} and~\cref{fig:ekf_rio_single_platform}. (Spinning radar example is taken from \href{URL}{https://navtechradar.com/).}}
  \label{fig:radars}
\end{figure}

\section{Radar-Based State Estimation Using Spinning Radars}\label{spin_radars}
Scanning radars rotate around their vertical axis while continuously emitting frequency-modulated radio-frequency waves. Along each azimuth angle (determined by the number of discrete azimuth bins) the sensor measures power returns at discrete range bins. These sensors possess a remarkable feature of being able to sense detections up to around \unit[500]{m} while spanning \unit[360]{\degree} in azimuth. Unlike \ac{soc} version, the spinning radars provide neither the elevation nor (typically) the Doppler information resulting in the planar perception. Given low scanning frequencies (\unit[4]{Hz} for NavTech CIR 304-H), the scans exhibit distortions due to the motion of the platform on which they are mounted \cite{radar_survey}. Some localization methods explicitly compensate for this effect as in \cite{burnett_compensate}. A number of radar-based localization and \ac{slam} methods have been presented in the scientific community in the past few years exploiting mechanically rotating \ac{fmcw} radars. These significant contributions have been made mostly in the autonomous driving and mining domains given the size, cost and power requirements of this sensor.

One of the pioneering works which sparked resurgence in research into \ac{slam} was \cite{cen2018precise} where the authors use a rotating radar to implement \ac{ro} estimating the ego-motion of a vehicle, making use of landmarks extraction based on estimating the signal noise statistics and subsequent matching of these landmarks exploiting local geometrical relationships between them within the scan. The estimated signal noise statistics are used to scale the power spectrum which is subsequently searched for peaks. Their matching algorithm does not require any prior guess of the transform between the scans and relies on exploiting the unary descriptors as well as the mutual geometrical relationships between the landmarks. Once the proposed set of matches is returned based on the unary descriptors, a positive-semi definite matrix is built based on the pairwise similarity of the proposed matches. In the next steps, the normalized eigenvector of the maximum eigenvalue is found and used within greedy search to find iteratively the best set of matches which serves to estimate the relative motion using the \ac{svd}. Once the relative motion is recovered, the scans can be re-aligned and a second round of association and motion estimation takes place. In the second round, the only difference is the kind of unary descriptors used to describe landmarks.

Authors in \cite{petillot} aim at achieving a \ac{slam} system able to function under extreme weather conditions. To this end, they propose a \ac{ro} system using a scanning radar incorporating pose tracking, local mapping, loop closure detection and pose graph optimization. The authors present a novel point cloud generation algorithm robust to noise in radar images which exploits modeling of the power along each azimuth by a Gaussian distribution. Pose tracking is achieved by point matching between the current frames and keyframes. Specifically, SURF features are extracted in both frames, which are subsequently matched using motion prior and pairwise consistency constraint (as in \cite{cen2018precise}). Matched points are used to calculate the relative transformation using \ac{svd}. Once a new keyframe is created, the mapping thread optimizes jointly the poses of keyframes and the belonging keypoints. Once a loop is detected, the pose graph optimization is performed on all keyframes and all keypoints are updated to form a global map. The approach is shown to outperform the results presented in \cite{cen2018precise}.

In \cite{barnes2020under}, a \ac{cnn}-based \ac{ro} is introduced capable of predicting robust features in radar scans, which are then used to estimate the optimal relative transformation. In particular, the authors design a deep learning architecture self-supervised with the odometry error. The architecture achieves self-supervision by embedding the differentiable \ac{svd} for point-based motion estimation. The output of the learned model are the keypoints, descriptors and scores. Owing to the learning problem formulation the learned keypoints are well-tailored to the localization task and assumptions-free as opposed to the hand-designed keypoints. The authors claim improvement over the method in \cite{cen2018precise} by \unit[45]{\%} in translation and \unit[29]{\%} in rotation respectively, and order of magnitude faster runtime. This result suggests that learned features enables surpassing the performances achieved with hand-designed features.

In \cite{park2020pharao} a direct \ac{ro} approach is shown using the entire radar scan as opposed to only extracted features. Namely, the application of Fourier-Mellin transform to Cartesian and log-polar scans is explored to obtain translation and rotation of the robot, respectively. Even though direct methods require heavy computations as they operate on full scans, the presented approach achieves real-time performance at \unit[10]{Hz} thanks to coarse-to-fine pre-processing of the radar images. Odometry is computed using pose graph optimization.

In \cite{9808131} the authors incorporate the vehicle dynamics of the platform in order to refine the landmarks association from \cite{cen2018precise}. Making assumptions on the platform motion allows for rejecting associations which do not conform to the constant-curvature motion constraints imposed by the assumed model. Found correspondences are used to calculate the relative motion between the scans using \ac{svd}. \cite{bar} presents an approach in which state estimation and features learning are combined within an \ac{ro} framework. As opposed to \cite{barnes2020under}, the features are learned without any supervision only using the onboard sensor data and the estimator is not required to be differentiable.

Authors in \cite{what_could_go_wrong} address the problem of uncertainty-driven fault detection in \ac{ro}. This is an important problem in automotive platforms where discontinuities in observed features can occur when the platform does not move on ideally planar surface, the distinct features become scarce or when motion of objects in the scene causes the features to disappear. The authors aim at providing an introspection capability to the \ac{ro} from \cite{cen2018precise} where the correspondence-finding algorithm delivers also its confidence in the matched features which is subsequently used to judge whether to accept it for updating the pose or not. In particular, the authors train a classifier to mark the \ac{ro} estimate as either good or bad based on the ground-truth labels from \ac{gnss} and \ac{imu} sensor streams and the input radar data. It is noted that the principal eigenvector components, which are used for retrieving matches within radar scans, are very distinct for the cases where the pose estimates are marked good and bad. This observation allows for training an \ac{svm} classifier predicting whether or not the association between two scans can be used for pose estimation. In \cite{burnett_lio_rio} the authors show a \unit[45]{\%} improvement in their continuous-time \ac{ro} framework using gaussian process motion prior by additionally incorporating an \ac{imu} sensor. 

\section{Radar-Inertial State Estimation Using SoC Radars}\label{soc_radars}
One of the expected advantages of small-sized \ac{uav}s and \ac{ugv}s is their ability to be deployed in complex and hostile environments such as disaster zones, areas with adverse weather conditions, or hazardous industrial zones to perform autonomous missions such as reconnaissance, inspection, search and rescue, and others. In order to operate autonomously in such environments, the localization system of the platform must make use of sensors robust towards phenomena such as fog, smoke or extreme illumination. Since recently, thanks to their physical properties, their miniaturization, and general advances in their technology, radar sensors are being increasingly investigated for their use in navigation of various types of autonomous robots requiring resilience towards unfavorable environment factors. Another key requirement of an onboard localization system on an autonomous mobile robot is the need to execute within the tight limits imposed by the real-time control loop and often on a resource-constrained computing platform.

While information-rich, the bulky and expensive scanning radars are inadequate for use on small-sized \ac{uav}s or \ac{ugv}s mostly due to their size and power consumption. Indeed, we focus more on the application of low-cost and small single-chip \ac{fmcw} radars on \ac{uav}, where the payload is of crucial relevance. \ac{soc} radars integrate antennas and all necessary signal processing hardware within a single printed circuit board. 4D \ac{soc} automotive radar sensors which are the ones mostly used in the robotics research can broadly be divided into higher-end sensors which are based on the cascaded design and which have much better sensing parameters in terms of point cloud resolution and maximum attainable range (Altos V2, Continental ARS548), and lower-end sensors (Texas Instruments AWR1843). The former are an order of magnitude costlier than the latter ones, while also being heavier and more power-greedy. Within this thesis we focus on the lower-end radar sensors in order to minimize the weight, size and power consumption of the designed localization systems so that they can be used even on small-size \ac{uav}s.

Pioneering works in the area of \ac{rio} for small autonomous robots using lower-end mmWave \ac{fmcw} \ac{soc} radar (Texas Instruments AWR1843) are introduced in \cite{doer2020ekf, doer2020radar, 9470842, 9843326} where the authors present a real-time \ac{rio} method in which the platform velocity estimated from a single radar scan is fused in a loosely-coupled manner with \ac{imu} readings in an \ac{ekf} framework. In \cite{doer2020radar} the authors extend the method presented in \cite{doer2020ekf} with online calibration of radar extrinsic parameters which boosts the accuracy and simplifies the usage of the approach. Nonetheless, the barometer sensor is additionally used in order to mitigate the vertical drift of the platform, hence the approach is not a pure \ac{rio} using a radar and an \ac{imu} only. In \cite{9470842} the same \ac{ekf} framework is further augmented with the yaw aiding using Manhattan assumptions on the environment thanks to which the method attains high accuracy with minimal run-time footprint. Higher accuracy is achieved at the expense on imposing constraints on the environment which is a limiting factor. In \cite{9843326} still the same method is equipped with \ac{gnss} sensor providing a versatile and precise outdoor localization solution, nonetheless at the expense of introducing a dependency on \ac{gnss} reception. In all of their approaches the authors of \cite{doer2020ekf, doer2020radar, 9470842, 9843326} decide not to use any form of 3D radar point cloud matching in their \ac{ekf}. They incorporate only Doppler radar measurements using \ac{ransac} based least-squares estimator. Radar processing and transmission delays are accounted for, in order to make the approach resilient to dynamic motion and usable for online-control, by hardware-triggering radar measurements and stochastically cloning the state corresponding to the triggered measurement which is then used for the ego-motion estimation. The approach to \ac{rio} presented in this thesis, in contrast, additionally makes use of the point correspondences between radar scans, which allows to reduce the cumulative drift with respect to Doppler-only methods as also mentioned in \cite{rio_huang}. Also, tracking point correspondences in time as done in this thesis enables the potential use of the \textit{persistent features} which can be further used to increase the accuracy of the estimators as explained in~\cref{chap:ekf_rio}.

Another early approach exploiting the Texas Instruments AWR1843 radar sensor is shown in \cite{almalioglu2020milli}. The authors demonstrate a \ac{rio} approach on a small indoor \ac{ugv}, interestingly, only making use of the radar 3D point cloud matching. Their system comprise a point association module and the following scan registration using \ac{ndt}. Relative displacements are subsequently fused with the \ac{imu} measurements in an \ac{ukf}. The authors test their approach in a small-scale, flat, indoor environment with \ac{ugv} performing a low-dynamics trajectory, it is therefore an open question how the proposed system would perform in dynamic flights of a \ac{uav} as is the focus of this thesis.

In \cite{kramer2020radar} and \cite{kramer_fog} an optimization-based approach is shown where a sliding window of past radar and \ac{imu} measurements is used to estimate the \ac{uav} velocity. In the constructed factor graph only Doppler factors are used from the radar output. The Doppler residuals are weighted by the normalized point intensity. The \ac{imu} residuals are formulated between two radar measurements only considering orientation, velocity and biases. The approach in \cite{kramer2020radar} attains comparable performance with the vision-based system and even outperforms it in visually degraded settings. Additionally, in \cite{kramer_fog} the authors add a comparison of their method with a lidar-based approach and propose a method to reduce the noise in the radar point clouds based on Deep Learning. The authors in both \cite{kramer2020radar} and \cite{kramer_fog} concentrate on estimating the ego-velocity instead of the full 6D pose. Also, both approaches have not yet been demonstrated in closed-loop flights. The factor graph approach presented in this thesis additionally adds tightly-coupled factors on point correspondences, Doppler velocities and persistent features.

In \cite{lessismore} the authors design a \ac{rio} system based on a higher-end automotive Continental ARS548 radar. In their implementation both Doppler velocity and point matches are used in a sliding-window nonlinear optimization backend estimator. Noise in the radar scans is addressed by measurements pre-processing. The latter involves removing points outside of the \ac{fov}, radius filter eliminating points excessively far from other points and \ac{imu}-aided outlier removal based on velocity. Apart from making use of the classic residual on Doppler velocity similar to other approaches in which a projection of the ego-velocity on the unit vector pointing towards a detection is used, the presented approach uses point matching guided by the \ac{rcs} value of detections, which assumes that big changes in \ac{rcs} within subsequent scans are inconsequential. Same authors in \cite{poin_uncert} augment their system from \cite{lessismore} by modeling the radar points uncertainty in polar coordinates and incorporating it in both the frontend and backend of their state estimation framework. This important idea has also been explored in \cite{const_vel_prior} where it has been noted that the uncertainty of detections is greatly influenced by the physics of radar sensing and the modeling in the Cartesian space tends to produce underconfident measurements at close ranges and overconfident measurements at far ranges. Proper physical modeling of the radar sensing uncertainty in \cite{poin_uncert} allows for fine-grained weighting of radar residuals in the backend, but also for guiding the 3D points association. Real-world experiments show the benefits of including their confidence model of radar measurements on the accuracy in the state estimation. As opposed to the work in this thesis, the above work employs a high-end automotive \ac{soc} radar and the tests are carried out using a small \ac{ugv}.

In \cite{9636014} the authors present an optimization-based continuous-time \ac{rio} method which is particularly well suited for multi-radar setups as the continuous representation of the vehicle trajectory allows sampling it at any given time, thus permitting efficient asynchronous fusion of multiple radars measurements with the \ac{imu}. The method in \cite{9636014} is demonstrated on an automotive platform with high-performance cascaded radars, an order of magnitude costlier, and still of greater size than the single-chip radar used in our work. 

In the domain of \ac{slam} for automotive systems using \ac{soc} radars, the authors in \cite{schuster_landmark_slam} exploit a setup of two radars giving planar coverage of approximately \unit[360]{\degree} and are mounted behind the vehicle bumpers. Both radars are used together with an \ac{imu} and wheel odometry to build residuals within an optimization problem. Radar residuals are built matching detected landmarks to the stored map. Matching is done by brut-force association of the Hamming distances of landmarks descriptors. \ac{ransac} is used on the matched landmarks to reject outliers. The approach  allows for consistent localization error below \unit[1]{m} over large-scale outdoor (parking lot) trajectories.

One of the best-performing \ac{rio} systems using \ac{soc} radars has been presented in \cite{iriom}. As opposed to the methods in this thesis, the presented system is in fact a full \ac{slam} relying on radar and an \ac{imu} sensors. Beyond the Doppler velocity, the presented approach uses distribution-to-multi-distribution distance concept to obtain matches between radar scans and submap. Both sources of information are fused within an iterated \ac{ekf} to obtain the full 3D pose of the platform. 
The system also contains a loop closure component, effectively reducing the accumulated drift. Rather than using \ac{ransac} as in majority of the aforementioned methods, the iterative reweighted least-squares are used to find the inlier set and consequently calculate the radar velocity. Scan-to-submap matching adopted by the authors ensures higher stability of matching by maintaining a local submap to which new scans are being registered. Sparsity of radar scans is countered within the update step partially by the distribution-to-distribution distance computation in the residual in which the local geometry of the matched point is considered. This is dictated by an assumption that due to the sparsity, of measurements, the points in the current scan will match exactly to the ones in the submap. The method is demonstrated on a small-sized \ac{ugv} with a high-end automotive ARS548 radar mounted, and exhibits performances close to those of Fast-LIO \ac{slam} \cite{fast_lio2}. An important observation is made in their work, namely that using the direct approach alone, that is, the Doppler velocity without scan matching is rendering radar-based estimators susceptible to vertical drift, thus evidencing the necessity of also incorporating point correspondences in the state estimation. 

In \cite{efear} another \ac{rio} method is presented exploiting both Doppler velocity as well as point matching as delivered by a high-end automotive \ac{soc} radar. The implemented approach consists in ego-velocity computation, feature extraction and scan-to-submap matching. For ego-velocity estimation gaussian filtering is applied followed DBSCAN clustering of both static and dynamic objects. Clustered static objects are then used for ego-velocity estimation using least-squares. Feature extraction is done by dividing the points into voxels whose representative points are being calculated along with covariances. Calculating the \ac{pca} for all voxels further permits pruning voxels with features having lesser stability. For matching, a sliding window of keyframes is constructed forming a submap, which after aligning the current scan using the estimated ego-velocity, is used to register current scan based on voxels. The method is shown to perform on-par with vision and lidar-based methods.

In \cite{drio} the authors address the problem of state estimation of an \ac{ugv} using \ac{rio} in highly dynamic scenes. Their method relies on the observation that the ground points are ever-present in the common \ac{ugv} scenarios and can be reliably used for odometry estimation. Due to the instability and sparsity of radar scans the ground detection methods from lidar-based approaches will not work from radar. The authors thus propose an algorithm which jointly estimates radar velocity and detects ground points. Estimated velocity together with the angular velocity from \ac{imu} enables planar displacement calculation. On a self-collected dataset the presented method outperforms the methods from \cite{iriom} (without mapping) and \cite{doer2020ekf}.

In \cite{4d_slam_preintegr} the authors present a radar-based \ac{slam} based on pose graph optimization. The primary contribution of the authors is the design of a ego-velocity pre-integration factor and the integration thereof within their pose graph optimization framework. Besides the novel velocity pre-integration factor the authors use factors calculated based on \ac{ndt}-based scan registration. Loop closure factor is also included to curb the accumulation of the drift. Interestingly, the authors do not use \ac{imu} in their implementation, relying only an a high-end ZF FRGen21 automotive 4D Radar.

As can be seen from the mentioned works, approaches to state estimation using radars usually use solely Doppler velocity or a combination of it with some form of scan matching. Interestingly, authors in \cite{kubel} argue that when industry-grade sensors are used and only odometry is needed, scan matching is no longer needed and the Doppler measurements suffice. This statement is disputed in \cite{lessismore, poin_uncert, iriom, rio_huang} where the authors claim the vital importance of the point matching in their \ac{rio} system despite using automotive-grade radar. At any rate, still in many systems the price, weight and power consumption requirements mandate the use of consumer-grade radar chips as the one we use in the present paper where using point matches boosts estimation accuracy. What can also be seen is that depending on the system constraints, various \ac{soc} radar types can be used with broadly varying measurement accuracy and point cloud density. In this thesis we are interested in implementing \ac{rio} with only low-end consumer-grade radar which imposes more challenges on the state estimation as compared to systems using high-end \ac{soc} radars, yet given the limited weight, size and power consumption permits its use in small-size \ac{uav}s.

%% file: chapters/radar_fundamentals.tex
\chapter[The Fundamentals Of Millimeter-Wave FMCW Radar Sensing][The Fundamentals Of Millimeter-Wave FMCW Radar Sensing]{The Fundamentals Of Millimeter-Wave FMCW Radar Sensing}\label{chap:fund_radar}

\bigskip

\noindent This chapter aims at clarifying the sensing principles of millimeter-wave \ac{fmcw} \ac{soc} radar sensors and at explaining the signal processing pipeline which leads from the raw radar data cube (see~\cref{fig:fund_cube}) to 3D point clouds and Doppler velocities. We begin by outlining the sensing mechanisms behind the millimeter-wave \ac{fmcw} radar in~\cref{sec:fund_sens}, how the radar data cube is formed and filtered against the noise to obtain radar targets. We also explain some radar terminology needed for understanding the following chapters. In~\cref{sec:fund_proc}, we discuss the signal processing needed to extract higher level measurements such as 3D points and Doppler velocities from low-level radar data. Although the mechanically spinning radar sensors also use the \ac{fmcw} technology, we limit the content of this chapter to the \ac{soc} radar sensors since this is the sensor used within this thesis.

\section{Millimeter-Wave FMCW radar sensing}\label{sec:fund_sens}
\begin{figure}[thpb]
  \centering
  \includegraphics[width=1.0\columnwidth]{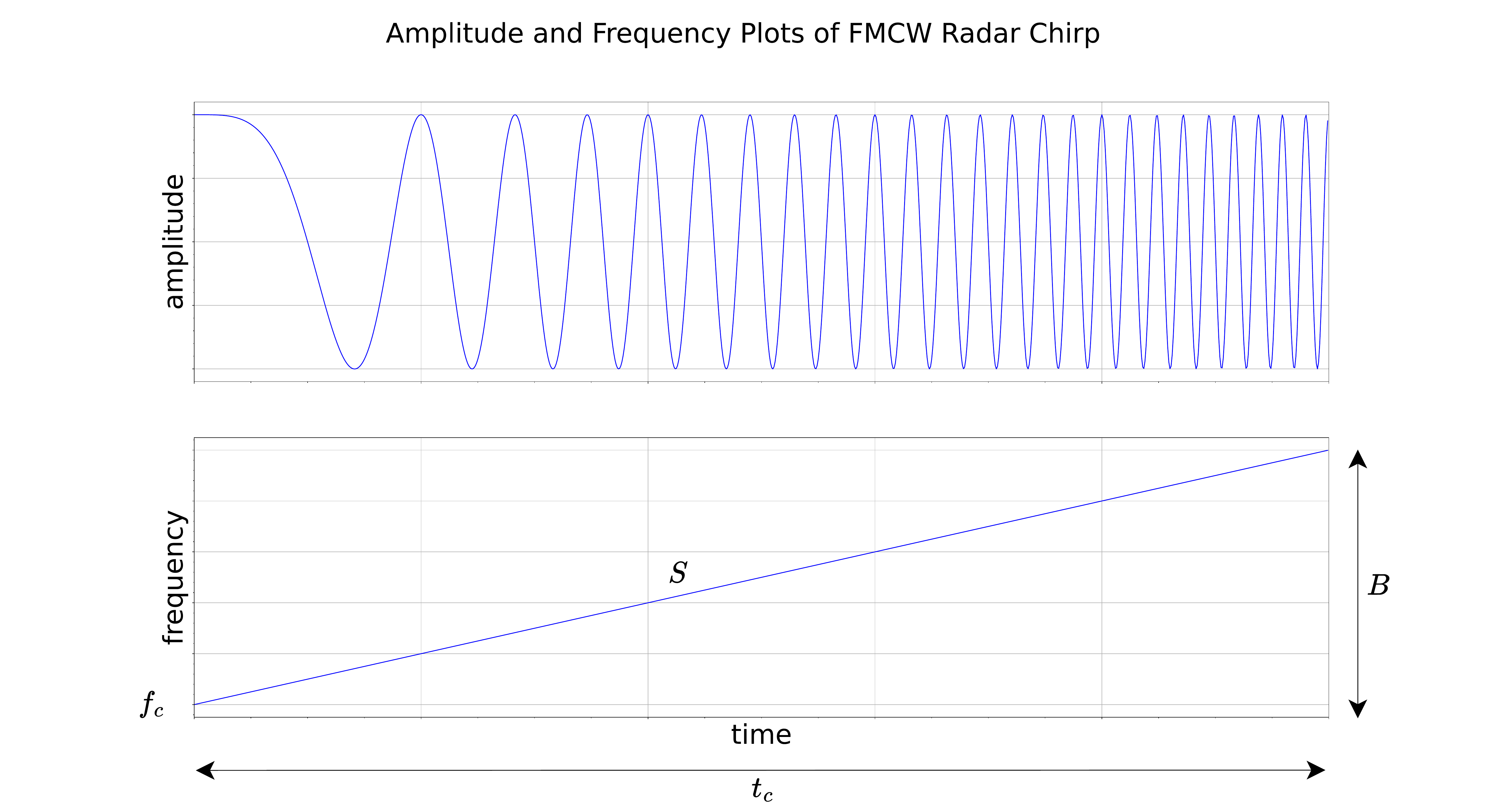}
  \caption[Single FMCW linear chirp]{Single chirp generated by \ac{fmcw} radar.}
  \label{fig:fund_chirp}
\end{figure}

\begin{figure}[thpb]
  \centering
  \includegraphics[width=1.0\columnwidth]{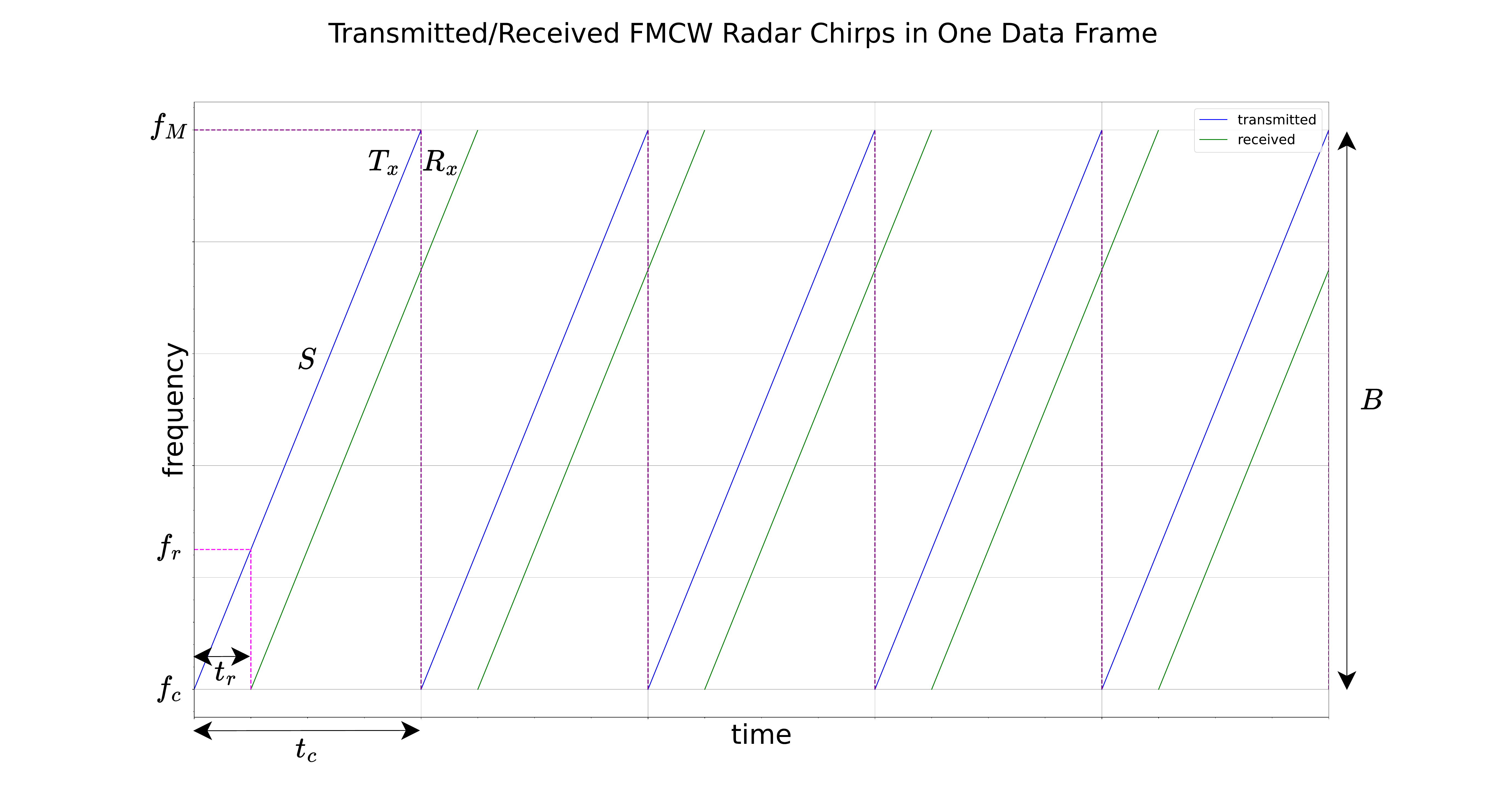}
  \caption[Radar frame consisting of several FMCW linear chirps]{Within a single cycle multiple chirps are generated and sent out. Difference between $f_r$ and $f_c$ allows determining the $t_r$ which is the time needed to travel back and forth between the transmitter and the reflected object. Using the phase difference between the chirps we can determine the Doppler velocity of the reflected objects.}
  \label{fig:fund_chirps}
\end{figure}

\begin{figure}[thpb]
  \centering
  \includegraphics[width=1.0\columnwidth]{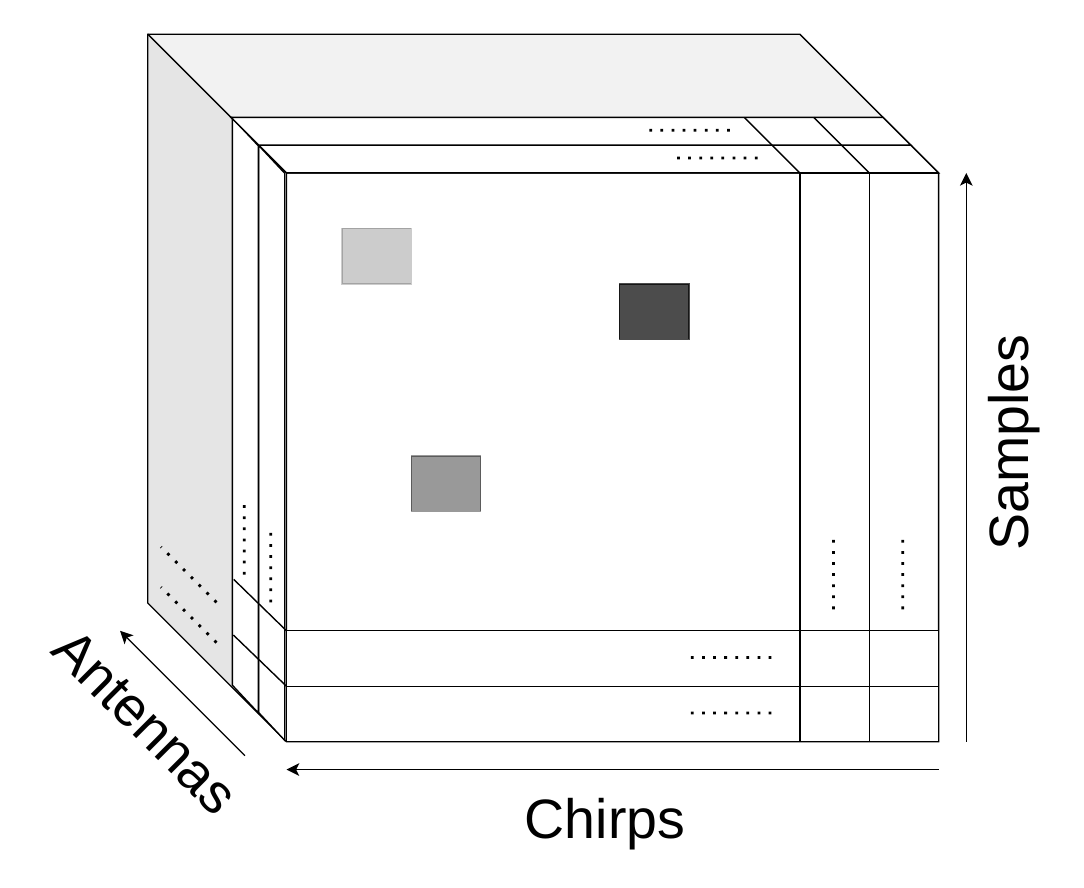}
  \caption[Radar cube consisting of measurements across multiple antennas]{Measuring the reflections from multiple chirps at multiple receivers forms the radar data cube. The entries are complex samples of mixed chirps.}
  \label{fig:fund_cube}
\end{figure}

Millimeter-wave \ac{fmcw} radar sensors transmit and receive electromagnetic waves of length between \unit[1-10]{mm} and mostly within the frequency spectrum of \unit[76-81]{GHz}.
They operate at increased frequency spectrum with respect to the previous generation (increase from \unit[25]{GHz} to above \unit[60]{GHz}). This increase offers better performance of the sensors in terms of accuracy of measurements and also allows the chip size reduction at the expense of causing more scattering effects of the reflected waves \cite{mmwave_perc_survey}.
\ac{fmcw} means that the radar transmitter sends out a signal which is modulated in frequency by another signal. The modulating signal is linear in most of the robotic and automotive applications. A single linearly modulated electromagnetic pulse is called a \emph{chirp}. A chirp gets reflected off objects in the environment (sometimes called \emph{scatterers}) and comes back to the receiver. Chirps are characterized by the starting frequency $f_c$, duration time $t_c$, bandwidth $B$ and slope $S$ as shown in the~\cref{fig:fund_chirp}. The reflected and emitted chirp signals are mixed to obtain the so-called \emph{intermediate frequency} (\ac{if}). While with \ac{if} originating from a single chirp it is possible to determine distances of reflecting objects, a radar typically sends multiple chirps during one cycle as shown in the~\cref{fig:fund_chirps}. Multiple chirps are needed to measure the radial velocity of the reflecting objects using the Doppler shift. Moreover, it is also possible to determine the bearing and azimuth angles of the reflecting objects by properly laying out multiple antennas on the chip. Thus, raw data sensed by the \ac{fmcw} radar can be thought of as the \emph{radar data cube} as shown in~\cref{fig:fund_cube}. To form a point cloud from a radar data cube, a radar detector needs to be applied in order to find the reflecting targets within the range-Doppler maps among the background noise. Range-Doppler maps are obtained from the data cube using methods described in the~\cref{sec:fund_proc}. Currently \ac{cfar} (and its flavors~\cite{cfar_fast, cfar_fast_2, cfar_fast_3}) is the most commonly used detector~\cite{richards2005fundamentals}. As input to \ac{cfar} the range-Doppler map is used which is the result of averaging all range-Doppler map values across the antennas dimension see~\cref{fig:fund_cube}).

\begin{figure}[thpb]
  \centering
  \includegraphics[width=1.0\columnwidth]{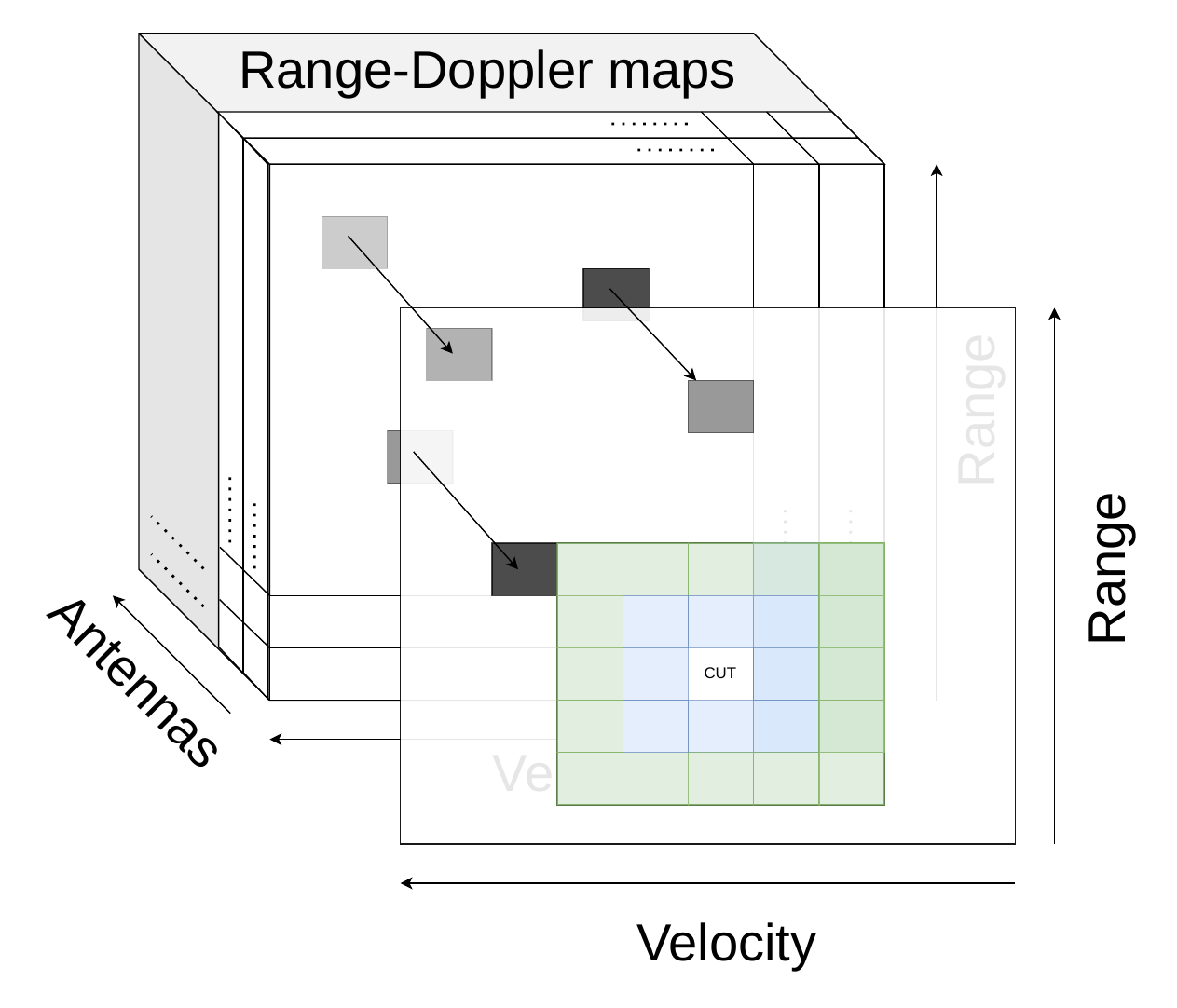}
  \caption[CFAR algorithm]{Target detection using \ac{cfar} proceeds by shifting a 2D window through the range-Doppler map built from averaged values from range-Doppler maps from all the antennas. Values in the green entries within the window are averaged and used to for the comparison threshold. Values in the blue entries are so called \emph{guard cells} and do not participate in the threshold calculation. CUT stands for Cell-Under-Test.}
  \label{fig:fund_cfar}
\end{figure}

The objective of \ac{cfar} is to identify the valid radar targets from the background noise using an automatically adjusting detection threshold. We briefly explain the basic idea behind \ac{cfar} using the CA-\ac{cfar} (Cell-Averaging \ac{cfar}) variant. Using radar data cube shown in the~\cref{fig:fund_cube} we produce a range-Doppler map by applying the \ac{fft} operations as explained in the~\cref{sec:fund_proc}. Next, for every cell we average the cell values to produce a single range-Doppler map. Range-Doppler map entries are complex numbers, yet for \ac{cfar} we compute a square of the magnitude for each cell. Using a sliding window as shown in the~\cref{fig:fund_cfar}, we estimate the background noise value by first computing the average of cells in the window excluding the Cell-Under-Test as well as the guard cells. Next, we compute the threshold as $T = \alpha \tilde{P}$, where $\tilde{P}$ is the average value computed from the entries in the window, and $\alpha$ is the threshold scaling factor calculated as $\alpha = N_r(P_{fa}^{\frac{-1}{N_r}} - 1)$, where $N_r$ is the number of cells used for averaging and $P_{fa}$ is the desired false alarm probability. The decision whether the Cell-Under-Test is a target is made by comparing its value to the calculated threshold $T$.

It is important to note that there exists evidence that \ac{cfar} is not very suitable for prevalent cluttered environments where autonomous systems operate nowadays and there is an active ongoing research into alternative methods for generating radar targets~\cite{novel_radar_pc_generation, cfar}.

By consecutively applying the \ac{fft} across each of the three dimensions of the radar cube shown in the~\cref{fig:fund_cube} we can calculate distances, Doppler velocities and both bearing and azimuth angles, depending on the spatial layout of the antennas on the chip.

\section{FMCW Radar Signal Processing}\label{sec:fund_proc}
Having in mind the sensing model outlined in the previous section we can now delineate the signal processing necessary to obtain the 4D radar point cloud, that is, a 3D point cloud and Doppler velocity associated with each point.

\subsection{Range}
In our setting, we use a millimeter wave \ac{fmcw} radar emitting a linear chirp signal. In such a case, the range to a reflecting object can be determined from the mixed transmitter and receiver signals.

Linear chirp (as seen in the upper subplot in the~\cref{fig:fund_chirp}) can be described in the time domain as (we omit the amplitude for clarity):

\begin{equation}\label{eq:fund_chirp}
    x(t) = e^{j(\pi S t^2 + 2\pi f_{c}t + \theta)}
\end{equation}

The instantaneous frequency is the time derivative of the argument (phase):

\begin{equation}\label{eq:fund_chirp_der}
    f(t) = \frac{d}{dt}\frac{1}{2\pi}(j(\pi S t^2 + 2\pi f_{c}t + \theta)) = St + f_{c}
\end{equation}

Where $S = \frac{B}{T_c}$ is the chirp rate (see~\cref{fig:fund_chirp}). Time delay $t_r$ with which a reflected chirp comes back to the receiver creates a difference in frequency $f_r$ which is called the \emph{beat frequency} or \emph{intermediate frequency}. Time delay $t_r$ is proportional to the beat frequency as seen in the~\cref{eq:fund_chirp_der}. We can see from the plot in the~\cref{fig:fund_chirps} that $f_r = S t_r = \frac{B}{T_c}t_r$, which means that $t_r = \frac{T_c}{B}f_r$ which is the time needed for the reflected chirp to travel to and from the reflecting object. We can also calculate the beat frequency as a result of complex mixing of signals:

\begin{equation}\label{eq:fund_chirp_mix}
    x(t) = e^{j(\pi S (t - t_r)^2 + 2\pi f_{c}(t - t_r) + \theta)}e^{-j(\pi S t^2 + 2\pi f_{c}t + \theta)} = e^{j(-2\pi S t_r t + \pi S t^{2}_{r} - 2 \pi f_{c} t_r)} 
\end{equation}

Since only the term containing $t$ determines the frequency of the mixed signals we have that (we can ignore the minus sign since the distance is only positive) $2 \pi f_r = 2 \pi S t_r$ and $f_r = S t_r = \frac{B}{T_c}t_r$. In order to determine the beat frequency we apply the \ac{fft} over the mixed signal and infer from it the time delay subsequently used for calculating the range: 

\begin{equation}
    d = \frac{f_{r}c_0}{2S}
\end{equation}

Where $c_0$ is the speed of light.

Two tones are distinguishable within the spectrum as long as their frequency difference is greater than the inverse of the chirp time ($\Delta f > \frac{1}{t_c}$). The range resolution therefore, that is, the smallest range difference at which two radar targets are still distinguishable one from the other, depends only on the chirp bandwidth and is given by:

\begin{equation}
    \Delta_r = \frac{c_0}{2B}
\end{equation}

The maximum range is: 

\begin{equation}
    r_{max} = \frac{N_s c_0}{4B}
\end{equation}

Where $N_s$ is the number of samples in the range \ac{fft}.

Note that to resolve ranges only the real part of the \ac{fft} is necessary since what we need are the magnitudes of the spectrum over the frequency axis. Phase information is not needed.

\subsection{Radial Velocity}
In order to calculate the radial velocity of an object we can exploit the rate of change of the phase of chirps. Namely, assuming that the reflected object moved between two chirps with velocity $v$, the time delay will increase slightly to $t_r = t_r + t_d$ where $t_d$ comes from the objects velocity. Now using this new increased time delay in the~\cref{eq:fund_chirp_mix} we can calculate the phase change of the object caused by $t_d$. Leaving out all negligible small terms we have that the phase change depends significantly only on the time delay caused by the motion of the object between two chirps:   

\begin{equation}\label{eq:fund_chirp_phase}
    \Delta \phi = 2\pi f_c t_d
\end{equation}

Where $t_d$ is the time delay of the chirp caused by the reflecting object traveling the distance $d_d$ during one chirp, therefore:

\begin{equation}\label{eq:fund_chirp_vel}
    \Delta \phi = 2\pi f_c \frac{2 d_d}{c_0} = 4 \pi \frac{d_d}{\lambda} = 2 \pi \frac{2 v t_c}{\lambda}
\end{equation}

Where we substituted $f = \frac{c_0}{\lambda}$, it follows then that:

\begin{equation}\label{eq:fund_chirp_velo}
    v = \frac{\lambda \Delta \phi}{4 \pi t_c}
\end{equation}

We can then calculate the velocity of objects by tracking the change of phase across chirps. Since the phase evolving in every range bin across all chirps forms a complex sinusoid, by performing the \ac{fft} over range bins across the series of chirps, the relative radial velocity of the object can be retrieved. We call the result of such 2D \ac{fft} process a \emph{range-Doppler} map.

The velocity resolution is given by:

\begin{equation}
    \Delta_v = \frac{\lambda}{2 N_c t_c}
\end{equation}

Where $N_c$ is the number of chirps within one frame. In words, the longer the frame time ($t_f = N_c * t_c$) the better the velocity resolution.

\subsection{Bearing Angle}
When multiple receiver antennas are available, calculation of angular information of reflecting objects is possible. Due to the spatial separation of the receiver antennas, the measured signals at each receiving antenna will be shifted in time: 

\begin{equation}
    t_a = \frac{d \sin \theta}{c_0}
    \label{eq:phase_shift1}
\end{equation}

\begin{figure}[th!pb]
  \centering
  \includegraphics[width=1.0\columnwidth]{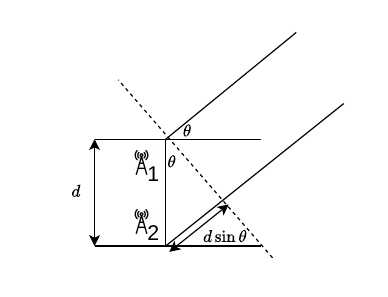}
  \caption[Angles resolution in FMCW radars using multiple antennas]{Multiple antennas on the chip allows for angle resolution using \ac{fmcw} radar. The dashed line represents the front of the reflected wave arriving at the two antennas.}
  \label{fig:fund_angles}
\end{figure}

Where $d$ is the separation distance between receiving antennas as seen in the~\cref{fig:fund_angles}. Exploiting once again the phase shift we can write:

\begin{equation}\label{eq:fund_angle}
    \Delta \phi = 2\pi f_c \frac{d \sin \theta}{c_0} = 2 \pi \frac{d}{\lambda} \sin \theta \approx 2 \pi \frac{d}{\lambda} \theta
\end{equation}

Now, measuring the phase shift using the \ac{fft} as in the case of velocities in the previous sub-section, but this time across all antennas we can retrieve the $\theta$ angle. Based on the antennas geometrical arrangement on the chip, both azimuth and elevation angles of objects with respect to the radar can be determined using the principle outlined in this subsection. To determine the elevation angle there must be receivers placed vertically and for calculating the azimuth angle there must be receivers placed horizontally. The angular resolution of a \ac{fmcw} radar is mostly limited by the number of receiver antennas on the chip. The greater the number of receivers on the chip, the greater the angular resolution. Nonetheless, every receiver entails a separate processing chain, which increases the chip size. Common solution to this problem is the Multiple-Input-Multiple-Output MIMO radar configuration, in which multiple transceivers are placed on the chip at known distances from each other, thus enabling the reception of phase-shifted signals. Temporal interleaving of transmitted signals leads to \textit{multiplicative} creation of the \textit{virtual} antennas increasing the angular resolution, as in:

\begin{equation}
    N_a = N_r \times N_t
    \label{eq:phase_mimo}
\end{equation}

Where $N_a$ is the number of antennas,  $N_t$ number of transceivers, and $N_r$ number of receivers.

Although \ac{fmcw} radars are thus capable of providing a 3D position of a reflecting object along with its relative Doppler velocity, the distance to the object and its relative Doppler velocity are the most precise parts of the information in small, light, and cost-efficient \ac{fmcw} radars.

Note that the presence of the sine function in the formula makes the relationship between the phase shift and the angle nonlinear, that is to say, that for larger angles the small angle approximation $\theta \approx \sin \theta$ does not hold anymore. 

The angular resolution is:

\begin{equation}
    \Delta \theta = \frac{\lambda}{N_a d \cos \theta}
    \label{eq:phase_shift1}
\end{equation}

\subsection{Radar Cross Section}
An important notion in radar engineering is that of the \ac{rcs}. \ac{rcs} can be defined as a property of an object illuminated by the electromagnetic waves emitted by radar, and quantifying a fictitious area describing its reflectivity back at the radar receiver \cite{rcs}. For simple objects such as a sphere the \ac{rcs} can be found analytically, while for more complex objects its value must be determined experimentally. The \ac{rcs} is defined as:

\begin{equation}\label{eq:rcs_proper}
    \sigma = 4 \pi \lim_{R\to\infty} R^2\frac{|E_s|^2}{|E_i|^2} 
\end{equation}

Where $R$ is the distance to the reflecting object, and $E_s$, $E_i$ are the backscattered, and incident electric fields, respectively. In practice, often what is used to estimate the \ac{rcs} is the so-called radar equation \cite{schoffmann_rcs}:

\begin{equation}\label{eq:rcs_pseudo}
    \frac{P_R}{P_E} = \frac{c G_E G_R \sigma}{(4 \pi)^2 R^4}
\end{equation}

Where $P_R$ and $P_E$ are the received and transmitted signals power, $G_R$ and $G_E$ are the receiver and transmitter gains, $c$ is a constant coefficient and $R$ is the distance to the reflected object.

\section{Summary}\label{sec:fund_conc}
In this chapter, we outlined the sensing principles behind millimeter-wave \ac{fmcw} \ac{soc} radars. We explained how the signal processing pipeline based on the successive application of the \ac{fft} allows for calculating the range, azimuth and elevation angles as well as the radial velocity which enables the construction of the 4D point clouds (3D points and Doppler velocity of each 3D point). We also defined the \ac{rcs} as a kind of \textit{visibility} of an object to the radar.

%% file: chapters/ekf_rio.tex
\chapter{Radar-Inertial Odometry Using The Extended Kalman Filter}\label{chap:ekf_rio}

\begin{sloppypar}
\emph{The present chapter contains results that were peer-reviewed and published in the International Conference on Unmanned Aerial Systems (ICUAS)~\cite{9836130}, IEEE/RSJ International Conference on Intelligent Robots and Systems (IROS)~\cite{previous_iros}, IEEE International Conference on Robotics and Automation (ICRA)~\cite{jan_icra} and the
IEEE International Symposium on Safety, Security, and Rescue Robotics (SSRR)~\cite{jan_fg}}
\end{sloppypar}

\bigskip

\noindent This chapter introduces three different approaches to estimating the navigation states of an \ac{uav} using the \ac{imu} and the \ac{fmcw} radar, all based on the \ac{ekf} algorithm. In \cref{sec:ekf_anchors} we introduce the first method which, although constrained to the usage of fixed anchors, allowed us to confirm the purposefulness of using the distance measurements from an \ac{fmcw} radar within the \ac{ekf} update step. In \cref{sec:ekf_rio_single} we expand on the previous approach by introducing the data association between the current and the previous set of the detected 3D points which generalizes the method to unknown environments. Moreover, we formulate the \ac{ekf} update step so as to additionally include the Doppler velocity besides the distance measurements. In \cref{sec:ekf_rio_multi} we present the final expansion of the \ac{ekf}-based \ac{rio} which includes persistent features, trails of measurements and past \ac{uav} poses as well as self-calibration capabilities. We also show how the final \ac{ekf} \ac{rio} implementation performs in closed-loop flights and in the simulated fog environment. In all our \ac{ekf} formulations we choose the tightly-coupled fusion which means we do not pre-process the radar measurements before fusing them in the \ac{ekf} but rather feed them in the estimator as they come delivered by the sensor. 

\section{EKF RIO using distance measurements to fixed anchors}\label{sec:ekf_anchors}
In this section, we present a preliminary \ac{rio} method which uses a tightly-coupled \ac{ekf} formulation in which we fuse measurements from an \ac{imu} with radar range measurements of known targets in the environment. This approach enables fast and accurate provision of the \ac{uav} state estimate even in scenarios where \ac{vio} suffers from both low rate (relative to the trajectory agility) correction information and image blur adversely affecting the state estimates. Also, a tightly-coupled formulation allows for state updates having sparse measurements, that is, contrarily to loosely-coupled formulations, ours does not require a prior 3D triangulation step for which distances to at least three features must be measured simultaneously. The method developed in this section can be considered a proof-of-concept for approaches shown later in the chapter (\cref{sec:ekf_rio_single} and~\cref{sec:ekf_rio_multi}). 
The work outlined in this section contributes a method for fast and accurate state estimation suitable for a rapidly moving \ac{uav} using a single small, lightweight, and low-cost \ac{fmcw} radar measuring ranges to so-called \textit{corner reflectors} \cite{doerry2008reflectors} in the environment and an \ac{imu} sensor. The approach uses reflectors at known positions and self-calibrates system extrinsics, \ac{imu} extrinsics, and estimates the 6DoF pose with 3D velocity. We show that with this method we can outperform \ac{vio} when sharp and aggressive motions are executed. Moreover, the presented approach can be used in environmental conditions completely prohibitive for \ac{vio} like fog or lack of light.
We evaluate our approach experimentally and compare against the state-of-the-art \ac{vio} algorithm in~\cite{geneva2020openvins}, using the platform shown in Fig.~\ref{fig:platform}.

\subsection{Methodology}\label{subsec:ekf_anchors_meth}
\subsubsection{System Overview}
We consider our approach a tightly-coupled one since we use the one-dimensional distance measurement from the onboard radar sensor to a radar target as information for \ac{imu} integration correction. We do not pre-process several distance measurements from several target to triangulate first a 3D position to only then use this result as position correction. Compared to the pre-triangulation method, our tightly coupled approach has several advantages: first, a quick non-linear observability analysis shows that all motion states including \ac{imu} biases are locally observable if the system is excited in acceleration and angular velocity while observing only one target (intuition to this fact can be gained from \cite{martinellivisual}). Thus, the radar does not need to observe all targets at once like in the triangulation pre-processing approach. Note that, at least sequentially, it needs to observe at least two different targets to eliminate any gauge freedom. Second, omitting the pre-processing step from the distance measurement to a 3D position measurement reduces the distortion of the noise characteristics that may otherwise impact the \ac{ekf} assumptions on Gaussian distributions. Third, we can process measurements as they arrive and do not need any sort of synchronization between the targets. Fourth, the only geometric condition to avoid singularities (in this case to eliminate the gauge freedom in 3D position and yaw) is that at least two targets have different positions in a direction perpendicular to gravity (i.e. in the xy-plane).

Our estimator setup is such that at least two of the targets need to be placed at known location to eliminate said gauge freedom (3D position and yaw), positions of additional targets can seamlessly be integrated in the estimation process. For the initialization of additional targets online, the approach in \cite{bluemlbias} could be adapted to use initial distance measurements from a newly observed target. That said, in this paper we only focus on the radar-inertial estimation of the motion states as proof-of-concept. 

The targets generate a clear signal in the radar receiver. We perform an \ac{fft} to extract the distances to each target and then use this information in the \ac{ekf} framework. The data association is done by matching the measured range closest to the expected range calculated from the current vehicle pose and a given target location. The following sections detail each step further.

\subsubsection{Radar Signal Processing}
\begin{figure}[h!]
  \centering
  \includegraphics[width=0.7\columnwidth]{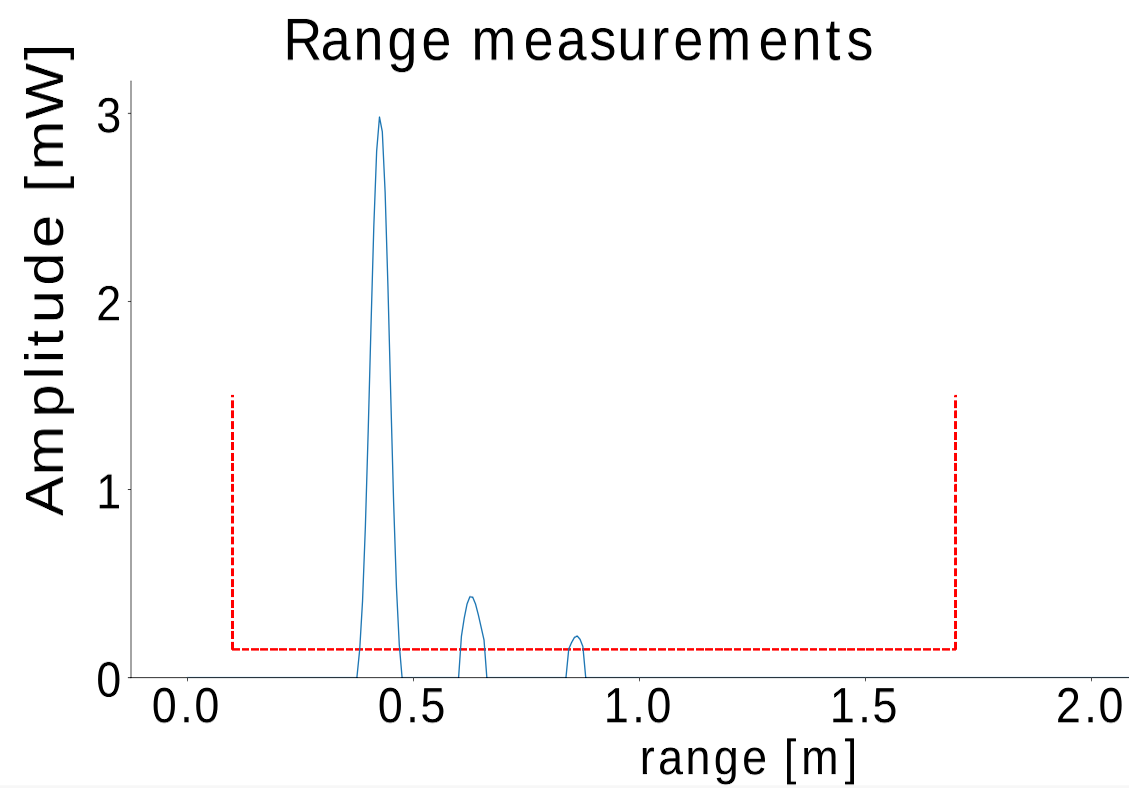}
  \caption[Range-amplitude plot from FFT processing]{Range-amplitude plot after performing \ac{fft} of the radar raw signal. The dashed red lines depict thresholds assumed for cropping the radar signal in frequency (mapped to range) and for application of the threshold above which targets are identified.}
  \label{fig:rangesonly}
\end{figure}

The radar sensor used to validate the approach presented in this section can be seen in the~\cref{fig:platform}. This miniature \ac{soc} \ac{fmcw} radar delivers raw signal in the form of the mixture of sinusoids for each chirp. Therefore, in order to retrieve the distance measurements to be used in the \ac{ekf} an amount of signal pre-processing must be done according to the principles outlined in the~\cref{chap:fund_radar}. As the method described in this section makes use of the radar delivering the raw data (as opposed to the radar sensor we use later on in~\cref{sec:ekf_rio_single} and~\cref{sec:ekf_rio_multi}), the pre-processing also includes convolving the receiver signal with a Hamming window to minimize spectral leakage. Very low and high frequency contributions which are not in the region of interest are filtered out using a bandpass filter. The bandpass filter was designed with cut-off frequencies mapped to ranges $r_{c,L} = 0.1$~m and $r_{c,H} = 1.7$~m to suppress frequency peaks caused by noise and multipath reflections taking into account the approximate flight altitude range. After filtering, the signal is thresholded above an experimentally obtained value of $c_{\text{noise}} = 0.15$ and the resulting signal is searched for clusters. The frequency value corresponding to the maximum value of each cluster is returned and mapped to a range value (\cref{fig:rangesonly}).

\subsubsection{Data association}
With the output of the processing module, we obtained range values of all targets in the field of view of the radar sensor. Compared to obtaining Doppler velocity information or angular information, the range values only need a single \ac{fft} pass for calculation. This enables very fast measurements and subsequent propagation correction in the estimator framework. However, this lightweight data processing comes at the difficulty of data association. Each identified peak (i.e. measurement) in the range-amplitude plot (\cref{fig:rangesonly}) needs to be associated with a target in the world to derive the necessary correction in the \ac{ekf} formulation. For this association, we first compute estimates of the ranges between the current vehicle position and the targets from the information we have at the current time step. Then, a greedy search is minimizing the error of a given measurement with the computed possible range estimates. If the previous radar scan already had a similar range measurement, this information is included as a prior. Since the radar scans are very fast compared to the vehicle motion, this element helps speeding up the data association process. The matched range measurement with a target forms then an update information pair for the \ac{ekf}. If a range measurement differs more than \unit[3]{cm} to any possible target range, then the measurement is discarded.

\subsubsection{Tightly coupled Extended Kalman Filter formulation}
The \ac{ekf} framework uses an \ac{imu} for the propagation of the state formulation defined by Eq.~\eqref{eq:ekf_anchors_states}. The states are the position of the \ac{imu}/body frame~$\reference{{\vp}}{G}{I}{}$ and velocity~$\reference{{\vv}}{G}{I}{}$ expressed with respect to the world frame, the orientation of the \ac{imu} in the world frame~$\reference{\bar{\vq}}{G}{I}{}$, gyroscopic~bias~$\vb_{\bomega}$ and accelerometer~bias~$\vb_{\va}$. The 3D translation between the onboard radar senor and the \ac{imu} is expressed as the calibration state $\reference{{\vp}}{I}{R}{}$ in the \ac{imu} frame. $\referencet{{\vp}}{G}{A}{}{i}$ represents the $i$-th corner reflector (i.e. radar target) 3D position in the world frame with $i=1, 2, ... K$. Note that at least two such targets need to be known and fixed to eliminate the gauge freedom. The full state vector $\vx$ is then defined as follows:

\begin{equation}\label{eq:ekf_anchors_states}
\begin{aligned}
    \vx &= \left[ \vx_{\cN}; \vx_{\cA} \right] \\
    &= \left[ \left[\reference{{\vp}}{G}{I}{}; \reference{\bar{\vq}}{G}{I}{}; \reference{{\vv}}{G}{I}{}; \vb_{\va}; \vb_{\bomega}; \reference{{\vp}}{I}{R}{}\right]; \left[\referencet{{\vp}}{G}{A}{}{1}; \referencet{{\vp}}{G}{A}{}{2}; \dots; \referencet{{\vp}}{G}{A}{}{K} \right]\right]
\end{aligned}
\end{equation}

The system dynamics of the core states are defined according to~\cite{weiss2012versatile}. The dynamics and process noise of the calibration state $\reference{{\vp}}{I}{R}{}$ are assumed to be zero because of the rigid body assumption. In our setup, $\referencet{{\vp}}{G}{A}{}{i}$ are kept fix as known values. The \ac{ekf} framework uses a regular error-state (including error quaternion) definition. We use the Hamilton notation for the quaternion representation~\cite{sommer2018and}.

An estimated range measurement $\hat{z}^i$ from the radar sensor to a target $i$ and the corresponding error using the true measurement $z^i$ from the data association process can then be defined as

\begin{align}
    \hat{z}^i &= \|\referencet{{\vp}}{G}{A}{}{i} - (\reference{{\vp}}{G}{I}{} + R\{\reference{\bar{\vq}}{G}{I}{}\}\reference{{\vp}}{I}{R}{})\|\\
    \tilde{z}^i &= z^i -\hat{z}^i
    \label{eq:rangemeasurement}
\end{align}

Where $R\{\reference{\bar{\vq}}{G}{I}{}\}$ rotates a vector from the IMU frame to the world frame using the quaternion $\reference{\bar{\vq}}{G}{I}{}$. This renders the update step very efficient for any number of currently visible targets by the radar sensor. With a standard computer, we achieve an update rate of about 90~Hz in our setup described below.

\subsection{Experiments}\label{subsec:ekf_anchors_exp}
The above described method enables a simple, yet computationally very efficient radar-inertial state estimation with self-calibration capabilities (\ac{imu} intrinsics, radar-\ac{imu} extrinsics). In the following, we test our method on real platforms with real data.

\subsubsection{Experimental setup}

\begin{figure}[thpb]
  \centering
  \includegraphics[scale=0.15]{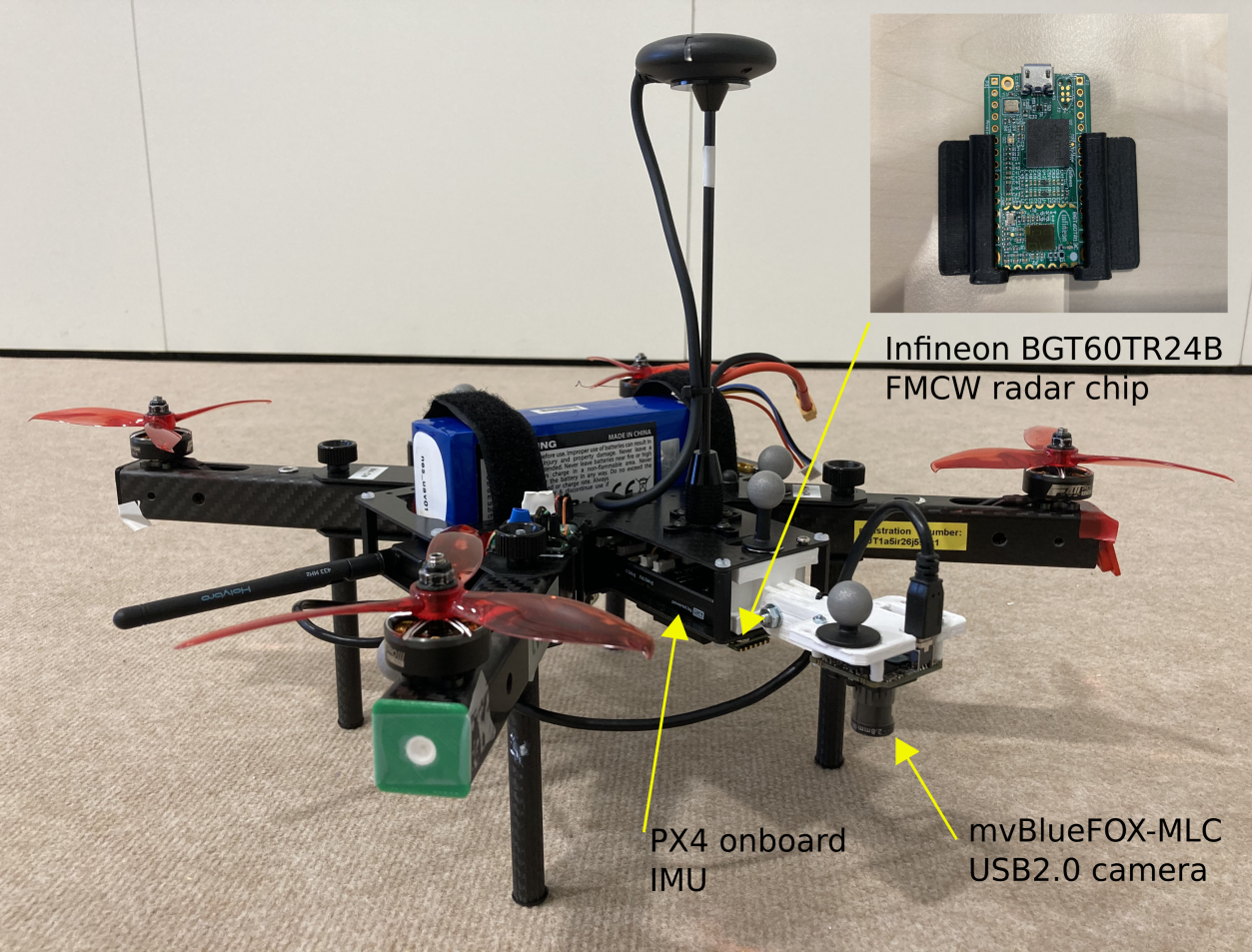}
  \caption[Experimental UAV platform used for state estimation using fixed anchors]{Experimental platform used in this work and the \ac{fmcw} radar sensor mounted in its custom-made housing.}
  \label{fig:platform}
\end{figure}
\begin{figure}[thpb]
  \centering
  \includegraphics[width=0.5\columnwidth]{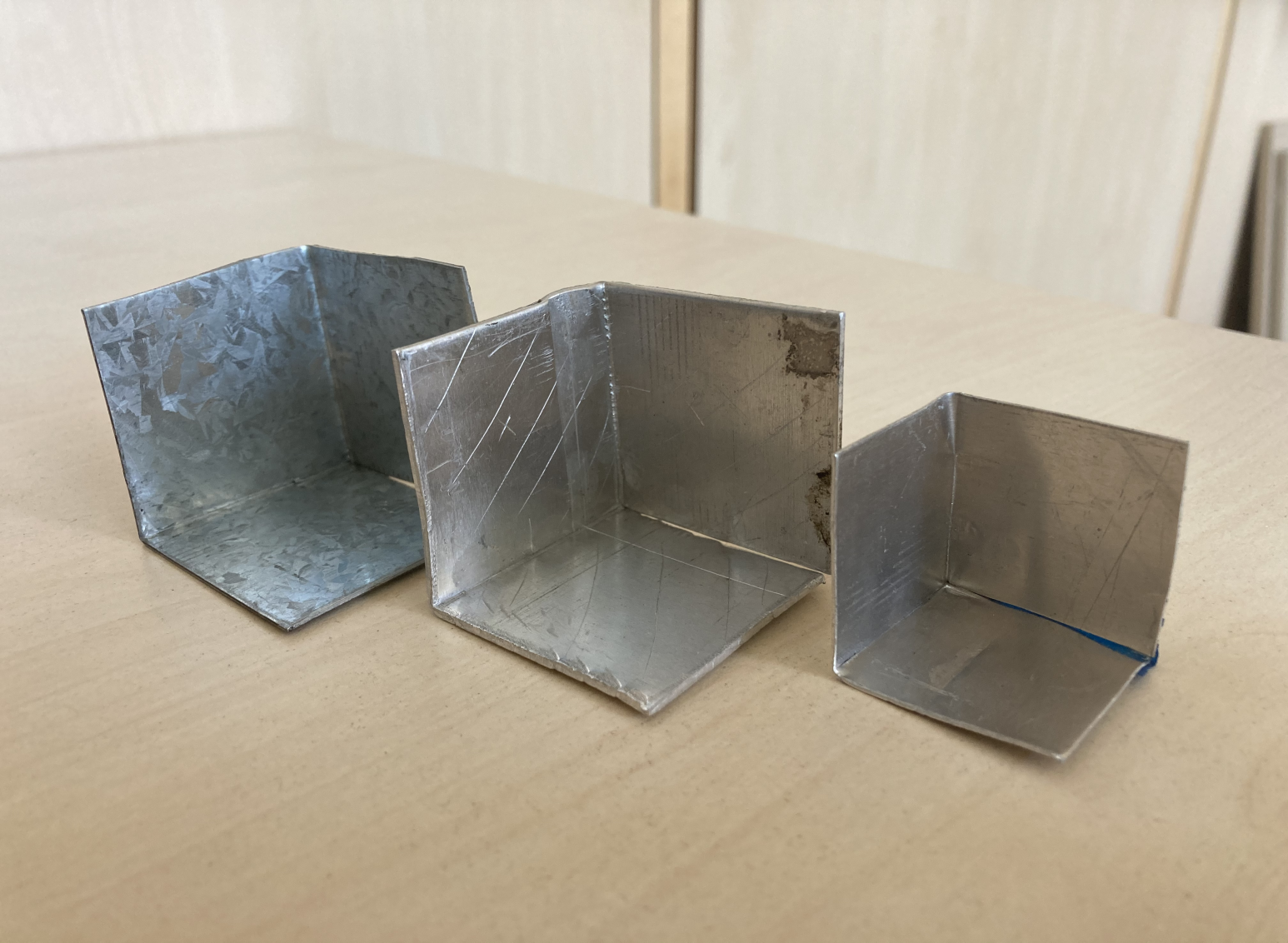}
  \caption[Corner reflectors used as fixed anchors]{Corner reflectors used in our experiments.}
  \label{fig:corners}
\end{figure}

The sensor used for the experiments is the lightweight \unit[60]{GHz} multi-channel \ac{fmcw} radar transceiver BGT60TR24B from Infineon, shown attached to the \ac{uav} in~\cref{fig:platform}, mounted on the Infineon XMC4500 board equipped with a USB interface. The frequency spectrum of chirps generated by the radar is between \unit[$f_l = 57$]{GHz} and \unit[$f_u = 63$]{GHz}. We set the sampling frequency to \unit[$f_{sr} = 2$]{kHz}, the number of samples to $N_s = 200$ and the number of chirps to $N_c = 20$. For inertial measurements we use the \ac{imu} supplied on the PX4 platform. We set the sampling rate of the \ac{imu} to \unit[$f_{si} = 200$]{Hz}. For comparison with \ac{vio}, we use images grabbed by the onboard mvBlueFOX-MLC camera connected over USB. We set the frequency of the camera to \unit[$f_{sc} = 20$]{Hz} which is a reasonable choice from the potential on-board processing viewpoint. Although \ac{vio} can benefit from the stereo-camera setup, we do not make use of it because of the limited payload of the \ac{uav}. We set the camera exposure time to \unit[$e = 8$]{ms} for a typical indoor scene.

We placed three corner reflectors (\cref{fig:corners}) as radar targets at the locations $A_{1} = [x=0.23, y=0.65, z=0.36]$m, $A_{2} = [x=0.52, y=0.53, z=0.48]$m and $A_{3} = [x=0.48, y=0.86, z=0.59]$m.

\begin{figure}[h!tbp]
  \centering
  \includegraphics[width=0.46\columnwidth]{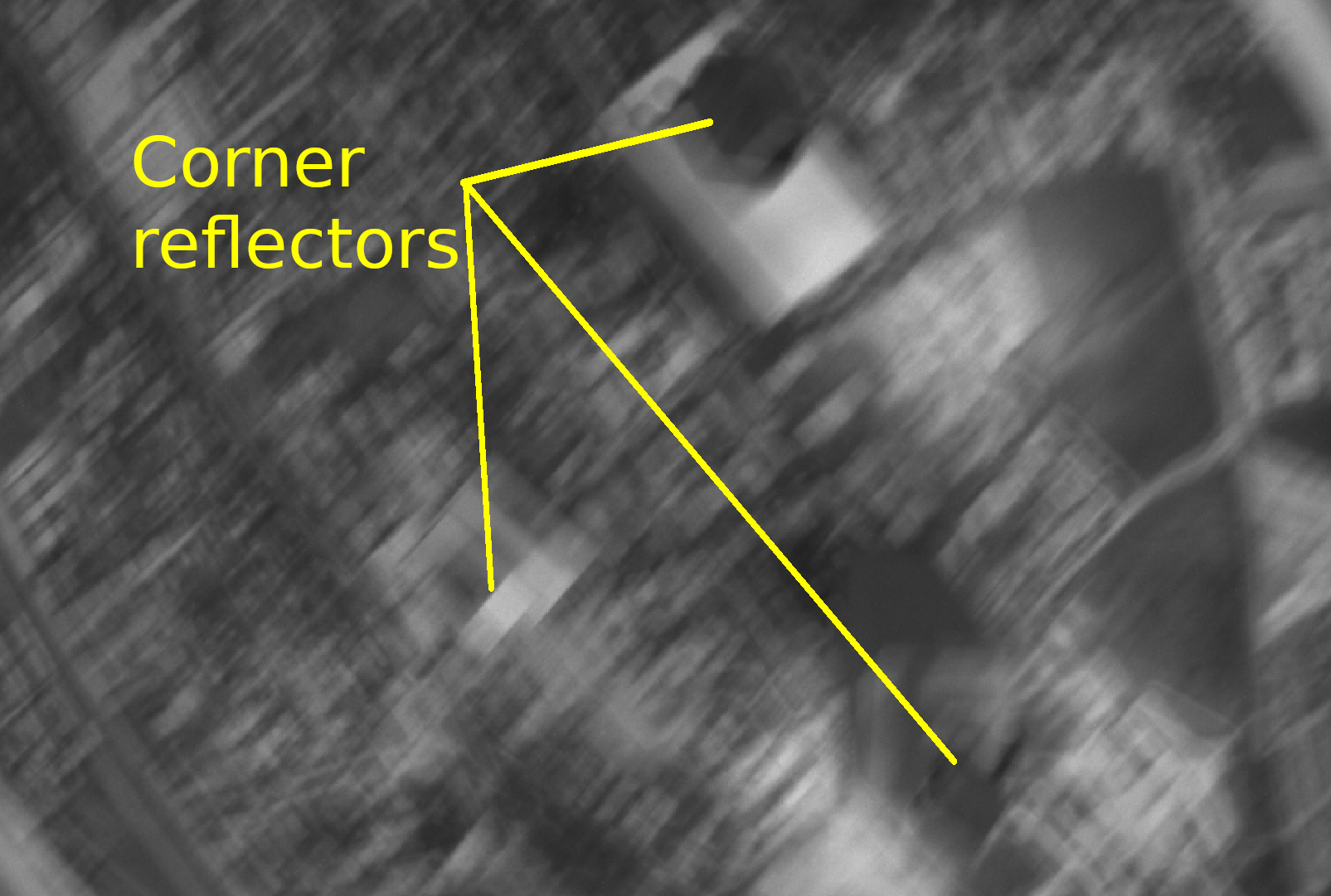}
  \includegraphics[width=0.43\columnwidth]{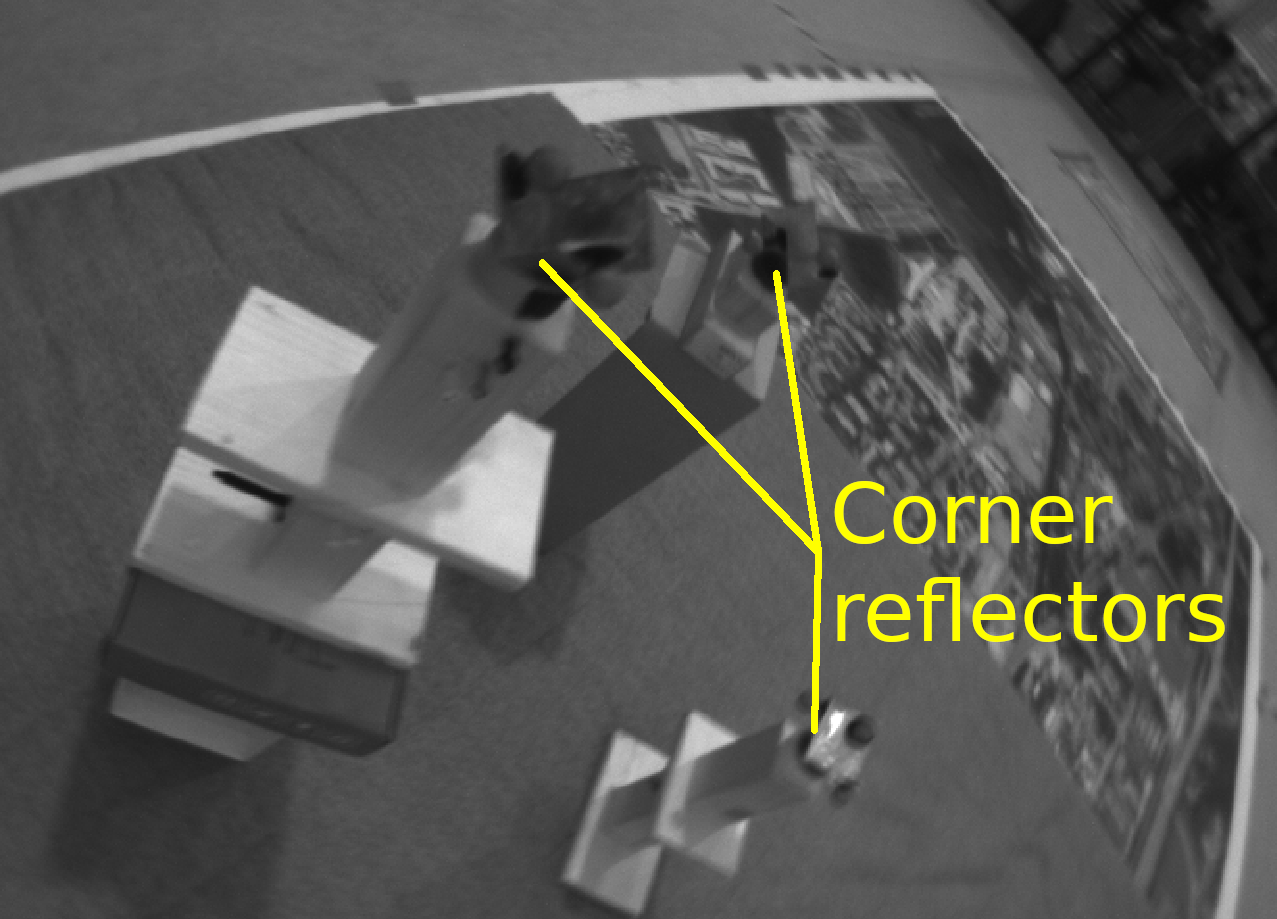}
  \caption[Corner reflectors as seen from the onboard camera]{Radar target setup shown from the onboard camera view. Left: above well textured area during a high-speed motion causing significant motion blur. Right: still phase above low textured area.}
  \label{fig:targetsetup}
\end{figure}

\begin{figure}[h!tbp]
  \centering
  \includegraphics[width=0.95\columnwidth]{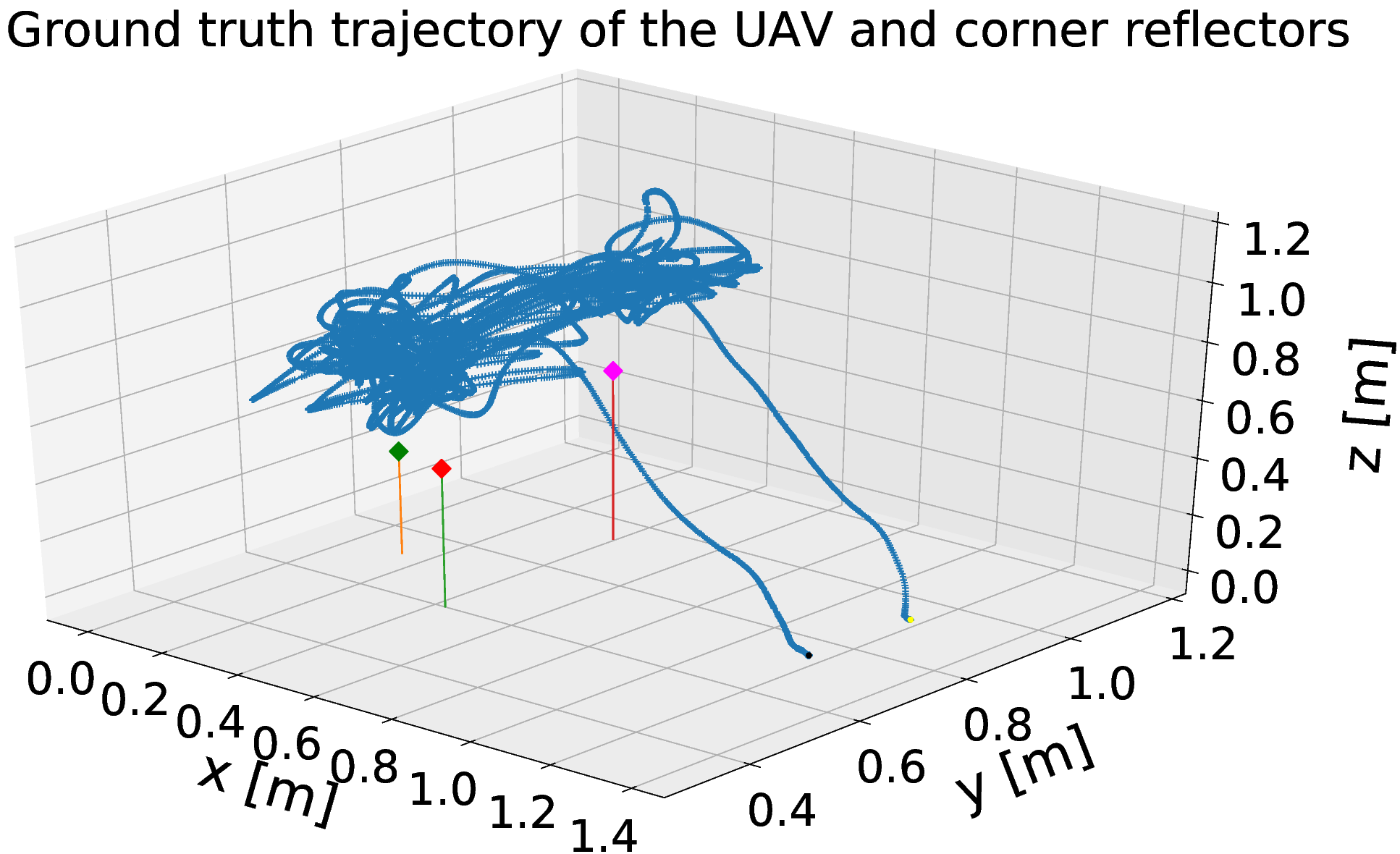}
  \caption[Ground truth trajectory of the UAV above corner reflectors]{Ground truth trajectory of the \ac{uav} above the set of three corner reflectors. Trajectory of the \ac{uav} is plotted in blue, green, magenta and red diamonds are the corner reflectors placed on the floor, the black and yellow dots are final and initial \ac{uav} positions respectively.}
  \label{fig:trajectory}
\end{figure}

\begin{figure}[h!]
  \centering
  \includegraphics[width=0.95\columnwidth]{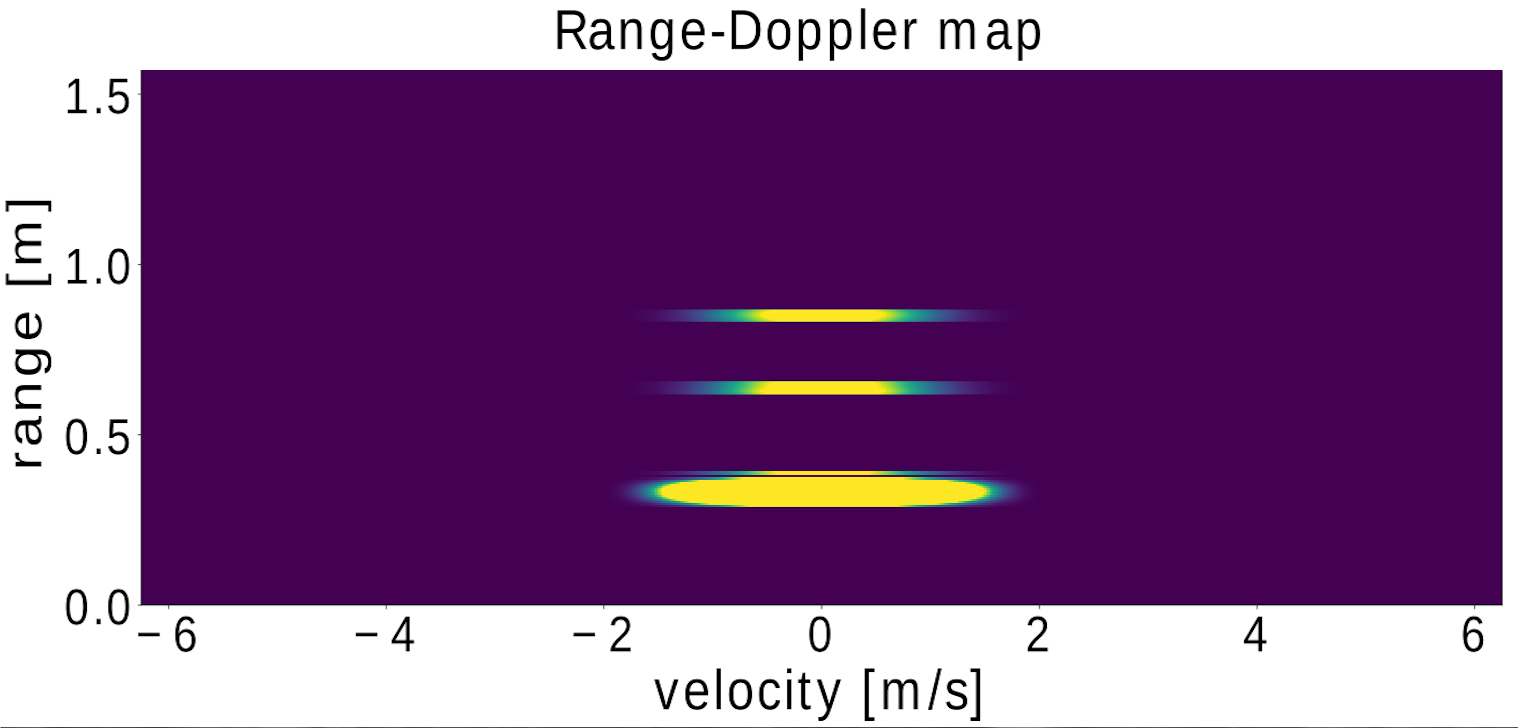}
  \caption[Range-Doppler map generated from sensing corner reflectors]{Range-Doppler map generated from a typical scene as shown e.g. in~\cref{fig:targetsetup} from the three corner reflectors. The ellipses in the velocity dimension are fairly wide which reflects high uncertainty of the Doppler velocity measurement. Thus, and because of the lighter computation, we only use range measurements in our approach.}
  \label{fig:rangedoppler}
\end{figure}

We then acquired two datasets performing highly aggressive hand-held trajectories with the platform in~\cref{fig:platform}. Each acquisition involved very sharp and aggressive movements (norm of max. angular velocity \unit[$\omega = 12.6$]{$\frac{rad}{s}$}, norm of max. linear acceleration \unit[$a = 33.0$]{$\frac{m}{s^{2}}$}) of the sensor rig such that high motion blur was affecting the camera sensor. Fig.~\ref{fig:targetsetup} depicts the setup from the onboard camera view with motion blur above the well textured area (left) and a sharp image on the low textured area (right). The trajectories were carried out just above the set of three corner reflectors (\cref{fig:trajectory}) allowing the same (visual) feature set to be visible during the entire experiment and thus allowing the \ac{vio} to leverage persistent features for locally non-drifting estimation. A sample range-Doppler reading by the radar of such a scene is depicted in~\cref{fig:rangedoppler}. The two acquired datasets are recorded with feature-poor and feature-rich backgrounds in order to see the effect the background has on the \ac{vio} performance. 

During acquisition, we recorded sensor readings from the camera, \ac{imu}, and radar together with the poses of the \ac{uav} and corner reflectors streamed by the motion capture system for the ground truthing. Both estimators, our \ac{ekf}-based radar-inertial state estimator and OpenVins \cite{geneva2020openvins} as the state-of-the-art reference \ac{vio} are run offline on the recorded sensor data on an Intel core i7 vPRO laptop with 15 GB RAM. For a fair comparison of the proposed methodology with the \ac{vio}, we carefully fine tuned the feature tracker (incl. persistent features) of OpenVins to get the best possible result out of the \ac{vio} algorithm given the high aggressiveness of the performed movements and the challenging amount of motion blur contained in the camera images.

\begin{figure}[h!]
  \centering
  \includegraphics[width=0.95\columnwidth]{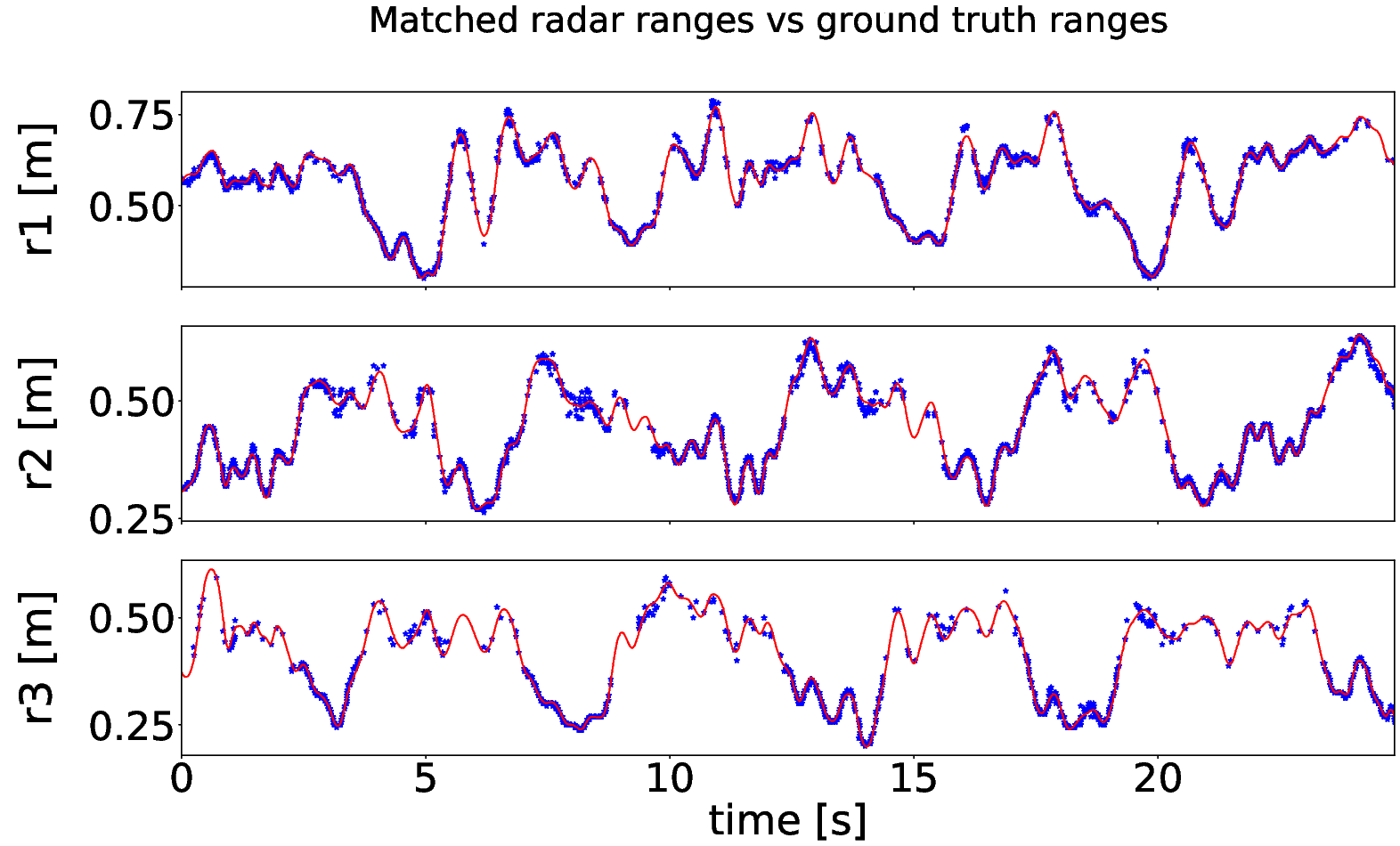}
  \caption[Ranges measured matched to ground truth ranges]{Ranges measured with radar matched to the ground truth. Ground truth is plotted in red, matched radar range in blue.}
  \label{fig:matchedranges}
\end{figure}

\subsubsection{Evaluation}
Firstly, we evaluate if our proposed data association yields adequate results to be used as update information in our \ac{ekf} estimator framework. For this, we compare the measured ranges using the approach described in~\cref{subsec:ekf_anchors_meth} to the ground truth range computed from the motion capture measurements of the radar sensor and the corresponding target 3D position. Fig.~\ref{fig:matchedranges} depicts in red the ground truth ranges during a 25 seconds long test and shows in blue the measurements obtained with our method. Note that often, our approach could not clearly associate the measured ranges to a target (gaps between the blue stars). This is due to the noise in the radar reading and our gating approach such that any measurement differing more than \unit[3]{cm} to any possible target range is directly discarded (see~\cref{subsec:ekf_anchors_meth}). That said, our tightly coupled estimator approach is resilient to intermittent target losses as any available measurement is seamlessly used as 1D measurement whenever available. Fig.~\ref{fig:matchedranges} shows this particularly well around \unit[t=15]{s} where only the first target yields measurements. From this data we measured a standard deviation of \unit[$\sigma_{m} = 1$]{cm} for the measurements. This noise value was included in the \ac{ekf} process. 

As~\cref{fig:rangedoppler} shows, Doppler velocity could additionally be extracted from the radar signal. However, it significantly increased computational load for its calculation, as it requires to compute the \ac{fft} across all ranges for each chirp, and the fact that velocity is already observable with only range measurements are all factors that do not motivate to use the Doppler velocity as measurement. With our implementation we achieve an average range measurement rate of about \unit[90]{Hz}.

\begin{figure}[H]
  \centering
  \includegraphics[width=0.8\columnwidth]{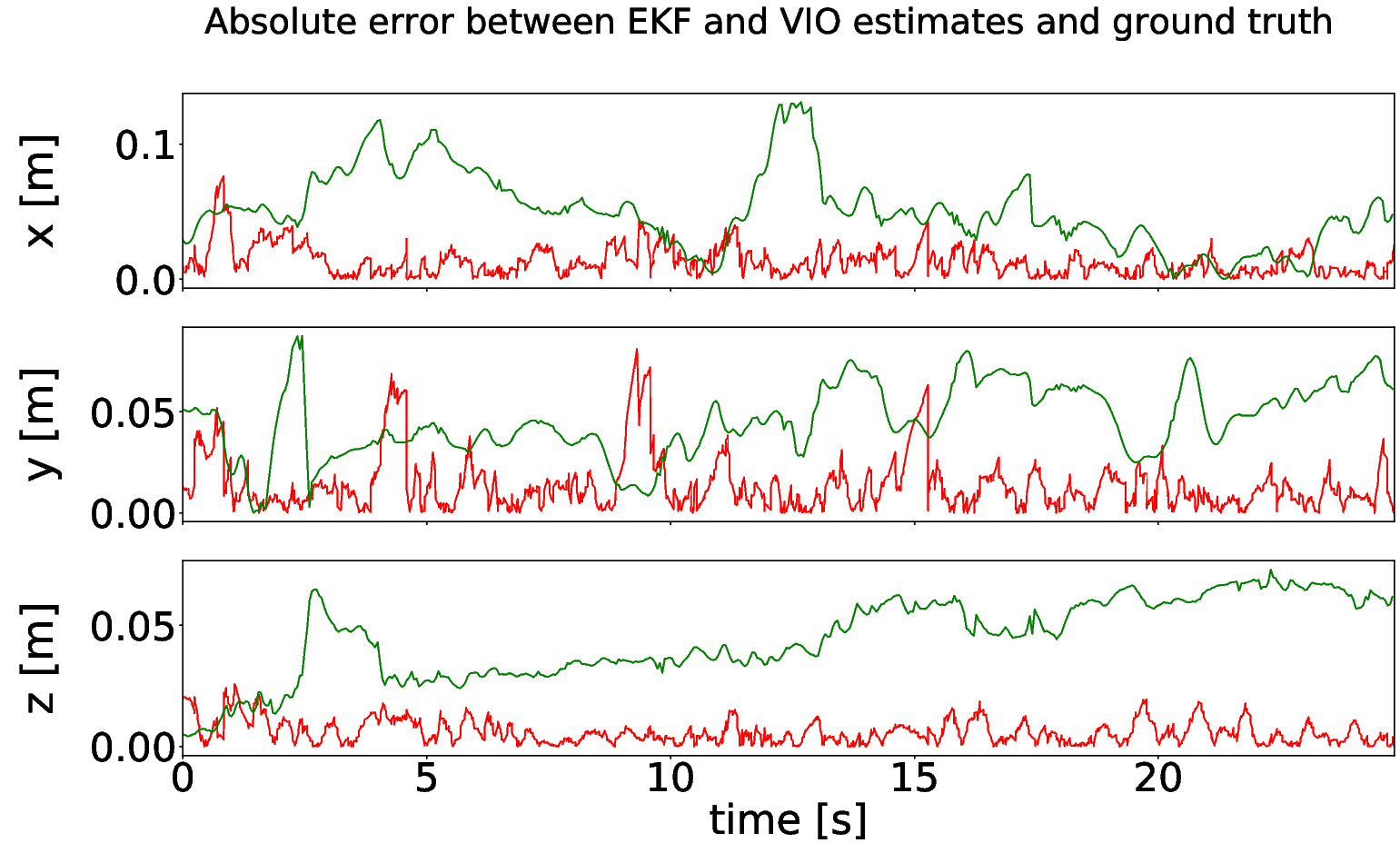}
  \caption[Absolute position estimation error in feature-rich scenario]{Absolute position estimation error for feature-rich scenario. Errors for VIO are plotted in green, for our radar based approach in red. The position drift of the \ac{vio} is more than \unit[13]{\%} showing the impact of the challenging data.}
  \label{fig:aee5}
\end{figure}

\begin{figure}[H]
  \centering
  \includegraphics[width=0.8\columnwidth]{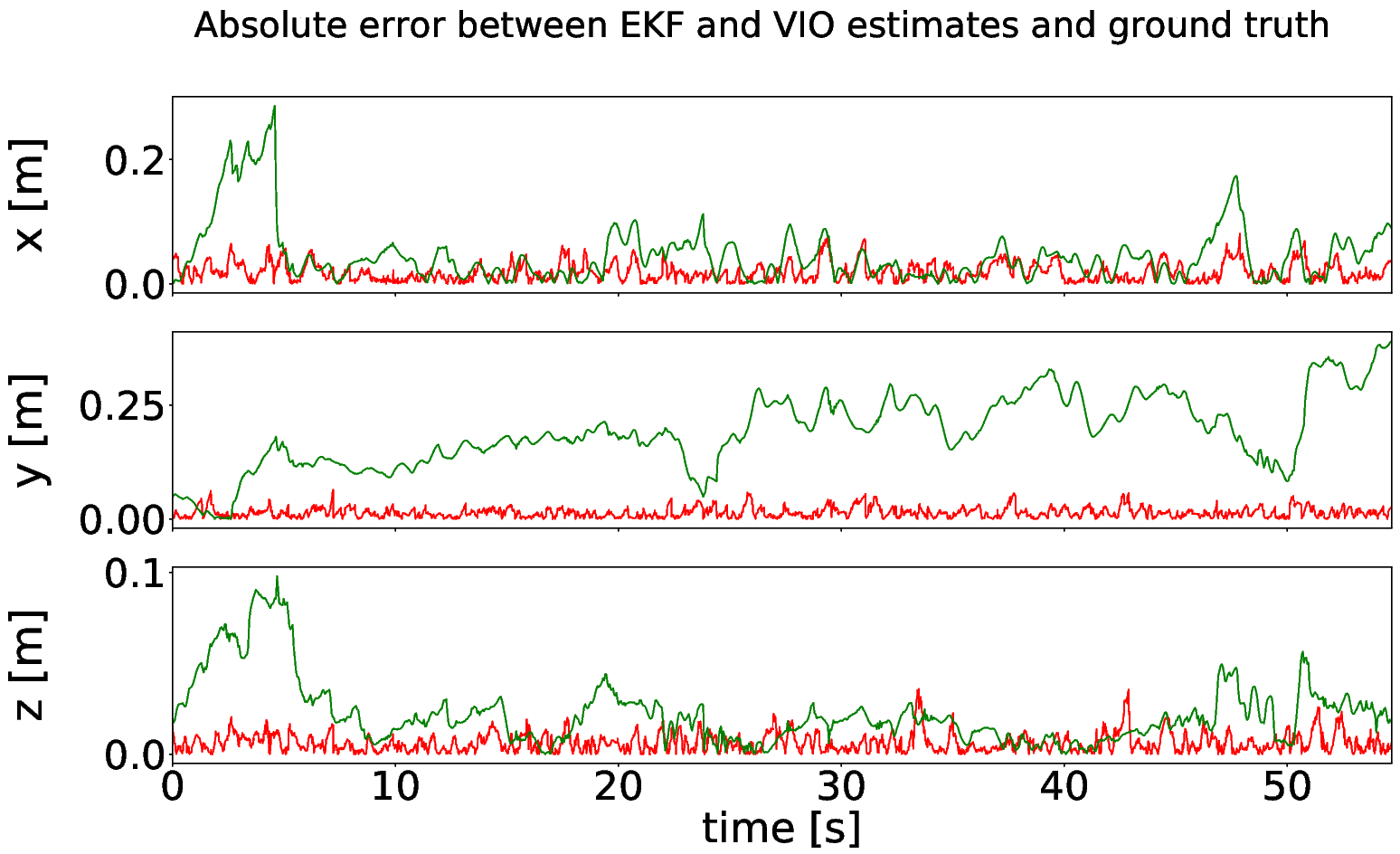}
  \caption[Absolute position estimation error in feature-poor scenario]{Absolute position estimation error for feature-poor scenario. Errors for \ac{vio} are plotted in green, for our radar based approach in red. With our best tuning efforts, we managed to get non-diverging results and a position drift of about \unit[20]{\%} for \ac{vio}.}
  \label{fig:aee12}
\end{figure}

\begin{figure}[H]
  \centering
  \includegraphics[width=0.8\columnwidth]{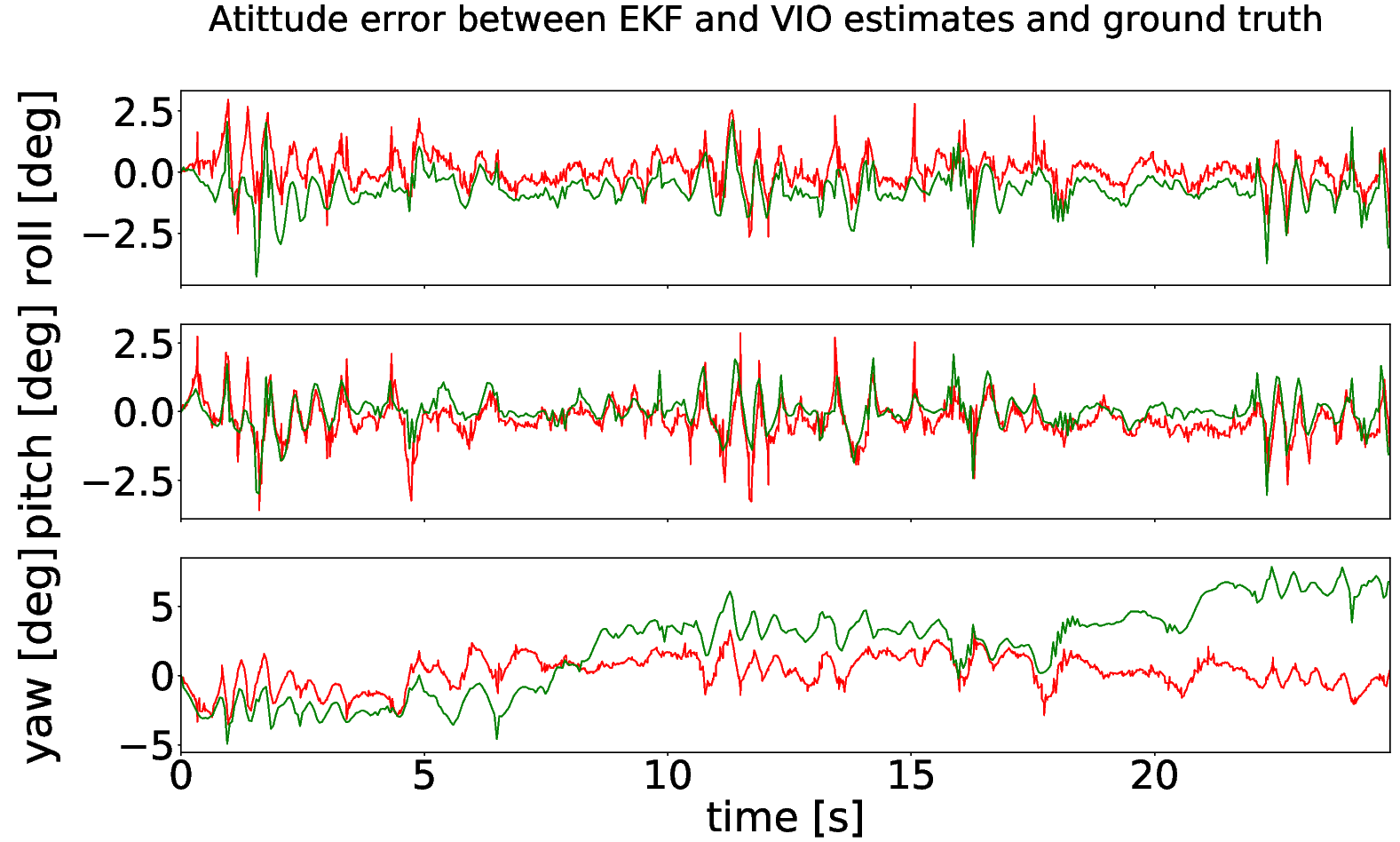}
  \caption[Attitude estimation error in feature-rich scenario]{Attitude estimation error for the feature-rich scenario. Errors for \ac{vio} are plotted in green, for our radar based approach in red. The challenging motion clearly affects the (unobservable) yaw drift of the \ac{vio}.}
  \label{fig:aee5attitude}
\end{figure}

Secondly, we test the estimator performance when the extracted ranges are fused together with IMU in an EKF framework. We compare this radar-inertial estimation with visual-inertial odometry for highly aggressive maneuvers. Fig.~\ref{fig:aee5} and~\cref{fig:aee12} show the absolute position error plots for our approach (red) and \ac{vio} (green) compared to the Optitrack ground truth in well-textured and poorly-textured scenes respectively. With the well tuned \ac{vio}, we managed to achieve non-diverging results for the \ac{vio} algorithm in both cases. A non-negligible drift (final error divided by overall path length), however, persists for both cases: about \unit[13]{\%} for the well-textured scene, and nearly \unit[20]{\%} for the poorly-textured one (see~\cref{tab:errors} for details). Fig.~\ref{fig:aee5attitude} shows a similar plot for the attitude error in the well-textured test. 
The unobservable yaw is most affected by the challenging data since the persistent features are lost due to the motion blur. We encourage to not take the figures and~\cref{tab:errors} as direct comparisons between the two algorithms since this would compare unobservable states in \ac{vio} against observable ones in our approach. Rather, they show the behavior of the approaches in challenging situations: for \ac{vio} they show an order of magnitude higher drift than usually reported in literature discussing well-behaving scenarios despite dataset specific tuning of the \ac{vio} algorithm. The increase in drift is caused by the \ac{vio} not being able to consistently keep the persistent features because the high motion blur. With the lack of their locally non-drifting information, the algorithm thus goes back to the mode in which only odometry information can be used. For our approach, the figures show an \ac{rmse} below \unit[3]{cm} in position despite the very agile motion and only using sequential 1D range measurements for IMU integration correction. 

\begin{figure}[H]
  \centering
  \includegraphics[width=0.8\columnwidth]{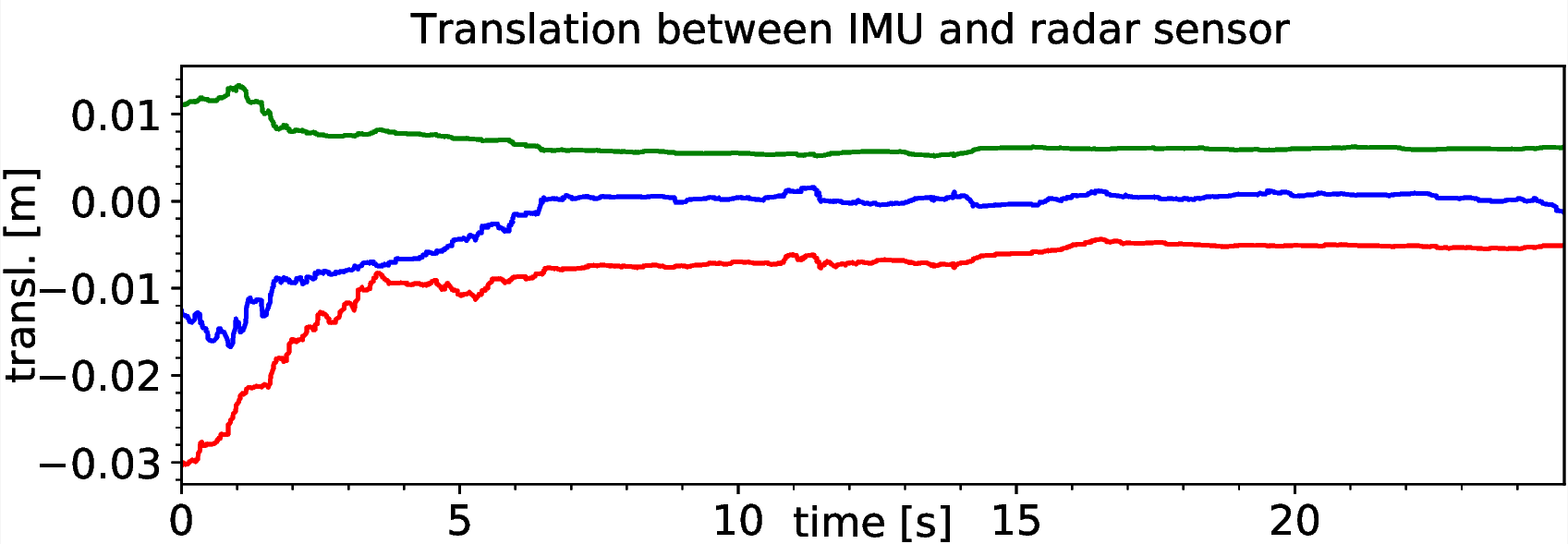}
  \caption[Estimated IMU-radar translation]{Estimated translation $\reference{{\vp}}{I}{R}{}$ between radar and \ac{imu} sensors for one of the experiments (feature-rich scenario). x, y and z coordinates are plotted in red, green and blue respectively.}
  \label{fig:aee5calibration}
\end{figure}

Third, we evaluated the self-calibration capability of the proposed estimator. Fig.~\ref{fig:aee5calibration} shows the evolution of the extrinsic calibration state $\reference{{\vp}}{I}{R}{}$ representing the 3D translation between the onboard IMU and radar sensor. After a wrong initialization, the state converges well.  \\

\begin{table}
\caption{Experiments and \ac{rmse} after convergence}
\label{tab:errors}
\centering
\begin{tabular}{|l|c|c|c|c|c|}
\hline
 & pos [m] & roll [\degree] & pitch [\degree] & yaw [\degree] & drift [\%] \\
\hline
 \multicolumn{6}{|c|}{well textured scene} \\
\hline
 Ours & ${0.0268}$ & ${0.7861}$ & ${1.3476}$ & ${1.2105}$ & ${7.0985}$\\
\hline
 \ac{vio} & $0.0894$ & $0.9272$ & $0.6389$ & $3.7205$ & $13.2782$\\
\hline
 \multicolumn{6}{|c|}{poorly textured scene} \\
\hline
Ours & ${0.0311}$ & ${1.1901}$ & ${1.3047}$ & ${1.4743}$ & ${5.1859}$\\
\hline
\ac{vio} & $0.2183$ & $1.1429$ & $0.7438$ & $8.9878$ & $19.9222$\\
\hline
\end{tabular}
\end{table}


\subsection{Conclusions}\label{subsec:ekf_anchors_con}
In this section, we presented a method which uses a tightly-coupled formulation of an Extended Kalman Filter in which we fuse measurements from an \ac{imu} with range measurements from a low-cost and lightweight \ac{fmcw} radar. We presented a data association method to match targets in the environment with the signals in a radar scan. To eliminate any gauge freedom, we fixed at least two targets defining the gravity aligned world frame and can seamlessly include additional target positions due to the tightly coupled estimator formulation. We showed that this approach enables fast and accurate estimation of the \ac{uav} pose even in scenarios where \ac{vio} suffers from the image blur adversely affecting its accuracy (despite using persistent features for locally non-drifting information). Our computationally simple approach only extracting range information from the raw radar signals and use them as sequential 1D measurements in an EKF formulation enabled estimator update rates of about \unit[90]{Hz}. Our \ac{ekf} formulation is capable of estimating the navigation states, \ac{imu} intrinsics, and radar sensor extrinsics. The fast and motion-blur free measurements are particularly relevant for \ac{uav}s performing aggressive manoeuvres. As the next steps, moving from this rather area-bound proof-of-concept presented here, we wish to integrate radar sensors with longer range and different measurement profile which will allow us to detect features in the environment and match them between consecutive scans. 

\section{RIO EKF using distance and Doppler velocity measurements of 3D points}\label{sec:ekf_rio_single}
As shown in the previous section, the \ac{fmcw} \ac{soc} radar can be successfully used to correct the \ac{imu} drift within an \ac{ekf}. Bearing in mind this conclusion, in this section, we show a \ac{rio} method which employs a different radar sensor model and is more flexible in that it does not require any prior knowledge of the environment (like reflective anchors positions in the previous chapter). Thus, in this section we present a novel \ac{rio} method which employs stochastic cloning \cite{roumeliotis2002stochastic} to enable matching of the measured 3D points from the previous radar scan to the ones in the current scan. In addition to these relative distance measurements of matched 3D points, we also use Doppler velocity information measured from all features in the current scan. We fuse all measurements in a tightly-coupled formulation in our \ac{ekf} setup. The tight coupling enables the incorporation of single distance and velocity measurements in the update step. This property relieves us from any constraints on required minimal number of matches (as it is e.g., needed for a prior \ac{icp} and subsequent loose coupling of the resulting delta-pose in the \ac{ekf}). This is a particularly strong advantage in view of robustness and accuracy over loosely coupled approaches since, e.g., \ac{icp} \cite{121791} works poorly on noisy and sparse \ac{fmcw} radar point clouds. Our \ac{rio} method makes no assumptions on the environment and makes use of no other sensors than \ac{imu} and a lightweight millimeter-wave \ac{fmcw} radar.

\subsection{System Overview}\label{subsec:ekf_rio_single_ov} 
\begin{figure}[ht!]
  \centering 
  \includegraphics[trim={0 0 0 0},clip,width=1.0\linewidth]{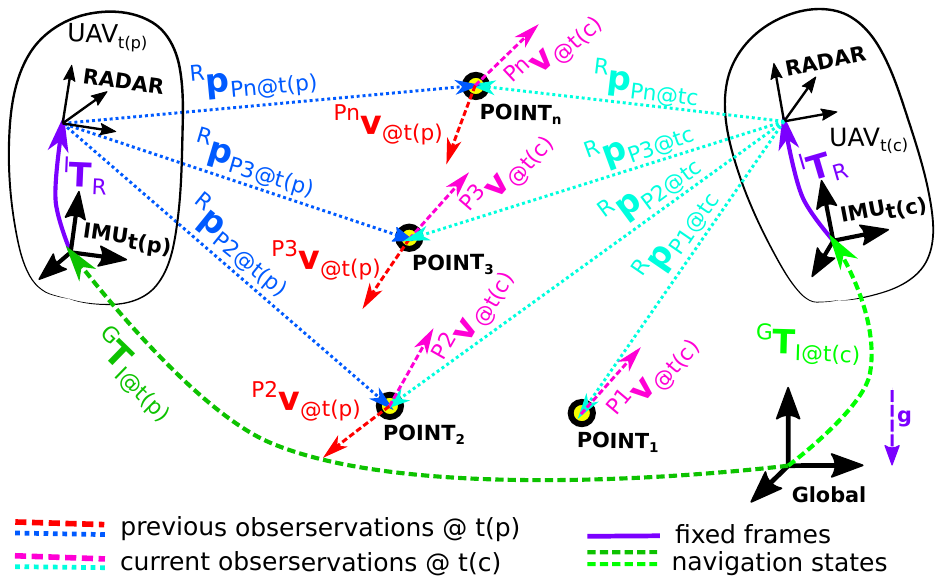}
  \caption[Spatial frames arrangement in the two-frame RIO method]{Two subsequent \ac{uav} poses are used in the distance and velocity measurement models. In the velocity measurement model, only the readings from the current pose are used.} \label{fig:rigid-body-configuration}
\end{figure}

In our \ac{rio} method we use error-state \ac{ekf} formulation~\cite{maybeck1979stochastic} in which \ac{imu} is a core sensor used for the system state propagation. Updates are performed with the \ac{fmcw} radar measurements, which provide both position and relative radial velocity of reflecting objects. Every time a radar measurement is taken, we augment the state of our \ac{ekf} estimator with the pose of the robot at which the measurement took place using stochastic cloning. Once a subsequent radar measurement is taken, we use the stored pose together with the current one in order to spatially align the radar scans and match the corresponding points across them. Distances to matches are used to form the residual vector in the \ac{ekf}.
Next, we use projections of the current robot velocity onto normal vectors to all points detected in the current radar scan together with their measured velocities to further augment the residual vector. Residual vectors are then used in the update step to estimate the mean of the error-state, which is injected into the regular state. The coordinate frames arrangement for measurements in our system is shown in~\cref{fig:rigid-body-configuration}.

\subsection{Radar-Inertial State Estimation} \label{subsec:ekf_rio_single_est}
The state vector $\vx$ in our filter is defined as follows:

\begin{equation}\label{eq:ekf_rio_single_states}
\begin{aligned}
    \vx &= \left[ \vx_{\cN}; \vx_{\cC} \right] \\
    &= \left[ \left[\reference{{\vp}}{G}{I}{}; \reference{\bar{\vq}}{G}{I}{}; \reference{{\vv}}{G}{I}{}; \vb_{\va}; \vb_{\bomega}\right]; \left[\reference{{\vp}}{G}{I_1}{}; \reference{\bar{\vq}}{G}{I_1}{} \right]\right]
\end{aligned}
\end{equation}

with the navigation state $\vx_{\cN}$ and the stochastic clone state $\vx_{\cC}$ of the \ac{imu} pose corresponding to the previous radar measurement as described later on in this section. The previous radar measurement is not part of the state vector. $\reference{{\vp}}{G}{I}{}$, $\reference{{\vv}}{G}{I}{}$, and $\reference{{\bar{\vq}}}{G}{I}{}$ are the position, velocity, and orientation of the \ac{imu}/body frame $\{I\}$  with respect to the navigation frame $\{G\}$, respectively. $\vb_{\bomega}$ and $\vb_{\va}$ are the measurement biases of the gyroscope and accelerometer, respectively. $\reference{{\vp}}{G}{I_1}{}$ and $\reference{\bar{\vq}}{G}{I_1}{}$ define the pose of the \ac{imu} frame $\{I_1\}$ corresponding to the last radar measurement with respect to the navigation frame $\{G\}$. We will use this frame later on in this section for ad-hoc point correspondence generation such that we do not need to keep 3D points in the state vector in order to use distance based measurements.

The evolution of the state is expressed by the following differential equations:
\begin{equation}\label{eq:ekf_rio_single_system_dynamics}
\begin{aligned}
    \reference{\dot{\vp}}{G}{I}{} &= \reference{\vv}{G}{I}{},\\ \reference{\dot{\vv}}{G}{I}{} &= \reference{\vR}{G}{I}{} \left( \reference{\va}{}{}{I}^{\#} - \vb_{\va} - \vn_{\va} \right) + \reference{\vg}{}{}{G}, \\
    \reference{\dot{\vR}}{G}{I}{} &= \reference{\vR}{G}{I}{}\skewmat{\reference{\bomega}{}{}{I}^{\#} - \vb_{\bomega} - \vn_{\bomega}}, \\ 
    \dot{\vb}_{\va} &= \vn_{\vb_{\va}},
    \dot{\vb}_{\bomega} = \vn_{\vb_{\bomega}}, \reference{\dot{\vp}}{G}{R}{} = \bZero, \reference{\dot{\vR}}{G}{R}{} = \bZero
\end{aligned}
\end{equation}
where $\reference{\va}{}{}{I}^{\#}$ and  $\reference{\bomega}{}{}{I}^{\#}$ are the accelerometer and gyroscope measurements of the \ac{imu} with a white measurement noise $\vn_{\va}$ and $\vn_{\bomega}$. $\vn_{\vb_{\va}} $ and $\vn_{\vb_{\bomega}}$ are assumed to be white Gaussian noise to model the bias change over time as a random process. The gravity vector is assumed to be aligned with the z-axis of the navigation frame $\reference{\vg}{}{}{G} = \left[ 0, 0, 9.81 \right]^\transpose$.

Since we use an error-state \ac{ekf} formulation we introduce the following error state vector from the states defined in~\cref{eq:ekf_rio_single_states}:

\begin{equation}\label{eq:ekf_rio_single_error_states}
\begin{aligned}
    \tilde{\vx} &= \left[ \tilde{\vx}_{\cN}; \tilde{\vx}_{\cC} \right] \\
    &= \left[ \left[ \reference{\tilde{\vp}}{G}{I}{}; \reference{\tilde{\btheta}}{\cG}{I}{}; \reference{\tilde{\vv}}{G}{I}{}; \tilde{\vb}_{\va}; \tilde{\vb}_{\bomega} \right]; \left[ \reference{\tilde{\vp}}{G}{I_1}{}; \reference{\tilde{\btheta}}{G}{I_1}{} \right] \right].
\end{aligned}
\end{equation}

For translational components, e.g., the position, the error is defined as $\reference{\tilde{\vp}}{G}{I}{} = \reference{\hat{\vp}}{G}{I}{} - \reference{{\vp}}{G}{I}{}$, while for rotations/quaternions it is defined as $\tilde{\bar{\vq}} = \hat{\bar{\vq}}^\inverse \otimes \bar{\vq} = \left[ 1; \frac{1}{2}\tilde{\btheta}\right]$, with $\otimes$ and $\tilde{\btheta}$ being quaternion product and small angle approximation, respectively.

\subsubsection{Stochastic Cloning}
In order to process relative measurements relating to estimates at different time instances, Roumeliotis and Burdick introduce the concept of Stochastic Cloning (SC) in~\cite{roumeliotis2002stochastic}. To appropriately consider the correlations/interdependencies between the estimates from different time instances, an identical copy of the required states and their uncertainties is used to augment the state vector and the corresponding error-state covariance matrix. Given the error-state definition in~\cref{eq:ekf_rio_single_error_states}, $\tilde{\vx}_{\cC}$ is defined as the error state of the stochastic clone of the \ac{imu} pose state ${\vx}_{\cI} = \left[\reference{{\vp}}{G}{I_1}{}; \reference{{\vq}}{G}{I_1}{}\right]$ and ${\vx}_{\cO} = \left[\reference{{\vv}}{G}{I}{}; \vb_{\va}; \vb_{\bomega} \right]$ are the other states of the navigation state. As cloned state is fully correlated with the \ac{imu} pose, it leads to the following  stacked/augmented covariance matrix of the corresponding error-state:

\begin{equation}\label{eq:ekf_rio_single_stacked_error_cov}
    \tilde{\vx} = \begin{bmatrix}
      \tilde{\vx}_{\cI} \\ \tilde{\vx}_{\cO} \\ \tilde{\vx}_{\cC}
    \end{bmatrix}, \bSigma = \begin{bmatrix}
      \bSigma_{\cI} & \bSigma_{\cI\cO} & \bSigma_{\cI} \\
      \bSigma_{\cO\cI} & \bSigma_{\cO} & \bSigma_{\cO\cI} \\
      \bSigma_{\cI} & \bSigma_{\cI\cO} & \bSigma_{\cC}
    \end{bmatrix}
\end{equation}

with $\bSigma_{\cN} = \begin{bmatrix}
      \bSigma_{\cI} & \bSigma_{\cI\cO} \\
      \bSigma_{\cO\cI} & \bSigma_{\cO}  
    \end{bmatrix} $ being the $15\times 15$ uncertainty of the navigation state $\tilde{\vx}_{\cN}$, and $\bSigma_{\cC} = \bSigma_{\cI}$ being the $6\times 6$ uncertainty of the cloned \ac{imu} pose error state $\tilde{\vx}_{\cI}$.
    
The cloned pose does not evolve with time, meaning no state transition (i.e., $\bPhi_{\cC}^{k+1|k} = \vI$) and no process noise (i.e., $\vG_{\cC}^{k+1|k} = \bZero$) is applied, while the original state estimate propagates as usual. From this, the error state propagation can be derived as 
\begin{equation}
    \begin{aligned}
    \tilde{\vx}^{k+1} &= \bPhi^{k+1|k} \tilde{\vx}^{k} + \vG^{k+1|k} \vw^{k}, \\
    \begin{bmatrix}  \tilde{\vx}_{\cN}^{k+1} \\ \tilde{\vx}_{\cC}^{k+1} \end{bmatrix} 
      &= \begin{bmatrix} \bPhi_{\cN}^{k+1|k} & \bZero \\ \bZero & \bPhi_{\cC}^{k+1|k} \end{bmatrix} \begin{bmatrix}  \tilde{\vx}_{\cN}^{k} \\ \tilde{\vx}_{\cC}^{k} \end{bmatrix} + \begin{bmatrix}  \vG_{\cN}^{k+1|k} \\ \vG_{\cC}^{k+1|k} \end{bmatrix} \vw^{k} \\
       &= \begin{bmatrix} \bPhi_{\cN}^{k+1|k} & \bZero \\ \bZero & \vI \end{bmatrix} \begin{bmatrix}  \tilde{\vx}_{\cN}^{k} \\ \tilde{\vx}_{\cC}^{k} \end{bmatrix} + \begin{bmatrix}  \vG_{\cN}^{k+1|k} \\ \bZero \end{bmatrix} \vw^{k}
    \end{aligned}        
\end{equation}
with the linearized state transition matrix $\bPhi$ and the linearized perturbation matrix $\vG$ computed as explained by Weiss in \cite{5979982} or related work.
The full error-state uncertainty of~\cref{eq:ekf_rio_single_stacked_error_cov} can then be propagated as

\begin{equation}\label{eq:ekf_rio_single_covprop}
    \begin{aligned}
    \bSigma^{k+1} &=  \bPhi^{k+1|k}\bSigma^{k}(\bPhi^{k+1|k})^\transpose +  \vG^{k+1|k} \vQ^{k} (\vG^{k+1|k})^\transpose \\
    &= \begin{bmatrix}
    \bSigma_{\cN}^{k+1}  & \bPhi_{\cN}^{k+1|k}\bSigma_{\cN\cC}^{k} \vI \\
    \vI \bSigma_{\cC\cN}^{k} (\bPhi_{\cN}^{k+1|k})^\transpose & \bSigma_{\cC}^{k}
    \end{bmatrix}
    \end{aligned}       
\end{equation}

with $\vI$ being the identity matrix (since cloned states do not evolve in time), $\vQ$ being the discretized process noise matrix, $\bSigma_{\cC\cN}^{k} = (\bSigma_{\cN\cC}^{k})^\transpose$ the cross-covariance between the navigation error-state and the stochastic clone error-state, and $\bPhi_{\cN}^{k+1|k}$ the error-state transition matrix of the navigation error-state $\tilde{\vx}_{\cN}$. This propagation allows us to rigorously reflect the cross-correlations between the cloned state and the evolved state in our error-state formulation. The above described formalism enables us to correctly use the state variables in order to align the previous radar scan to the current one prior to point matching.

\subsubsection{3D Point Matching}
In order to estimate the distance to detected points using our measurement model, we need to perform point matching between the current and the previous radar scan aligned to the current \ac{uav} pose. This is roughly following the idea of \cite{4209642} in order to avoid tracking 3D points in the state vector. With a single past pose in the state vector, we can find point correspondences in an ad-hoc fashion, as follows.

We base our point matching algorithm on work described in \cite{cen2018precise} and \cite{almalioglu2020milli} for 2D ground vehicle setups and extend it to our 3D \ac{uav} setting. Having two consecutive radar scans which are aligned using the pose information stored in the state vector, as the first step, we solve the linear sum assignment problem using the Munkres algorithm \cite{munkres1957algorithms}. We pose the problem as follows:

\begin{align}
    min\sum_{i}\sum_{j}\vC_{i,j}\vX_{i,j} 
    \label{eq:linsum}
\end{align}

Where $\vX$ is a boolean matrix where $\vX_{i,j}=1$ iff row $i$ is assigned to column $j$. Constraints of the problem are such that each row is assigned to at most one column and each column to at most one row. Entries of the $\vC$ matrix are computed as Euclidean distances between all points $\reference{\vp}{R}{P}{}$ from a previous radar scan at time instance $t_{p}$ and from a current radar scan at $t_{c}$:

\begin{align}\label{eq:scorval_C}
    \vC_{i,j}= \| \referencet{\vp}{R}{P_{i}}{}{t_{c}} - \referencet{\vp}{R}{P_{j}}{}{t_{p}} \| 
\end{align}

In the second step, using the proposed potential matches from the previous step, we build a matrix $\vS$ of scores where each entry is computed as:

\begin{align}\label{eq:scorval_s}
    s_{i,j}= \frac{1}{1 + \| \referencet{\vp}{R}{P_{i}}{}{t_{c}} - \referencet{\vp}{R}{P_{j}}{}{t_{p}} \|}
\end{align}

unless the value of reflection intensity in the current scan is below a certain threshold or the  Euclidean distance is above certain maximum threshold. If either of the two aforementioned conditions holds true, the entry is set to $s_{i,j}=0$. In the third step, a greedy search is performed on the pairs of points whose corresponding entries in $\vS$ are non-zero. If a point in the previous scan has more than one candidate for a match in the current scan, then, from among the candidates, we choose the one which minimizes the following expression:

\begin{align}\label{eq:relmin}
    d_{i,j}= | \sum_{k}\|\referencet{\vp}{R}{P_{i}}{}{t_{c}} - \referencet{\vp}{R}{P_{k}}{}{t_{c}}\| - \sum_{k}\|\referencet{\vp}{R}{P_{j}}{}{t_{p}} - \referencet{\vp}{R}{P_{k}}{}{t_{p}}\||
\end{align}

\begin{figure}[h!tbp]
  \centering
  \includegraphics[width=1.\columnwidth]{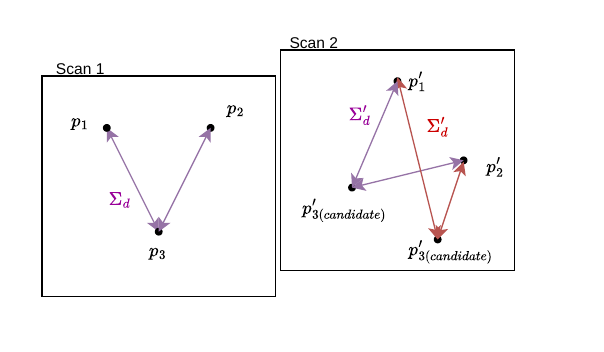}
  \caption[Relative coherence between 3D points in subsequent scans]{Points in subsequent scans should exhibit relative geometrical coherence. We exploit this observation (also used in \cite{cen2018precise, petillot}) to guide the point correspondences search. Intuitively, if there are to candidates for a match, then its distance to some neighboring already matched points in the first scan should be close to the its distance to the matches of those points in the second scan.}
  \label{fig:rel_matchings}
\end{figure}

\begin{figure}[h!tbp]
  \centering
  \includegraphics[width=1.\columnwidth]{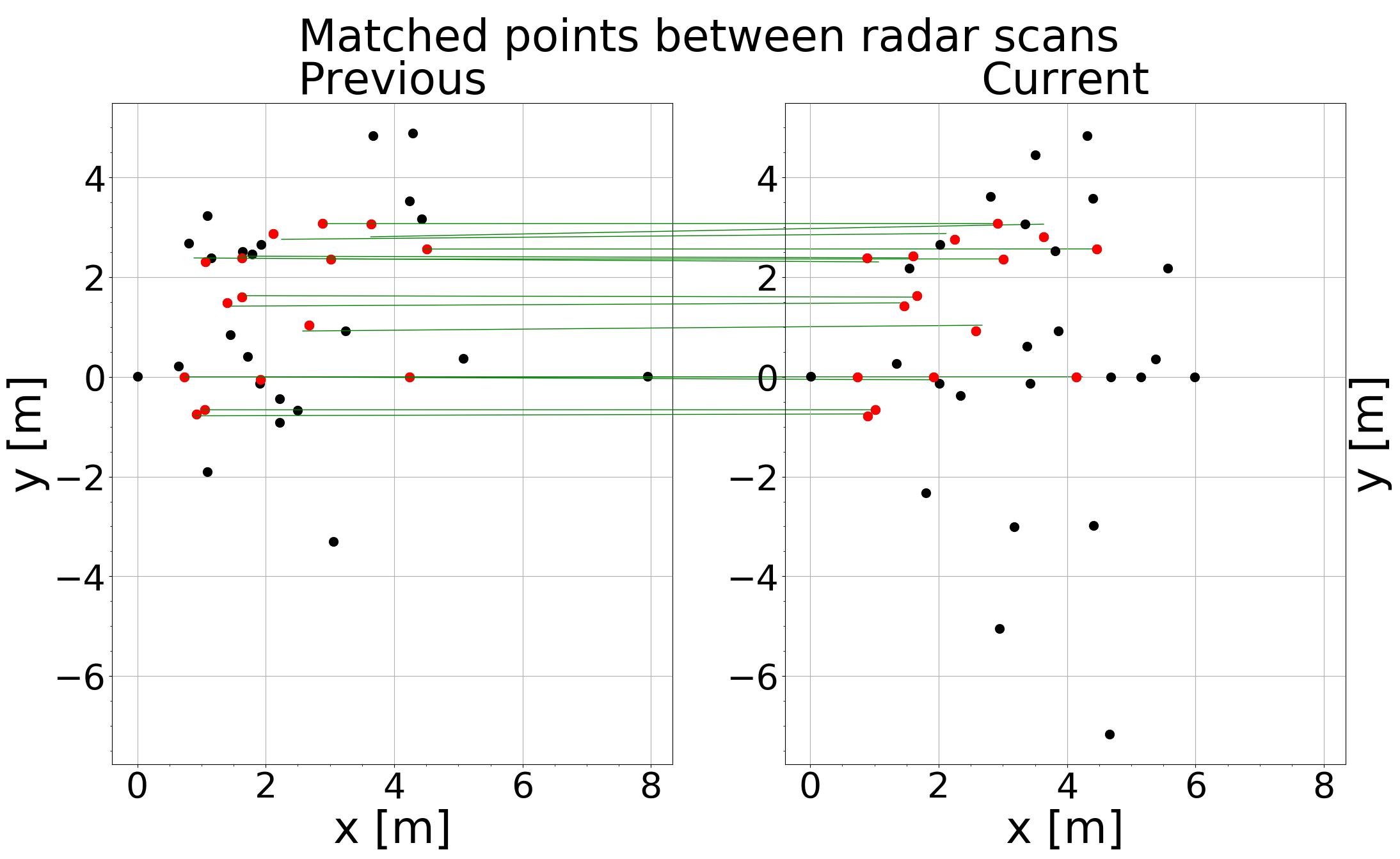}
  \caption[Matched 3D points between two scans]{Matched points between subsequent scans projected onto the xy plane. Our matching algorithm proves to be robust in the face of outliers and sparse, noisy 3D point clouds.}
  \label{fig:matchings}
\end{figure}

Where $\referencet{\vp}{R}{P_{k}}{}{\{t_{c}, t_{p}\}}$ are already matched points in the current and previous scans respectively. This idea has been exploited in \cite{cen2018precise} and relies on the fact that subsequent radar scans should ideally keep the relative arrangements between the constituting points (see~\cref{fig:rel_matchings}).
The result of this 3D point matching part is a set of 3D point correspondences (see~\cref{fig:matchings}) between a previous and the current radar scan. We can then compare the current radar measurements with the estimated distance value computed from the previous radar scan and the estimated state variables as described in the following section.

\subsubsection{Distance Measurement Model}
In order to estimate the distances to the matched 3D points in the current scan, we transform the corresponding 3D matches $\referencet{\vp}{R}{\cP_{j}}{}{t_{p}}$ from the previous radar scan at time instance $t_{p}$ to the current radar reference frame, considering the robot's spatial evolution:
\begin{equation}
\begin{aligned}\label{eq:ekf_rio_single_propagated_distance}
  \referencet{{\vp'}}{R}{P_{j}}{}{t_{p}} = & \referencet{\vR}{I}{R}{}{\transpose} \left( -\reference{\vp}{I}{R}{} + (\referencet{\vR}{G}{I}{}{t_c})^\transpose \left(-\referencet{\vp}{G}{I}{}{t_{c}} + \right.  \right. \\
    & \left. \left. \referencet{\vp}{G}{I}{}{t_{p}} + \referencet{\vR}{G}{I}{}{t_{p}} \left( \reference{\vp}{I}{R}{} + \reference{\vR}{I}{R}{} \referencet{\vp}{R}{P_{j}}{}{t_{p}} \right) \right) \right)
\end{aligned}
\end{equation}
where $\reference{\vR}{I}{R}{}$ and $\reference{\vp}{I}{R}{}$ is the constant pose (orientation and position) of the radar frame with respect to the \ac{imu} frame. $\referencet{\vR}{G}{I}{}{\{t_{c}, t_{p}\}}$ and $\referencet{\vp}{G}{I}{}{\{t_{c}, t_{p}\}}$ are the \ac{imu} orientation and position corresponding to the previous and current radar scans at $t_{p}$ and $t_{c}$, respectively, with respect to the navigation frame $\{G\}$.

Note that, at this point, we could already formulate a measurement for the matched 3D point in the past with the currently measured one. However, as mentioned in~\cref{chap:fund_radar}, low-cost \ac{fmcw} radars have fairly precise measurements of the object's distance and Doppler velocity, but heavily lack of precision in azimuth and elevation. Thus, we transform the 3D point from Cartesian space to Spherical coordinates and only use the most informative dimension, the distance. Additional measurement formulations for azimuth and elevation could be included with higher measurement uncertainty. The low information versus added complexity and the non-Gaussian noise distribution in these dimensions are, however, arguments to not include them in our \ac{rio} framework.

The estimated distance, which is compared to the current distance measurement, is calculated for each point as the norm of the transformed point from $t_{p}$:

\begin{equation}
\begin{aligned}\label{eq:ekf_rio_single_distmeasmodel}
    d_{P_{j}}= {} & \Big\| \referencet{{\vp'}}{R}{P_{j}}{}{t_{p}} \Big\|
\end{aligned}
\end{equation}

where $d_{P_{j}}$ is the distance to a single matched 3D point $\referencet{\vp'}{R}{P_{j}}{}{t_{p}}$ in the previous radar scan at $t_{p}$ aligned to the current radar pose at $t_{c}$. Since this measurement relates to states from pastime instances, stochastic cloning is necessary as introduced earlier in this section.

\subsubsection{Velocity Measurement Model}
In order to estimate the velocities of the detected radar 3D points $\reference{\vv}{R}{\cP_i}{}$ in the current scan at $t_{c}$, we transform the current robot ego-velocity from the \ac{imu} frame into the current radar frame and subsequently project it onto the direction vector pointing towards the corresponding 3D point. This is expressed by the following measurement model:

\begin{equation}
\begin{aligned}\label{eq:ekf_rio_single_velmeasmodel}
    \reference{\vv}{R}{P_i}{} =& {\frac{\vr^\transpose}{\|\vr\|}} \left( \referencet{\vR}{I}{R}{}{\transpose} \referencet{\vR}{G}{I}{}{\transpose} \reference{\vv}{G}{I}{} + \right. \\
    & \left. \referencet{\vR}{I}{R}{}{\transpose} \left( \reference{\bomega}{}{}{I} \times \reference{\vp}{I}{R}{}  \right)  \right)
\end{aligned}
\end{equation}

where $\vr = \reference{{\vp}}{R}{P_{i}}{} $ is the 3D point detected in the current scan,  $\reference{\bomega}{}{}{I}$ is the current angular velocity of the \ac{imu} in the \ac{imu} frame, and $\reference{\vv}{G}{I}{}$ is the current linear velocity of the \ac{imu} in the navigation frame. In order to reject outliers, we apply a chi-squared test to each measurement's residual, in which we check if the Mahalanobis distance corresponding to the residual is contained within the interval defined by the thresholds associated with a chosen percentile of the $\chi^{2}$ distribution.

\begin{sloppypar}
\subsubsection{Estimator Summary}\label{subsec:estimator_summary}
In summary, our \ac{ekf}-based \ac{rio} approach consists of~\cref{eq:ekf_rio_single_system_dynamics} and~\cref{eq:ekf_rio_single_covprop} to propagate the state and its covariance using the \ac{imu} measurements. We then use a tightly-coupled formulation to compare the distances of matched features with current radar distance measurements using~\cref{eq:ekf_rio_single_propagated_distance} and~\cref{eq:ekf_rio_single_distmeasmodel}, and also include in a tightly-coupled fashion the velocity information the radar sensor provides using~\cref{eq:ekf_rio_single_velmeasmodel} to correct \ac{imu} integration errors. The inclusion of both the point distance measurements and current point velocity information in a tightly-coupled fashion is key to the improved performance of our approach compared to state-of-the-art methods.

Although used in the position and velocity updates (\cref{eq:ekf_rio_single_propagated_distance} and~\cref{eq:ekf_rio_single_velmeasmodel}), we do not keep 3D points in the state vector. This idea is borrowed from \cite{4209642} where 3D points are triangulated from images on-the-fly without inclusion in the state vector. Our adaptation to \ac{rio} and highly simplified implementation of this idea suffers from reduced estimation consistency, but results in less complexity. A thorough analysis of the statistical impact of this simplified implementation can be tackled in future work.
\end{sloppypar}

\subsection{Experiments}\label{subsec:experiments}
The above described approach enables a simple, yet computationally efficient \ac{rio} method. In the following, we test our method on a real platform with real data.

\begin{sloppypar}
\subsubsection{Experimental Setup}
\begin{figure}[thpb]
  \centering
  \includegraphics[width=0.95\columnwidth]{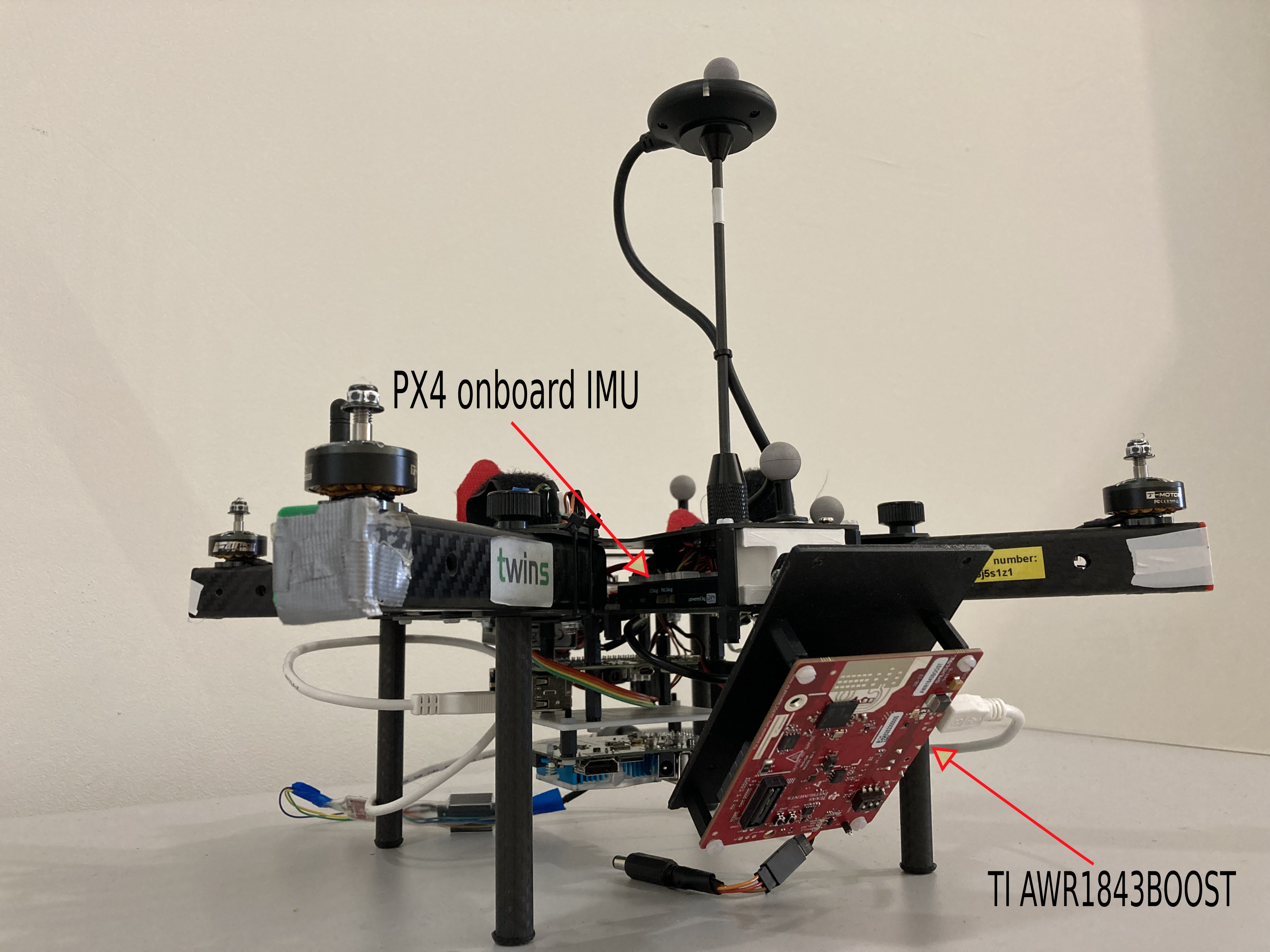}
  \caption[Experimental platform with TI radar mounted]{Experimental platform used in this work with the \ac{fmcw} radar sensor mounted in its custom-made housing tilted at 45\textdegree{} angle.}
  \label{fig:ekf_rio_single_platform}
\end{figure}

The sensor used for the experiments is the lightweight and inexpensive $\unit[77]{GHz}$ multichannel millimeter-wave \ac{fmcw} radar transceiver manufactured by Texas Instruments integrated on an evaluation board AWR1843BOOST, shown attached to the \ac{uav} in~\cref{fig:ekf_rio_single_platform}, equipped with a USB interface and powered with $\unit[2.5]{V}$. The frequency spectrum of chirps generated by the radar is between $f_l = \unit[77]{GHz}$ and $f_u = \unit[81]{GHz}$. The radio frequency (RF) signals propagate in a \ac{fov} of $\unit[120]{\degree}$ in azimuth and $\unit[30]{\degree}$ in elevation. Measurements are obtained at the rate of $f_{m} = \unit[20]{Hz}$. The radar is attached to one extremity of the experimental platform facing forward by a tilt of about $\unit[45]{\degree}$ with respect to the horizontal plane as shown in~\cref{fig:ekf_rio_single_platform}. This improves the velocity readings compared to nadir view while keeping point measurements on the ground and thus at a reasonable distance. For inertial measurements, we use the \ac{imu} of the Pixhawk 4 flight controller unit (FCU) with a sampling rate of $f_{si} = \unit[200]{Hz}$. We manually calibrate the transformation between the radar and \ac{imu} sensors, which is used as a constant spatial offset in the \ac{ekf}. The initial navigation states of the filter are set to the ground truth values with a random offset drawn from the states initial uncertainties as listed in~\cref{tab:S1-initial-values}. The above described platform is moved in a hand-held manner across a spacious room, as shown in~\cref{fig:ekf_rio_single_scene}, repeatedly performing five times the same rectangular-shaped trajectory of approximate dimensions of slightly more than $[\unit[4.5]{m} \times \unit[5.5]{m}]$ traversing the total distance of $\unit[116.4]{m}$. The scene was augmented with some arbitrary reflective clutter since the test environment was otherwise a clutter-less clean lab space. No position information from the added objects of any sort was used in our approach other than what the onboard radar sensor perceived by itself.
\end{sloppypar}

\bgroup
\def\arraystretch{1.5}
{
\begin{table}[t]
\caption[Initial standard deviation of the navigation states in the two-frame method experiments]{Initial standard deviation of the navigation states}
\label{tab:S1-initial-values}
\resizebox{1\linewidth}{!}{%
\begin{tabular}{llllll}
\cline{2-6}
\multicolumn{1}{l|}{}               & \multicolumn{1}{l|}{$\reference{\vp}{G}{I}{G}$} & \multicolumn{1}{l|}{$\reference{\vv}{G}{I}{G}$} & \multicolumn{1}{l|}{$\reference{\vq}{G}{I}{G}$} & \multicolumn{1}{l|}{$\reference{\vb_{a}}{}{}{I}$} & \multicolumn{1}{l|}{$\reference{\vb_{\omega}}{}{}{I}$} \\ \hline
\multicolumn{1}{|l|}{$\bsigma^{0}$} & \multicolumn{1}{l|}{$\unit[0.32]{cm}$}                 & \multicolumn{1}{l|}{$\unit[0.32]{cm/s}$}              & \multicolumn{1}{l|}{$\unit[0.1]{rad}$}             & \multicolumn{1}{l|}{$\unit[0.71]{m/s^2}$}           & \multicolumn{1}{l|}{$\unit[0.1]{rad/s}$} \\ \hline
\end{tabular}%
}
\end{table}
} 

\begin{figure}[thpb]
  \centering
  \includegraphics[width=0.85\columnwidth]{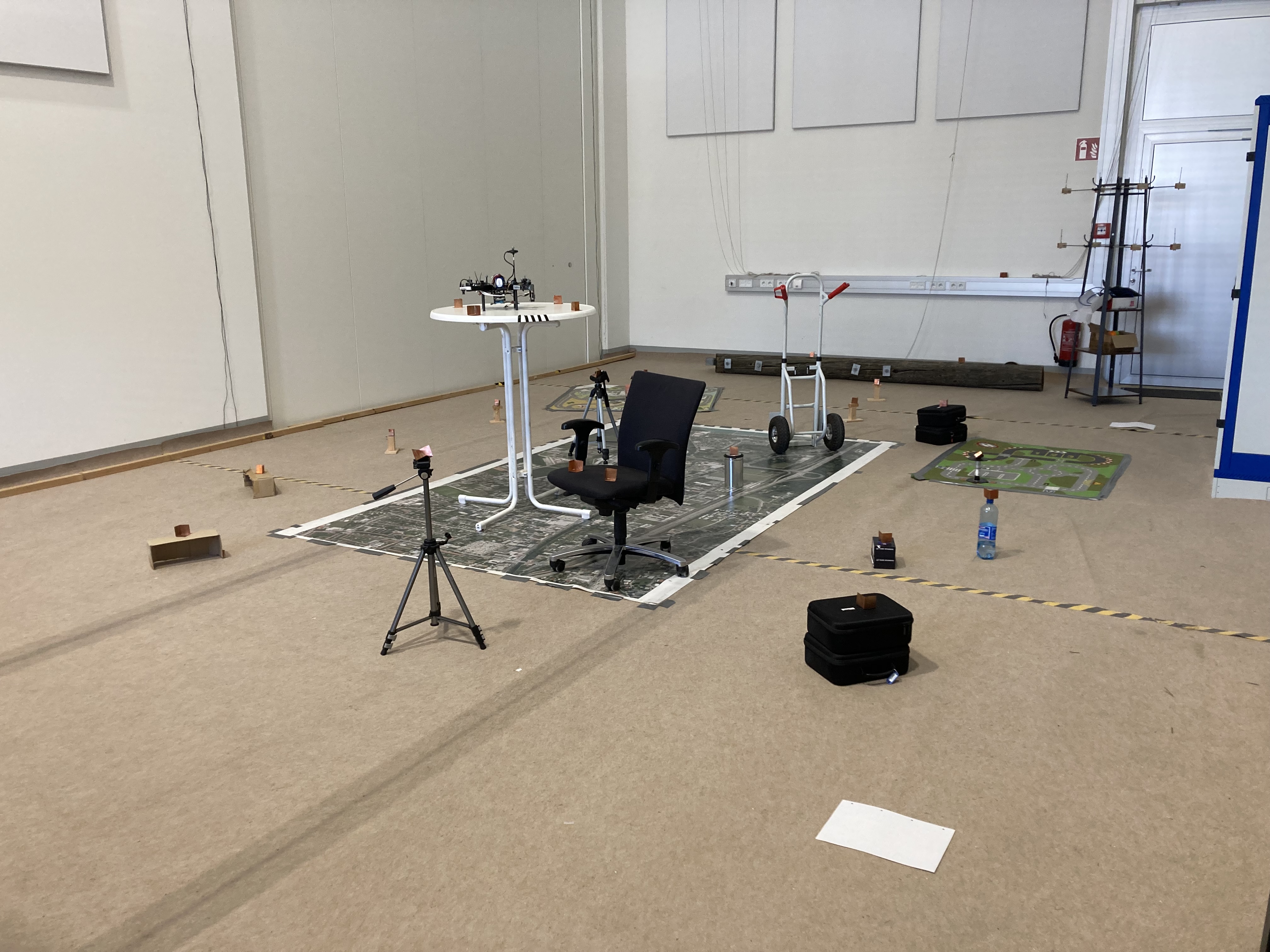}
  \caption[Indoor space used for experiments]{Indoor space where experiments were performed with reflecting clutter scattered on the scene. The objects were placed randomly and no global position (nor attitude) information of any sort was used in our approach.}
  \label{fig:ekf_rio_single_scene}
\end{figure}

We use a motion capture system to record the ground truth trajectories. During acquisition, we recorded sensor readings from the \ac{imu} and radar together with the poses of the \ac{uav} streamed by the motion capture system. Our \ac{ekf}-based \ac{rio} is executed offline but at real-time speed on the recorded sensor data on an Intel Core i7-10850H vPRO laptop with $\unit[16]{GB}$ RAM in a custom C++ framework.

\subsubsection{Evaluation}
We evaluate our \ac{rio} approach with the data recorded in an indoor space equipped with a motion capture system. Ground truth trajectories as well as the estimated ones can be seen in ~\cref{fig:ekf_rio_single_3dtop} to~\cref{fig:ekf_rio_single_3dside}. On these plots, one can observe drift of the estimate versus the ground truth in position and yaw, since these four dimensions are unobservable in a \ac{rio} framework. The amount of drift is a direct measure of quality for a given approach.

\begin{figure}[h!tbp]
  \centering
  \includegraphics[width=0.85\columnwidth]{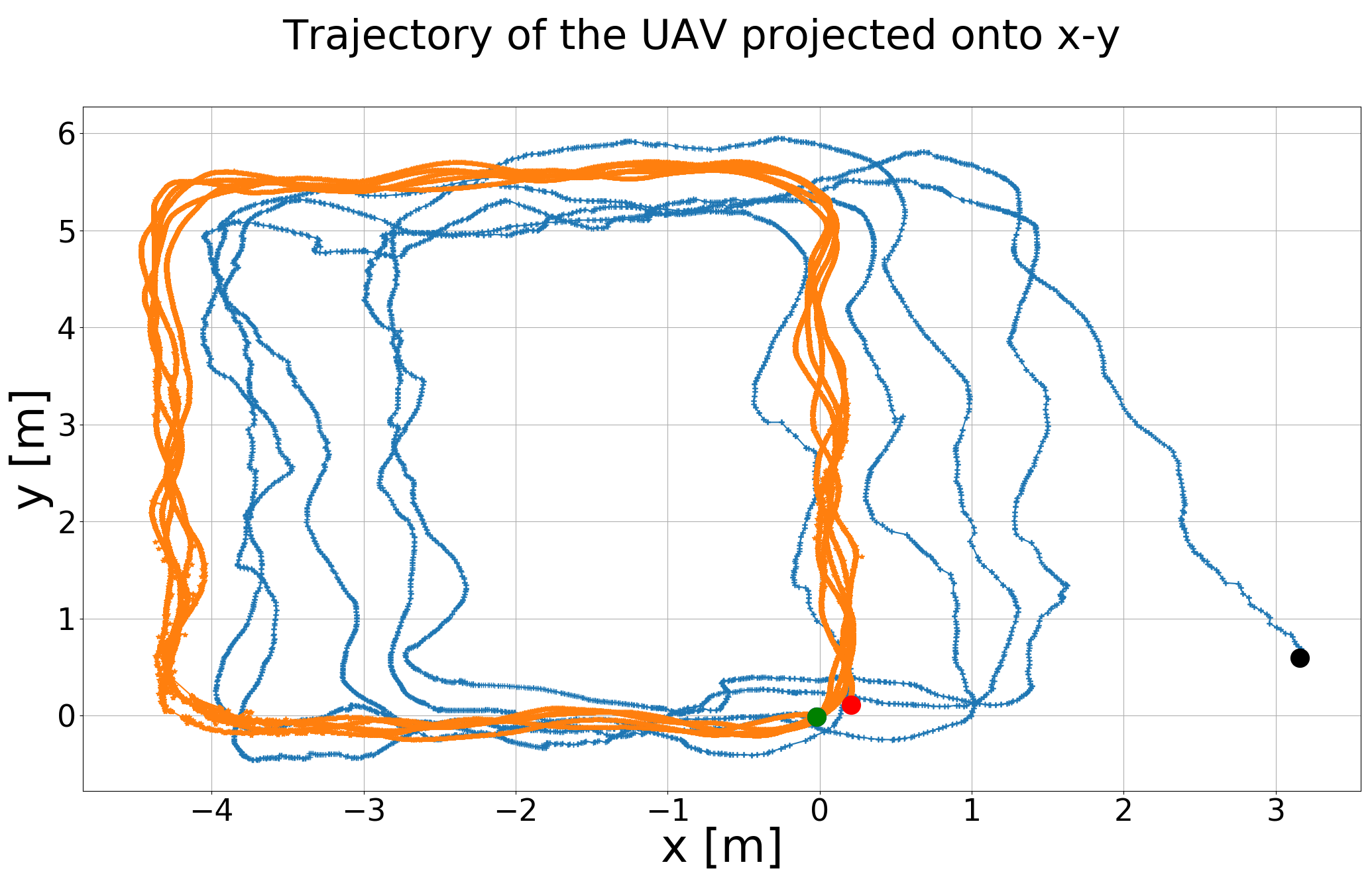}
  \caption[Top view of the \ac{uav} 3D trajectory]{Top view of the \ac{uav} 3D trajectory. The true trajectory is plotted in orange and the estimated one in blue. The biggest (angular) drift can be noted after the last turn. Note the coloured dots marking the end and the beginning of trajectories.}
  \label{fig:ekf_rio_single_3dtop}
\end{figure}

\begin{figure}[h!tbp]
  \centering
  \includegraphics[width=0.85\columnwidth]{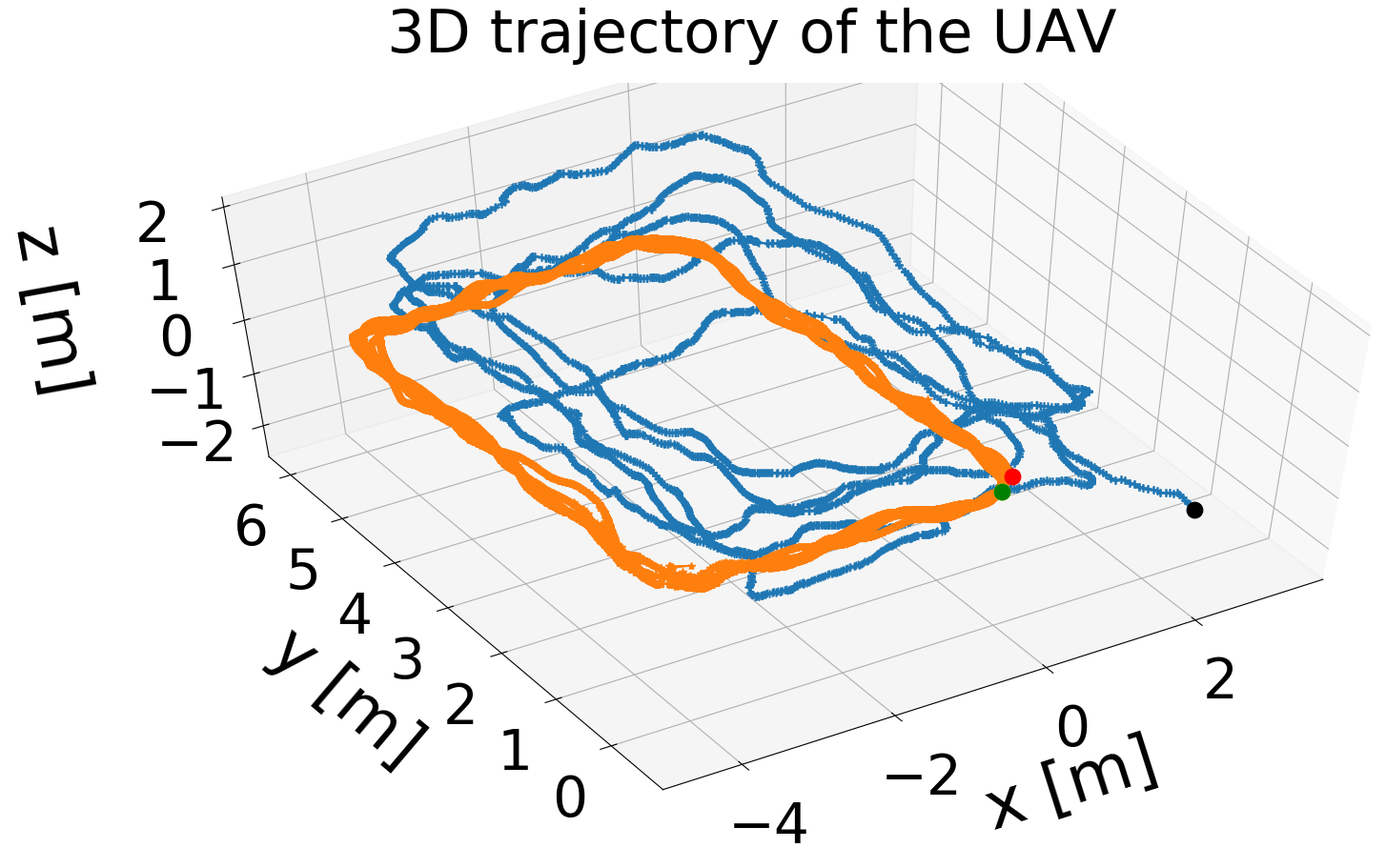}
  \caption[Side view of the \ac{uav} 3D trajectory]{Side view of the \ac{uav} 3D trajectory. The true trajectory is plotted in orange and the estimated one in blue. The total of $\unit[116.4]{m}$ distance is covered.}
  \label{fig:ekf_rio_single_3dside}
\end{figure}

Assessment of position, attitude, and velocity tracking is provided on plots~\cref{fig:ekf_rio_single_position},~\cref{fig:ekf_rio_single_attitude} and~\cref{fig:ekf_rio_single_velocity} respectively. For the position, one can clearly see the random walk behavior of this unobservable state. On this aspect, note that the provided metrics below are a snapshot of such a random walk (i.e., one of many realizations) -- as are the numbers in e.g., \cite{doer2020ekf}. Nevertheless, the vast improvement against state-of-the-art shows that our approach generally has some beneficial aspects. 

For the attitude, the drift in yaw is clearly visible. In this dataset we also observe an offset in pitch occurring after the first few seconds and remaining throughout the rest of the run. We assume a slight misalignment of the tracking system reference frame with respect to gravity. This value is observable in a \ac{rio} framework and will converge towards a gravity aligned reference frame. 

The 3D velocity state is observable and the plot, besides the noticeable jitters also in the ground truth data, does not show a particularly unexpected behavior. The ground truth was computed by numerically differentiating the tracking system's position signal. 

\begin{figure}[h!tbp]
  \centering
  \includegraphics[width=0.95\columnwidth]{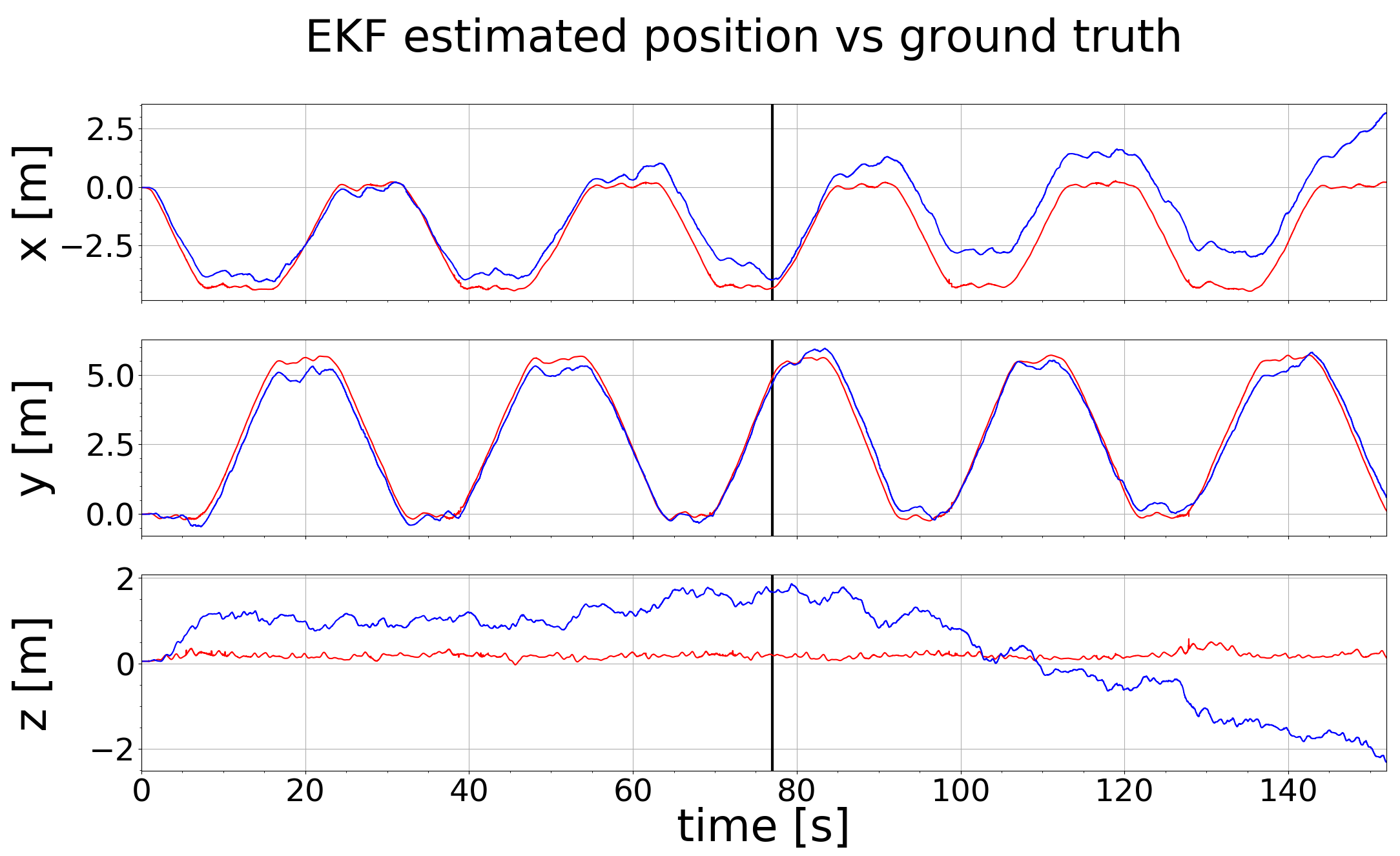}
  \caption[Estimated position in the approach with one past frame]{Estimated position plotted in blue against ground truth plotted in red. Note the thick black line passing through the time instant at which $\unit[60]{m}$ distance has been traversed (this is for better metrics comparison against state of the art detailed in the text).}
  \label{fig:ekf_rio_single_position}
\end{figure}

\begin{figure}[h!tbp]
  \centering
  \includegraphics[width=0.95\columnwidth]{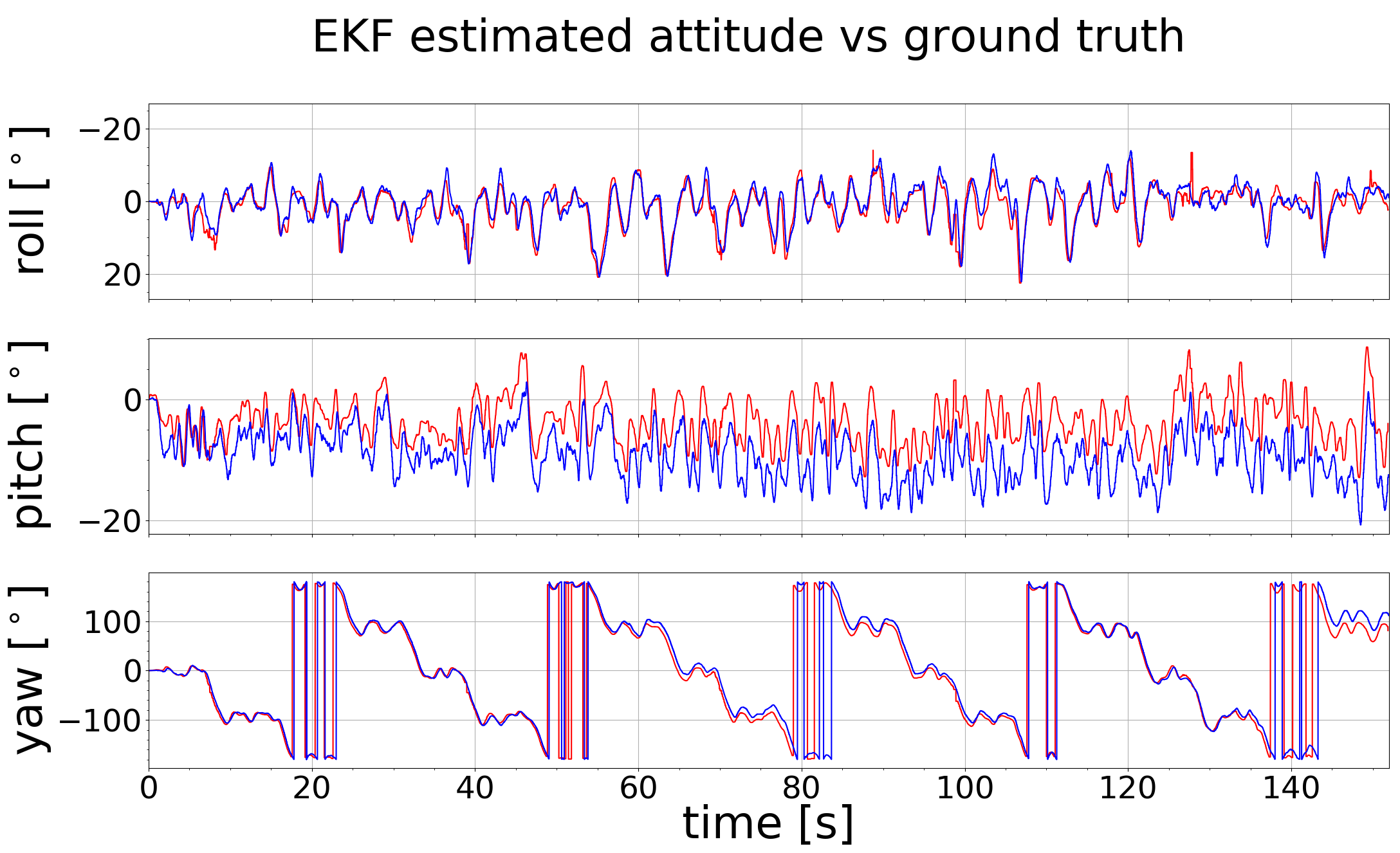}
  \caption[Estimated attitude in the approach with one past frame]{Estimated attitude plotted in blue against ground truth plotted in red. The offset in pitch angle is assumed to be caused by a slight misalignment of the motion capture system's reference frame and the gravity.}
  \label{fig:ekf_rio_single_attitude}
\end{figure}

\begin{figure}[h!tbp]
  \centering
  \includegraphics[width=0.95\columnwidth]{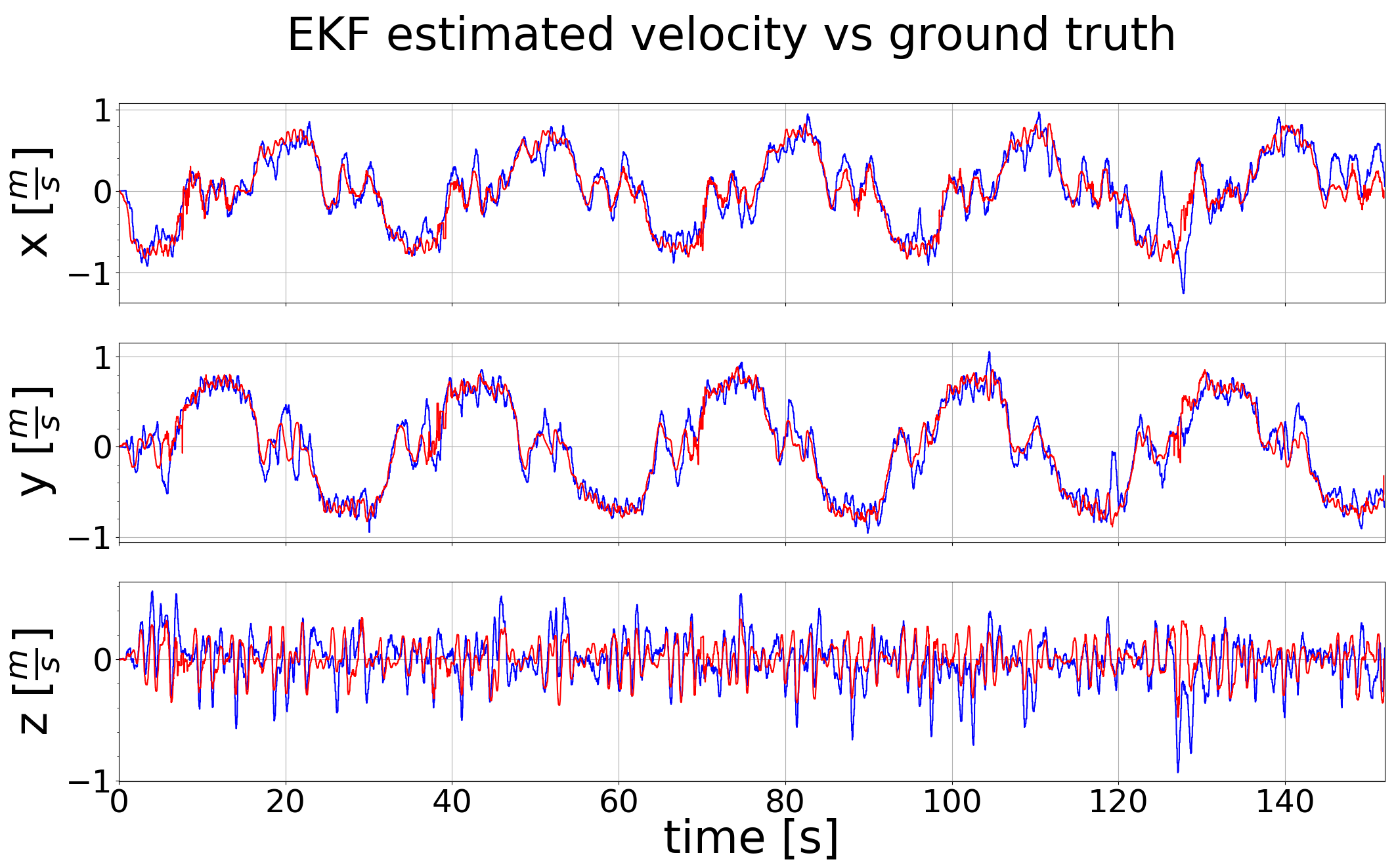}
  \caption[Estimated velocity in the approach with one past frame]{Estimated velocity plotted in blue against ground truth plotted in red. We compute the ground truth velocity by numerically differentiating the motion capture system's position.}
  \label{fig:ekf_rio_single_velocity}
\end{figure}

On~\cref{fig:ekf_rio_single_drift} we plot the drift from the true trajectory as percent of the traveled distance, drift in meters (norm of the position error) and the \ac{ttd}. 

\begin{figure}[h!tbp]
  \centering
  \includegraphics[width=0.95\columnwidth]{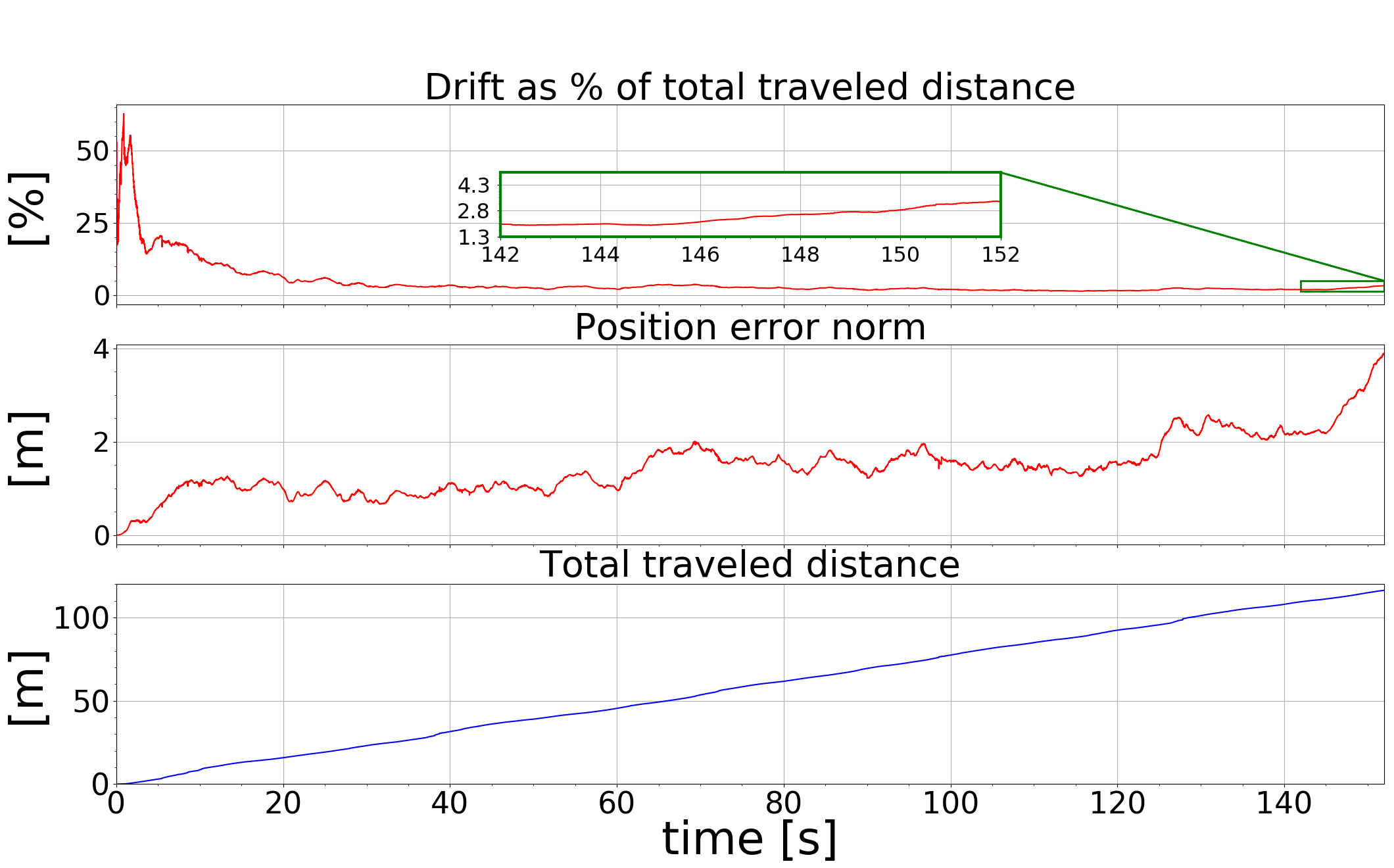}
  \caption[Drift over the TTD in the approach with one past frame]{Drift as $\unit[]{\%}$ of the \ac{ttd}, position error norm and the \ac{ttd}. The drift (position error norm) remains below $\unit[4]{m}$ throughout the executed trajectory and reaches $\unit[3.32]{\%}$ at the end of the \ac{ttd}.}
  \label{fig:ekf_rio_single_drift}
\end{figure}

The first $t_{init}=\unit[8]{s}$ of the experiment was the initialization period during which agile motions were performed in-place in order to initialize the filter before executing the trajectory. All values are plotted after this initialization phase. We choose to compare our work to the state-of-the-art approach in \cite{doer2020ekf} in which the authors use the same radar sensor to provide metrics for their loosely coupled radar-velocity based \ac{rio} approach using indoor hand-held data. In the following, we show the same metrics underlining the benefit of a) tight coupling and b) use of both distance and velocity information from the radar sensor. Note that in the same work, the authors also included a barometric pressure sensor in their \ac{rio} to further reduce drift. This would, however, be a different comparison.

Firstly, we evaluate final drift values. We can note in the~\cref{fig:ekf_rio_single_drift} that over the \ac{ttd} of $\unit[116.4]{m}$ we achieve a final drift below $\unit[4]{m}$. Authors in \cite{doer2020ekf} in their hand-held experiment achieve drift close to $\unit[4]{m}$ (or about $\unit[6.67]{\%}$) over the trajectory of $\unit[60]{m}$, which is roughly half of the distance covered in our experiment. Exact numbers for our approach are $\unit[3.86]{m}$ ($\unit[3.32]{\%}$ for \ac{ttd} equal to $\unit[116.4]{m}$) and $\unit[1.54]{m}$ ($\unit[2.56]{\%}$ for \ac{ttd} equal to $\unit[60]{m}$). Next, we compare the norm of \ac{mae} of position and velocity for \ac{ttd} of $\unit[60]{m}$. For our approach, these numbers are $\unit[1.05]{m}$ and $\unit[0.38]{\frac{m}{s}}$ respectively, against $\unit[1.95]{m}$ and $\unit[0.14]{\frac{m}{s}}$ for authors in \cite{doer2020ekf}. At the \ac{ttd} of $\unit[116.4]{m}$ the norm of \ac{mae} of position equals to $\unit[1.36]{m}$ and is still lower than the one from \cite{doer2020ekf} at $\unit[60]{m}$ despite the inherent drift of the unobservable position in \ac{rio} approaches. The norm of \ac{mae} of velocity for our method at \ac{ttd} of $\unit[116.4]{m}$ equals to $\unit[0.36]{\frac{m}{s}}$. The higher \ac{mae} in velocity compared to state of the art is an interesting fact: the velocity states are fully observable and should converge well. For ground truth velocity generation, we used the tracking system's position at $\unit[100]{Hz}$ and used a median filtered, delta-time scaled difference of this signal. This has highly limited precision and may have contributed to the relatively large \ac{mae} in velocity.

\subsection{Conclusions}\label{subsec:ekf_rio_single_conclusion}
In this section we presented a tightly-coupled \ac{ekf} based approach to \ac{rio} in which we fuse \ac{imu} readings with both velocity and distance measurements to 3D points detected by lightweight and inexpensive \ac{fmcw} radar. We make use of the past IMU pose and a rigid body assumption such that we can generate several point correspondences between two radar scans in an ad-hoc fashion. This requires only maintaining (through stochastic cloning) a 6DoF past pose in the state vector compared to tracking many 3D point vectors. In this context, we showed that our improved matching method can cope with noisy and sparse radar point clouds and generate reliably point correspondences between two scans. With this, we showed that using both distance and velocity measurements, we accomplish accurate 6D pose and 3D velocity estimation of a mobile platform (in this case a hand-held \ac{uav}). In particular, the use of the accurately measured distance information to a point is beneficial as it does not include the point's azimuth and elevation angular information which is generally very poorly measured by low-cost \ac{fmcw} radar. Moreover, we showed that using our method, we reduce the position drift compared to similar state-of-the-art approach. Last but not least, our method is applicable in a variety of environments where potentially \ac{gnss} systems are unavailable and vision sensors, commonly used for \ac{uav} navigation, cannot be relied upon. 

The approach to \ac{rio} shown in this section lends itself to be extended with powerful concepts from the camera-based state estimation, like exploiting multi-state constraints in the update step, as well as tracking the history of the detected 3D points in order to classify the ones consistently observed as persistent features. These aspects will be explored in the following sections.


\section{RIO Multi-State EKF using distance, Doppler velocity measurements of 3D points,  and persistent landmarks}\label{sec:ekf_rio_multi}
Building upon the method developed in the previous section, in this section, we show an approach which significantly reduces the final drift (by a factor of 4 in average) and increases the accuracy (in terms of \ac{mae} by a factor of 2) of \ac{rio} compared to the single-frame approach described in the previous section (and in \cite{previous_iros}) and demonstrate it in real flights. To this end, we employ the stochastic cloning \cite{roumeliotis2002stochastic} for augmenting the state with a chosen number of past robot poses and corresponding radar scans (3D point clouds) in a \ac{fifo} buffer from which measurement trails are constructed. Trails matched consistently over a given amount of time are promoted to persistent landmarks and added to the state vector. In addition to these distance measurements to landmarks and trails, we also use Doppler velocity of points from the current radar scan. We fuse all measurements in a tightly-coupled formulation in our \ac{ekf} setup. The tight coupling permits the integration of single distance and velocity measurements during update steps. This property obviates any limitations on required minimal number of matches (as it is e.g., needed for a prior \ac{icp} and subsequent loose coupling of the resulting delta-pose in the \ac{ekf}). This is a particularly strong advantage in view of robustness and accuracy over loosely coupled approaches since, e.g., \ac{icp} \cite{121791} works poorly on noisy and sparse \ac{fmcw} radar point clouds. Note that our \ac{rio} method makes no assumptions on the environment and makes use of no other sensors than \ac{imu} and a lightweight millimeter-wave \ac{fmcw} radar providing sparse and noisy 3D point clouds along with Doppler velocities of the detected points. It is suitable for a \ac{uav} and real-time capable.

Note that the leveraged techniques from the vision community \cite{MSCKF-Paper}, the sparse and noisy radar 3D (versus 2D in vision) measurements require important enhancements. This includes a different definition and treatment of 3D instead of 2D trails, a measurement definition along the most precise dimension of the sensor (i.e., radial distance), and the inclusion of Doppler velocity measurements. On the other hand, the multi-state approach for radar inherently handles hovering situations where the 2D vision measurements require special treatment.

\subsection{System Overview}\label{subsec:ekf_rio_multi_overview} 
Our \ac{rio} method is based on an error-state \ac{ekf} formulation~\cite{maybeck1979stochastic} which uses an \ac{imu} as the primary sensor for the state propagation. Updates are performed with the \ac{fmcw} radar measurements, which consist of sparse and noisy 3D pointclouds and relative radial velocities of detected points. The principle of sensing of the radar we use is explained in~\cref{chap:fund_radar}. Every time a radar measurement is obtained, we augment the state of our \ac{ekf} filter with the pose of the robot at which the measurement took place using stochastic cloning as described in~\cref{subsec:ekf_rio_multi_est}. New poses are appended to the buffer of past poses in a \ac{fifo} fashion. The maximum number of cloned poses is defined by the parameter $N$. From taken measurements, we construct and maintain a set of trails which record the continuous detectability of 3D points by the radar sensor, which have maximum trail length $N$. Meaning that, every matched point keeps a history of its detected positions and every element of this history refers to a cloned robot pose at which the detection was taken. Such a point with a history of detections is referred to as a trail and it is kept in memory as long as it is actively matched to a point in the current scan following our sparse radar point cloud matching method described in the~\cref{subsec:ekf_rio_single_ov}. If it is not matched, it is inactive and thus removed. For 3D point matching, we use the latest detection of each trail along with the robot pose at which it was taken, together with the pose at which the current scan was taken. Specifically, we use these two poses to spatially align the current radar scan with the trails and trigger the matching procedure on such aligned matches. Once matched, the whole trail history is used to form the residual vector in the \ac{ekf} update.

It is important to note that as opposed to the way the features seen from multiple sensor poses are treated in \cite{MSCKF-Paper}, in our method we can continuously use all matched trails in every update step as long as they are visible, not merely once. That is because the radar sensor directly gives 3D points as measurements, hence no triangulation step is needed, which makes feature position calculations dependent on the state. Also, given the sparse nature of measurements emitted by the \ac{soc} \ac{fmcw} radar sensor employed in our method, it is not necessary to carry out measurement compression with the QR decomposition as in \cite{MSCKF-Paper}.

Next, we use projections of the current robot velocity onto normal vectors to all points detected in the current radar scan together with their measured velocities to further augment the residual vector. The final component of the residual vector comes from using persistent landmarks. Namely, in the case a trail has been continuously seen for $N$ times, it is removed from the set of trails, added to the state vector as persistent landmark and matched to the detections. Residual vectors are then used in the update step to estimate the mean of the error-state, which is injected into the regular state. The coordinate frames arrangement for measurements in our system is shown in~\cref{fig:ekf_rio_multi_rigid_body_configuration}.

\begin{figure}[]
  \centering 
  \includegraphics[trim={10 5 20 7},clip,width=0.55\columnwidth]{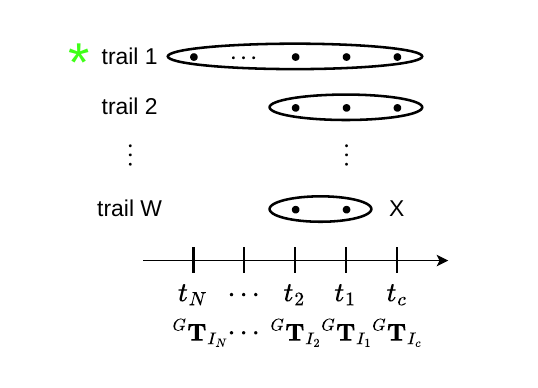}
  \caption[Measurement trails used in multi-frame approach]{Measurement trails used to record a history of detections. The green star marks a trail that was classified as a persistent landmark. 
  X means that the trail W has not been matched to any point in the current scan at $t_{c}$ and will thus be deleted. Dot in each trail is a single past detection. As shown below time instant $t$, points in the trail have associated cloned robot poses at which the measurements were taken.} \label{fig:ekf_rio_multi_trails_mng}
\end{figure}
\begin{figure}[]
  \centering 
  \includegraphics[trim={0 0 0 0},clip,width=1.0\linewidth]{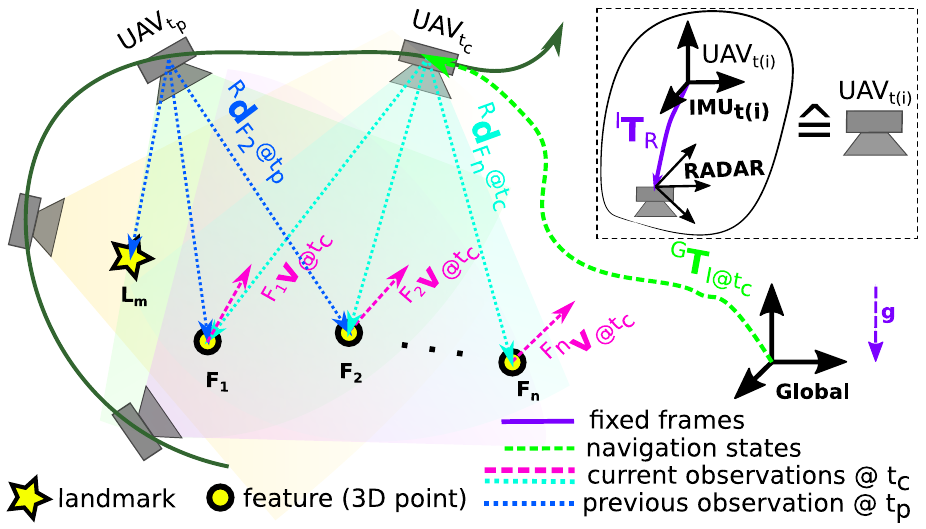}
  \caption[Spatial arrangement of frames in the multi-state EKF RIO]{Multiple consecutive past \ac{uav} poses are used in the distance measurements models employing persistent landmarks and trails of measurements. In the velocity measurement model, only the readings from the current pose are used.} \label{fig:ekf_rio_multi_rigid_body_configuration}
\end{figure}

\subsection{Multi-State Radar-Inertial State Estimation With Persistent Landmarks}\label{subsec:ekf_rio_multi_est}
The state vector $\vx$ in our filter is defined as follows:

\begin{equation}\label{eq:ekf_rio_multi_states}
\begin{aligned}
    \vx =& \left[ \vx_{\cI}; \vx_{\cC}; \vx_{\cL} \right] =  \big[ \big.  \left[\reference{{\vp}}{G}{I}{}; \reference{\bar{\vq}}{G}{I}{}; \reference{{\vv}}{G}{I}{}; \vb_{\va}; \vb_{\bomega}\right];\\
    & \left[\reference{{\vp}}{G}{I_{1}}{}; \reference{\bar{\vq}}{G}{I_{1}}{}; \ldots ;  \reference{{\vp}}{G}{I_{N}}{}; \reference{\bar{\vq}}{G}{I_{N}}{}\right]; \left[ \reference{{\vp}}{G}{L_{1}}{}; \ldots ; \reference{{\vp}}{G}{L_{M}}{} \right] \big. \big] 
\end{aligned}
\end{equation}

with the \ac{imu} state $\vx_{\cI}$, the stochastically cloned states $\vx_{\cC}$ of the \ac{imu} poses corresponding to the previous radar measurements as described later on, and the set of persistent landmarks $\vx_{\cL}$. The previous radar measurements (point cloud of reflecting objects and their Doppler velocities) are not part of the state vector. $\reference{{\vp}}{G}{I}{}$, $\reference{{\vv}}{G}{I}{}$, and $\reference{{\bar{\vq}}}{G}{I}{}$ are the position, velocity, and orientation of the \ac{imu}/body frame $\{I\}$  with respect to the navigation frame $\{G\}$, respectively. $\vb_{\bomega}$ and $\vb_{\va}$ are the measurement biases of the gyroscope and accelerometer, respectively. $\left[\reference{{\vp}}{G}{I_{n}}{}, \reference{\bar{\vq}}{G}{I_{n}}{}\right]$ with  $n=1, \ldots, N$ define a set of past \ac{imu} poses with respect to the navigation frame $\{G\}$ at the moments of past radar measurements. $\reference{{\vp}}{G}{L_{m}}{}$ with $m=1, \ldots, M$ define the position of persistent landmarks $\cL$ with respect to the navigation frame $\{G\}$. We use $\left[\reference{{\vp}}{G}{I_1}{}, \reference{\bar{\vq}}{G}{I_{1}}{}\right]$ (corresponding to the newest coordinates of the trails) for ad-hoc point correspondence generation such that we do not need to keep 3D points in the state vector in order to use distance based measurements.

The evolution of the state is expressed by the following differential equations:

\begin{equation}\label{eq:ekf_rio_multi_system_dynamics}
\begin{aligned}
    \reference{\dot{\vp}}{G}{I}{} &= \reference{\vv}{G}{I}{},\\ \reference{\dot{\vv}}{G}{I}{} &= \reference{\vR}{G}{I}{} \left( \reference{\va}{}{}{I}^{\#} - \vb_{\va} - \vn_{\va} \right) + \reference{\vg}{}{}{G}, \\
    \reference{\dot{\vR}}{G}{I}{} &= \reference{\vR}{G}{I}{}\skewmat{\reference{\bomega}{}{}{I}^{\#} - \vb_{\bomega} - \vn_{\bomega}}, \\ 
    \dot{\vb}_{\va} &= \vn_{\vb_{\va}},
    \dot{\vb}_{\bomega} = \vn_{\vb_{\bomega}}, \reference{\dot{\vp}}{G}{I_n}{} = \bZero, \reference{\dot{\vR}}{G}{I_n}{} = \bZero, \\
    \reference{\dot{\vp}}{G}{L_m}{} &= \bZero
\end{aligned}
\end{equation}

where $n=1, .., N$ refers to the most recent past \ac{imu} poses which are not changing in time, $m=1, .., M$ refer to $M$ most recent estimated positions of landmarks, $\reference{\va}{}{}{I}^{\#}$ and $\reference{\bomega}{}{}{I}^{\#}$ are the accelerometer and gyroscope measurements of the \ac{imu} with a white measurement noise $\vn_{\va} $ and $\vn_{\bomega}$. $\vn_{\vb_{\va}} $ and $\vn_{\vb_{\bomega}}$ are assumed to be white Gaussian noise to model the bias change over time as a random process. The gravity vector is assumed to be aligned with the z-axis of the navigation frame $\reference{\vg}{}{}{G} = \left[ 0, 0, 9.81 \right]^\transpose$.

Since we use an error-state \ac{ekf} formulation we introduce the error state vector from the states defined in \cref{eq:ekf_rio_multi_states}:

\begin{equation}\label{eq:ekf_rio_multi_error_states}
\begin{aligned}
    \tilde{\vx} =& \left[ \tilde{\vx}_{\cI}; \tilde{\vx}_{\cC} ; \tilde{\vx}_{\cL}\right] = \Big[ \Big. \left[ \reference{\tilde{\vp}}{G}{I}{}; \reference{\tilde{\btheta}}{G}{I}{}; \reference{\tilde{\vv}}{G}{I}{}; \tilde{\vb}_{\va}; \tilde{\vb}_{\bomega} \right];\\
    & \left[ \reference{\tilde{\vp}}{G}{I_{1}}{}; \reference{\tilde{\btheta}}{G}{I_{1}}{}; \ldots; \reference{\tilde{\vp}}{G}{I_{N}}{}; \reference{\tilde{\btheta}}{G}{I_{N}}{} \right]; \left[ \reference{\tilde{\vp}}{G}{L_{1}}{}; ...; \reference{\tilde{\vp}}{G}{L_{M}}{} \right]  \Big. \Big]
\end{aligned}
\end{equation}

For translational components, e.g., the position, the error is defined as $\reference{\tilde{\vp}}{G}{I}{} = \reference{\hat{\vp}}{G}{I}{} - \reference{{\vp}}{G}{I}{} $, while for rotations/quaternions it is defined as $\tilde{\bar{\vq}} = \hat{\bar{\vq}}^\inverse  \otimes \bar{\vq} = \left[ 1;  \frac{1}{2}\tilde{\btheta}\right]$, with $\otimes$ and $\tilde{\btheta}$ being quaternion product and small angle approximation, respectively.

\subsubsection{State Augmentation}
In order to process relative measurements relating to estimates at different time instances, Roumeliotis and Burdick introduce the concept of Stochastic Cloning (SC) in~\cite{roumeliotis2002stochastic}. To appropriately consider the correlations/interdependencies between the estimates from different time instances, an identical copy of the required states and their uncertainties is used to augment the state vector and the corresponding error-state covariance matrix. Given the error-state definition in~\cref{eq:ekf_rio_multi_error_states}, $\tilde{\vx}_{\cC_i}$ is defined as the error-state of the $i$-th stochastic clone of the \ac{imu} pose $\left[\reference{{\vp}}{G}{I_i}{}; \reference{{\vq}}{G}{I_i}{}\right]$. As the newest cloned state is fully correlated with the \ac{imu} pose and remaining cloned states are correlated with each other, it leads to the following  augmented covariance matrix of the corresponding error-state:

\begin{equation}\label{eq:ekf_rio_multi_state_mean}
    \tilde{\vx} = \left[ \tilde{\vx}_{\cI};  \tilde{\vx}_{\cC_{1}};  \ldots ; \tilde{\vx}_{\cC_{n}};  \tilde{\vx}_{\cL} \right],
\end{equation}
\begin{equation}\label{eq:ekf_rio_multi_stacked_covariance}
     \bSigma = \begin{bmatrix}
      \bSigma_{\cI}   & \bSigma_{\cI\cC_{1}}     & \cdots    & \bSigma_{\cI\cC_{n}}     & \bSigma_{\cI\cL}      \\
      \bullet       & \bSigma_{\cC_{1}}      & \cdots    & \bSigma_{\cC_{1}\cC_{n}} & \bSigma_{\cC_{1}\cL}  \\ 
      \bullet       & \bullet              & \ddots    & \vdots               & \vdots            \\
      \bullet       & \bullet              & \bullet   & \bSigma_{\cC_{n}}      & \bSigma_{\cC_{n}\cL}  \\
      \bullet       & \bullet              & \bullet   &  \bullet             & \bSigma_{\cL}       \\
    \end{bmatrix}
\end{equation}

with $\bSigma_{\cI} $ being the $15\times 15$ uncertainty of the \ac{imu} error-state $\tilde{\vx}_{\cI}$. $\bSigma_{\cL} $ is the $3M \times 3M$ uncertainty of a set of $M$ landmark error-states $\tilde{\vx}_{\cL} = \left[\tilde{\vx}_{L_1}; \ldots; \tilde{\vx}_{L_m} \right]$. $\bSigma_{\cC_1} = \bSigma_{\cI_{\{\tilde{\vp}, \tilde{\btheta}\}}}$ is the $6\times 6$ uncertainty of the newly cloned \ac{imu} pose error state (which is fully correlated, thus $\bSigma_{\cI\cC_{1}}=\bSigma_{\cI\cI_{\{\tilde{\vp}, \tilde{\btheta}\}}}$). All cross-covariances of the current \ac{imu} pose are assigned to the cross-covariances of the newly cloned state: $\bSigma_{\cC_{1}\cC_{i}} = \bSigma_{\cI_{\{\tilde{\vp}, \tilde{\btheta}\}}\cC_{i}}$ with $i = 2, \ldots, n$ and $\bSigma_{\cC_{1}\cL} = \bSigma_{\cI_{\{\tilde{\vp}, \tilde{\btheta}\}}\cL}$. $\bSigma_{\cC_{n}}$ is the $6\times 6$ uncertainty of the oldest cloned \ac{imu} pose. $\bSigma_{\cC_{i}\cC_{j}}$ is the cross-correlation between $i$-th and $j$-th cloned \ac{imu} pose, $\bSigma_{\cI\cC_{i}}$ is the cross-correlation between the current \ac{imu} state and the $i$-th cloned \ac{imu} pose, and $\bSigma_{\cC_{i}\cL}$ are the cross-correlations between the cloned \ac{imu} poses and the landmarks. 

The cloned poses and landmarks do not evolve with time, meaning no state transition and no process noise (i.e., $\bPhi_{C_n}^{k+1|k} = \vI$, $\vG_{C_n}^{k+1|k} = \bZero$ with $n= 1,..., N$ and $\bPhi_{L_m}^{k+1|k} = \vI$, $\vG_{C_n}^{k+1|k} = \bZero$ with $m= 1,..., M$) is applied, while the navigation states evolve with the \ac{imu} measurements. The linearized error state propagation can be derived as:

\begin{equation}
    \begin{aligned}
    \tilde{\vx}^{k+1} =&~\bPhi^{k+1|k} \tilde{\vx}^{k} + \vG^{k+1|k} \vw^{k}, \\
    \begin{bmatrix}  \tilde{\vx}_{\cI}^{k+1} \\ \tilde{\vx}_{\cC}^{k+1} \\ \tilde{\vx}_{\cL}^{k+1} \end{bmatrix} 
      =&\begin{bmatrix} 
      \bPhi_{\cI}^{k+1|k} & \bZero            & \bZero \\ 
      \bZero            & \bPhi_{\cC}^{k+1|k} & \bZero \\
      \bZero            & \bZero            & \bPhi_{\cL}^{k+1|k}
      \end{bmatrix} 
      \begin{bmatrix}  \tilde{\vx}_{\cI}^{k} \\  \tilde{\vx}_{\cC}^{k} \\ \tilde{\vx}_{\cL}^{k} \end{bmatrix} \\
       &+ \begin{bmatrix}  \vG_{\cI}^{k+1|k} \\ \vG_{\cC}^{k+1|k}\\ \vG_{\cL}^{k+1|k} \end{bmatrix} \vw^{k} \\
       =&\begin{bmatrix} 
            \bPhi_{\cI}^{k+1|k} & \bZero & \bZero\\ 
            \bZero & \vI & \bZero \\
            \bZero & \bZero & \vI 
          \end{bmatrix} 
          \begin{bmatrix}  \tilde{\vx}_{\cI}^{k} \\ \tilde{\vx}_{\cC}^{k} \\ \tilde{\vx}_{\cL}^{k} \end{bmatrix} + \begin{bmatrix}  \vG_{\cI}^{k+1|k} \\ \bZero \\ \bZero \end{bmatrix} \vw^{k}
    \end{aligned}        
\end{equation}

with the linearized state transition matrix $\bPhi$ and the linearized perturbation matrix $\vG$ computed as explained by Weiss in \cite{5979982} or related work.
The full error-state uncertainty of~\cref{eq:ekf_rio_multi_stacked_covariance} can then be propagated as:

\begin{equation}\label{eq:ekf_rio_multi_covprop}
    \begin{aligned}
    \bSigma^{k+1} &=  \bPhi^{k+1|k}\bSigma^{k}(\bPhi^{k+1|k})^\transpose +  \vG^{k+1|k} \vQ^{k} (\vG^{k+1|k})^\transpose \\
    &= \begin{bmatrix}
        \bSigma_{\cI}^{k+1}   & \bPhi_{\cI}^{k+1|k}\bSigma_{\cI\cC}^{k} \vI  & \bPhi_{\cI}^{k+1|k}\bSigma_{\cI\cL}^{k} \vI \\
        \bullet             & \bSigma_{\cC}^{k}                        & \bSigma_{\cC\cL}^{k} \\
        \bullet             & \bullet                                & \bSigma_{\cL}^{k}    
    \end{bmatrix}
    \end{aligned}       
\end{equation}

with $\vQ$ being the discretized process noise matrix and $\bPhi_{\cI}^{k+1|k}$ the error-state transition matrix of the \ac{imu} error-state $\tilde{\vx}_{\cI}$.  
This propagation allows us to rigorously reflect the cross-correlations between the landmark, the cloned states, and the evolved \ac{imu} states in our error-state formulation. The above described formalism enables us to correctly use the state variables in order to align the trails to the current scan prior to point matching as well as compute residuals during the update.

\subsubsection{Multi-State Update With Measurement Trails}
Given a set of matched 3D point-trails as in the~\cref{fig:ekf_rio_multi_trails_mng}, we now want to estimate the distances to the matched points in the current scan across all points contained in the trails history. For a single matched trail, using cloned poses in the buffer, we transform all points $\referencet{\vp}{R}{P_{j}}{}{t_{p}}$ from the trail history at time instance $t_{p}$, where $p=1, \ldots, V$ and $V$ is the length of the matched trail, to the current radar reference frame, considering the robot's spatial evolution:

\begin{equation}
\begin{aligned}\label{eq:ekf_rio_multi_propagated_distance}
  \referencet{{\vp'}}{R}{P_{j}}{}{t_{p}} = & \referencet{\vR}{I}{R}{}{\transpose} \left( -\reference{\vp}{I}{R}{} + (\referencet{\vR}{G}{I}{}{t_c})^\transpose \left(-\referencet{\vp}{G}{I}{}{t_{c}} + \right.  \right. \\
    & \left. \left. \referencet{\vp}{G}{I}{}{t_{p}} + \referencet{\vR}{G}{I}{}{t_{p}} \left( \reference{\vp}{I}{R}{} + \reference{\vR}{I}{R}{} \referencet{\vp}{R}{P_{j}}{}{t_{p}}  \right) \right) \right)
\end{aligned}
\end{equation}

where $\reference{\vR}{I}{R}{}$ and $\reference{\vp}{I}{R}{}$ is the constant pose (orientation and position) of the radar frame with respect to the \ac{imu} frame. $\referencet{\vR}{G}{I}{}{\{t_{c}, t_{p}\}}$ and $\referencet{\vp}{G}{I}{}{\{t_{c}, t_{p}\}}$ are the \ac{imu} orientation and position corresponding to the trail history element at time $t_{p}$ and current radar scan at $t_{c}$, with respect to the navigation frame $\{G\}$. Similarly to the method described in~\cref{sec:ekf_rio_single}, we transform the 3D point from Cartesian space to Spherical coordinates and only use the most informative dimension, the distance for residual construction.

The estimated distance, which is compared to the current distance measurement, is calculated for each point in the trail history as the norm of the transformed point from $t_{p}$:

\begin{equation}
\begin{aligned}\label{eq:ekf_rio_multi_distmeasmodel}
    d_{P_{j}}= {} & \Big\| \referencet{{\vp'}}{R}{P_{j}}{}{t_{p}} \Big\|
\end{aligned}
\end{equation}

where $d_{P_{j}}$ is the distance to a single point in the matched trail history $\referencet{\vp'}{R}{P_{j}}{}{t_{p}}$ at $t_{p}$ aligned to the current radar pose at $t_{c}$.    
Since this measurement relates to states from past time instances, stochastic cloning is necessary as introduced earlier in this section.

\subsubsection{Update With Persistent Landmarks}
When a trail has been continuously matched for a predefined amount of times in the past, it is promoted to a persistent landmark and added as such to the state vector. Specifically, after each update, the set of trails is scanned for elements which have been matched consecutively for $N$ times. When a trail meets this criterion, it is used to initialize a persistent landmark in the state vector and the covariance matrix is augmented according to \cite{1545392}. For convenience, we introduce $\vx_{\cD} = \left[ \vx_{\cI}; \vx_{\cC} \right]$. Blocks needed for augmenting the state vector and error-state covariance are shown in the Fig.~\ref{fig:ekf_rio_multi_state_and_cov} and are computed as:

\begin{figure}[thpb]
  \centering
  \includegraphics[width=0.75\columnwidth]{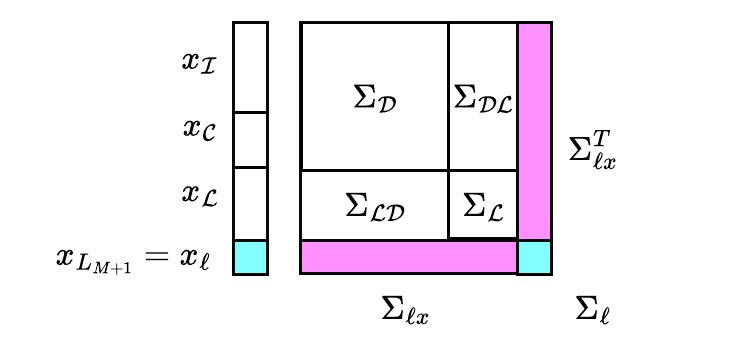}
  \caption[Augmentation of the state and covariance with a persistent landmark]{Augmented nominal state and error-state covariance after adding a persistent landmark.}
  \label{fig:ekf_rio_multi_state_and_cov}
\end{figure}

\begin{equation}\label{eq:ekf_rio_multi_covland1}
    \begin{aligned}
    \bSigma_{\ell} &=  \vH_{\cD}\bSigma_{{\cD}}\vH_{\cD}^{\transpose} + \vH_{\ell}\vR\vH_{\ell}^{\transpose},
    \end{aligned}       
\end{equation}

\begin{equation}\label{eq:ekf_rio_multi_covland2}
    \begin{aligned}
    \bSigma_{\ell x} &=  \vH_{\cD}\bSigma_{\cD x}
    \end{aligned}       
\end{equation}

with $\bSigma_{\cD x} = \left[\bSigma_{\cD}, \bSigma_{\cD \cL}\right]$, and $\bSigma_{\cD \cL}$ being the cross-covariance between the \ac{imu} and \ac{imu} clones error-state vector segment, and the persistent landmarks. $\vR$ is the covariance matrix of the measurement noise, $\vH_{\cD} = \frac{\partial p}{\partial \tilde{\vx}_{\cD}}$ and $\vH_{{\ell}} = \frac{\partial p}{\partial \tilde{\vx}_{\ell}}$ are the Jacobians of the inverse observation model of a 3D point radar measurement, $p$ (Eq.~\ref{eq:ekf_rio_multi_covlandd}), with respect to the \ac{imu} and \ac{imu} clones error-state variables $\tilde{\vx}_{\cD}$, and the error-state variable $\tilde{\vx}_{\ell}$ of the newly added landmark $\vx_{\ell} = \vx_{L_{M+1}}$, respectively. The inverse observation model of the 3D radar point in the navigation frame $\{G\}$ is expressed as:

\begin{equation}\label{eq:ekf_rio_multi_covlandd}
    \begin{aligned}
    \reference{\vp}{G}{L_{m}}{} &=  p\left(\vx, \vz \right) = \referencet{\vR}{G}{I}{}{} \big( \big. \referencet{\vR}{I}{R}{}{} \referencet{\vp}{R}{L_{m}}{}{} + \reference{\vp}{I}{R}{} \big. \big) + \reference{\vp}{G}{I}{}
    \end{aligned}       
\end{equation}

with $\reference{\vR}{I}{R}{}$ and $\reference{\vp}{I}{R}{}{}$ being the pose between the \ac{imu} and radar sensor (which is assumed to be rigid and known a-priori), $\referencet{\vp}{R}{L_{m}}{}{}{}$ is the radar observation of the trail point in the current radar reference frame $\{R\}$ with which an $m$-th landmark will be initialized, and  $\reference{\vR}{G}{I}{}{}$ and $\reference{\vp}{G}{I}{}{}$ being the current pose of the \ac{imu} in the navigation frame.
For readability, the estimate of the $m$-th landmark is abbreviated by ${\vl}_{m} = \reference{\vp}{G}{L_{m}}{}{}$. 

When a persistent landmark does not have a match within the current radar scan, then it is discarded from the state vector and the covariance matrix is shrunk accordingly.

Finally, the estimated distance used for the update is computed according to:

\begin{equation}\label{eq:ekf_rio_multi_landmarkup}
    \begin{aligned}
    {\vl'}_{m} &= \reference{\vp}{R}{L_{m}}{} =  \referencet{\vR}{I}{R}{}{\transpose} \big( \big. \referencet{\vR}{G}{I}{}{\transpose} 
    \big( \big. {\vl}_{m} - \reference{\vp}{G}{I}{}{} \big. \big)   - \reference{\vp}{I}{R}{}{} \big. \big),
    \end{aligned}
\end{equation}

\begin{equation}\label{eq:ekf_rio_multi_landmarkdist}
\begin{aligned}
    d_{\vl_{m}}= {} & \Big\| {\vl'}_{m} \Big\|
\end{aligned}
\end{equation}

\subsubsection{Estimator Summary}\label{subsec:estimator_summary}
Summarizing, in our \ac{rio} method we propagate the state and its covariance according to~\cref{eq:ekf_rio_multi_system_dynamics} and~\cref{eq:ekf_rio_multi_covprop}. The update step of our tightly-coupled \ac{ekf} consists of three components - the first one makes use of distances to points in the history elements of trails compared to current radar measurements (\cref{eq:ekf_rio_multi_propagated_distance} and~\cref{eq:ekf_rio_multi_distmeasmodel}), the second one compares distances of persistent landmarks to current radar measurements (\cref{eq:ekf_rio_multi_landmarkup} and~\cref{eq:ekf_rio_multi_landmarkdist}), Finally, the third component employs Doppler velocities as in~\cref{sec:ekf_rio_single} reduced to the inlier set using 3-point RANSAC as detailed in \cite{doer2020ekf}. For all components, we apply outlier rejection using the chi-squared test.

\subsection{Experiments}\label{subsec:ekf_rio_multi_experiments}
In the following, we outline the setup we used and the experiments we performed to validate our method on a real platform with the data from real flights as well as the results of the evaluation.
\subsubsection{Experimental Setup}
The sensor used for the experiments is a lightweight and inexpensive \ac{fmcw} radar manufactured by Texas Instruments integrated on an evaluation board AWR1843BOOST, shown attached to the \ac{uav} in~\cref{fig:ekf_rio_single_platform}, equipped with a USB interface and powered with $\unit[2.5]{V}$. The frequency spectrum of chirps generated by the radar is between $f_l = \unit[77]{GHz}$ and $f_u = \unit[81]{GHz}$. The \ac{fov} is $\unit[120]{\degree}$ in azimuth and $\unit[30]{\degree}$ in elevation. Measurements are obtained at the rate of $f_{m} = \unit[15]{Hz}$. The radar is affixed to one extremity of the experimental platform facing forward by a tilt of about $\unit[45]{\degree}$ with respect to the horizontal plane as shown in~\cref{fig:ekf_rio_single_platform}. This improves the velocity readings compared to nadir view while keeping point measurements on the ground and thus at a reasonable distance. For inertial measurements, we use the \ac{imu} of the Pixhawk 4 flight controller unit (FCU) with a sampling rate of $f_{si} = \unit[200]{Hz}$. We manually calibrate the transformation between the radar and \ac{imu} sensors, which is used as a constant spatial offset in the \ac{ekf}. The initial navigation states of the filter are set to the ground truth values. $N$ was set to 7. We placed some arbitrary reflective clutter in the scene since the test environment was otherwise a clutter-less clean lab space. No position information from the added objects of any sort was measured or used in our approach other than what the onboard radar sensor perceived by itself. We use a motion capture system to record the ground truth trajectories. During acquisition, we recorded sensor readings from the \ac{imu} and radar together with the poses of the \ac{uav} streamed by the motion capture system as ground truth. Our \ac{ekf}-based \ac{rio} is executed offline on the recorded sensor data on an Intel Core i7-10850H vPRO laptop with $\unit[16]{GB}$ RAM in a custom C++ framework compiled with gcc 9.4.0 at -O3 optimization level. In the case of the closed-loop flight in~\cref{subsec:ekf_rio_multi_cl}, we recorded the pose estimates calculated (and fed to the controller) onboard in real-time by our \ac{rio}. Our \ac{rio} was executed on a raspberry pi 4 with $\unit[4]{GB}$ RAM mounted on the platform. Execution timings for the aforementioned machines are shown in~\cref{tab:ekf_rio_multi_timings} and~\cref{tab:cl_timings}, and confirm the real-time capability of the implementation.

\subsubsection{Evaluation}\label{subsec:ekf_rio_multi_eval}
For evaluation of the presented \ac{rio} approach, we use the data recorded in an indoor space shown in~\cref{fig:ekf_rio_multi_platform} during seven manually-controlled \ac{uav} flights. The flown trajectories were not pre-planned and included pronounced motions in all three dimensions. One of the executed trajectories can be seen in the~\cref{fig:ekf_rio_multi_3d_trajectory}. We choose to measure the quality of our estimator using the norm of \ac{mae} of position and the final pose drift in percent (without yaw alignment) in order to easily compare against the state-of-the-art. In~\cref{fig:ekf_rio_multi_maes}, one can observe the mean of the $||\ac{mae}||$ across the flown trajectories. The $||\ac{mae}||$ increases steadily with the flown distance as the pose drift builds up as expected. For comparison, we show that the $||\ac{mae}||$ is reduced by a factor of 2 with respect to our previous results presented in~\cref{sec:ekf_rio_single}, and by a factor of 4 with respect to the state-of-the-art shown in \cite{doer2020ekf} where only an \ac{fmcw} radar and \ac{imu} is used and no assumptions on environment are made. The sample based 1$\sigma$ bounds grow to a value of about $\sigma = \unit[0.25]{m}$. Regarding the final drift, as shown in Tab.~\ref{tab:ekf_rio_multi_results}, on average we achieve $\unit[0.81]{\%}$ on trajectories ranging from $\unit[127.5]{m}$ to $\unit[175.0]{m}$ against $\unit[5.0]{\%}$ reported for a $\unit[60]{m}$ long flown trajectory in \cite{doer2020ekf} for a similar setup as ours consisting of only \ac{fmcw} radar and \ac{imu}. Authors in \cite{doer2020ekf, doer2020radar} also report the final drift values for the same setups, yet additionally augmented with a barometer sensor to reduce the vertical drift - even in this case our results with no additional sensor remain comparable - $\unit[0.81]{\%}$ (ours) against $\unit[0.60]{\%}$ \cite{doer2020ekf} and $\unit[0.36]{\%}$ \cite{doer2020radar}. Interestingly, the minimum value of the final drift in our experiments - $\unit[0.17]{\%}$ - is even lower than the lowest value reported in \cite{doer2020ekf, doer2020radar} obtained flying the same distance but with additional sensors and much lower excitation in their case. Compared to $\unit[3.32]{\%}$ final drift for a hand-held trajectory of $\unit[116.4]{m}$ with low excitation in~\cref{sec:ekf_rio_single} , our method outperforms it by a factor of 4 in a real \ac{imu} flight with high excitation. 

\begin{figure}[thpb]
  \centering
  \includegraphics[width=0.95\columnwidth]{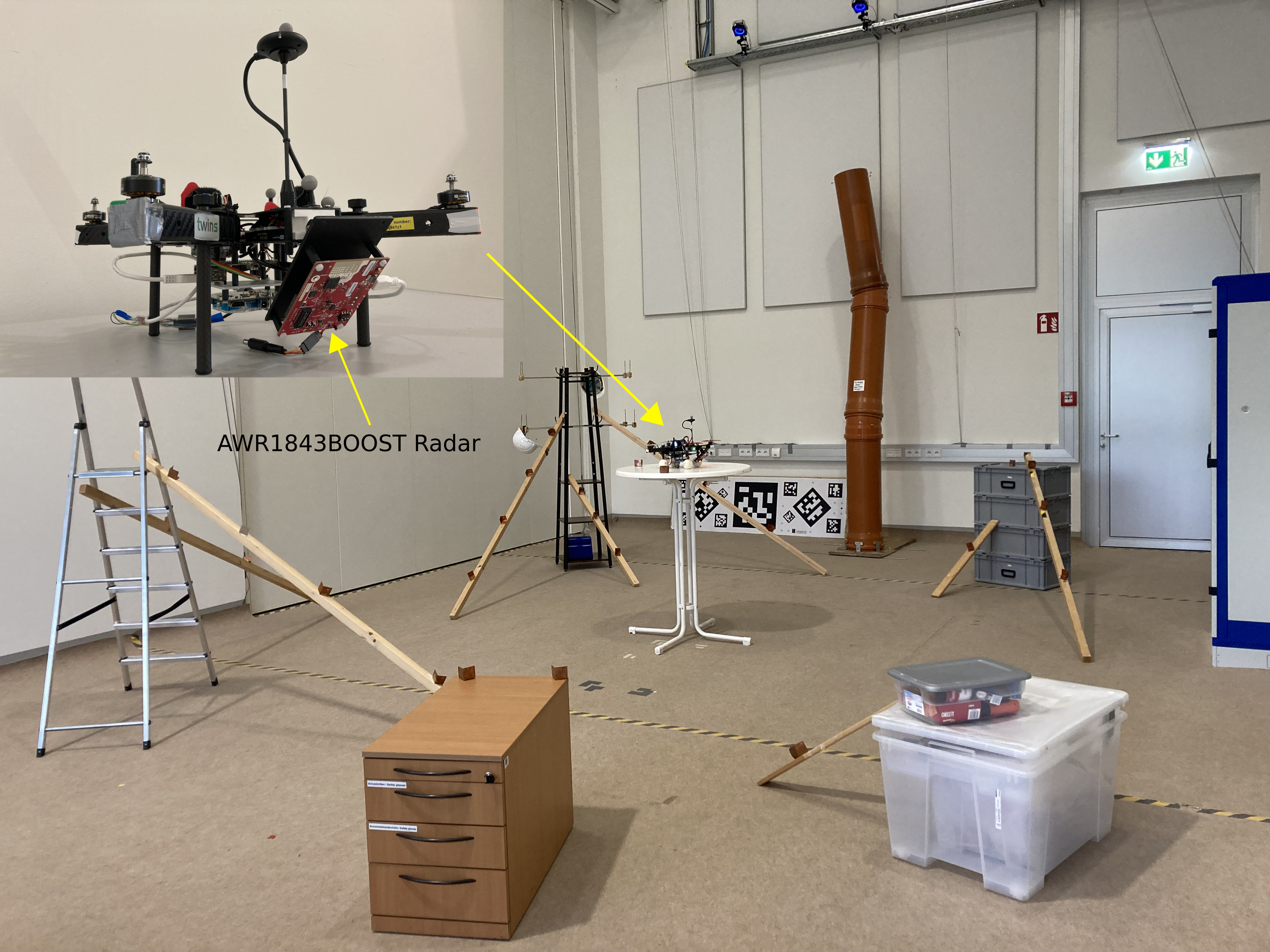}
  \caption[Experimental platform and the scene used for evaluation of the multi-state approach]{Experimental platform used in this work with the \ac{fmcw} radar sensor and the indoor space where the experiments were conducted. Note the mounting of the sensor tilted at 45\textdegree{} angle. Reflective clutter was scattered randomly on the scene. No global position (nor attitude) information of any sort about scattered objects was used in our approach.}
  \label{fig:ekf_rio_multi_platform}
\end{figure}

\begin{figure}[thpb]
  \centering
  \includegraphics[width=0.8\columnwidth]{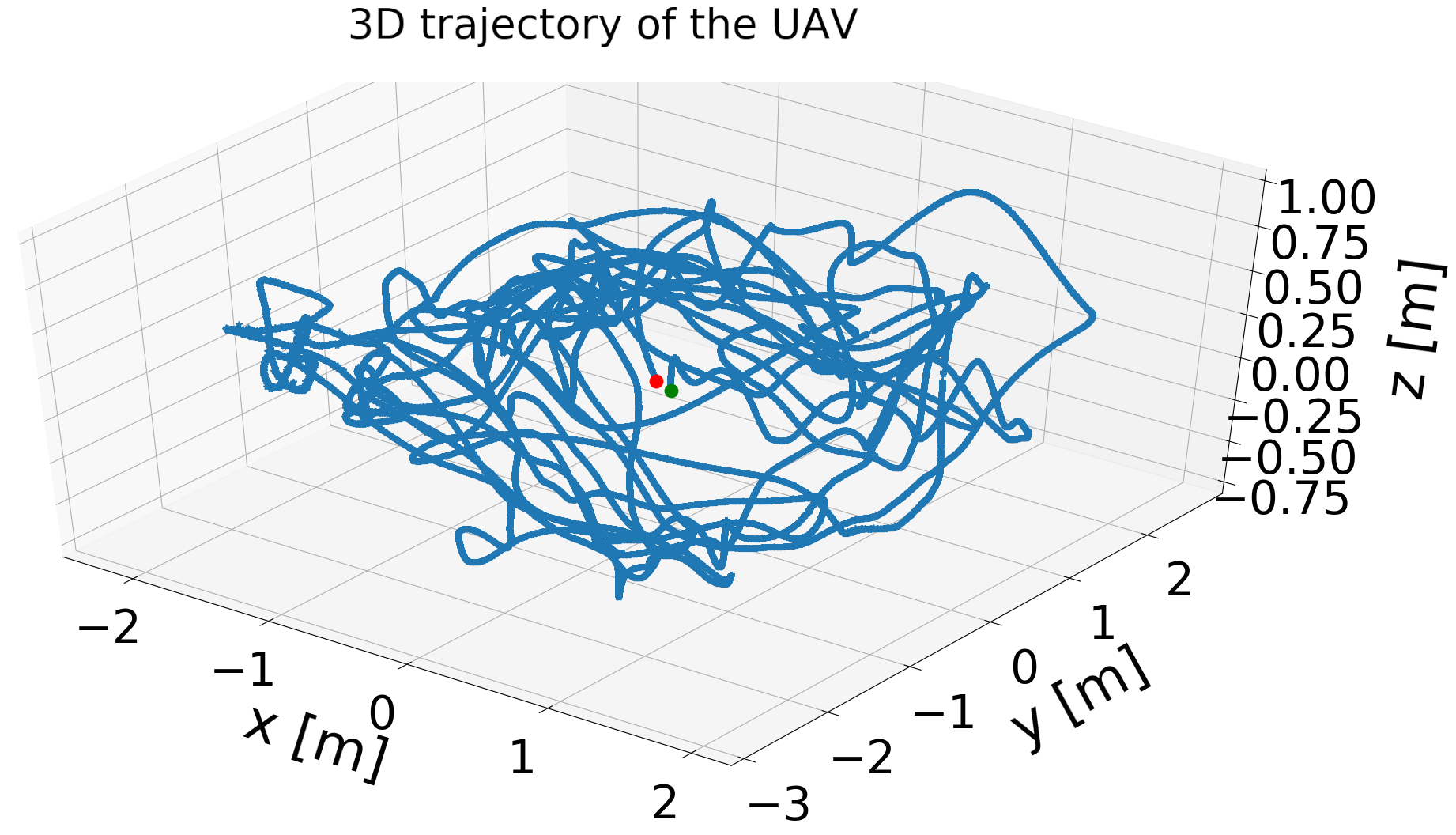}
  \caption[Sample of the executed trajectories for evaluation of the multi-state approach]{One of the executed trajectories. As can be seen, we tried to perform pronounced motions in all three dimensions. Green and red dots represent the take-off and landing positions on the white table seen in the~\cref{fig:ekf_rio_multi_platform}.}
  \label{fig:ekf_rio_multi_3d_trajectory}
\end{figure}

\begin{figure}[thpb]
  \centering
  \includegraphics[width=1.0\columnwidth]{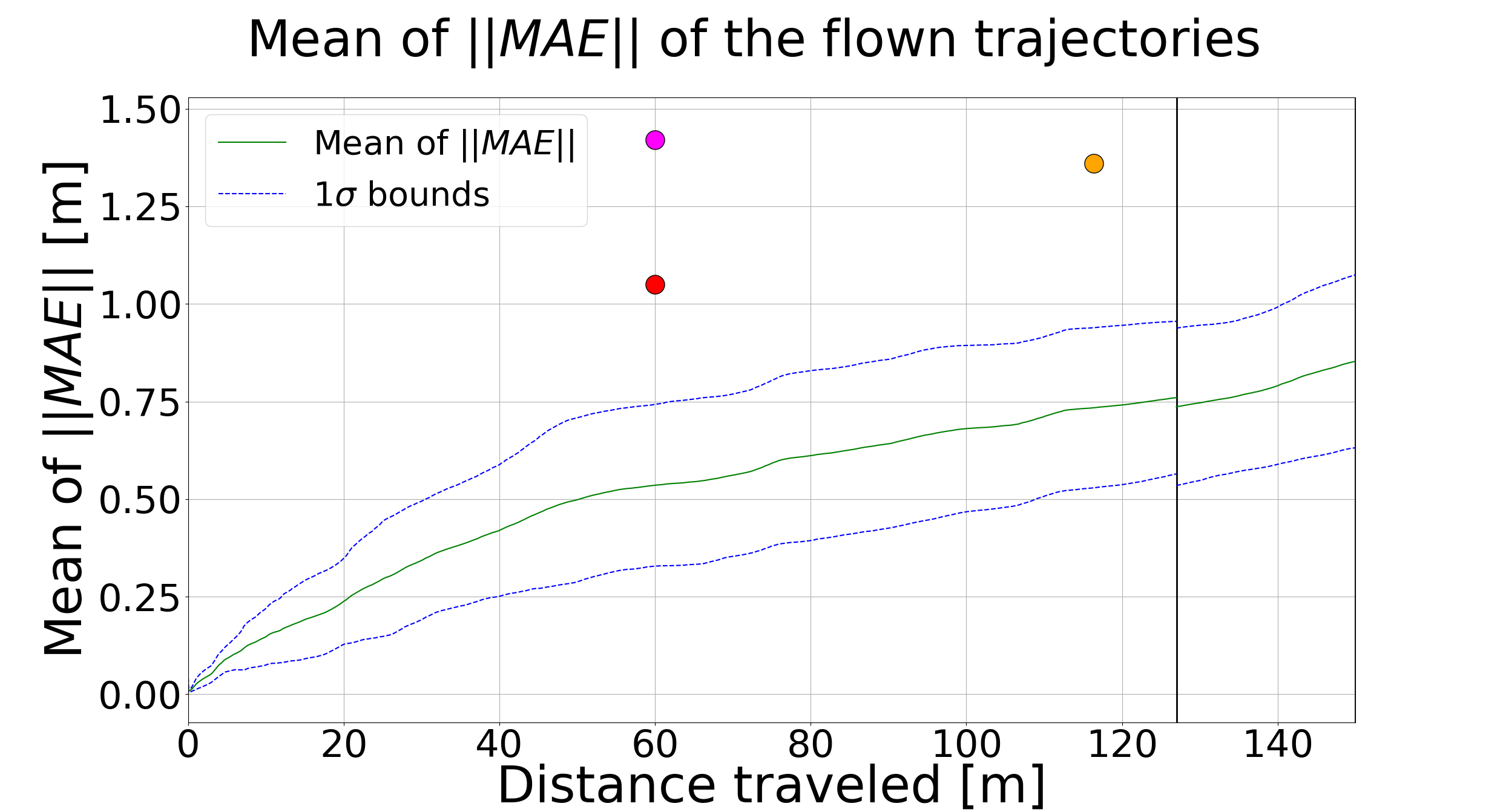}
  \caption[Mean of the norm of MAE for the multi-state approach]{Mean of the norm of \ac{mae} for all flown trajectories. Up to $\unit[127]{m}$ the mean of all trajectories is taken. From
  $\unit[127]{m}$ to $\unit[150]{m}$ mean of six trajectories is taken (since trajectory 1 ended). Blue dashed lines are the sample based 1$\sigma$ bounds. For comparison, red and orange dots represent results presented in~\cref{sec:ekf_rio_single} and the magenta dot represent the state-of-the-art result in \cite{doer2020ekf} for the flown trajectory.}
  \label{fig:ekf_rio_multi_maes}
\end{figure}

\bgroup
\def\arraystretch{1.15}
{
\begin{table}[thpb]
\centering
\caption[Evaluation metrics for the multi-state approach]{Metrics gathered across all flown trajectories}
\label{tab:ekf_rio_multi_results}
\resizebox{0.8\linewidth}{!}{%
\begin{tabular}{|cc|c|c|}
\hline
\multicolumn{1}{|c|}{Trajectory} & Length {[}m{]} & Norm of MAE at \unit[127]{m} {[}m{]} & Final drift {[}\%{]} \\ \hline
\multicolumn{1}{|c|}{1} & 127.5 & 0.90 & \textbf{0.17} \\ \hline
\multicolumn{1}{|c|}{2} & 150.5 & 1.10 & 1.62 \\ \hline
\multicolumn{1}{|c|}{3} & 158.4 & 0.68 & 0.80 \\ \hline
\multicolumn{1}{|c|}{4} & 160.4 & 0.71 & 0.61 \\ \hline
\multicolumn{1}{|c|}{5} & 166.7 & 0.59 & 0.78 \\ \hline
\multicolumn{1}{|c|}{6} & 168.1 & 0.84 & 0.91 \\ \hline
\multicolumn{1}{|c|}{7} & 175.0 & \textbf{0.45} & 0.81 \\ \hline
\multicolumn{2}{|c|}{Average} & \textbf{0.75} & \textbf{0.81} \\ \hline
\end{tabular}
} 
\end{table}
} 
\bgroup
\def\arraystretch{1.15}
{
\begin{table}[thpb]
\centering
\caption[Execution timings for the multi-state approach]{Execution timings for a single trajectory}
\label{tab:ekf_rio_multi_timings}
\resizebox{0.8\linewidth}{!}{%
\begin{tabular}{c|cc|cc}
\cline{2-3}
 & \multicolumn{2}{c|}{Average time {[}ms{]}} &  &  \\ \cline{2-5} 
 & \multicolumn{1}{c|}{Propagation} & Update & \multicolumn{1}{c|}{Duration {[}s{]}} & \multicolumn{1}{c|}{Realtime factor} \\ \hline
\multicolumn{1}{|c|}{Trajectory 7} & \multicolumn{1}{c|}{0.08} & 2.07 & \multicolumn{1}{c|}{436} & \multicolumn{1}{c|}{25.36} \\ \hline
\end{tabular}
}
\end{table}
}

\subsection{Online Calibration and its Impact on the Accuracy and Consistency of The Multi-State EKF RIO}\label{subsec:ekf_rio_multi_online_calib}
In this section we introduce the online estimation of the extrinsic calibration parameters and thus simplifying the use of the framework as well as improving its accuracy and consistency. With online extrinsic calibration parameters the state vector looks as follows:

\begin{equation}\label{eq:ekf_rio_multi_calib_states}
\begin{aligned}
    \vx =& \left[ \vx_{\cI}; \vx_{\cR}; \vx_{\cC}; \vx_{\cL} \right] =  \big[ \big.  \left[\reference{{\vp}}{G}{I}{}; \reference{\bar{\vq}}{G}{I}{}; \reference{{\vv}}{G}{I}{}; \vb_{\va}; \vb_{\bomega}\right];\\
    & \left[\reference{{\vp}}{I}{R}{}; \reference{\bar{\vq}}{I}{R}{}\right];
     \left[\reference{{\vp}}{G}{I_{1}}{}; \reference{\bar{\vq}}{G}{I_{1}}{}; \ldots ; 
     \reference{{\vp}}{G}{I_{N}}{}; \reference{\bar{\vq}}{G}{I_{N}}{}\right]; \\ &\left[ \reference{{\vp}}{G}{L_{1}}{}; \ldots ; \reference{{\vp}}{G}{L_{M}}{} \right] \big. \big] 
\end{aligned}
\end{equation}

where the newly added state variables $\left[\reference{{\vp}}{I}{R}{}; \reference{\bar{\vq}}{I}{R}{}\right]$ describe the 3D pose of the radar sensor with respect to the \ac{imu} sensor.
After the addition of the sensor extrinsic calibration parameters as variables to the state vector the stochastic cloning procedure remains analogous to the one reported in~\cref{sec:ekf_rio_multi}, with the only difference being, that the covariance matrix blocks containing the cross-covariance terms copied upon the cloning of a new \ac{imu} state are bigger in size. Namely, the increase in size corresponds to the two $3\times3$ blocks which correspond to the cross-covariance values between the state-to-be-cloned and the calibration error states - position and orientation. Naturally, during propagation the calibration states do not evolve in time hence, we add the following equations to~\cref{eq:ekf_rio_multi_system_dynamics}:

\begin{equation}\label{eq:ekf_rio_multi_calib_dynamics}
\begin{aligned}
    \reference{\dot{\vp}}{I}{R}{} &= \bZero \\
    \reference{\dot{\vR}}{I}{R}{} &= \bZero
\end{aligned}
\end{equation}

Estimation accuracy is impacted by the precision of the extrinsic calibration parameters of the sensors. Also, as stated in \cite{li_consistency}, the unmodelled uncertainty of these parameters when treated as static negatively impacts the filter's consistency. Therefore, we propose to estimate these parameters online during the operation of the algorithm. This renders the platform easier to use as the calibration parameters are not needed to be measured manually nor any complex calibration procedure is required. Online calibration estimation also increases the accuracy which can be seen in Fig.~\ref{fig:ekf_rio_multi_calib_maes} where we note the reduction of the $\|MAE\|$ for the same dataset used in~\cref{sec:ekf_rio_multi}, but with the here proposed online calibration implemented. Additionally, based on the analysis done in \cite{jones_sukhatme} and \cite{kelly_sukhatme}, we know that these parameters are observable for general trajectories. 

\begin{figure}[h!tbp]
  \centering
  \includegraphics[width=0.95\columnwidth]{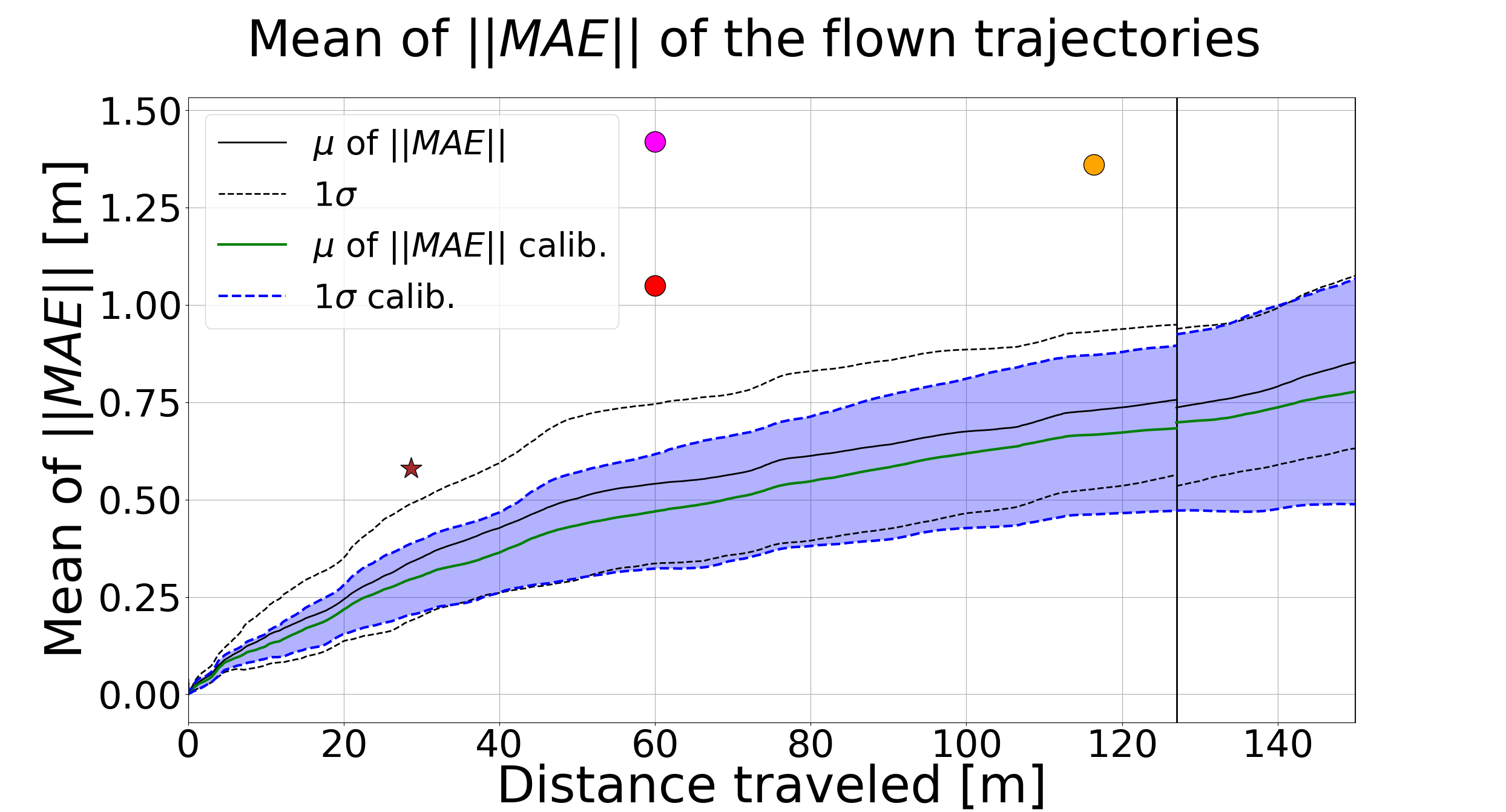}
  \caption[Mean of the norm of \ac{mae} for the flown dataset with online calibration]{Mean of the norm of \ac{mae} for the flown dataset of seven trajectories from~\cref{sec:ekf_rio_multi}. Black dashed lines are from~\cref{sec:ekf_rio_multi} whereas the coloured lines are the values obtained for the same dataset but with the online calibration implemented. As can be noticed there is an increase in accuracy (mean of $\|MAE\|$ reduced). Marked with the brown star is the mean value obtained for the closed-loop flights presented in the~\cref{subsec:ekf_rio_multi_cl} (Tab.~\ref{tab:cl_means}). For comparison, red and orange dots represent results presented in~\cref{sec:ekf_rio_single} and the magenta dot represent the state-of-the-art result in \cite{doer2020ekf} for the flown trajectory. From \unit[127]{m} to \unit[150]{m} six trajectories are used to compute the mean since trajectory 1 ended at \unit[127]{m} (solid black vertical line).}
  \label{fig:ekf_rio_multi_calib_maes}
\end{figure}

In the Fig.~\ref{fig:ekf_rio_multi_calib_pos_error} and Fig.~\ref{fig:ekf_rio_multi_calib_nees}, we can see comparisons of the position estimation errors (with $\pm3\sigma$ envelopes) and the position \ac{nees} respectively for the case of manual and online calibration for one of the trajectories from the above mentioned dataset. In Fig.~\ref{fig:ekf_rio_multi_calib_pos_error}, we can observe that our system with online calibration exhibits better characterization of the actual uncertainty as the estimation errors are bounded by the $\pm3\sigma$ envelopes across the vast majority of the trajectory and the errors oscillate closer to zero. The bounded segments are less frequent for the case of the manual calibration and the errors are shifted further away from zero. These aspects are also reflected in Fig.~\ref{fig:ekf_rio_multi_calib_nees} which depicts the comparison of the \ac{nees} for the same data as for the position errors above. The above comparisons indicate much better consistency for the case with the online calibration since the \ac{nees} is greatly reduced (Bar-Shalom et al. \cite{barshalom}). In the case of manual calibration, we paid close attention to measuring as accurately as possible the extrinsic parameters of the sensors. Manually measured calibration parameters were used as the initial values for the case of online calibration.

\begin{figure}[h!tbp]
  \centering
  \includegraphics[width=0.95\columnwidth]{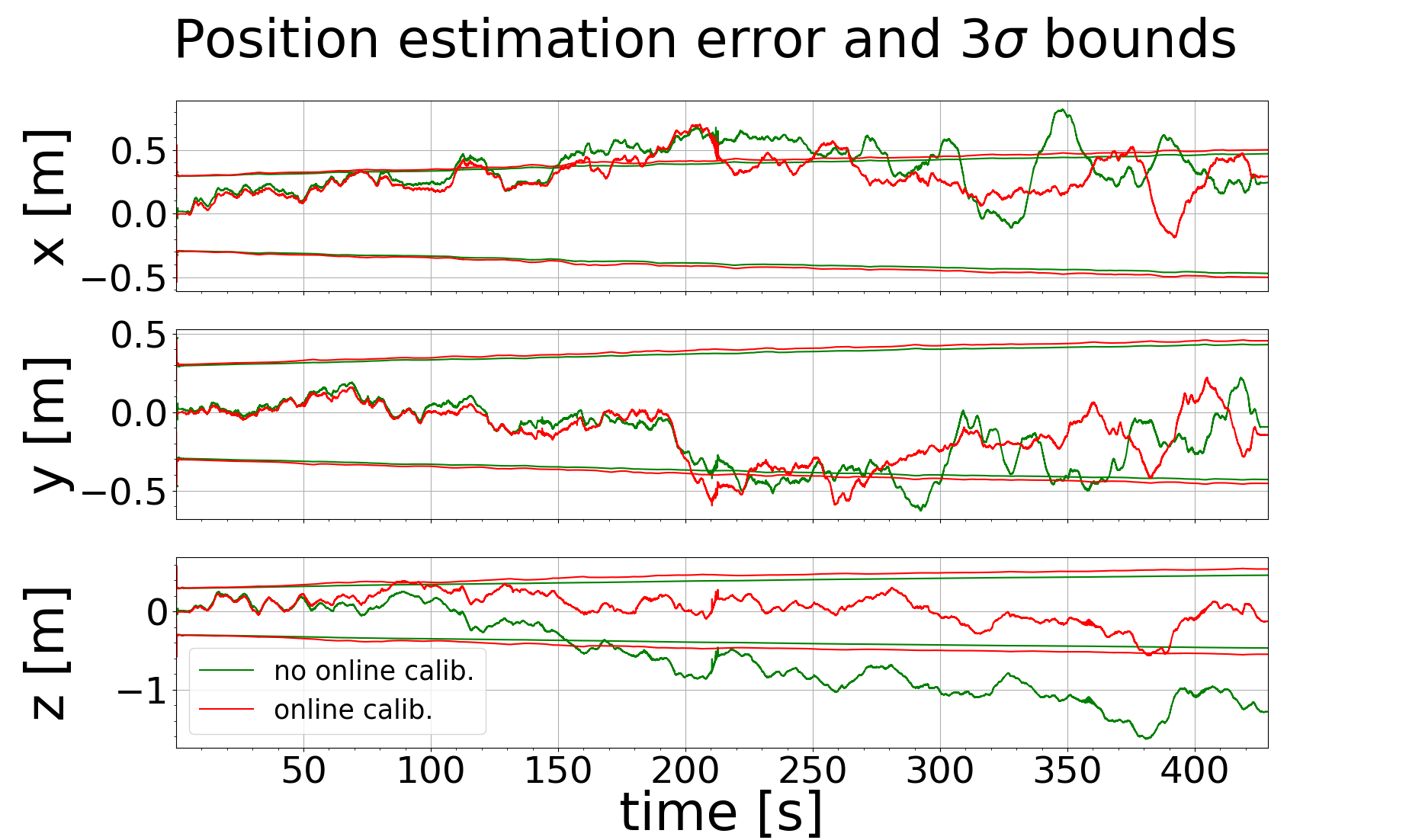}
  \caption[Comparison of the position estimation errors and the $\pm3\sigma$ uncertainty envelopes]{Comparison of the position estimation errors and the $\pm3\sigma$ uncertainty envelopes for one of the trajectories from the dataset in~\cref{sec:ekf_rio_multi} with the manual and the online calibration (green and red respectively). As can be seen in the case of the online calibration, the errors are more thoroughly bounded by the uncertainty across the trajectory than in the case of the manual calibration. Since the estimator is inherently inconsistent \cite{li_consistency}, one cannot expect the errors to be bounded $\unit[99.7]{\%}$ of the time.}
  \label{fig:ekf_rio_multi_calib_pos_error}
\end{figure}

\begin{figure}[h!tbp]
  \centering
  \includegraphics[width=0.95\columnwidth]{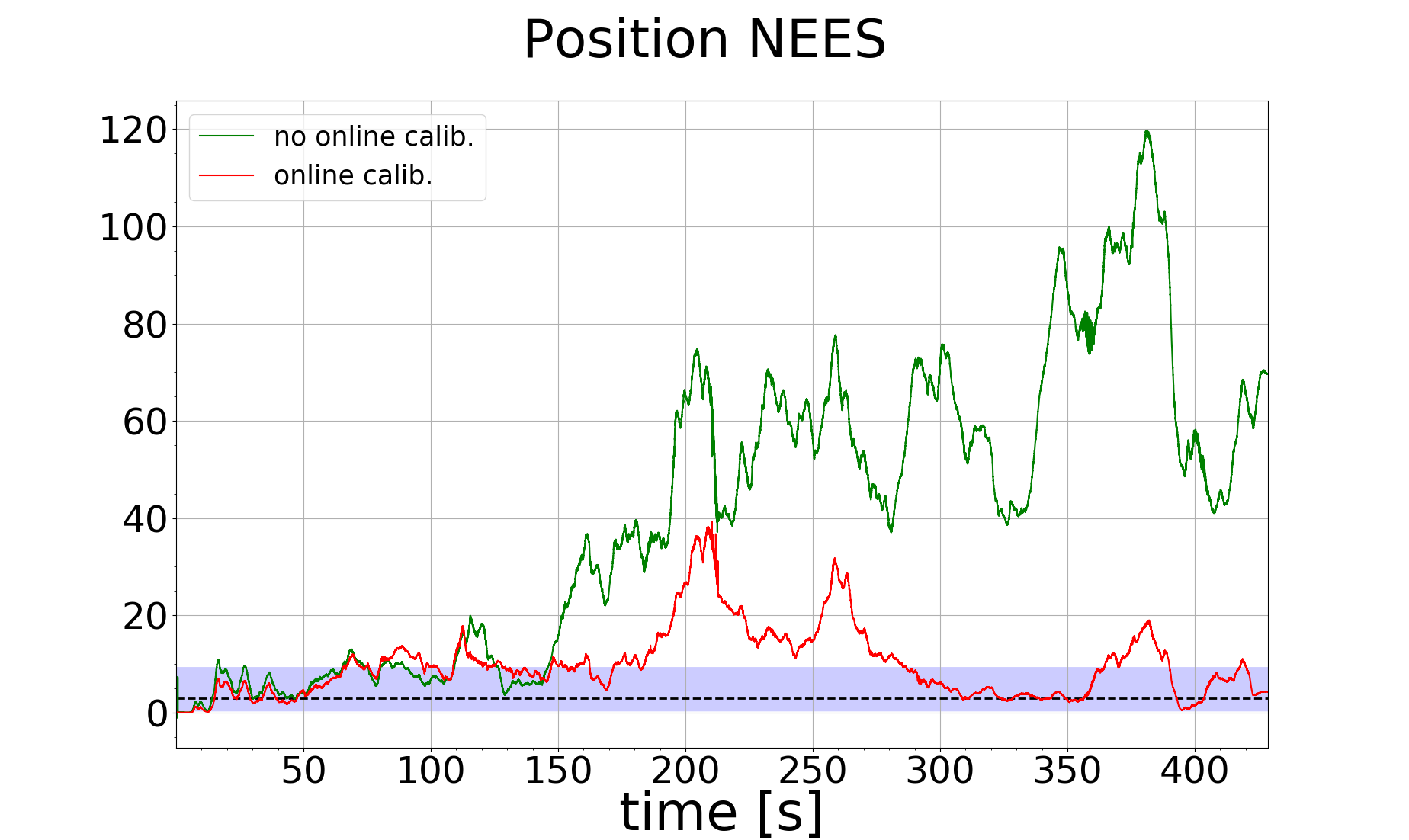}
  \caption[Comparison of the position NEES]{Comparison of the position \ac{nees} for one of the trajectories with the manual and the online calibration (green and red respectively). The filter consistency is greatly improved with the online calibration which can be seen in the significant reduction of the \ac{nees}, which in the case of the online calibration does not continue to grow and more often remains bounded within the two-sided \unit[95]{\%} probability concentration region (shaded blue) than in the case of manual calibration.}
  \label{fig:ekf_rio_multi_calib_nees}
\end{figure}

In the~\cref{fig:ekf_rio_multi_calib_calib_orient} and~\cref{fig:ekf_rio_multi_calib_calib_pos} we plot the convergence of the extrinsic calibration parameters to their measured values for one of the trajectories in the case where the initial extrinsic calibration parameters were initialized with vastly erroneous values, that is, $\vc_{trans}^{init} = [x=\unit[-40]{cm}, y=\unit[-40]{cm}, z=\unit[-40]{cm}]$ in position and $\vc_{rot}^{init} = [roll=\unit[20]{\degree}, pitch=\unit[20]{\degree}, yaw=\unit[20]{\degree}]$ in orientation. Extrinsic calibration parameters which we measured manually with simple tools such as a ruler and a protractor were equal to $\vc_{trans}^{meas} = [\unit[7.5]{cm}, \unit[-1.0]{cm}, \unit[-4.0]{cm}]$ and $\vc_{rot}^{meas} = [\unit[0]{\degree}, \unit[47]{\degree}, \unit[0]{\degree}]$. Given the lack of ground truth for these parameters, we consider the manually measured values the true ones. As observed, our filter successfully recovers from such high initial calibration errors and the parameters converge very close to their manually measured values. The difference between measured and estimated values are $\vc_{trans}^{diff} = [\unit[6.4]{cm}, \unit[2.6]{cm}, \unit[5.0]{cm}]$ and $\vc_{rot}^{diff} = [\unit[0.087]{\degree}, \unit[2.50]{\degree}, \unit[3.27]{\degree}]$ in position and rotation respectively for this particular run which is representative for the majority of the runs. These numbers are comparable to \cite{ewise} where the authors used synthetic data with a dedicated calibration procedure while we use real data during a regular \ac{uav} flight with no dedicated calibration procedure.

\begin{figure}[h!tbp]
  \centering
  \includegraphics[width=1.0\columnwidth]{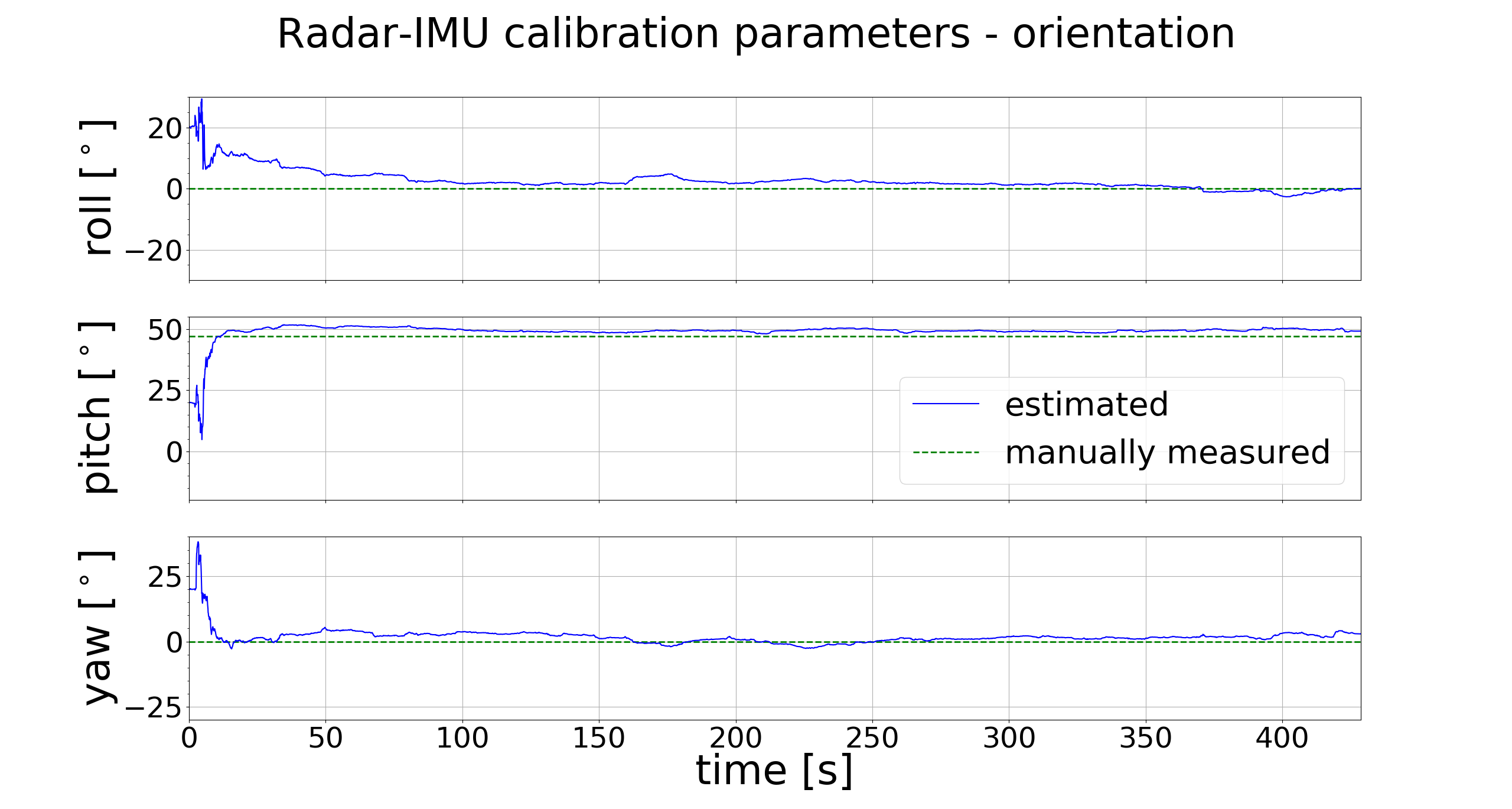}
  \caption[Convergence of the rotational extrinsic calibration parameters in the multi-state approach]{Convergence of the rotational extrinsic calibration parameters for the system with extremely high initial calibration errors. Including the extrinsic calibration parameters in the state vector enables the system to recover from extremely inaccurate initial calibration values and eventually converge to the real values reflecting the tilt of the radar of about \unit[45]{\degree} around the y axis (we manually measured \unit[47]{\degree}) around the y-axis. Green dashed lines are used to mark the manually measured extrinsic calibration parameters values which we consider the true values since we do not have a precise ground truth on these quantities.}
  \label{fig:ekf_rio_multi_calib_calib_orient}
\end{figure}

\begin{figure}[h!tbp]
  \centering
  \includegraphics[width=1.0\columnwidth]{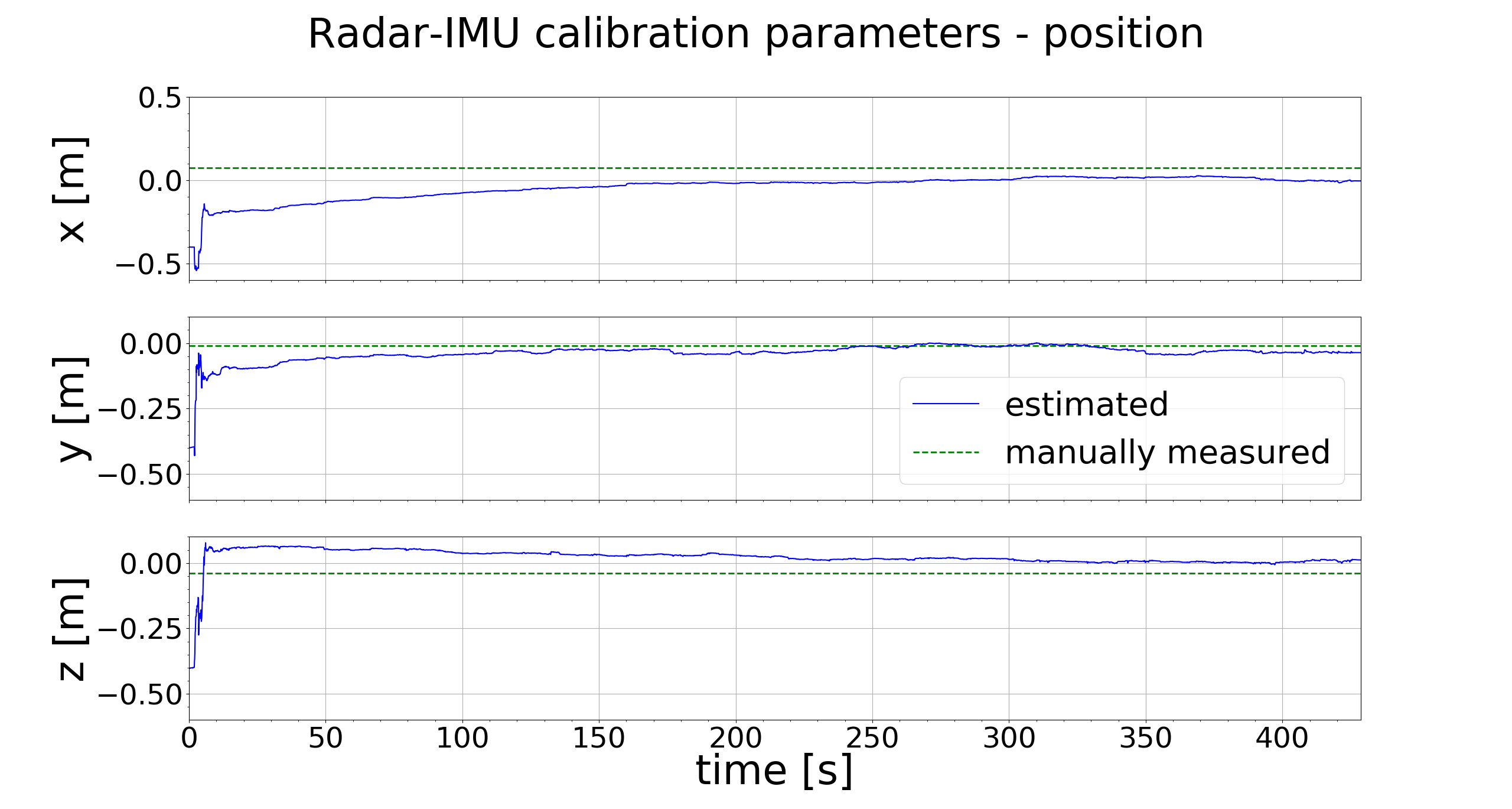}
  \caption[Convergence of the translation extrinsic calibration parameters in the multi-state approach]{Convergence of the translation extrinsic calibration parameters for the system with extremely high initial calibration errors. Green dashed lines are used to mark the manually measured extrinsic calibration parameters values which we consider the ground truth.}
  \label{fig:ekf_rio_multi_calib_calib_pos}
\end{figure}

Even though the estimated and measured values are similar in the estimator's asymptotic behavior, the previously discussed consistency improvement is noticeable. This stems from the filter's ability to include the calibration states for its energy dissipation rather than being hindered to adapt these states online.

For completeness of the discussion about our estimator consistency, we note that we have not implemented yet the approach suggested in \cite{fej}. Thus we expect further consistency improvement by doing so in future work.

\subsection{Multi-State EKF RIO Closed-Loop Flights}\label{subsec:ekf_rio_multi_cl}
For evaluation of the presented \ac{rio} approach, we design a 3D trajectory such that the \ac{uav} can execute it within the motion capture covered area. The top view of the trajectory can be seen in green in the~\cref{fig:cl_traj_topview}. The \ac{uav} then executes this trajectory twice within a single mission using the state estimates computed by our \ac{rio} implementation onboard in real-time. We executed three missions and computed the mean values of the final drift and the norm of \ac{mae} for each of them. Our estimator exhibited stable behaviour in all performed flights. Computed mean values are $\unit[2.84]{\%}$ for the final drift and  $\unit[0.58]{m}$ for $\|MAE\|$ (see~\cref{tab:cl_means}). The mean traveled distance across trajectories is $\unit[28.62]{m}$. As can be seen in the~\cref{fig:ekf_rio_multi_calib_maes}, the accuracy (measured by $\|MAE\|$) obtained in closed-loop flights, marked with a brown star, lies slightly above the mean calculated over the manually flown and subsequently offline processed dataset of seven trajectories collected in indoor space(\cref{fig:ekf_rio_multi_platform}). We attribute this slim discrepancy to the online closed-loop execution of the estimator and controller on a resource-constrained hardware during which some of the measurements are dropped due to CPU load spikes. We note that these are precisely the challenges when demonstrating estimation approaches in closed-loop autonomous flights and that these are important factors our \ac{rio} approach is able to mitigate well. In the~\cref{fig:cl_2dpos} we plot the 2D coordinates of the estimated and true position. In~\cref{tab:cl_timings} we show the average values of the time it takes to execute the entire \ac{imu} and radar processing functions on the onboard computer mentioned in~\cref{subsec:ekf_rio_multi_experiments}.

\bgroup
\def\arraystretch{1.15}
{
\begin{table}[h!tbp]
\centering
\caption[Metrics for the closed-loop flights evaluation]{Metrics computed across closed-loop flights}
\label{tab:cl_means}
\begin{tabular}{|c|c|c|c|}
\hline
Trajectory & Distance {[}m{]} & \multicolumn{1}{l|}{\begin{tabular}[c]{@{}l@{}}Norm of \ac{mae} at\\ full distance {[}m{]}\end{tabular}} & \multicolumn{1}{l|}{Final drift {[}\%{]}} \\ \hline
1 & 28.43 & 0.64 & 3.41 \\ \hline
2 & 28.88 & 0.37 & 2.05 \\ \hline
3 & 28.55 & 0.73 & 3.07 \\ \hline
Average & \textbf{28.62} & \textbf{0.58} & \textbf{2.84} \\ \hline
\end{tabular}
\end{table}
}

\begin{figure}[thpb]
  \centering
  \includegraphics[width=0.95\columnwidth]{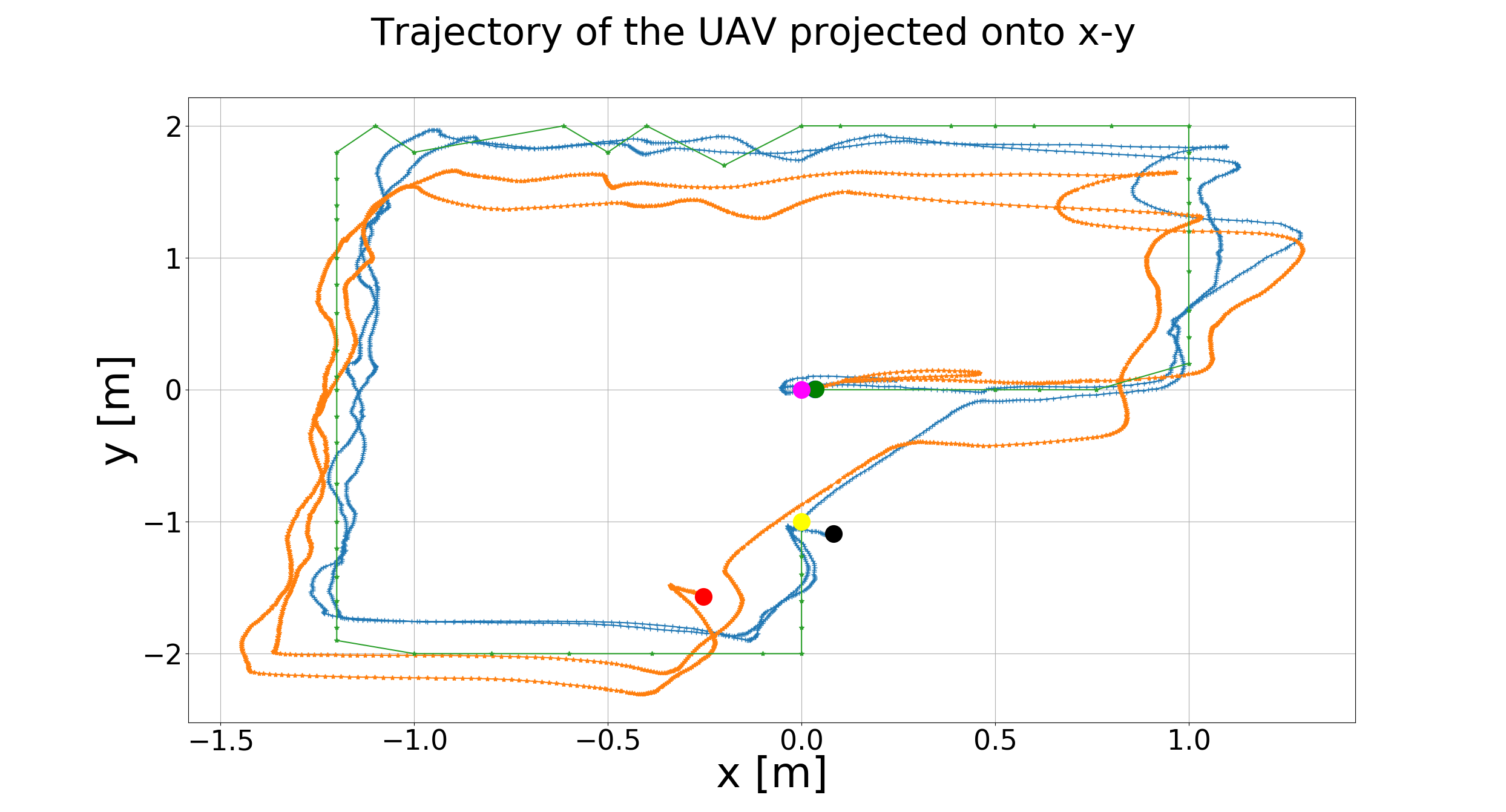}
  \caption[Closed-loop flight 3D plot top-view]{Top view of one of the executed closed-loop flights. Marked in blue and orange are the estimated and true trajectories respectively. Plotted in green is the trajectory input to the controller. Coloured dots are the starting and end points. The magenta-coloured dot marks the starting point of the input trajectory, whereas the dots marking the starting points of the ground truth and estimate coincide.}
  \label{fig:cl_traj_topview}
\end{figure}

\begin{figure}[h!tbp]
  \centering
  \includegraphics[width=0.95\columnwidth]{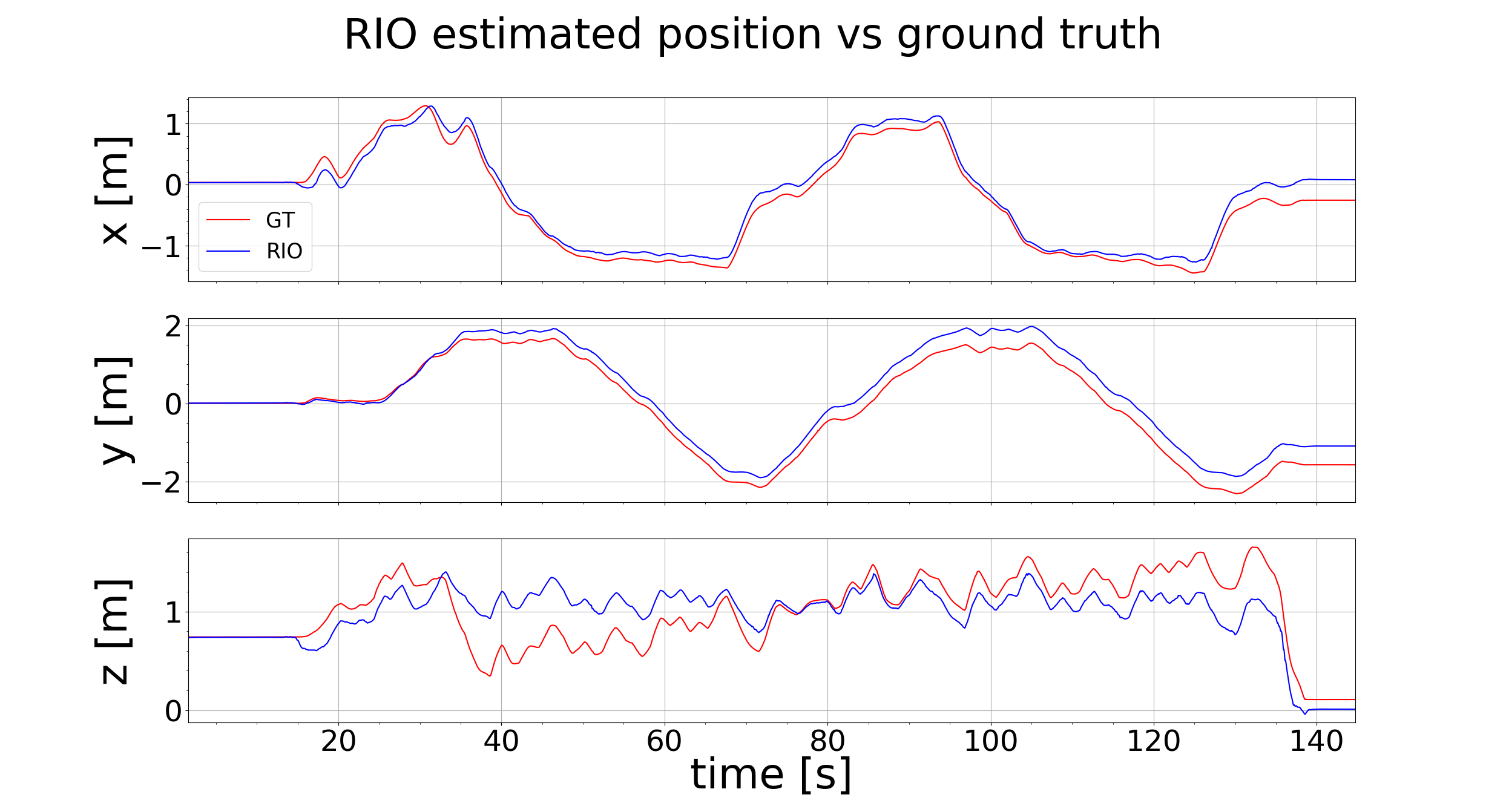}
  \caption[Closed-loop flight 2D plot]{Position coordinates of the true and estimated trajectories plotted against time for one of the closed-loop flights. The take-off and landing points were not on the same plane which causes vertical shift in the z coordinate. Note that no special handling in the filter is needed in the steady initial phase when the robot does not move since no feature triangulation is required but direct depth information is provided by the radar -- an important advantage over vision based approaches.}
  \label{fig:cl_2dpos}
\end{figure}

\bgroup
\def\arraystretch{1.15}
{
\begin{table}[h!tbp]
\centering
\caption[Onboard execution timings]{Onboard execution timings}
\label{tab:cl_timings}
\begin{tabular}{|cc|l}
\cline{1-2}
\multicolumn{2}{|c|}{Average time {[}ms{]}} &  \\ \cline{1-2}
\multicolumn{1}{|c|}{\ac{imu} processing} & Radar processing &  \\ \cline{1-2}
\multicolumn{1}{|c|}{1.04} & 16.97 &  \\ \cline{1-2}
\end{tabular}
\end{table}
}

\subsection{Evaluation of the State Estimation in the Artificial Fog}\label{subsec:cl_fog}
As depicted in the~\cref{fig:cl_fog_platform}, we evaluate our estimation framework in dense artificial fog in order to showcase the benefits of the radar-based navigation in visually degraded conditions. To perform the comparison, we moved the \ac{uav} seen in the~\cref{fig:cl_fog_platform} arbitrarily in a hand-held fashion through a dense artificial fog. We moved the platform manually instead of flying, since as our experiments showed, with the propellers turned on, the amount of the artificial fog we were able to create with the fog machine was quickly dispersed and the evaluation of the impact of the fog became impossible. We further compare the performance of our \ac{rio} to a state-of-the-art \ac{ekf}-based \ac{vio} implementation, \emph{OpenVINS} \cite{geneva2020openvins}. For \ac{vio}, a Matrixvision mvBlueFOX-MLC USB2.0 camera is mounted rigidly next to the radar sensor (seen in the Fig.~\ref{fig:cl_fog_platform}) providing images at \unit[20]{Hz}. For closed-loop computation performance reasons, the \ac{vio} is post-processed offline on the desktop hardware described in the~\cref{subsec:ekf_rio_multi_experiments}. 

Sequence of the results produced by the feature tracker of OpenVINS is shown in~\cref{fig:cl_vio_timelapse}. As expected, the tracker struggles to find sufficient features and to track the few found ones correctly.

The performance of the two approaches is shown in~\cref{fig:cl_rio_vio_comparision}. \ac{vio} diverges almost immediately when the vehicle starts moving as it cannot track features correctly in the dense fog. In comparison, our \ac{rio} framework successfully continues to estimate the robot pose despite the fog since the radar waves can penetrate it. Thus, in environments such as dense fog, darkness or strong light our \ac{rio} approach can be successfully used to estimate the vehicle states correctly and thus be employed to close the feedback control loop in autonomous flights in settings such as disaster zones.

\begin{figure}[h!tbp]
    \centering
    \includegraphics[width=\textwidth]{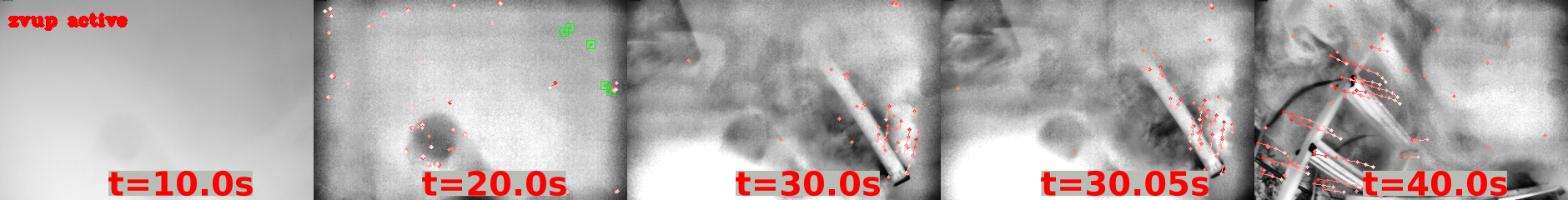}
    \caption[Sequence of the images used by the OpenVINS to track features]{Sequence of the images used by the \acs{vio} framework OpenVINS for tracking features and state estimation. Due to the fog, features can barely be tracked in the first 30 seconds. Thus, OpenVINS struggles to keep track of persistent features (marked in green) and to perform proper state estimation.}
    \label{fig:cl_vio_timelapse}
\end{figure}

\begin{figure}[h!tbp]
    \centering
    \includegraphics[width=\linewidth]{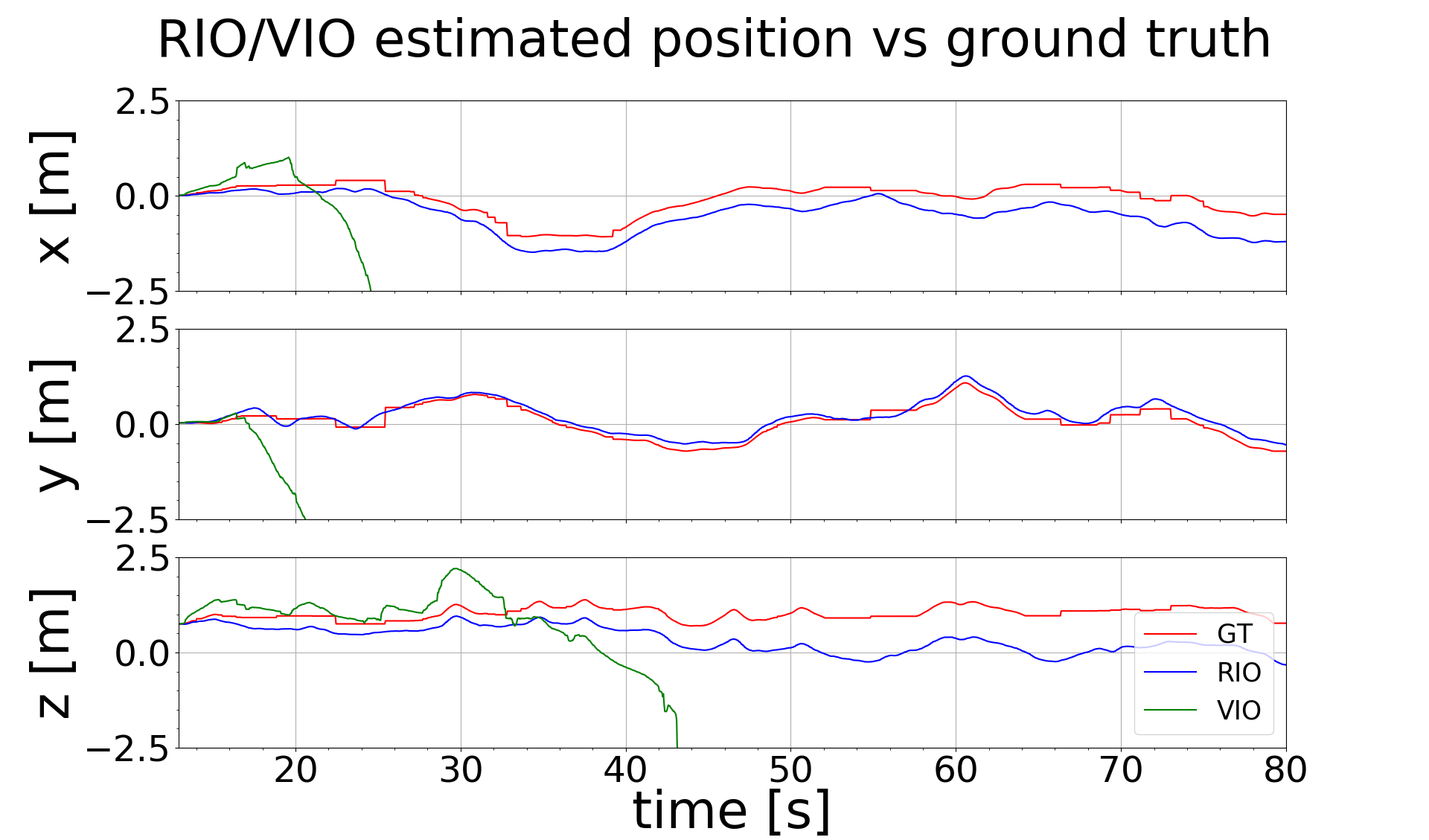}
    \caption[Comparison of position estimates for the flight in the fog]{Comparison of position estimates of a \ac{vio} implementation \cite{geneva2020openvins} and our \ac{rio} approach. Note the almost immediate divergence of the \ac{vio} when faced with the dense artificial fog. As can also be seen, our \ac{rio} method proves to be robust against it.}
    \label{fig:cl_rio_vio_comparision}
\end{figure}

\begin{figure}[h!tbp]
  \centering
  \includegraphics[width=0.95\columnwidth]{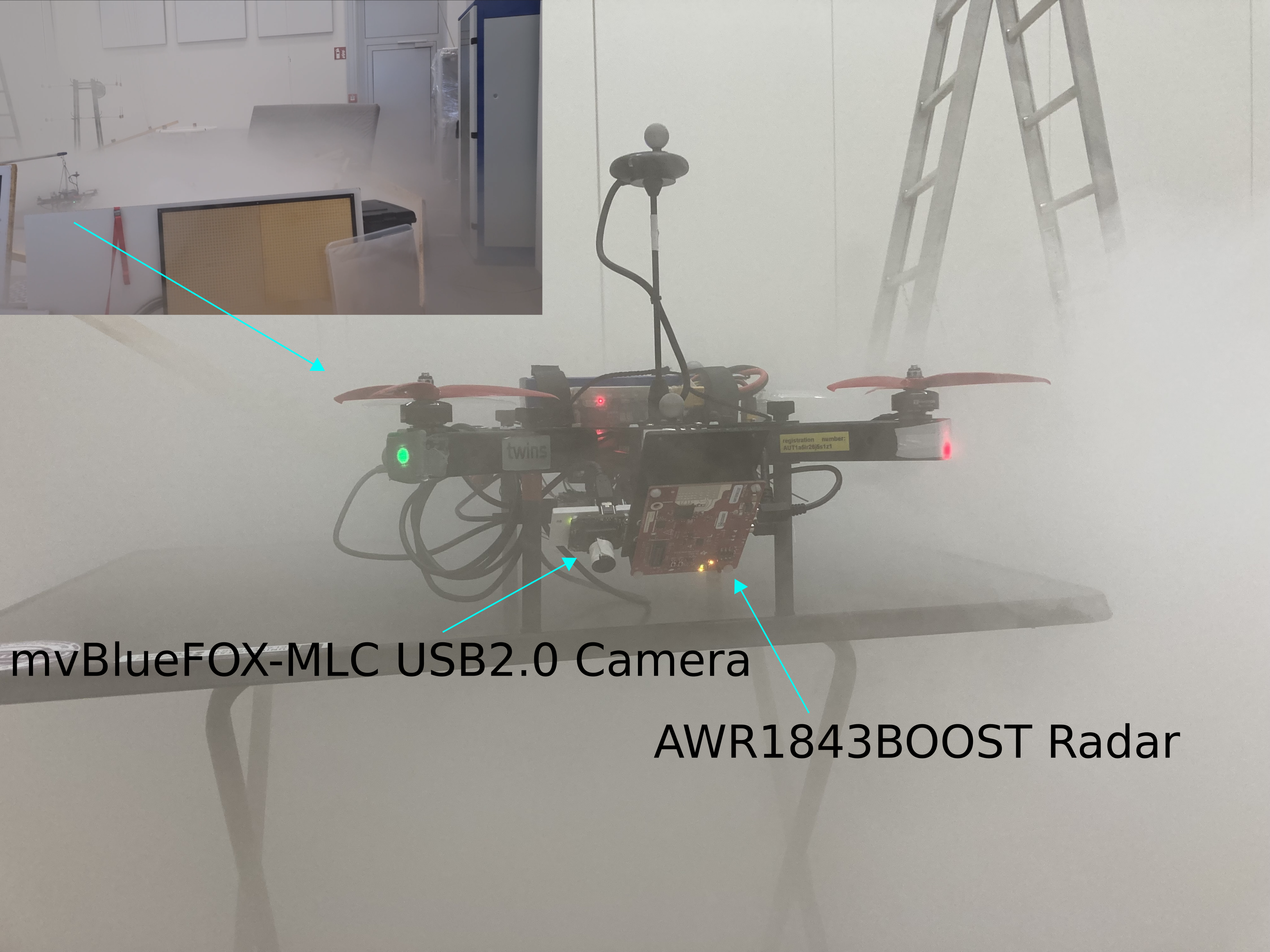}
  \caption[Experimental platform and the scene used in the flight in the fog evaluation]{Experimental platform used in this work with the \ac{fmcw} radar and camera sensors placed in the take-off position in dense artificial fog used to simulate a disaster site conditions. Note that the camera has been used only for comparison with \ac{vio} in the experiments with the artificial fog (see the~\cref{subsec:cl_fog}).}
  \label{fig:cl_fog_platform}
\end{figure}

\subsection{Conclusions}\label{subsec:ekf_rio_multi_conclusion}
In this section we presented a tightly-coupled and real-time capable \ac{ekf} \ac{rio} method which builds upon the approach presented in~\cref{sec:ekf_rio_single} but effectively leverages multi-state and persistent landmarks aspects from the vision community enhanced for the noisy, inaccurate, and sparse radar signals. In the presented framework, for correcting the drift of the \ac{imu}, during the update step we exploit lightweight and inexpensive \ac{fmcw} radar distance measurements to 3D points taken at several time instants in the past, distance measurements to persistent landmarks as well as Doppler velocity measurements. It is important to note that in our design we exploit the distance to 3D points to build residuals as opposed to just using 3D point locations, this is dictated by the observation that range (distance) measurements are very accurate in \ac{fmcw} radars. are We showed in real-world flight experiments that our method exhibits solid improvements over state-of-the-art in terms of accuracy and that it has execution times suitable for real time control. Indeed, closed-loop experiments showed that our method scales well towards commonplace issues related to real-time onboard operation of state estimation frameworks on resource-modest platforms such as sporadic measurement losses due to CPU load spikes. We also showed how the online calibration impacts the accuracy and the consistency of our state estimator. 

It is to be noted that as opposed to the \ac{vio} approaches, in our method the metric scale is directly observable thanks to the radar sensing principles (see~\cref{chap:fund_radar}). Hence, no special initialization involving acceleration excitation is needed as in \ac{vio} \cite{vins_shen}. This is advantageous since it renders the whole system easier to use.

Moreover, the presented odometry approach makes no assumptions on the environment and can be deployed in \ac{gnss}-denied settings. Additionally, given the employed sensor suite it is largely unaffected by conditions deemed challenging for other sensors used in \ac{uav} navigation, thus making it promising for applications such as search and rescue operations.  

\begin{sloppypar}
\section{Conclusions}\label{sec:ekf_rio_multi_conclusion_fin}
In this chapter we presented our progressive research on \ac{ekf}-based \ac{rio}. We started with a proof-of-concept environment-bound version where strongly reflective anchors whose position in space is known beforehand are used. We progressed towards a two-frame \ac{rio} where shifting to more powerful radar sensor allowed us to utilize the 4D radar point clouds to form residuals both on the Doppler velocities as well as feature distances. Finally, inspired by the approach in \cite{MSCKF-Paper}, we added support for several past frames, persistent features and online extrinsic calibration. We demonstrated our final approach in closed-loop flights and also in unfavorable environmental conditions.

The contents presented in this chapter put together into the final multi-state estimator bring the state-of-the-art in \ac{rio} close to the state of \ac{vio} thanks to the elements like persistent features, multi-state constraints update, and online extrinsic calibration. Remarkably, the achieved accuracy is similar to that obtained with \ac{vio}, with the added inherent benefits of the radar sensor in harsh environments which incapacitate cameras. All in all, the presented approach is highly performant and complimentary to \ac{vio}.

Since leveraging the features from \ac{vio} brings benefits to the \ac{rio} \ac{ekf} framework, it seems purposeful to use the same elements to implement a \ac{rio} estimator using the factor graph formalism to still boost the overall performance. This research avenue will be explored in the following chapter.
\end{sloppypar}

%% file: chapters/fg_rio.tex
\chapter[Radar-Inertial Odometry Using Factor Graphs][Radar-Inertial Odometry Using Factor Graphs]{Radar-Inertial Odometry Using Factor Graphs}\label{chap:fg_rio}

\begin{sloppypar}
\emph{The present chapter contains results that have been peer-reviewed and published in the IEEE/RSJ International Conference on Intelligent Robots and Systems (IROS)~\cite{jan_fg}.}
\end{sloppypar}

\bigskip

\begin{figure}[thpb]
  \centering
  \includegraphics[width=1.0\columnwidth]{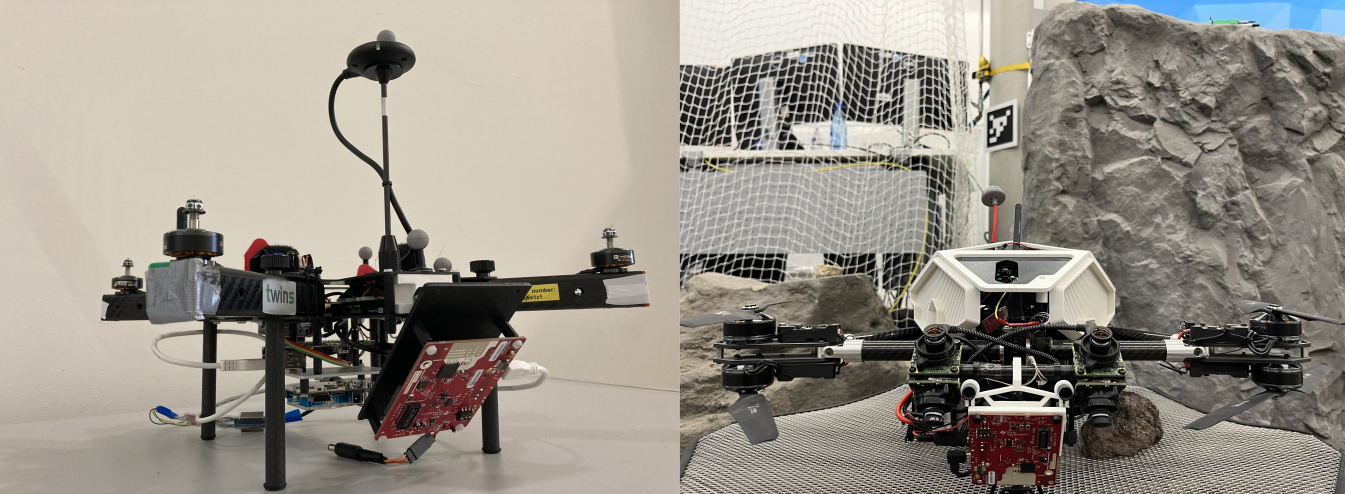}
  \caption[Experimental platforms CNS UAV and ARDEA-X]{Experimental platforms used in this work. CNS-UAV on the left and DLR's ARDEA-X on the right. The red chip mounted at \unit[45]{\degree} inclination on each platform is the TI AWR1843BOOST \ac{fmcw} radar, which outputs highly noisy and sparse 4D pointclouds (3D points and Doppler velocities).}
  \label{fig:fg_rio_platform}
\end{figure}

\noindent In this chapter, we present a novel \ac{rio} method based on the nonlinear optimization of factor graphs \cite{dellaert2017factor} allowing the estimation of the full 6DoF state of a small \ac{uav} using only \ac{imu} sensor and a light-weight, inexpensive, low-power Texas Instruments AWR1843BOOST \ac{fmcw} \ac{soc} radar (see~\cref{fig:fg_rio_platform}) in unknown and unprepared environments. The optimization problem underlying the state estimation task is maintained computationally tractable in real-time by employing a sliding window of states with the partial marginalization of oldest states using the Schur Complement technique. In our formulation we construct tightly-coupled radar factors from the distances to the 3D points matched between subsequent radar scans, instantaneous relative (Doppler) velocities and the distances to persistent landmarks. Tight coupling allows the construction of factors from single measurements, thus bypassing the necessity of computing pose increments from noisy and sparse radar point clouds using a method such as \ac{icp}, which in such case is prone to fail \cite{121791}. In an attempt to make an exact \textit{one-to-one} comparison with the state-of-the-art multi-state \ac{ekf}-based \ac{rio} approach in~\cref{sec:ekf_rio_multi}, we implement both methods in a single custom C++ framework where all front-end features such as point matching, measurement trails construction, velocity-based point pruning based on RANSAC etc. are shared with only estimation back-ends changing. To our knowledge it is the first such comparison of these two common estimation back-ends in the \ac{rio} context.

\subsection{Estimator Overview}\label{subsec:fg_rio_sys_overview}
In our \ac{rio} approach at every iteration of the estimator an optimization problem is formulated as a nonlinear factor graph over a sliding window of $N$ successive \ac{imu} states corresponding to time instants at which radar measurements were taken and a set of $L$ persistent landmarks.
We define the state variables of our system as follows: 

\begin{equation}\label{eq:fg_rio_state_imu}
\begin{aligned}
    \vx_{\cI} =& \left[\reference{{\vp}}{G}{I}{}; \reference{\bar{\vq}}{G}{I}{}; \reference{{\vv}}{G}{I}{}; \vb_{\va}; \vb_{\bomega}\right] \\
    \vx_{\cL} =& \left[ \reference{{\vp}}{G}{L}{} \right] \\
    \vX =& \left[ \vx_{\cI_{1}}; \ldots; \vx_{\cI_{N}}; \vx_{\cL_{1}}; \ldots; \vx_{\cL_{M}} \right]   
\end{aligned}
\end{equation}

with the \ac{imu} state $\vx_{\cI}$ and a state of a persistent landmark $\vx_{\cL}$. $\reference{{\vp}}{G}{I}{}$, $\reference{{\vv}}{G}{I}{}$, and $\reference{{\bar{\vq}}}{G}{I}{}$ are the position, velocity, and orientation of the \ac{imu}/body frame $\{I\}$  with respect to the navigation frame $\{G\}$, respectively. $\vb_{\bomega}$ and $\vb_{\va}$ are the measurement biases of the gyroscope and accelerometer, respectively. $\reference{{\vp}}{G}{L}{}$ define the position of a persistent landmark $\cL$ with respect to the navigation frame $\{G\}$. $\vX$ is the set of all states contained within a single sliding window. In order for the optimizer to solve the problem we must compute the jacobians of observation models with respect to the error-state which we define as follows:

\begin{equation}\label{eq:fg_rio_error_states}
\begin{aligned}
     \tilde{\vx}_{\cI} =& \left[ \reference{\tilde{\vp}}{G}{I}{}; \reference{\tilde{\btheta}}{G}{I}{}; \reference{\tilde{\vv}}{G}{I}{}; \tilde{\vb}_{\va}; \tilde{\vb}_{\bomega} \right]\\
     \tilde{\vx}_{\cL} =& \left[ \reference{\tilde{\vp}}{G}{L}{} \right]  
\end{aligned}
\end{equation}

For translational components, e.g., the position, the error is defined as $\reference{\tilde{\vp}}{G}{I}{} = \reference{\hat{\vp}}{G}{I}{} - \reference{{\vp}}{G}{I}{} $, while for rotations/quaternions it is defined as $\tilde{\bar{\vq}} = \hat{\bar{\vq}}^\inverse  \otimes \bar{\vq} = \left[ 1;  \frac{1}{2}\tilde{\btheta}\right]$, with $\otimes$ and $\tilde{\btheta}$ being quaternion product and small angle approximation, respectively. 

\begin{figure}[thpb]
  \centering
  \includegraphics[width=1.0\columnwidth]{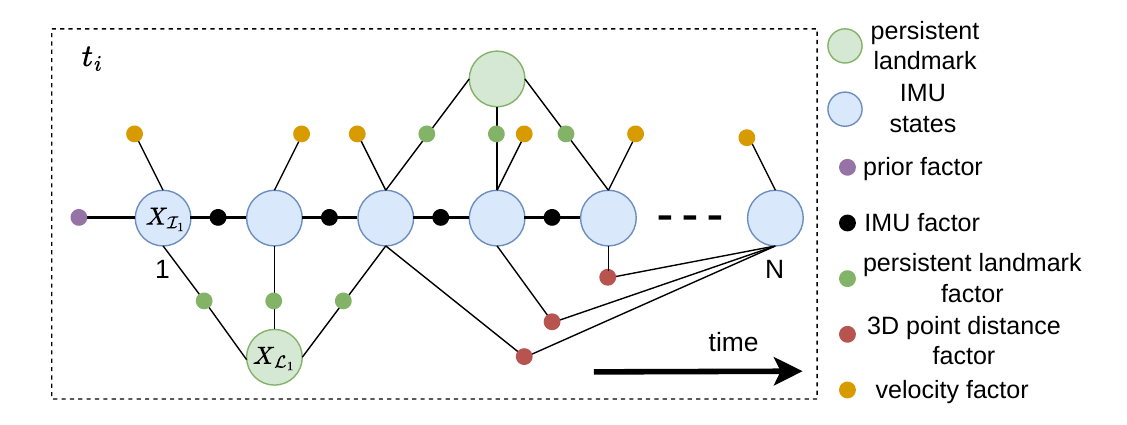}
  \caption[Snapshot of the sliding window of states and measurements and the corresponding factor graph]{Snapshot of the sliding window of states and measurements and the corresponding factor graph. Note the tightly-coupled nature of the graph, in which single measurements corresponding to Doppler velocities of 3D points, matched 3D points between radar scans and matched persisted landmarks are used to construct each factor. The tight coupling allows for maximal exploitation of sensor information. Note that the number of factors in the graph is only illustrative - in a real graph in our system there are many more factors involved.}
  \label{fig:fg_rio_graph_window}
\end{figure}
\begin{figure}[thpb]
  \centering
  \includegraphics[width=0.7\columnwidth]{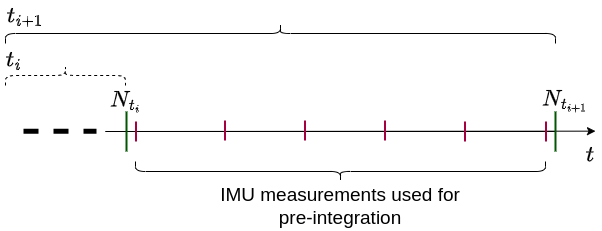}
  \caption[Initialization of the newest \ac{imu} state in the sliding window]{Initialization of the newest \ac{imu} state in the sliding window. Dashed and solid overbraces mark the sliding window before and after the shift forwards which occurs upon acquiring a new radar measurement (green vertical lines). The new \ac{imu} state $\vx_{\cI_{N}}(t_{i+1})$ in the window will be initialized with the prediction from the previous solution $\vx_{\cI_{N}}(t_{i})$ using the pre-integrated new \ac{imu} measurements (marked with red vertical lines) between the time instants of the previous and the current radar measurements.}
  \label{fig:fg_rio_integ}
\end{figure}

We depict the representation of a single factor graph corresponding to one sliding window of states in the Fig.~\ref{fig:fg_rio_graph_window}. The edges in the graph represent factors and nodes the estimated states. 
Each time a new radar measurement is received we form a new graph and solve it to obtain the estimate of $\vX$. To accommodate the new \ac{imu} state in the graph, we first marginalize out the oldest one and use it to form the new prior (see~\cref{subsec:fg_rio_marg}). We populate the new graph with factors corresponding to the \ac{imu} states which remained after the marginalization and the persistent landmarks seen from them. Then, we initialize all states in the window $\vX$ with their previous solutions. The new \ac{imu} state in $\vX$ is initialized from the last \ac{imu} state in the previous solution by applying the delta-motion from the pre-integrated \ac{imu} measurements from the timespan between the previous measurement (corresponding to the last estimated state) and the current one (see~\cref{fig:fg_rio_integ}). We solve the graph using Levenberg-Marquardt algorithm with the GTSAM package \cite{gtsam}.

The overall cost function for the proposed factor graph is as follows:

\begin{equation}\label{eq:fg_rio_cost}
\begin{aligned}
     \vX^{*} =&   \operatorname*{argmin}_{\vX} e_{I} + e_{P} + e_{D} + e_{V} + e_{L}
\end{aligned}
\end{equation}
where, $e_{I}$, $e_{P}$, $e_{D}$, $e_{V}$, $e_{L}$ are the cost contributions from all of the \ac{imu} pre-integration, prior, radar 3D distance, velocity, persistent landmarks factors present in the graph, respectively.  

\subsection{Factors}\label{subsec:fg_rio_factors}
In factor graph formulations, factors represent probabilistic constraints on the variables involved in the estimation and are obtained from measurements or prior knowledge. To define a factor we typically define a probabilistic measurement model constraining a subset of the state variables and upon creation, supply the corresponding measurement. Moreover, for solvers using the "\textit{lift-solve-retract}" paradigm \cite{Forster2015OnManifoldPF} we must provide a jacobian of the model with respect to the error-state defined in the tangent space of the state manifold. In~\cref{fig:fg_rio_graph_window}, we show all types of factors included in our factor graph approach. \ac{imu} pre-integration factor constrains two estimated \ac{imu} states and we construct it from all the \ac{imu} measurements obtained between the two constrained states. Using \ac{imu} pre-integration factor is necessary since the \ac{imu} measurements come at high frequency and not pre-integrating them would lead to a huge number of variables in the optimization. We use the factor formulated in \cite{Forster2015OnManifoldPF}.

Our tightly-coupled relative velocity factor is a unary factor, which means that it introduces constraints on the subset of the variables in only one \ac{imu} state. Hence, in the~\cref{fig:fg_rio_graph_window} these factors have connections only to one node representing the state at which the velocity measurements were taken. As seen in the~\cref{eq:fg_rio_velmeasmodel}, the constrained state variables are  $\reference{\bar{\vq}}{G}{I}{}$ and $\reference{{\vv}}{G}{I}{}$. The factor expresses the projection of the current robot ego-velocity transformed into the radar frame onto the direction vector pointing towards the corresponding 3D point:

\begin{equation}
\begin{aligned}\label{eq:fg_rio_velmeasmodel}
    \reference{\vv}{R}{P_i}{} =& -{\frac{\vr^\transpose}{\|\vr\|}} \left( \referencet{\vR}{I}{R}{}{\transpose} \referencet{\vR}{G}{I}{}{\transpose} \reference{\vv}{G}{I}{} + \right. \\
    & \left. \referencet{\vR}{I}{R}{}{\transpose} \left( \reference{\bomega}{}{}{I} \times \reference{\vp}{I}{R}{} \right) \right)
\end{aligned}
\end{equation}

where $\vr = \reference{{\vp}}{R}{P_{i}}{} $ is the 3D point detected in the current scan,  $\reference{\bomega}{}{}{I}$ is the current angular velocity of the \ac{imu} in the \ac{imu} frame, and $\reference{\vv}{G}{I}{}$ is the current linear velocity of the \ac{imu} in the navigation frame. In order to reject outliers, we use \ac{dcs} robust kernel function \cite{dcs} in this factor.

In order to further constrain the graph variables and make maximal use of the rich information provided by the radar, we build tightly-coupled factors for persistent landmarks and 3D point distance measurements organized in measurements trails. As we aim at an exact \textit{one-to-one} comparison with the method in~\cref{sec:ekf_rio_multi}, we use the exact same code for 3D point matching, obtaining measurement trails and persistent landmarks as in the chapter~\cref{chap:ekf_rio}.

We construct 3D point distance factors from a set of point trails which contain a history of continuous detections of the same 3D points (see~\cref{sec:ekf_rio_multi} for details). For all points in a trail we use the \ac{imu} states to transform all points $\referencet{\vp}{R}{P_{j}}{}{t_{p}}$ from the trail history at time instance $t_{p}$, where $p=1, \ldots, V$ and $V$ is the length of the matched trail, to the current radar reference frame:

\begin{equation}
\begin{aligned}\label{eq:fg_rio_propagated_distance}
  \referencet{{\vp'}}{R}{P_{j}}{}{t_{p}} = & \referencet{\vR}{I}{R}{}{\transpose} \left( -\reference{\vp}{I}{R}{} + (\referencet{\vR}{G}{I}{}{t_c})^\transpose \left(-\referencet{\vp}{G}{I}{}{t_{c}} + \right. \right. \\
    & \left. \left. \referencet{\vp}{G}{I}{}{t_{p}} + \referencet{\vR}{G}{I}{}{t_{p}} \left( \reference{\vp}{I}{R}{} + \reference{\vR}{I}{R}{} \referencet{\vp}{R}{P_{j}}{}{t_{p}}  \right) \right) \right)
\end{aligned}
\end{equation}

where $\reference{\vR}{I}{R}{}$ and $\reference{\vp}{I}{R}{}$ is the constant pose (orientation and position) of the radar frame with respect to the \ac{imu} frame. $\referencet{\vR}{G}{I}{}{\{t_{c}, t_{p}\}}$ and $\referencet{\vp}{G}{I}{}{\{t_{c}, t_{p}\}}$ are the \ac{imu} orientation and position corresponding to the trail history element at time $t_{p}$ and current radar scan at $t_{c}$, with respect to the navigation frame $\{G\}$. For factor construction we use the distance to the transformed matched point:

\begin{equation}
\begin{aligned}\label{eq:fg_rio_distmeasmodel}
    d_{P_{j}}= {} & \Big\| \referencet{{\vp'}}{R}{P_{j}}{}{t_{p}} \Big\|
\end{aligned}
\end{equation}

where $d_{P_{j}}$ is the distance to a single point $j$ in the matched trail history $\referencet{\vp'}{R}{P_{j}}{}{t_{p}}$ at $t_{p}$ aligned to the current radar pose at $t_{c}$. 

We also use persistent landmarks to build factors which introduce constraints between the landmarks and the \ac{imu} states from which these landmarks have been seen (\cref{fig:fg_rio_graph_window}). Promotion of trails with sufficiently long history detection to persistent landmarks is described in details in~\cref{sec:ekf_rio_multi}. Measurement model used in the factor is:

\begin{equation}\label{eq:fg_rio_landmarkup}
    \begin{aligned}
    {\vl'}_{m} &= \reference{\vp}{R}{L_{m}}{} =  \referencet{\vR}{I}{R}{}{\transpose} \big( \big. \referencet{\vR}{G}{I}{}{\transpose} 
    \big( \big. {\vl}_{m} - \reference{\vp}{G}{I}{}{} \big. \big) - \reference{\vp}{I}{R}{}{} \big. \big),
    \end{aligned}
\end{equation}

\begin{equation}
\begin{aligned}\label{eq:fg_rio_landmarkdist}
    d_{\vl_{m}}= {} & \Big\| {\vl'}_{m} \Big\|
\end{aligned}
\end{equation}

As for the velocity factor, for both 3D points and persistent landmark factors, we use \ac{dcs} for outlier rejection. Compared to the chi-squared-based outlier rejection in~\cref{sec:ekf_rio_multi}, using the \ac{dcs} does not remove the factors judged as out-of-distribution based on their residuals, only down-weigh them.

All 3D points delivered by radar and used for constructing factors are pruned for outliers using RANSAC similarly to \cite{doer2020ekf}.

\subsection{Partial Marginalization}\label{subsec:fg_rio_marg}
\begin{figure}[thpb]
  \centering
  \includegraphics[width=1.0\columnwidth]{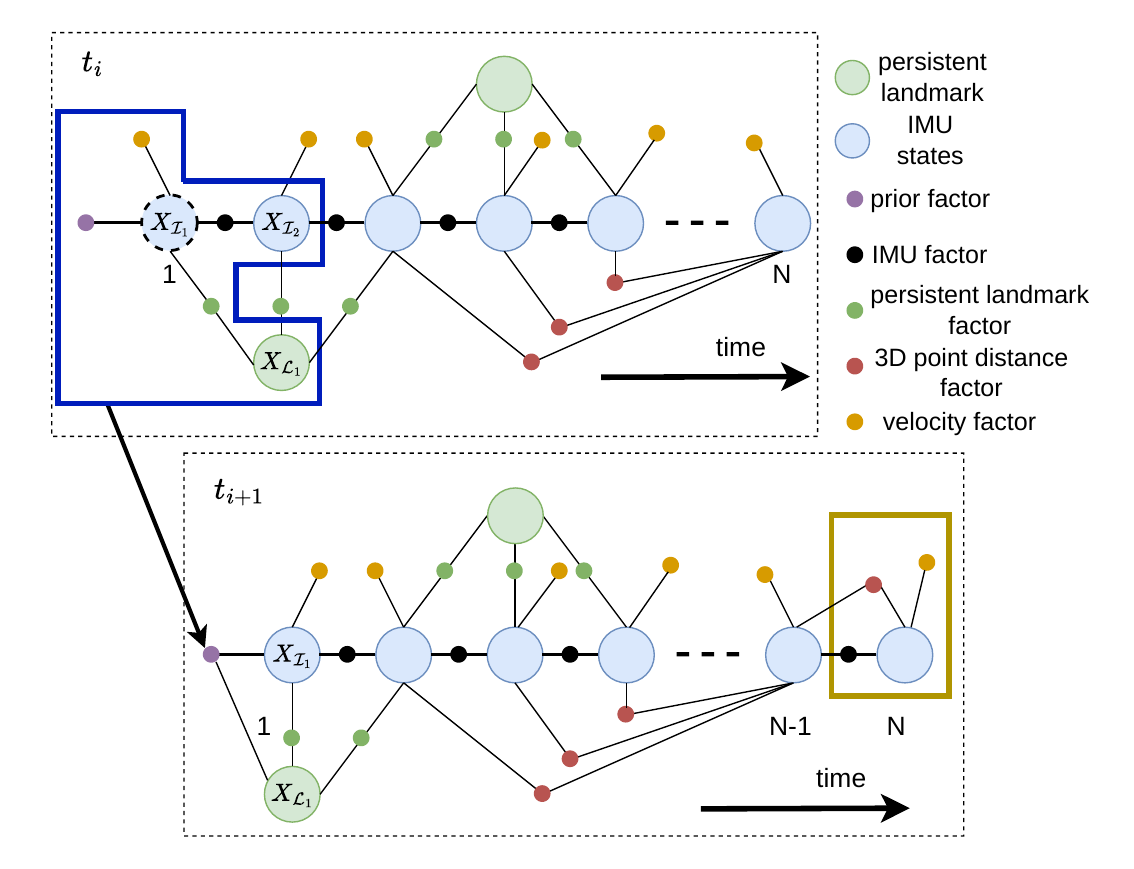}
  \caption[Marginalization of the oldest \ac{imu} state]{Depiction of the marginalization of the oldest \ac{imu} state (circle with dashed black line). Sub-graph (marked with the blue rectangle) is formed from the state to marginalize out and the states it is connected to (\textit{Markov blanket}). We linearize the sub-graph and in the obtained Gauss-Newton system of equations we apply the Schur Complement to the marginalized out variables to form the new prior factor. Newly added states and factors are marked with the yellow rectangle. Note that the indices adjust in the window from time $t_i$ to $t_{i+1}$ (e.g., $\vx_{\cI_{k}}(t_{i})$ becomes $\vx_{\cI_{k-1}}(t_{i+1})$) as we only keep a maximum of $N$ \ac{imu} states.}
  \label{fig:fg_rio_marg_graph}
\end{figure}

As the robot evolves in its environment, new \ac{imu} states and persistent landmarks are being added to the state vector. Nonetheless, to keep the state estimation task computationally feasible, we must bound the number of variables in the underlying optimization problem. Hence, upon the addition of new states we must remove the oldest ones by marginalizing them out. In our system we achieve marginalization with the Schur Complement technique (see~\cref{fig:fg_rio_marg_graph}). Namely, when forming a new graph upon obtaining a new radar measurement, due to conditional independence \cite{Leutenegger}, we consider a sub-graph containing only the states to be marginalized out and the states connected to these states (sometimes called \textit{Markov blanket}). We linearize the resulting sub-graph around the previous solution (current estimate) to obtain its hessian matrix and gradient vector:

\begin{equation}
\begin{aligned}\label{eq:fg_rio_schur}
\begin{bmatrix} \vH_{\mu\mu} & \vH_{\mu\lambda} \\ \vH_{\lambda\mu} & \vH_{\lambda\lambda} \end{bmatrix} \begin{bmatrix} \tilde{\vx}_{\mu} \\ \tilde{\vx}_{\lambda} \end{bmatrix} = \begin{bmatrix} \vb_{\mu} \\ \vb_{\lambda} \end{bmatrix}
\end{aligned}
\end{equation}

Where $\mu$ and $\lambda$ denote the sets of states to marginalize out and states connected to those states, respectively. We calculate the Schur Complement of the states to marginalize out in the hessian and the corresponding gradient:

\begin{equation}
\begin{aligned}\label{eq:fg_rio_schur1}
 \vH_{\lambda\lambda}^{*} =& \vH_{\lambda\lambda} - \vH_{\lambda\mu}\vH_{\mu\mu}^{-1}\vH_{\mu\lambda} \\
 \vb_{\lambda}^{*} =& \vb_{\lambda} - \vH_{\lambda\mu}\vH_{\mu\mu}^{-1}\vb_{\mu}\\
\end{aligned}
\end{equation}

and use the resulting matrix and vector to form the new prior factor. 

It is important to note that the marginalization allows for retaining the information from residuals that depend on the removed variables, nevertheless during optimization the linearization points of the marginalized out variables are not updated. Similarly, variables within the Markov blanket do not get their linearization points updated in order not to destroy the nullspaces of the respective Jacobian matrices which would otherwise introduce inconsistencies \cite{marg_cremers}. As noted in \cite{vins_shen} early fixing of the linearization points may result in suboptimal estimates. Since the current marginalization prior will be utilized to calculate the subsequent prior, the linearization points of the marginalized out variables will remain in the prior for the time the estimator operates, as opposed to filtering approaches, which are able to perfectly marginalize out old variables.

\section{Results}\label{sec:fg_rio_results}
\subsection{Experimental Setups}\label{subsec:fg_rio_platforms}
We perform real-world experiments with two different platforms (\cref{fig:fg_rio_platform}) to demonstrate our \ac{rio} method on different systems. One of the platforms (CNS-\ac{uav}) is described thoroughly in~\cref{sec:ekf_rio_multi}. The other one is the ARDEA-X \ac{uav} designed and built from ground up at the German Aerospace Center's (DLR) Institute of Robotics and Mechatronics. The system was designed for autonomous exploration of unknown regions in the context of planetary space robotics. An Intel NUC for higher-level and a Pixhawk for lower-level tasks make up ARDEA's two primary navigational components. The radar module is connected to the Intel NUC. The \ac{imu} data is obtained from the Pixhawk. On both platforms the same \ac{soc} \ac{fmcw} radar chip is mounted (Texas Instruments AWR1843) and configured in the same way (as in ~\cref{sec:ekf_rio_multi}). We gather datasets with each platform in which we record the ground truth trajectories using motion capture system and sensor readings from the \ac{imu} and radar. In the case of ARDEA-X we record the \ac{ekf} \ac{rio} estimates computed onboard. For CNS-\ac{uav} both factor graph and \ac{ekf} are executed offline on the recorded sensor data on an Intel Core i7-10850H vPRO laptop with $\unit[16]{GB}$ RAM in a custom C++ framework compiled with gcc 9.4.0 at -O3 optimization level. In the case of CNS-\ac{uav}, for both factor graph and \ac{ekf} \ac{rio} we manually calibrate the extrinsic parameters between the \ac{imu} and the radar. In the case of the ARDEA-X, for \ac{ekf} \ac{rio} these parameters are estimated online. Both estimators use exactly same front-end parameters and in both cases the sliding window size is set to $N = 10$. Execution time for our factor graph \ac{rio} on the above mentioned desktop PC is $\unit[16.15]{ms}$ on average which proves its real-time capability. Compared with $\unit[2.15]{ms}$ (propagation and update) for the \ac{ekf} \ac{rio} in~\cref{sec:ekf_rio_multi} the latter is (as expected) performing better.

\subsection{Evaluation}\label{subsec:fg_rio_comparison}
\begin{figure}[thpb]
  \centering
  \includegraphics[width=1.0\columnwidth]{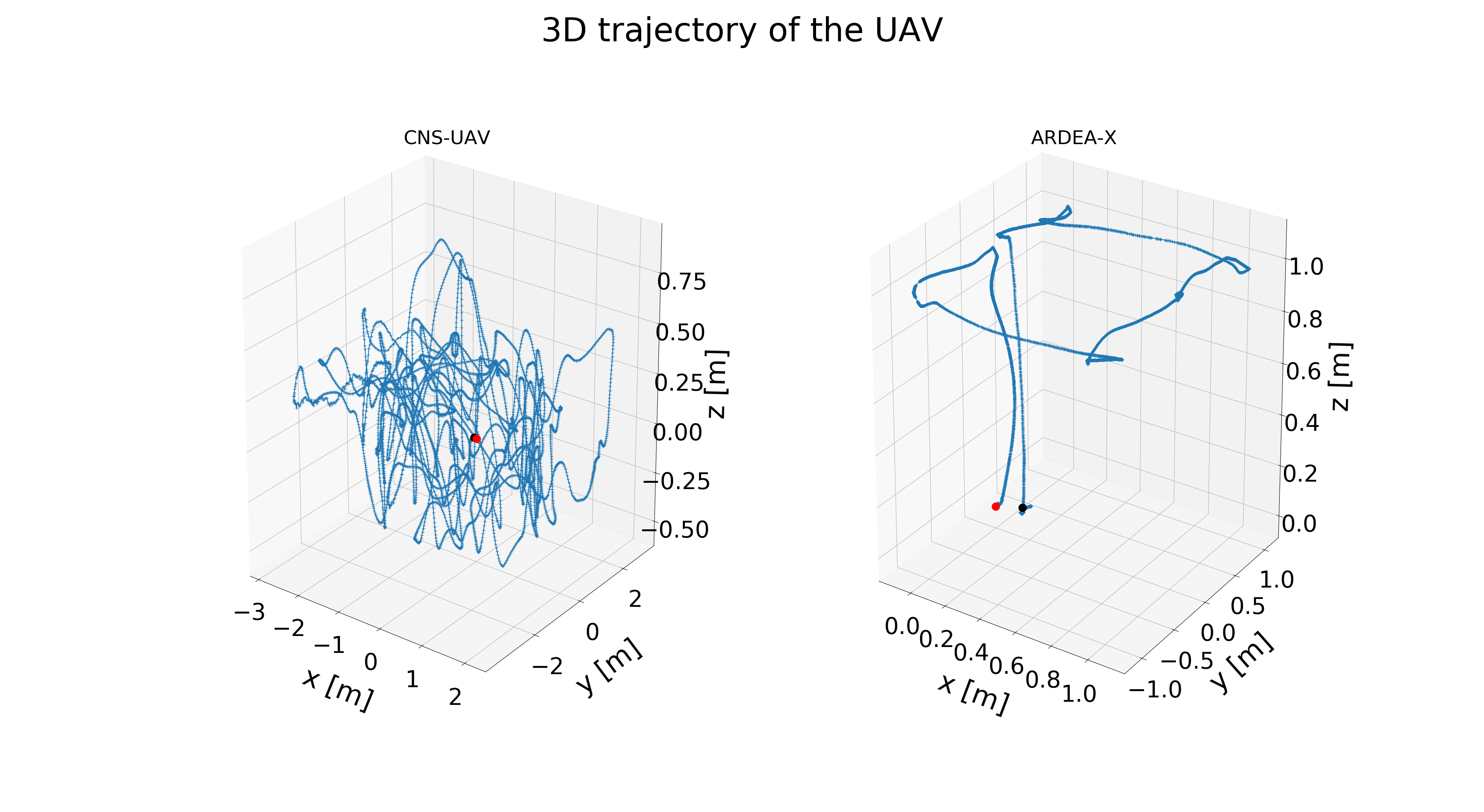}
  \caption[Sample trajectories from each dataset with the take-off and landing points]{Sample trajectories from each dataset with the take-off and landing points marked in red and black respectively. Trajectories collected with the CNS-\ac{uav} are not pre-planned, longer and in general more challenging in terms of dynamics.}
  \label{fig:fg_rio_3d}
\end{figure}

We use the data recorded with the two platforms described in~\cref{subsec:fg_rio_platforms} for evaluation of our factor graph \ac{rio} approach and for comparison with the \ac{ekf} \ac{rio} method from~\cref{sec:ekf_rio_multi}. With each of the platforms we create a dataset of several flown trajectories (five in case of ARDEA-X and three in case of CNS-\ac{uav}). Within the CNS-\ac{uav} dataset the trajectories are not pre-planned, manually flown, are between \unit[150]{m} - \unit[175]{m} long and include pronounced motions in all three dimensions. ARDEA-X dataset contains two pre-planned waypoint-based and three not pre-planned manually flown shorter trajectories. Sample trajectories from each dataset can be seen in the~\cref{fig:fg_rio_3d}. For every flight in each dataset we compute the norm of the \ac{rmse} of position together with the mean and standard deviation of the \ac{rmse} values (see~\cref{tab:fg_rio_rmses}). Our comparison shows that both methods perform similarly on average as seen in the~\cref{tab:fg_rio_rmses}. This is perhaps a counter-intuitive conclusion since the optimization of factor graphs is often considered to provide superior accuracy thanks to successive linearizations performed during the optimization. Nevertheless, in the case where the linearization point is well-determined, this advantage turns out not to be the crucial factor in determining the accuracy. Indeed, in the case of our experiments, the system is either initialized with the knowledge of the ground truth, or with its initial pose being the frame of reference for the estimator. Such settings leave very little room for any transient behaviours of the estimator. Thus, diminishing the value added from possible multiple linearizations.

\begin{figure}[thpb]
  \centering
  \includegraphics[width=1.0\columnwidth]{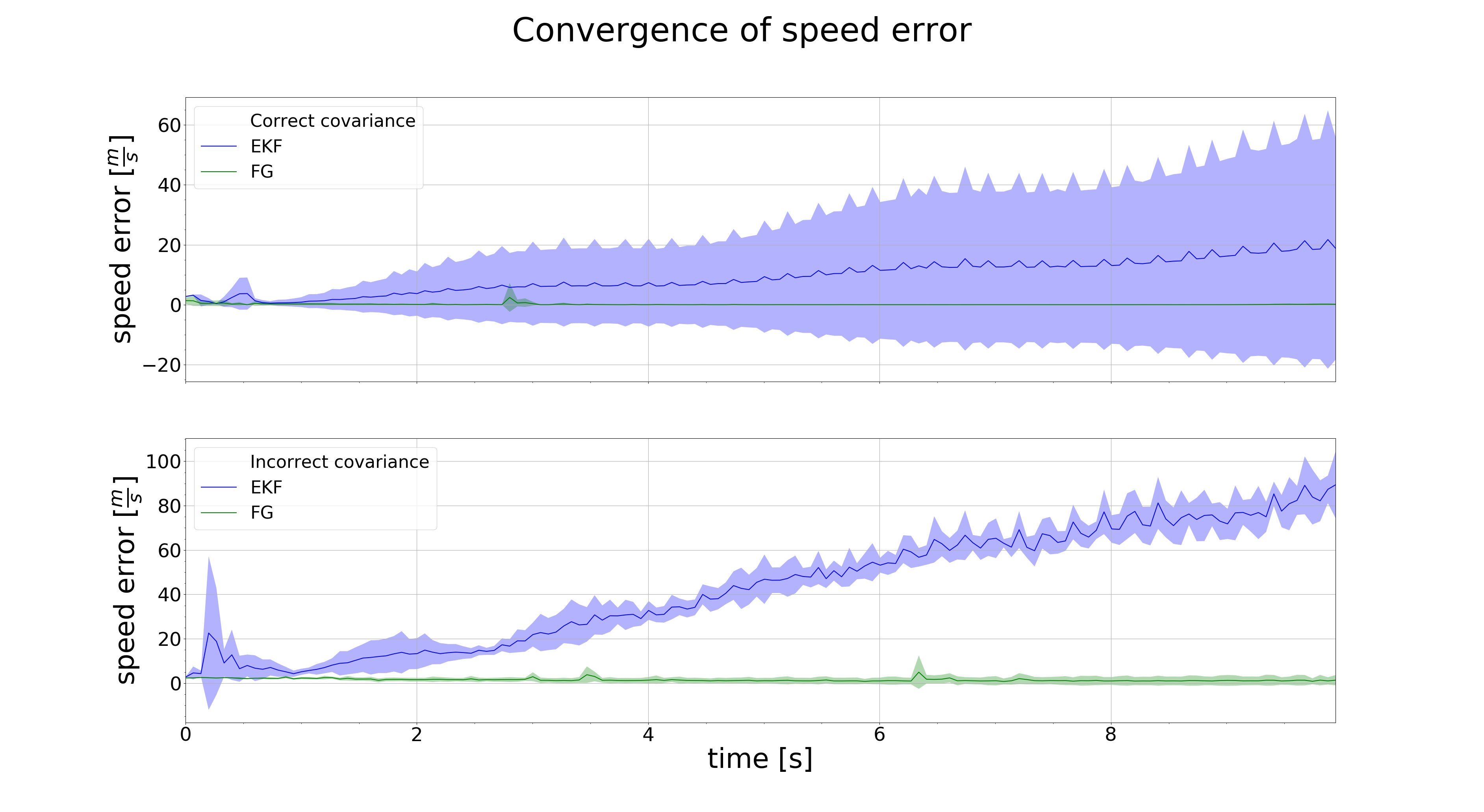}
  \caption[Plot of convergence of the mean of errors in speed (norm of the velocity) in the case of incorrect velocity initialization for both methods]{Plot of convergence of the mean of errors in speed (norm of the velocity) in the case of incorrect velocity initialization for both methods. The mean is taken across the whole ARDEA-X dataset during the first \unit[10]{s} of the data recording when the \ac{uav} stands still on the ground before the take-off. In the upper sub-plot, we initialize incorrectly the velocity, yet we account for this uncertainty by setting accordingly large corresponding covariance entries. The factor graph approach is unaffected and converges quickly to the correct value (\unit[0]{$\frac{m}{s}$}), whereas the \ac{ekf} never reaches convergence. In the lower sub-plot, for the same wrong initialization, we do not adjust the covariance settings, effectively misleading the estimator into believing it is initialized correctly. In this case the \ac{ekf} diverges catastrophically, while factor graph approach still converges although to a slightly offset value. Shaded regions denote the 1$\sigma$ bounds of the plotted means.}
  \label{fig:fg_rio_convergence}
\end{figure}

To demonstrate the benefits of iterative linearizations in the factor graphs during transient phases, we initialize both the \ac{ekf} and the factor graph-based estimators with wrong initial velocity set to [2.0, 2.0, 0.0]$\frac{m}{s}$ (whereas the true one is equal to [0.0, 0.0, 0.0]$\frac{m}{s}$). In the Fig.~\ref{fig:fg_rio_convergence} we show that in both cases when we do, and do not account for the wrong velocity initialization in the initial covariance, the factor graph-based approach always outperforms the \ac{ekf}.

\begin{figure}[thpb]
  \centering
  \includegraphics[width=1.0\columnwidth]{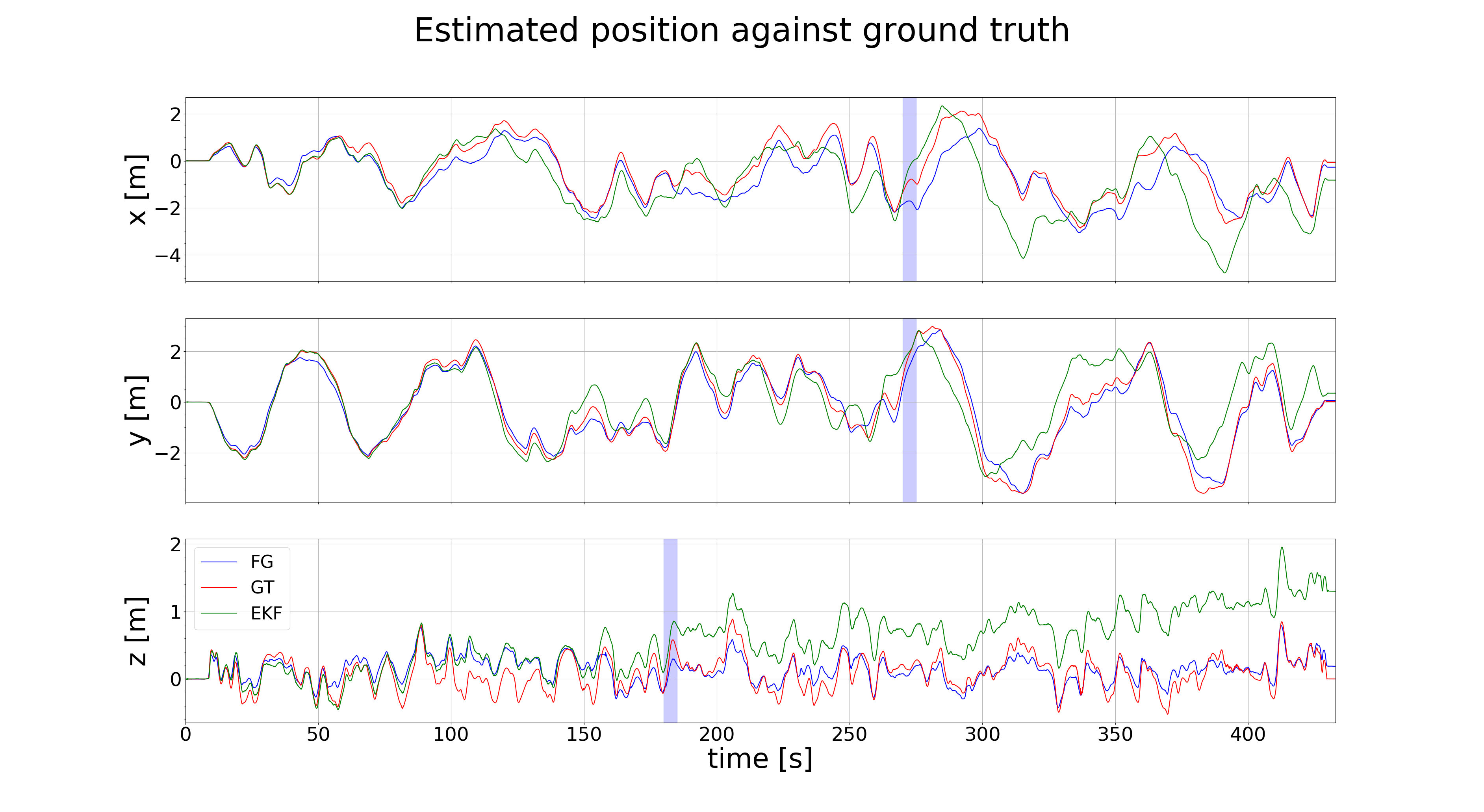}
  \caption[Estimated position of the UAV using both FG and EKF methods - trajectory 1]{Estimated position of the \ac{uav} using both methods for the first trajectory from the CNS-\ac{uav} dataset. The factor graph-based method performs visibly better with the norm of \ac{rmse} less than the half of the corresponding \ac{rmse} value for the \ac{ekf}-based method. The thin shaded regions mark points in time where the \ac{ekf}-based estimator started acquiring very heavy drift due to outliers.}
  \label{fig:fg_rio_awr7}
\end{figure}

\begin{figure}[thpb]
  \centering
  \includegraphics[width=1.0\columnwidth]{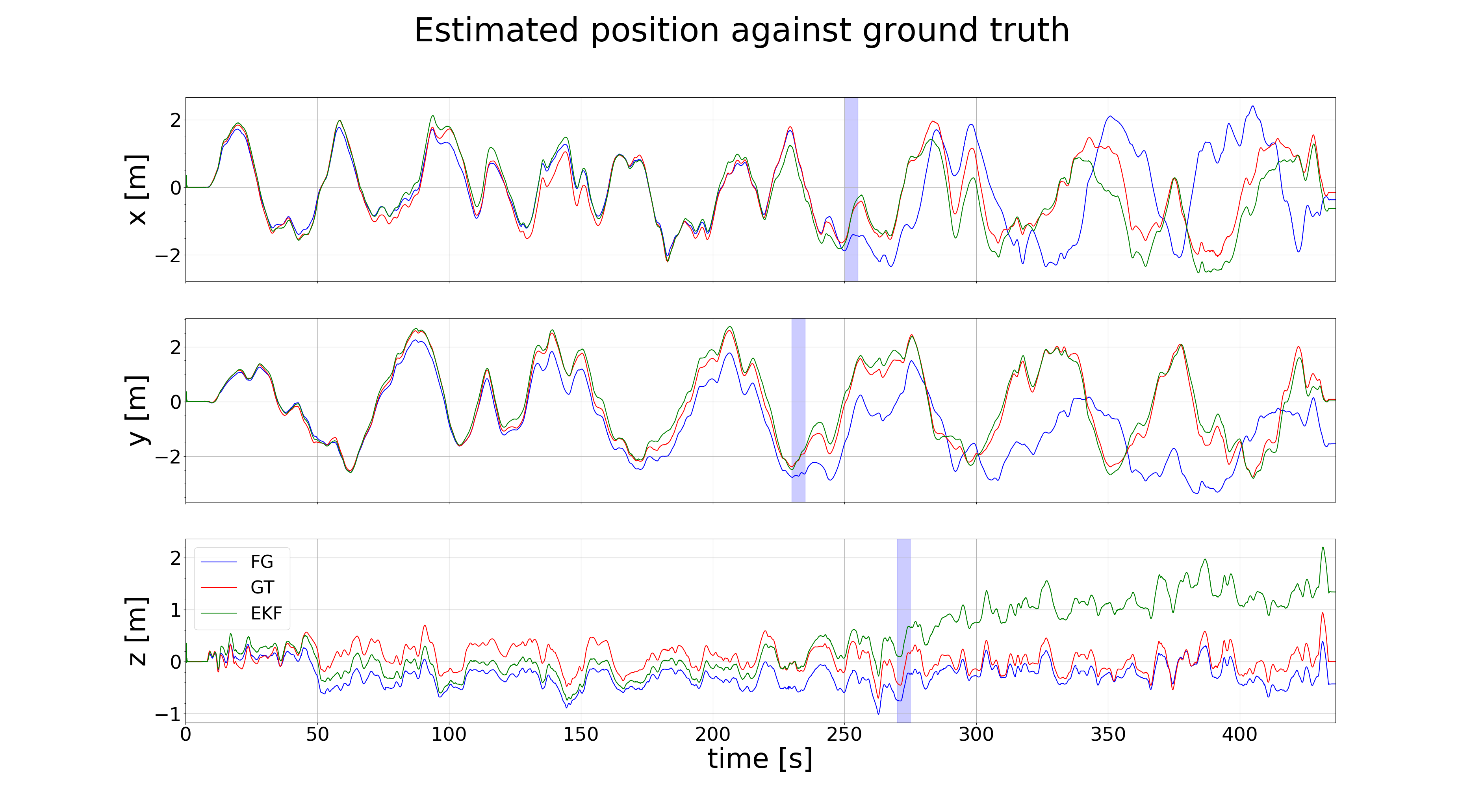}
  \caption[Estimation results for the position of the UAV using both FG and EKF methods - trajectory 2]{Estimation results for the position of the \ac{uav} using both methods for the second trajectory from the CNS-\ac{uav} dataset. Thin shaded regions are marked as areas where the estimators failed to reject outliers. This resulted in major degradation of the accuracy. We see that in the first two sub-plots factor-graph based method acquires very strong yaw drift (x and y axes seem swapped) from which it does not recover. For the \ac{ekf} the concentration of the outliers in the marked region in the third sub-plot results in a strong vertical drift.}
  \label{fig:fg_rio_awr8}
\end{figure}

\begin{figure}[thpb]
  \centering
  \includegraphics[width=1.0\columnwidth]{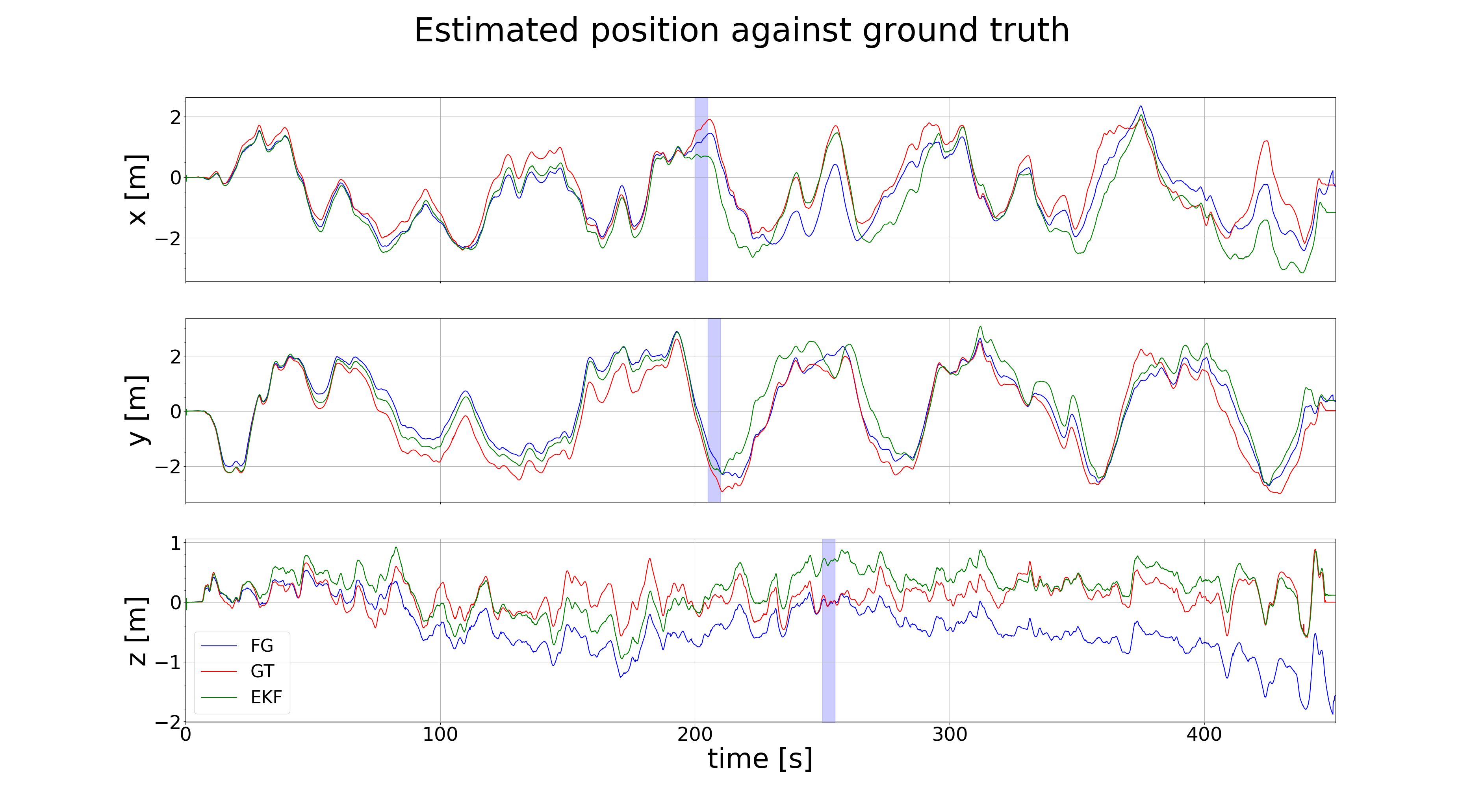}
  \caption[Plot of the estimates of position of the UAV using both FG and EKF methods - trajectory 3]{Plot of the estimates of position of the \ac{uav} using both methods for the third trajectory from the CNS-\ac{uav} dataset. In this case both estimators perform on-par in terms of \ac{rmse} with the outliers affecting the vertical drift for the factor graph-based estimator and $x, y$ coordinates for the \ac{ekf}-based one.}
  \label{fig:fg_rio_awr10}
\end{figure}

In our comparison we note the importance of the front-end in any state estimation system. While both \ac{rio} methods perform similarly on average, it turns out that there are qualitative differences between the estimation results of particular \ac{uav} trajectories (See~\cref{fig:fg_rio_awr7,fig:fg_rio_awr8,fig:fg_rio_awr10}). These discrepancies are attributed to different outlier rejection strategies between the \ac{ekf} and the factor graph-based \ac{rio}.

\begin{table}[]
\centering
\caption[Evaluation of position estimates for both EKF and FG methods]{Norm of \ac{rmse} values of position across flights performed with ARDEA-X and CNS-\ac{uav} for both methods}
\label{tab:fg_rio_rmses}
\begin{tabular}{|ccc|}
\hline
\multicolumn{3}{|c|}{ARDEA-X dataset} \\ \hline
\multicolumn{1}{|c|}{Nr} & \multicolumn{1}{c|}{\ac{rmse} Norm \ac{ekf}} & \ac{rmse} Norm FG \\ \hline
\multicolumn{1}{|c|}{1} & \multicolumn{1}{c|}{0.136} & 0.213 \\ \hline
\multicolumn{1}{|c|}{2} & \multicolumn{1}{c|}{0.083} & 0.153 \\ \hline
\multicolumn{1}{|c|}{3} & \multicolumn{1}{c|}{0.377} & 0.446 \\ \hline
\multicolumn{1}{|c|}{4} & \multicolumn{1}{c|}{0.698} & 0.711 \\ \hline
\multicolumn{1}{|c|}{5} & \multicolumn{1}{c|}{0.265} & 0.279 \\ \hline
\multicolumn{1}{|c|}{Average} & \multicolumn{1}{c|}{\textbf{0.312}} & \textbf{0.360} \\ \hline
\multicolumn{1}{|c|}{Std. dev.} & \multicolumn{1}{c|}{\textbf{0.218}} & \textbf{0.200} \\ \hline
\multicolumn{3}{|c|}{CNS-\ac{uav} dataset} \\ \hline
\multicolumn{1}{|c|}{1} & \multicolumn{1}{c|}{1.417} & 0.625 \\ \hline
\multicolumn{1}{|c|}{2} & \multicolumn{1}{c|}{0.877} & 1.660 \\ \hline
\multicolumn{1}{|c|}{3} & \multicolumn{1}{c|}{1.077} & 1.008 \\ \hline
\multicolumn{1}{|c|}{Average} & \multicolumn{1}{c|}{\textbf{1.124}} & \textbf{1.098} \\ \hline
\multicolumn{1}{|c|}{Std. dev.} & \multicolumn{1}{c|}{\textbf{0.223}} & \textbf{0.426} \\ \hline
\end{tabular}
\end{table}


\section{Conclusions}\label{sec:fg_rio_conclusion}
In this chapter we presented a novel tightly-coupled \ac{rio} method based on the nonlinear optimization of factor graphs in a sliding window of states and measurements. The used sensor suite is light-weight, inexpensive, low-power and consumer-grade, hence widely accessible. The presented approach is rendered real-time capable thanks to the application of the partial marginalization of oldest states in order to bound the state vector size and form an informative prior for the estimator. We performed a \textit{one-to-one} comparison with a state-of-the-art multi-state \ac{ekf} \ac{rio} method on the in-house datasets collected with two different \ac{uav} platforms in order to demonstrate the soundness of our framework. Comparing the two methods reveals that they perform on-par in terms of accuracy when the linearization point is not far from the true state. In terms of CPU load the comparison shows that the \ac{ekf} \ac{rio} is less resource-demanding. We demonstrated the advantages that successive linearizations in the factor graph-based method bring to the convergence of the estimator in transient phases (when the linearization point is far from the true state). We provided this demonstration in both the case when the wrong initialization is, and when it is not reflected in the uncertainty of the initial state (which is often the case in practice). The whole comparison was performed in such way that both \ac{ekf} and factor graph back-ends were implemented in the same software framework, effectively sharing exactly the same front-end features and parameters. We make the software framework and datasets used in the present paper open-source in order to facilitate further research and comparisons. 

The results in this chapter and in the recent research in~\cite{chuchu_iter} lead to an important question. Namely, \textit{do we need iterations in state estimators?} The accuracy comparison shows that on average, in nominal conditions, that is, when the linearization point is not far from the true state, the iterations within the factor graph bring no added value and are rather a waste of computational resources, since the accuracy attained with factor graph is comparable to non-iterated \ac{ekf}. The picture changes when disturbances from the true state are introduced, in which case the iterations allow faster convergence. It is to be noted that there exists a growing body of research within the area of equivariant filters which exhibit remarkable convergence properties without iterations~\cite{alessandro_eqf}.

Another interesting open research question appears when comparing the uncertainty representation in both \ac{ekf} and factor graph formulation and its potential impact on the accuracy on consistency of the estimation. Namely, when calculating the prior factor using the Schur complement technique, as far as the cross-correlations between the \ac{imu} poses in the sliding window are concerned, from one optimization run to another, only cross-terms between the marginalized pose and the one next to it are considered in the prior Hessian matrix. All other poses have no cross-terms unless a landmark seen from them has been marginalized out. This is strikingly different from the multi-state filter approach where stochastic cloning introduces cross-terms in the covariance matrix between the \ac{imu} poses within the maintained estimation window.

%% file: chapters/dl_matching.tex
\chapter[Sparse And Noisy 3D Radar Point Cloud Matching Using Deep Learning][Sparse And Noisy 3D Radar Point Cloud Matching Using Deep Learning]{Sparse And Noisy 3D Radar Point Cloud Matching Using Deep Learning}\label{chap:dl_matching}

\begin{sloppypar}
\emph{The present chapter contains results that have been peer-reviewed and accepted to the IEEE/RSJ International Conference on Intelligent Robots and Systems (IROS) in Hangzhou, China (2025)~\cite{DLmatching}.}
\end{sloppypar}

\bigskip

\noindent As demonstrated in the previous chapters, using 3D point correspondences is vital for state estimation using \ac{rio} systems. There exist well-established method for this task yet they prove unsuitable when the quality of the point clouds drops, as is the case of many \ac{fmcw} radar sensors. In this chapter we present a novel learning framework for predicting robust point correspondences between \ac{fmcw} \ac{soc} radar 3D point clouds in \ac{rio} estimation. Our framework is inspired by recent advances in deep learning for dense 3D point clouds processing in \cite{pct, dct} and tailored for the sparse and noisy radar measurements.

\section{Learning 3D Point Correspondences In Radar 3D Point Clouds}\label{sec:dl_net} 
We base our network architecture on the one defined in \cite{dct} for registration of dense and noiseless 3D point clouds of shapes, and adapt it to our setting of learning correspondences in variable-length, sparse and noisy \ac{soc} radar 3D point clouds. In particular, our framework consists of the following steps (\cref{fig:dl_netw}): 

\begin{enumerate}
  \item Calculate the input embeddings in a \textit{per-point} manner for each of the two consecutive input point clouds using the \textit{PointNet} architecture \cite{pointnet}.
  \item Use two transformer \cite{att} sub-networks to predict a matrix whose rows and columns correspond to the points in the first and second input point clouds, respectively, and whose entries express the degree of likelihood that the corresponding row and column form a match.
  \item Solve the \ac{lsa} optimization problem on the predicted matrix to find the set of point correspondences, where each item contains the index into the first (row) and second (column) point cloud.
  \item During training, cast the problem into a multi-label classification setting by considering the column index from the \ac{lsa} solution the class label of a point in the second point cloud, which allows leveraging the cross-entropy loss function.
  \item During inference, apply acceptance and \ac{fov} thresholds to the set of matches found in step 3 to form the output. 
\end{enumerate}

We train and test our network on a real-world, self-collected dataset consisting of 13 manually and autonomously flown \ac{uav} trajectories from which we select 8 for training and 5 for testing.
To make the validation more thorough and comparable, we also test our approach with the public Coloradar dataset \cite{colo}. Evaluation of our method in an open-source state-of-the-art EKF-based \ac{rio} framework from~\cref{sec:ekf_rio_multi} (which does not use a learning-based matching algorithm to find 3D point matches) shows an increase in estimation accuracy by over \unit[14]{\%} for the self-collected dataset and by \unit[19]{\%} for the Coloradar dataset, in terms of position RMSE norm. We also note that, when deactivating the Doppler information and only keeping the 3D point matches as correction information for the IMU integration in \ac{rio}, we note a difference in accuracy of more than \unit[70]{\%} when using our method on the self-collected dataset. To our knowledge, this is the first framework for learning point correspondences in sparse and noisy 3D point clouds as available from inexpensive \ac{soc} radar sensors.

\subsection{Network Architecture}\label{subsec:dl_arch}
As seen in~\cref{fig:dl_netw}, the first step in our network applies the \textit{PointNet} sub-network to two consecutive input point clouds to embed them in a higher-dimensional space. We obtain the input point clouds by finding the length $N$ of the longest point cloud in our whole dataset and padding all point clouds to that length with zero vectors of size $1\times3$. We also append a zero vector to the beginning of each point cloud which is necessary to train our network as a multi-label classifier (see~\cref{subsec:dl_train}). That way, each of the resulting input point clouds has the shape $(N + 1)\times3$. We apply the embedding sub-network on a \textit{per-point} basis, which means that for a single point in the input represented by three coordinates $\{x, y, z\}$ we obtain an embedding vector of size $1 \times E$, where $E$ is the chosen embeddings size. \textit{PointNet} parameters are learned and shared among all points in the input. In the next step, we forward the embedded points in each point cloud to two transformer sub-networks. The role of each of the transformers is to compute new embeddings of each point cloud using the contextual information of both point clouds jointly. That way, the network can leverage the attention mechanism on both point clouds together which permits finding the embeddings tailored to the specific kind of point clouds, thus making the embeddings task-specific \cite{dct}. Specifically, in each transformer block apart from passing to the decoder the output embeddings of the encoder, we also pass in the other point cloud input embeddings. The final embeddings are calculated by summing the transformer output, which encodes the mutual information about the point clouds, with the initial embeddings. The output of the network is obtained by calculating the dot product of the final embeddings of each point in the first point cloud with the final embeddings of each point in the second point cloud. This operation yields an output matrix of shape $(N+1) \times (N+1)$. Entries in the output matrix express the affinity between points in each input point cloud, that is, the likelihood that a pair of points form a correspondence.

\subsection{Network Training And Inference}\label{subsec:dl_train}
In the case of noisy, sparse and variable-length \ac{soc} radar point clouds, we cannot conveniently train the network on the odometry error using the \ac{svd} as in \cite{barnes2020under} and \cite{dct}. We thus propose a different approach to calculating the loss in our network. Namely, we treat the index of every point in the point cloud as its class label and reserve the class label (and the index) "0" for any non-matched points. The output matrix of the network is structured as follows, 

\begin{landscape}
\begin{figure}[thpb]
  \centering
  \includegraphics[width=1\columnwidth]{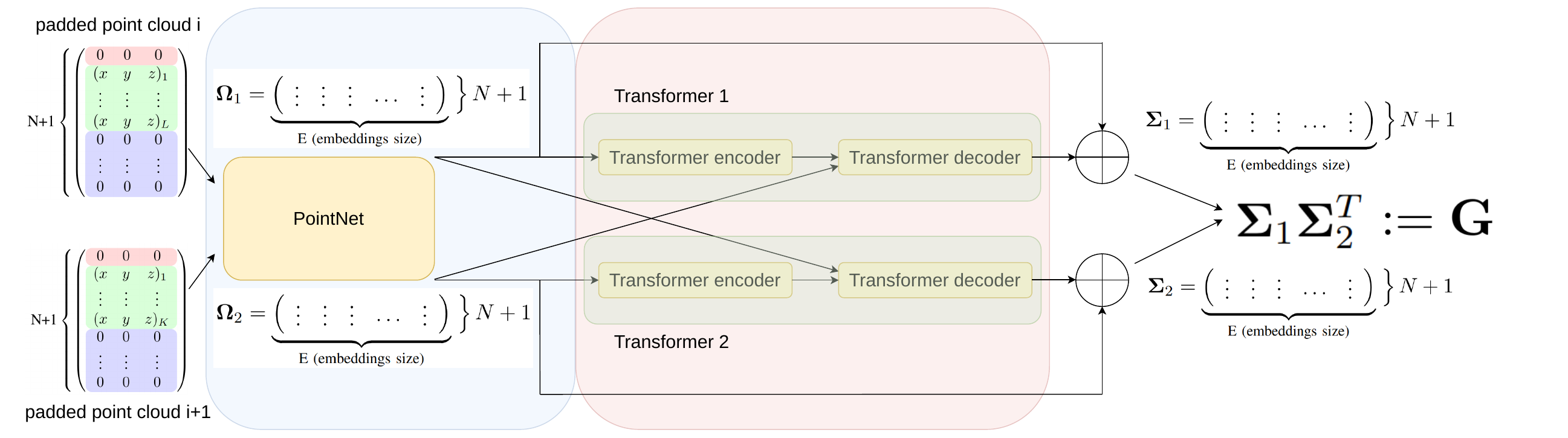}
  \caption[Architecture of the network for radar 3D point matching]{Our learning framework is based on the architecture proposed in \cite{dct} for dense and structured point clouds registration. We adapted it to our scenario with highly noisy and sparse consumer-grade \ac{soc} radar 3D point clouds. Note the zero padding of the input point clouds (\cref{subsec:dl_train}) and the pre-pended "0", which account for the variable-length input and the class label attributed to any point with no match, respectively. Input point clouds are passed to the embedding \textit{PointNet} sub-network (light blue block). Individual embeddings enter the transformer sub-networks (light red block) where the self and the reciprocal attention is computed for each point cloud. Output matrix $\vG$ representing the mutual affinity between points in the input point clouds is obtained by calculating the dot product of the final embeddings. Within the output we search for the set of matches by solving the \ac{lsa} (see~\cref{subsec:dl_train}).}
  \label{fig:dl_netw}
\end{figure}
\end{landscape}

\begin{equation}\label{eq:dl_output}
     \vG =  \underbrace{ \begin{pNiceMatrix}[margin]
      \bullet & \bullet & \cdots & \cdots & \cdots & \bullet & \bullet      \\
      \bullet & \Block[fill=green!15,rounded-corners]{4-4}{} c_{11}  & c_{12} & \cdots & c_{1K} & \cdots & \bullet \\
      \vdots  & c_{21}  & c_{22} & \cdots & c_{2K} & \cdots & \vdots \\
       \vdots & \vdots & \vdots & \ddots & \vdots & \cdots & \vdots \\
       \vdots & c_{L1} & c_{L2} & \cdots & c_{LK} & \cdots    & \vdots  \\ 
      \bullet       & \vdots   & \vdots & \vdots & \vdots &            \ddots                  & \bullet            \\
      \bullet       & \bullet              & \cdots  & \cdots & \cdots     & \bullet          & \bullet       \\
    \end{pNiceMatrix}}_{\displaystyle N+1}
    \left.\vphantom{
    \begin{pNiceMatrix}[margin]
      \bullet       & \bullet              & \cdots     & \bullet      \\
      \bullet       & \left[\begin{array}{cccc}
                            c_{11} & c_{12} & \cdots & c_{1K} \\
                             c_{21} & c_{22} & \cdots & c_{2K}\\
                             \vdots & \vdots & \ddots & \vdots \\
                             c_{L1} & c_{L2} & \cdots & c_{LK}
                       \end{array}\right]      & \cdots    & \bullet  \\ 
      \vdots       & \vdots              & \ddots                  & \bullet            \\
      \bullet       & \bullet              & \bullet               & \bullet       \\
    \end{pNiceMatrix}
    }\right\}\displaystyle{N+1}
\end{equation}

and is obtained by taking the dot product of the final embeddings $\bSigma_1$ and $\bSigma_2$ of the two consecutive input point clouds as shown in the Fig.~\ref{fig:dl_netw}. Entries in row $i$ express the likelihood that point $i$ in the first input point cloud is a correspondence to point $j$ in the second input point cloud, where $i = 1 ... L$, $j = 1 ... K$ and $L$, $K$ are lengths of the respective point clouds. Only the green sub-matrix in the output matrix in~\cref{eq:dl_output} carries useful information. All other elements result from adding the "0" (non-matched) class label and from padding the point clouds to equal length with zeros. In particular, appending the zero vector at the beginning of each input point cloud creates the $0$-th row and column in $\vG$. This is crucial during training, since we assign a "0" class ($0$-th index) in the ground-truth for every point in the first point cloud which does not have a match in the second point cloud.

During inference, since point clouds are usually of different lengths, we solve the \ac{lsa} problem on the green sub-matrix to find the optimal assignment,

\begin{align}
    min\sum_{i=1}^{L}\sum_{j=1}^{K}\vC_{i,j}\vX_{i,j} 
    \label{eq:dl_linsum}
\end{align}

Where $\vX$ is a boolean matrix where $\vX_{i,j}=1$ iff row $i$ is assigned to column $j$ and $L$, $K$ are lengths of the input point clouds. $\vC$ is the green sub-matrix from~\cref{eq:dl_output}. Constraints of the problem are such that each row is assigned to at most one column and each column to at most one row. For each entry in the solution we apply an experimentally chosen threshold to decide whether it is a match or not. \ac{lsa} is usually solved using Munkres algorithm \cite{munkres1957algorithms}.

During training, the structuring of the network output shown in~\cref{eq:dl_output} allows us to compute the cross-entropy loss between each row (the index of which is the index of a point in the first point cloud) and the ground-truth label (index of the matched point in the second point cloud or the "0" index for a non-match), as follows,

\begin{align}\label{eq:dl_cost}
    l_n= -\frac{1}{M}\sum_{i=1}^{M}log\left(\frac{exp(\vG(p_i, q_i))}{\sum_{j=1}^{N+1}exp(\vG(p_i, q_j))}\right) 
\end{align}

where $n = 1...B$ and $B$ is the mini-batch size, and $(p_i, q_i)$, $i = 1...M$ are the indices of the ground-truth matches in the first and second point cloud, respectively. $N+1$ is the length of each row (and column) of the $\vG$ matrix. 

Preparing the input data and ground-truth labels for training requires pre-processing. In order to generate the ground-truth point correspondences, we use the spatial transformation from the motion capture system between radar frames of every two consecutive radar measurements. Using the spatial information, we transform the 3D points from the first point cloud to the frame of the second point cloud and perform geometric matching by solving the \ac{lsa} optimization problem as in~\cref{eq:dl_output} but this time on a matrix whose entries are euclidean distances between points in both point clouds expressed in the second point cloud frame, as in,

\begin{align}\label{eq:dl_scorval_C}
    \vC_{i,j}= \| \referencet{\vp}{R_c}{P_{i}}{}{c} - (\referencet{\vR}{R_c}{R_p}{}{}\referencet{\vp}{R_p}{P_{j}}{}{p}+\reference{\vp}{R_c}{R_p}{}) \| 
\end{align}

where $\reference{\vp^{\{p, c\}}}{R_{\{p, c\}}}{P}{}$ are all points from the previous radar scan at time instance $t_{p}$ and from the current radar scan at $t_{c}$, in the previous and current radar frames, respectively. $\referencet{\vR}{R_c}{R_p}{}{}$ and $\reference{\vp}{R_c}{R_p}{}$ are rotation and translation parts of the spatial transform between the current and previous radar frames obtained from the motion capture system. That way, we obtain the ground-truth class labels (indices of matched points in each point cloud). We shift the obtained labels by one to account for the "0" class for every non-matched point. Pre-processing the input radar data consists only of removing the points outside of the \ac{fov} of the sensor and aforementioned zero padding.

\begin{figure}[thpb]
  \centering
  \includegraphics[width=1.0\columnwidth]{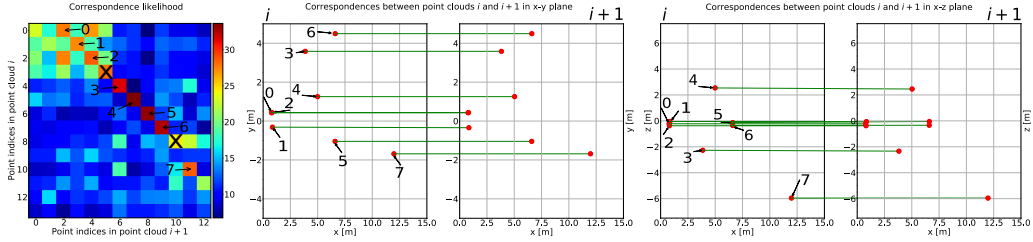}
  \caption[Inferred correspondence likelihood matrix and matches]{Leftmost part of the figure shows the correspondence likelihood matrix (the green part of the $\vG$ matrix in the~\cref{eq:dl_output}) inferred from the learned proposed model for two consecutive point clouds $i$ and $i+1$. Rightmost and the middle parts show the resulting 3D point correspondences for the same point clouds. Correspondences are shown on $xy$ (middle sub-plot) and $xz$ (right sub-plot) planes. Numbers marking the entries in the matrix are consistent with the indices of the matches in sub-plots. During inference, we prune the matches outside the \ac{fov} of the radar and with correspondence likelihoods below the empirically found acceptance threshold (shown as "x" in the leftmost plot). Thus, some entries in the matrix have no corresponding matches despite having relatively hot values. Only matched points are shown to not clutter the figure.}
  \label{fig:dl_corresp}
\end{figure}

\section{Results}\label{sec:dl_results}
\subsection{Experiments}\label{subsec:dl_exp}
\begin{figure}[thpb]
  \centering
  \includegraphics[width=1.0\columnwidth]{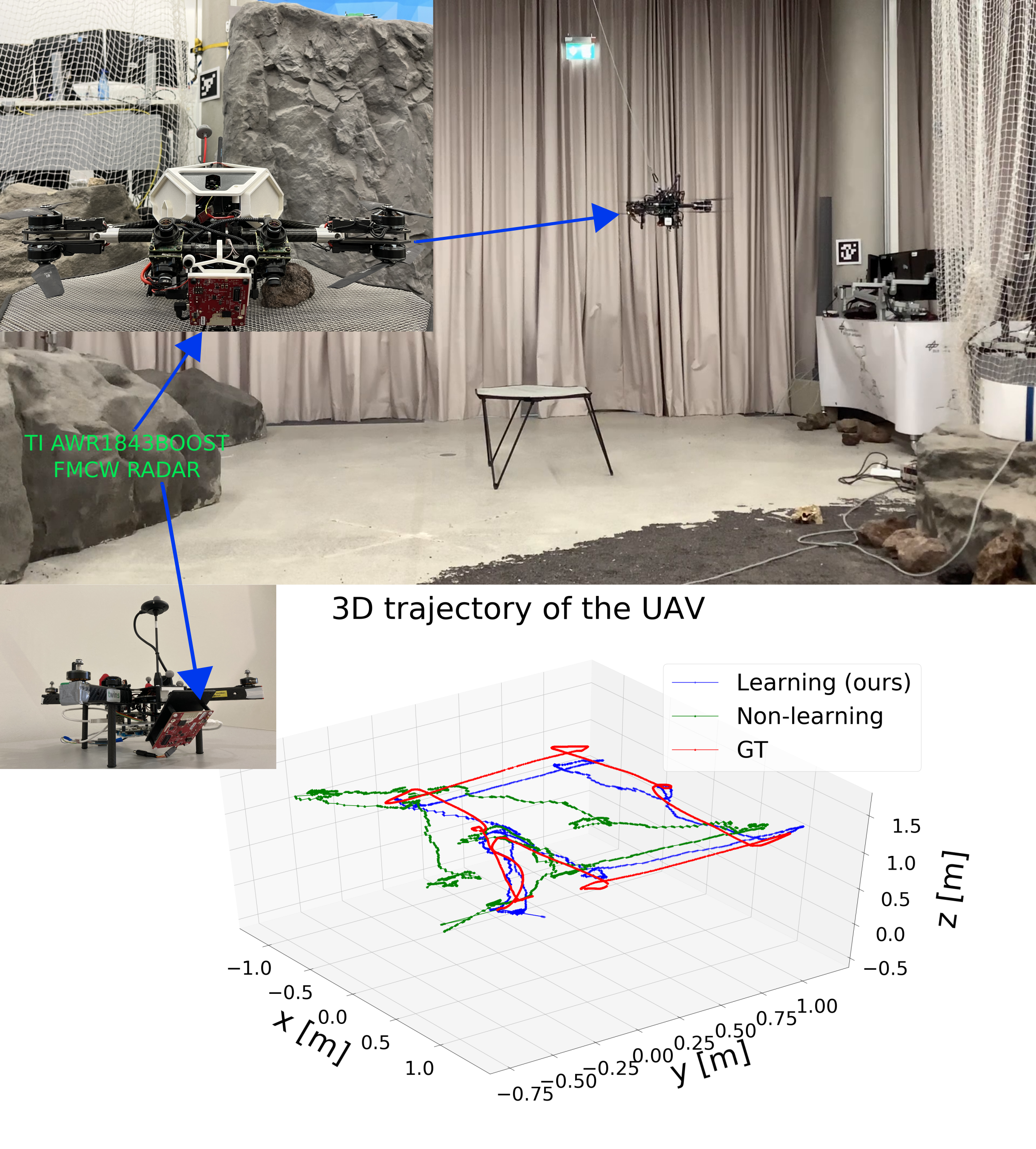}
  \caption[Platforms used in experiments and the Doppler-only trajectory]{ARDEA-X \cite{ardea} and CNS-\ac{uav} platforms used in this work with the mounted consumer-grade \ac{fmcw} \ac{soc} radar sensor. The radar chip that we use outputs highly noisy and sparse 4D point clouds (3D points and Doppler velocities). In the lower part of the figure we plot the estimated position for one of the validation flights using the EKF-based \ac{rio} framework from~\cref{sec:ekf_rio_multi} when switching between matching methods and using \textbf{\textit{solely}} 3D point matches in the update step. Note how the proposed method allows tracking the position of the \ac{uav} making \textbf{\textit{no}} use of the Doppler information, while the non-learning approach drifts considerably.}
  \label{fig:dl_platform}
\end{figure}

In order to train and validate our learning framework, we collect a dataset consisting of 13 \ac{uav} trajectories with two different platforms (ARDEA-X and CNS-\ac{uav}) described in~\cref{chap:fg_rio} (see~\cref{fig:dl_platform}). Both platforms use the same consumer-grade TI AWR1843BOOST \ac{fmcw} \ac{soc} radar chip mounted and configured in the same way, as well as the same pixhawk \ac{imu} sensor. We record radar and \ac{imu} sensor measurements as well as the ground truth pose of the \ac{uav} using a motion capture system. We divide this dataset into 8 training and 5 validation trajectories. The training dataset contains trajectories between \unit[150]{m} - \unit[180]{m} in length flown manually. The validation dataset contains shorter trajectories between \unit[11]{m} - \unit[38]{m}, among which some are manually flown while others are pre-planned, executed using specified waypoints. We also validate our method on five sequences from the public open-source Coloradar dataset \cite{colo}. Coloradar sequences are collected using hand-held sensor rig containing the same TI radar chip as ARDEA-X and CNS-\ac{uav} platforms. Coloradar sequences contain much more aggressive motion than the self-collected dataset, thus being more challenging. While in three sequences a motion capture system is used as the ground-truth, the other two contain high-precision \ac{lio} data. We train our network using PyTorch open-source package. We assess our 3D point matching framework qualitatively in an indirect way by plugging it into an open-source \ac{rio} framework from~\cref{sec:ekf_rio_multi} and comparing the accuracy of the obtained estimates to the case when the default (non-learning) matching algorithm is used. Between executions of the \ac{rio} estimator, we only exchange the matching algorithm, all other parameters and settings remain the same. The \ac{rio} which we use for validation is \ac{ekf}-based and in the update step uses three sources of information: 3D point matches, Doppler velocities and persistent features. For the self-collected dataset, we execute the \ac{rio} in two configurations: in the default configuration with both Doppler and point matches, and with only point matches enabled in the update step. For the Coloradar we only use default configuration (point matches and Doppler).
In all cases, we compile the \ac{rio} framework without persistent features in the update. In the case of our learning-based framework, we execute the inference on the learned model in a Python node before feeding it to the \ac{rio}. The inference with a non-optimized model takes on average \unit[0.0273]{s}, which means the optimized implementation would lend itself to real-time use. The non-learning matching algorithm is implemented within the \ac{rio} estimator as its default matching algorithm and described in~\cref{sec:ekf_rio_single}. Both \ac{rio} and inference node are executed offline on the recorded sensor data on an Intel Core i7-10850H vPRO laptop with 16 GB RAM.

\begin{figure}[thpb]
  \centering
  \includegraphics[width=1.0\columnwidth]{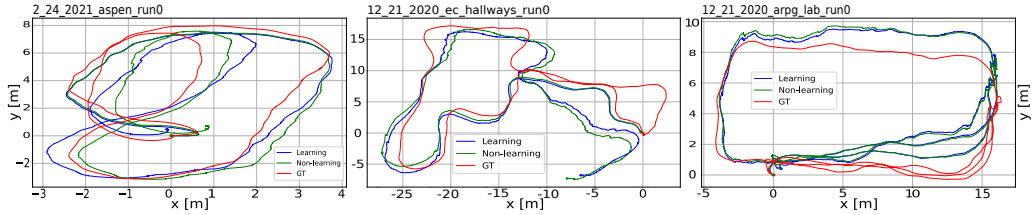}
  \caption[Method performance on coloradar dataset]{Three illustrative sequences (also used in \cite{poin_uncert}) out of five chosen from the Coloradar dataset for evaluation. From left to right: "2\_24\_2021\_aspen\_run0", "12\_21\_2020\_ec\_hallways\_run0", "12\_21\_2020\_arpg\_lab\_run0". In \textcolor{red}{red} we mark the ground-truth, and in \textcolor{green}{green} the non-learning and in \textcolor{blue}{blue} the learning (proposed) approaches, respectively. Across all used Coloradar sequences, using our learning-based matching method results in a decrease of the norm of position RMSE by \unit[19]{\%}.}
  \label{fig:dl_colo2d}
\end{figure}

\begin{figure}[thpb]
  \centering
  \includegraphics[width=1.\columnwidth]{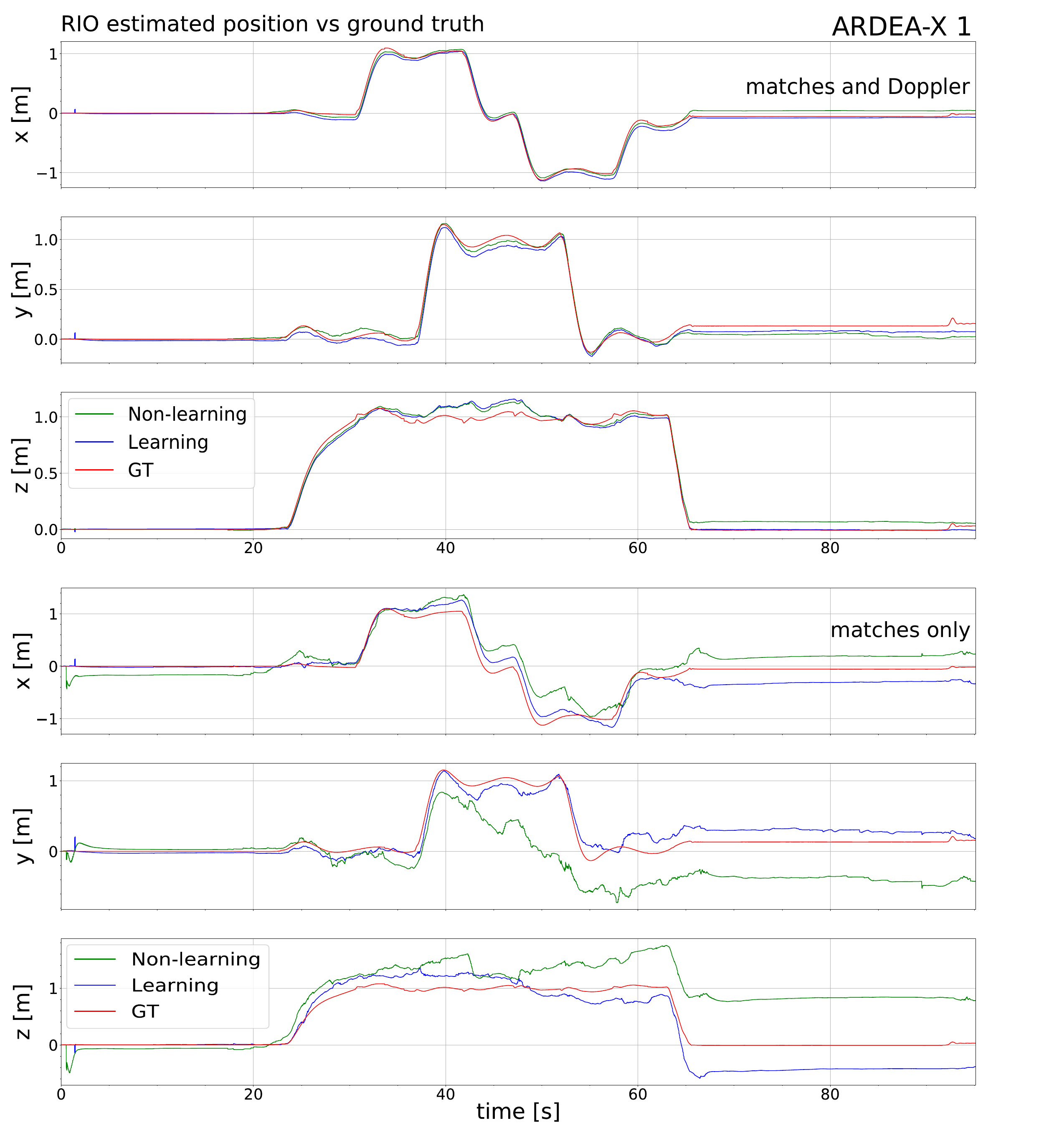}
  \caption[Position estimates for the ARDEA-X flight]{Estimated position of the ARDEA-X \ac{uav} for the flight $1$ from the~\cref{tab:dl_rmses_att} and~\cref{tab:dl_rmses}. We plot the $x, y, z$ coordinates of the estimate against the ground-truth for the configuration with matches and Doppler, and only matches used in the update step of the \ac{ekf} \ac{rio} framework used for validation. Each configuration is executed with the proposed learning-based and the default non-learning matching algorithm. In \textcolor{red}{red} the ground-truth, in \textcolor{green}{green} and \textcolor{blue}{blue} non-learning and learning approaches, respectively.}
  \label{fig:dl_2dpos}
\end{figure}

\begin{figure}[thpb]
  \centering
  \includegraphics[width=1.\columnwidth]{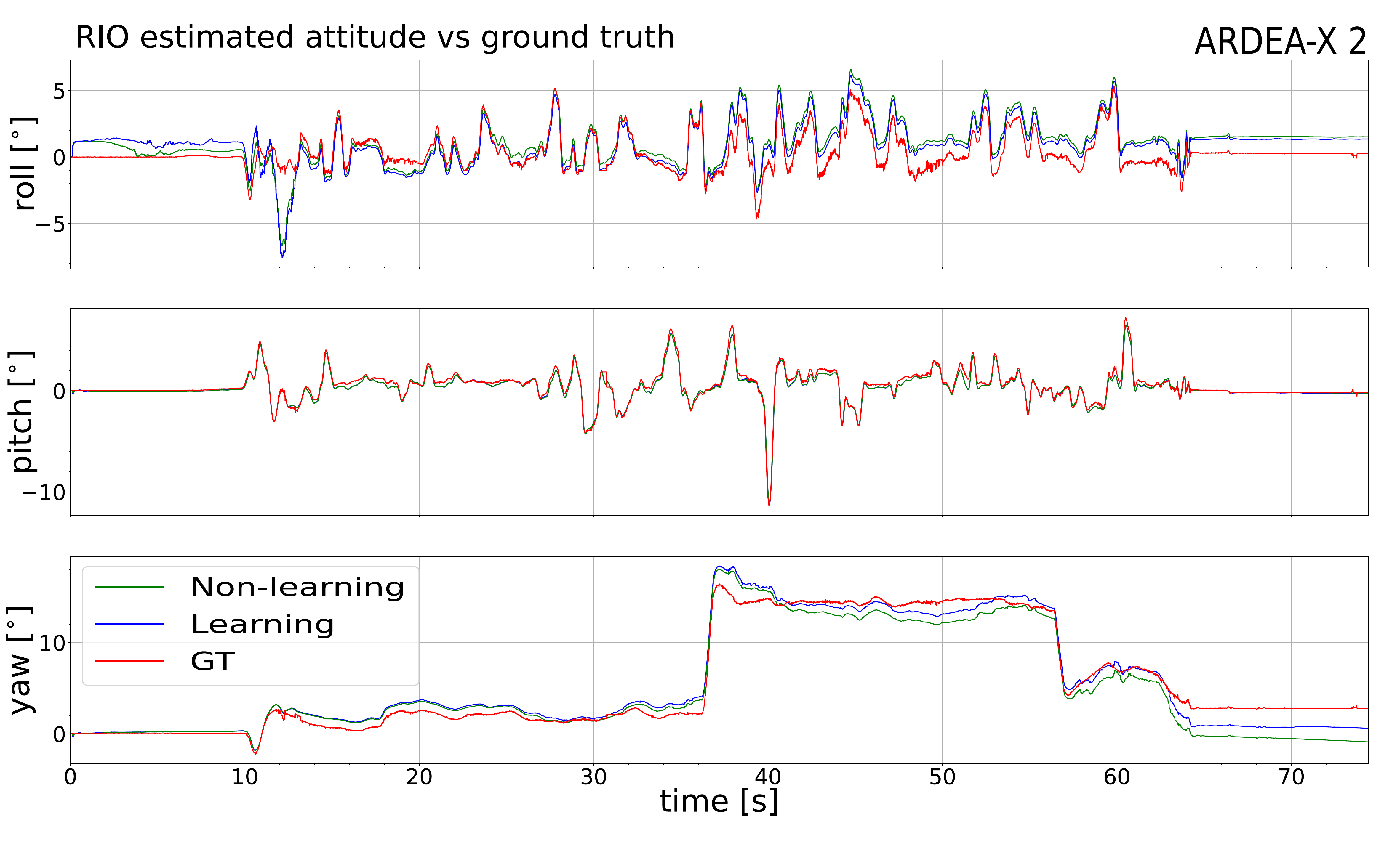}
  \caption[Attitude estimates for the ARDEA-X flight]{Estimated attitude of the ARDEA-X \ac{uav} for the flight $2$ from the~\cref{tab:dl_rmses_att} and~\cref{tab:dl_rmses}. We plot the roll, pitch and yaw angles of the estimate against the ground-truth for the configuration with matches and Doppler used in the update step of the EKF in \ac{rio} framework used for validation, when executed with the proposed learning-based (in \textcolor{blue}{blue}) and the default non-learning matching algorithm (in \textcolor{green}{green}). In \textcolor{red}{red} we plot the ground-truth attitude.}
  \label{fig:dl_2datt}
\end{figure}

\subsection{Evaluation}\label{subsec:dl_comparison}
For each trajectory from both validation datasets (self-collected and Coloradar), we compute the norm of \ac{rmse} of the position and attitude estimates along with the mean and the standard deviation of the obtained values when switching the matching algorithm inside the \ac{rio} estimator (see~\cref{tab:dl_rmses} and~\cref{tab:dl_rmses_att}). In the case of the self-collected dataset, we compute the norm of \ac{rmse} values in the case when both point matches and Doppler are used, and additionally, when only point matches residuals are used in the update step of the \ac{ekf} in the \ac{rio}. Our comparisons show that when only point matches are used in the update step, which is the most direct way of assessing the performance of our learning-based matching framework, we obtain a striking \unit[70.38]{\%} improvement in the position estimate accuracy on average. When compared to the state-of-the-art configuration of the \ac{rio}, that is, with both Doppler and point matches residuals enabled, we obtain a \unit[14.28]{\%} improvement in position accuracy on average. With the Coloradar dataset, we only execute the full configuration containing both Doppler and point matches residuals and obtain \unit[19.01]{\%} improvement in position accuracy on average. The motion in the Coloradar dataset is too aggressive for the configuration using only point matches to work properly (for both learning and non-learning). For our recorded dataset, as can be seen in~\cref{fig:dl_2dpos}, when point matches are used as the sole source of information for measurement updates in the \ac{ekf} \ac{rio}, the presented method allows for much more accurate estimation than the non-learning approach and when combined with the Doppler velocity measurements, greatly reduces the final error.

In~\cref{fig:dl_corresp}, we can see how the trained network infers the correspondences. Note how matches $0$ and $2$ lie very close geometrically and hence points involved in them have all high mutual affinities, nevertheless, the network still makes correct distinction between them. Points involved in match $1$, which also lie close to points involved in matches $0$ and $2$, do not have high affinity with points in $0$, $2$. This can be explained by looking at the the middle plot and observing that on the $xy$-plane, points in match $1$ are offset from points in $0$ and $2$. For points in matches $5$ and $6$, despite all of them being close on the $xz$-plane, our network correctly assigns the mutual affinities, since on the $xy$-plane, the points are clearly separated resulting in unambiguous matches. Points in matches $3, 4, 7$ are significantly spaced in both planes, thus all have strong unambiguous mutual affinity values. Points $(3, 5)$ are not considered a match despite their high mutual affinity because at least one of them is outside of the \ac{fov}. Similarly, the points $(8, 10)$ are not counted as a match since their correspondence likelihood value is below the empirically determined acceptance threshold.

In~\cref{fig:dl_colo2d} we plot the estimation results for three out of chosen five Coloradar sequences for both used matching methods. From the five sequences, "12\_21\_2020\_ec\_hallways\_run0", "2\_24\_2021\_aspen\_run0" and "12\_21\_2020\_arpg\_lab\_run0" are also chosen in the latest state-of-the-art work on \ac{rio} presented in \cite{poin_uncert} where the authors also provide the norm of \ac{rmse} of position and attitude estimate for their method. This allows us to note that for sequences "2\_24\_2021\_aspen\_run0" and "12\_21\_2020\_arpg\_lab\_run0" our method outperforms \cite{poin_uncert} $\unit[3.044]{m}$ (ours) to $\unit[3.820]{m}$ and $\unit[5.388]{m}$ (ours) to $\unit[6.101]{m}$, respectively in position and $\unit[13.372]{\degree}$ (ours) to $\unit[30.905]{\degree}$, and $\unit[8.287]{\degree}$ (ours) to $\unit[12.640]{\degree}$, respectively in attitude. For the sequence "12\_21\_2020\_ec\_hallways\_run0" our method performs worse, $\unit[9.927]{m}$ (ours) to $\unit[5.223]{m}$ in position as well as in attitude $\unit[18.357 ]{\degree}$ (ours) to $\unit[16.070]{\degree}$. 

Note, however, that in \cite{poin_uncert} the underlying estimator is different to the one we use here. Thus, we compare in~\cref{tab:dl_rmses} and~\cref{tab:dl_rmses_att} more rigorously the specific benefit of the learning-based matching.
For this, we implement the \ac{rio} framework of~\cref{sec:ekf_rio_multi} with Doppler and point matches as well as only point matches from our learning based approach and with the original non-learning approach.~\cref{tab:dl_rmses} clearly shows that our learning based approach improves the performance in the position estimation in almost all runs. When using both Doppler and point matches, the improvement using our approach is in average over \unit[14]{\%} or a bit over \unit[5]{cm} on our own datasets, and \unit[19]{\%} or \unit[8]{cm} on the Coloradar datasets. With a standard deviation much higher than the improvements, these results have limited statistical relevance. However, when using only point matches, the improvement is much higher with over \unit[70]{\%} or nearly \unit[1.5]{m} on our datasets. With a standard deviation of a bit over \unit[18]{cm}, this clearly underlines the estimation improvement due to our approach. The Coloradar dataset trajectories are too agile for the estimator to work properly when not including Doppler information, hence the 'x' in the lower right part of the table.

The benefit of our approach regarding the attitude estimation is less clear. Using both Doppler and point matches we observe in average a decrease in performance of \unit[5]{\%} or roughly half a degree on both our and the Coloradar datasets. When only using point matches, we observe in average a bit more than \unit[9]{\%} or nearly 1 degree performance drop on our datasets. Note, however, that these differences are barely statistically relevant since the standard deviation in all cases is higher than 5 degrees for our approach. Thus, we can conclude that while our method has a clearly positive impact on the position estimate, the attitude barely benefits from the new approach. The root cause of the reduced benefit in attitude is to be investigated further -- we assume a connection to the bad angular resolution and high angular noise that comes with this type of sensors.
We plot the attitude estimates for one of the trajectories from the self-collected dataset in the Fig.~\ref{fig:dl_2datt}. 

\begin{table}[]
\centering
\caption[Evaluation of position for both DL and classical matching methods]{\ac{rmse} norm values of position estimate for both matching methods across self-collected \ac{uav} flights and sequences from open-source Coloradar dataset.}
\label{tab:dl_rmses}
\begin{tabular}{|c|cccc|}
\hline
\multirow{3}{*}{Nr} & \multicolumn{4}{c|}{\begin{tabular}[c]{@{}c@{}}Self-collected dataset\\ $\|$RMSE$\|$ of position {[}m{]}\end{tabular}} \\ \cline{2-5} 
 & \multicolumn{2}{c|}{Doppler and matches} & \multicolumn{2}{c|}{Matches only} \\ \cline{2-5} 
 & \multicolumn{1}{c|}{\begin{tabular}[c]{@{}c@{}}Learning\\ (ours)\end{tabular}} & \multicolumn{1}{c|}{\begin{tabular}[c]{@{}c@{}}Non\\ learning\end{tabular}} & \multicolumn{1}{c|}{\begin{tabular}[c]{@{}c@{}}Learning\\ (ours)\end{tabular}} & \begin{tabular}[c]{@{}c@{}}Non\\ learning\end{tabular} \\ \hline
1 & \multicolumn{1}{c|}{\textbf{0.083}} & \multicolumn{1}{c|}{0.101} & \multicolumn{1}{c|}{\textbf{0.351}} & 0.748 \\ \hline
2 & \multicolumn{1}{c|}{\textbf{0.212}} & \multicolumn{1}{c|}{0.272} & \multicolumn{1}{c|}{\textbf{0.793}} & 1.795 \\ \hline
3 & \multicolumn{1}{c|}{0.714} & \multicolumn{1}{c|}{\textbf{0.704}} & \multicolumn{1}{c|}{\textbf{0.745}} & 1.608 \\ \hline
4 & \multicolumn{1}{c|}{0.380} & \multicolumn{1}{c|}{\textbf{0.338}} & \multicolumn{1}{c|}{\textbf{0.431}} & 1.074 \\ \hline
5 & \multicolumn{1}{c|}{\textbf{0.230}} & \multicolumn{1}{c|}{0.473} & \multicolumn{1}{c|}{\textbf{0.736}} & 5.088 \\ \hline
Average & \multicolumn{1}{c|}{0.324} & \multicolumn{1}{c|}{0.378} & \multicolumn{1}{c|}{0.611} & 2.063 \\ \hline
Std. dev. & \multicolumn{1}{c|}{0.217} & \multicolumn{1}{c|}{0.202} & \multicolumn{1}{c|}{0.183} & 1.558 \\ \hline
Sequence & \multicolumn{4}{c|}{\begin{tabular}[c]{@{}c@{}}Coloradar dataset\\ $\|$RMSE$\|$ of position {[}m{]}\end{tabular}} \\ \hline
aspen\_run0 & \multicolumn{1}{c|}{\textbf{3.044}} & \multicolumn{1}{c|}{5.327} & \multicolumn{1}{c|}{x} & x \\ \hline
arpg\_lab\_run0 & \multicolumn{1}{c|}{\textbf{5.388}} & \multicolumn{1}{c|}{6.080} & \multicolumn{1}{c|}{x} & x \\ \hline
\multicolumn{1}{|l|}{ec\_hallways\_run0} & \multicolumn{1}{c|}{\textbf{9.927}} & \multicolumn{1}{c|}{11.523} & \multicolumn{1}{c|}{x} & x \\ \hline
aspen\_run4 & \multicolumn{1}{c|}{\textbf{5.315}} & \multicolumn{1}{c|}{6.296} & \multicolumn{1}{c|}{x} & x \\ \hline
aspen\_run5 & \multicolumn{1}{c|}{\textbf{3.466}} & \multicolumn{1}{c|}{4.280} & \multicolumn{1}{c|}{x} & x \\ \hline
Average & \multicolumn{1}{c|}{5.428} & \multicolumn{1}{c|}{6.701} & \multicolumn{1}{c|}{x} & x \\ \hline
Std. dev. & \multicolumn{1}{c|}{2.728} & \multicolumn{1}{c|}{2.808} & \multicolumn{1}{c|}{x} & x \\ \hline
\end{tabular}
\end{table}

\begin{table}[]
\centering
\caption[Evaluation of attitude for both DL and classical matching methods]{\ac{rmse} norm values of attitude estimate for both matching methods across self-collected \ac{uav} flights and sequences from open-source Coloradar dataset.}
\label{tab:dl_rmses_att}
\begin{tabular}{ccccc}
\hline
\multicolumn{1}{|c|}{\multirow{3}{*}{Nr}} & \multicolumn{4}{c|}{\begin{tabular}[c]{@{}c@{}}Self-collected dataset\\ $\|$RMSE$\|$ of attitude {[}\degree{]}\end{tabular}} \\ \cline{2-5} 
\multicolumn{1}{|c|}{} & \multicolumn{2}{c|}{Matches and Doppler} & \multicolumn{2}{c|}{Matches only} \\ \cline{2-5} 
\multicolumn{1}{|c|}{} & \multicolumn{1}{c|}{\begin{tabular}[c]{@{}c@{}}Learning\\ (ours)\end{tabular}} & \multicolumn{1}{c|}{\begin{tabular}[c]{@{}c@{}}Non\\ learning\end{tabular}} & \multicolumn{1}{c|}{\begin{tabular}[c]{@{}c@{}}Learning\\ (ours)\end{tabular}} & \multicolumn{1}{c|}{\begin{tabular}[c]{@{}c@{}}Non\\ learning\end{tabular}} \\ \hline
\multicolumn{1}{|c|}{1} & \multicolumn{1}{c|}{3.996} & \multicolumn{1}{c|}{\textbf{3.203}} & \multicolumn{1}{c|}{\textbf{0.966}} & \multicolumn{1}{c|}{6.689} \\ \hline
\multicolumn{1}{|c|}{2} & \multicolumn{1}{c|}{\textbf{1.642}} & \multicolumn{1}{c|}{2.036} & \multicolumn{1}{c|}{4.730} & \multicolumn{1}{c|}{\textbf{3.597}} \\ \hline
\multicolumn{1}{|c|}{3} & \multicolumn{1}{c|}{17.068} & \multicolumn{1}{c|}{\textbf{16.945}} & \multicolumn{1}{c|}{17.198} & \multicolumn{1}{c|}{\textbf{17.067}} \\ \hline
\multicolumn{1}{|c|}{4} & \multicolumn{1}{c|}{10.058} & \multicolumn{1}{c|}{\textbf{9.712}} & \multicolumn{1}{c|}{8.547} & \multicolumn{1}{c|}{\textbf{6.234}} \\ \hline
\multicolumn{1}{|c|}{5} & \multicolumn{1}{c|}{18.158} & \multicolumn{1}{c|}{\textbf{16.551}} & \multicolumn{1}{c|}{21.108} & \multicolumn{1}{c|}{\textbf{14.305}} \\ \hline
\multicolumn{1}{|c|}{Average} & \multicolumn{1}{c|}{10.184} & \multicolumn{1}{c|}{9.689} & \multicolumn{1}{c|}{10.510} & \multicolumn{1}{c|}{9.578} \\ \hline
\multicolumn{1}{|c|}{Std. dev.} & \multicolumn{1}{c|}{7.453} & \multicolumn{1}{c|}{7.077} & \multicolumn{1}{c|}{8.446} & \multicolumn{1}{c|}{5.782} \\ \hline
\multicolumn{1}{|c|}{Sequence} & \multicolumn{4}{c|}{\begin{tabular}[c]{@{}c@{}}Coloradar dataset\\ $\|$RMSE$\|$ of attitude {[}\degree{]}\end{tabular}} \\ \hline
\multicolumn{1}{|c|}{aspen\_run0} & \multicolumn{1}{c|}{13.372} & \multicolumn{1}{c|}{\textbf{9.389}} & \multicolumn{1}{c|}{x} & \multicolumn{1}{c|}{x} \\ \hline
\multicolumn{1}{|c|}{arpg\_lab\_run0} & \multicolumn{1}{c|}{8.287} & \multicolumn{1}{c|}{\textbf{7.506}} & \multicolumn{1}{c|}{x} & \multicolumn{1}{c|}{x} \\ \hline
\multicolumn{1}{|c|}{ec\_hallways\_run0} & \multicolumn{1}{c|}{\textbf{18.357}} & \multicolumn{1}{c|}{21.666} & \multicolumn{1}{c|}{x} & \multicolumn{1}{c|}{x} \\ \hline
\multicolumn{1}{|c|}{aspen\_run4} & \multicolumn{1}{c|}{\textbf{7.776}} & \multicolumn{1}{c|}{8.117} & \multicolumn{1}{c|}{x} & \multicolumn{1}{c|}{x} \\ \hline
\multicolumn{1}{|c|}{aspen\_run5} & \multicolumn{1}{c|}{9.363} & \multicolumn{1}{c|}{\textbf{7.947}} & \multicolumn{1}{c|}{x} & \multicolumn{1}{c|}{x} \\ \hline
\multicolumn{1}{|c|}{Average} & \multicolumn{1}{c|}{11.431} & \multicolumn{1}{c|}{10.925} & \multicolumn{1}{c|}{x} & \multicolumn{1}{c|}{x} \\ \hline
\multicolumn{1}{|c|}{Std. dev.} & \multicolumn{1}{c|}{6.045} & \multicolumn{1}{c|}{4.451} & \multicolumn{1}{c|}{x} & \multicolumn{1}{c|}{x} \\ \hline
\multicolumn{1}{l}{} & \multicolumn{1}{l}{} & \multicolumn{1}{l}{} & \multicolumn{1}{l}{} & \multicolumn{1}{l}{} \\
\multicolumn{1}{l}{} & \multicolumn{1}{l}{} & \multicolumn{1}{l}{} & \multicolumn{1}{l}{} & \multicolumn{1}{l}{}
\end{tabular}
\end{table}

\section{Conclusions}\label{sec:dl_conclusion}
In this paper, we presented a novel self-supervised learning framework for finding 3D point correspondences in sparse and noisy point clouds from a \ac{soc} \ac{fmcw} radar. To our knowledge, this is the first learning approach addressing the problem of data association in the challenging setting of 3D point cloud measurements from low-cost, low-power, lightweight, consumer-grade radar chips. In our framework, we leverage the \textit{PointNet} architecture to compute individual point embeddings in each of the two consecutive input point clouds. Subsequently, using transformer architecture and its attention mechanism, we augment the initial embeddings with the reciprocal information from both inputs, to finally form the matching likelihood matrix by calculating the dot product of the augmented embeddings. We provide a self-supervision method using set-based multi-label classification cross-entropy loss, where the ground-truth set of matches is calculated by solving the \ac{lsa} optimization problem. Employing multi-label classification cross-entropy loss enables directly using  correspondences in training. This is crucial since training on odometry error using e.g. \ac{svd}, as used in methods for scanning radars or dense point clouds, is not feasible with the sparse and noisy measurements from the \ac{soc} radar sensor that we use in this work. We applied our framework to the task of \ac{rio} estimation on a small-sized \ac{uav} and showed that it outperforms the default non-learning 3D point matching method. In particular, in an open-source state-of-the-art \ac{rio} framework we switched the 3D point matching algorithm from the default non-learning one to the one presented in this paper while keeping all other settings and parameters unchanged. The reduction in the norm of \ac{rmse} of position estimate calculated over the whole real-world validation dataset in both cases when only matches, and matches with Doppler velocity are used in the estimator reveals that our learning-based method surpasses the non-learning one. We make both our framework and the datasets open-source for the benefit of the research community.

%% file: chapters/conclusion.tex
\chapter{Discussion and Conclusions}\label{chap:concl_chp}

\bigskip

\section{Discussion}

As \ac{slam} technologies get more and more mature, a pertinent challenge which appears, is that of being able to deploy them in various environmental conditions including difficult outdoor as well as indoor scenarios. Those might entail limited visibility due to air obscurants such as fog, smoke or fine particles in the air. Recent advances in millimeter-wave \ac{fmcw} radar technologies gave rise to a body of research directed at investigating localization methods utilizing small-sized \ac{soc} \ac{fmcw} radar sensors as a way to mitigate the impact of the harsh environmental conditions on the autonomous systems. This dissertation and the findings therein aim at advancing this research specifically in the direction of the applicability on small-size \ac{uav}s. To that end, we limited our work to lightweight, consumer-grade sensors having negligible payload as it is crucial for hardware onboard small-size \ac{uav}s. 

The proof-of-concept method presented in~\cref{sec:ekf_anchors} allowed us to conclude the suitability of radar sensors and their accurate distance measurements for \ac{uav} state estimation when fused with the \ac{imu} data. Motivated by our findings in this constrained environment context, we switched to a radar sensor providing 4D point clouds. This in turn allowed us to design a novel state-of-the-art \ac{rio} in which all 4D measurements are used to form residuals in the \ac{ekf} update step. Extending the proposed novel \ac{ekf}-based estimator by adding a buffer with several past radar poses along with their observed measurements enabled further accuracy improvement. Implementing online calibration capabilities significantly simplified the usage of the system while also reducing the \ac{rmse} and improving the consistency. Efficient software implementation made open-source permitted us to test our \ac{ekf}-based \ac{rio} in closed-loop flights and also in visually challenging conditions. Radar-based localization systems, as opposed to \ac{vio}, have no special requirements on initialization procedures since they provide directly metric scale measurements. 

In~\cref{chap:fg_rio} we turned our attention to another state estimation paradigm, that is, the factor graph optimization in order to compare it in a \textit{one-to-one} fashion with the \ac{ekf}-based algorithm. Based on the undertaken analysis we could draw conclusions that on average both approaches perform similarly in terms of position \ac{rmse}. The difference between the two estimators becomes starker when we induce initialization errors which bring the linearization points away from the true state values. In these conditions we observe the benefits of the iterations in the form of faster convergence of the factor graph approach. Nevertheless, it is not to be concluded that the factor graph optimization methods are univocally superior to filters. While exhibiting more resilience towards wrong initialization, they suffer from errors introduced by marginalization. Also, the emerging equivariant filter paradigm is poised to significantly improve the convergence properties of non-iterative filters \cite{mscqf}.

Identifying the front-end robustness in \ac{rio} systems as of crucial importance, we dedicate the~\cref{chap:dl_matching} to research into improving the 3D point correspondence finding within sparse and noisy \ac{fmcw} radar point clouds. To that end, we resort to the deep learning and the transformer architecture which lends itself to finding mutual affinities of points in subsequent 3D radar scans thanks to its attention mechanism.

All in all, we make a step forward on a path towards AI-backed \ac{rio} estimators with online calibration capabilities and deployable onboard in closed-loop flights in computationally constrained hardware, thus offering \ac{uav} localization system with strong environmental resilience using only consumer-grade sensors.

\section{Conclusions}
In conclusion, this dissertation offers successful research advancements in the area of \ac{rio} applied to small-sized \ac{uav}s using consumer-grade, lightweight and commonly available sensors and computing hardware. Several \ac{rio} estimator formulations presented therein together with their successful deployment on real-world platforms, insights into their respective strengths and weaknesses, as well as the application of deep learning methods to improve their front-end processing offer significant progress on the way to the utilization of radar sensors in the area of \ac{uav} perception.